\documentclass[10pt,twocolumn,letterpaper]{article}

\usepackage{wacv}
\makeatletter
\@namedef{ver@everyshi.sty}{}
\makeatother
\usepackage{tikz}

\usepackage{times}
\usepackage{epsfig}
\usepackage{graphicx}
\usepackage{amsmath}
\usepackage{amssymb}
\usepackage{booktabs}
\usepackage[accsupp]{axessibility}
\usepackage{amsmath}
\usepackage{bm}
\usepackage{amsthm}
\usepackage{amsfonts}

\usepackage{multirow}
\usepackage{booktabs}
\usepackage{makecell}
\usepackage{pifont} 
\usepackage{pgfplots}
\usepackage{verbatim}
\usepackage{tabularx}
\usepackage{makecell}
\usepackage{diagbox}
\usepackage{caption}
\usepackage{subcaption}

\usepackage{lipsum, babel}
\usepackage{graphicx}
\usepackage{pgfplots}
\usepackage{pgfplotstable}  
\usetikzlibrary{matrix}
\usepgfplotslibrary{groupplots}
\pgfplotsset{compat=newest}
\usepackage[norelsize, linesnumbered, ruled, lined, boxed, commentsnumbered]{algorithm2e}
\usepackage{algpseudocode}
\definecolor{commentcolor}{rgb}{0.294,0.604,0.608}   % define comment color
\definecolor{pycolor}{rgb}{0.858, 0.188, 0.478}

\newcommand{\cmark}{\ding{51}}
\newcommand{\xmark}{\ding{55}}

\definecolor{kyusuncolor}{rgb}{0.133, 0.545, 0.133}

\definecolor{jaehooncolor}{rgb}{0.133, 0.1, 0.9}

\usepackage{newfloat}
\usepackage{listings}

\newcommand{\figref}[1]{Fig. \ref{#1}}
\newcommand{\tabref}[1]{Table \ref{#1}}
\newcommand{\equref}[1]{(\ref{#1})}

\def\wacvPaperID{1605} % Enter the WACV Paper ID here

\wacvalgorithmstrack   % Uncomment this line if you are submitting to the Algorithms Track.
\wacvfinalcopy % *** Uncomment this line for the final submission

\ifwacvfinal
\usepackage[breaklinks=true,bookmarks=false]{hyperref}
\else
\usepackage[pagebackref=true,breaklinks=true,colorlinks,bookmarks=false]{hyperref}
\fi

\begin{document}

%%%%%%%%% TITLE
\title{3D GAN Inversion with Pose Optimization} % Temporary 

% \and
\author{Jaehoon Ko\thanks{Equal contribution.} $\,^\text{1}$~~~~Kyusun Cho\footnotemark[1] $\,^\text{1}$~~~~Daewon Choi$^\text{1}$~~~~Kwangrok Ryoo$^\text{1,2}$~~~~Seungryong Kim\thanks{Corresponding Author.} $\,^\text{1}$\\
$^\text{1}$Korea University~~~~~~$^\text{2}$LG AI Research\\
{\tt\small \{kjh9604, kyustorm7, daeone0920, seungryong\_kim\}@korea.ac.kr} \\ {\tt\small kwangrok21@lgresearch.ai}
}
% 저도 하와이 데려가요...........

\maketitle
% \thispagestyle{empty}

%%%%%%%%% ABSTRACT
\begin{abstract}
With the recent advances in NeRF-based 3D aware GANs quality, projecting an image into the latent space of these 3D-aware GANs has a natural advantage over 2D GAN inversion: not only does it allow multi-view consistent editing of the projected image, but it also enables 3D reconstruction and novel view synthesis when given only a single image. However, the explicit viewpoint control acts as a main hindrance in the 3D GAN inversion process, as both camera pose and latent code have to be optimized simultaneously to reconstruct the given image. Most works that explore the latent space of the 3D-aware GANs rely on ground-truth camera viewpoint or deformable 3D model, thus limiting their applicability. In this work, we introduce a generalizable 3D GAN inversion method that infers camera viewpoint and latent code simultaneously to enable multi-view consistent semantic image editing. The key to our approach is to leverage pre-trained estimators for better initialization and utilize the pixel-wise depth calculated from NeRF parameters to better reconstruct the given image. We conduct extensive experiments on image reconstruction and editing both quantitatively and qualitatively, and further compare our results with 2D GAN-based editing to demonstrate the advantages of utilizing the latent space of 3D GANs. Additional results and visualizations are available at \url{3dgan-inversion.github.io}
\end{abstract}

%With the recent advances in NeRF-based 3D aware GANs quality, projecting an image into the latent space of these 3D-aware GANs has a natural advantage over 2D GAN inversion: not only does it allow multi-view consistent editing of the projected image, but it also enables 3D reconstruction and novel view synthesis when given only a single image.  However, the explicit viewpoint control acts as a main hindrance in the 3D GAN inversion process, as both camera pose and latent code have to be optimized simultaneously to reconstruct the given image. Most works that explore the latent space of the 3D-aware GANs rely on ground-truth camera viewpoint or deformable 3D model, thus limiting their applicability. In this work, we introduce a generalizable 3D GAN inversion method that infers camera viewpoint and latent code simultaneously to enable multi-view consistent semantic image editing. The key to our approach is to utilize the pixel-wise depth calculated from NeRF parameters to better reconstruct the given image. We conduct extensive experiments on image reconstruction and editing both quantitatively and qualitatively, and further compare our results with 2D GAN-based editing to demonstrate the advantages of utilizing the latent space of 3D GANs. 

%%%%%%%%% BODY TEXT
\section{Introduction}
Recent Generative Adversarial Network (GAN)~\cite{goodfellow2014generative} architectures show incredible results in synthesizing unconditional images with a diverse range of attributes. Especially, StyleGAN~\cite{karras2019style, Karras_2020_improving} has achieved photorealistic visual quality on high-resolution images. Moreover, several works have explored the latent space $\mathcal{W}$ and found its disentangled properties, which enable the control of certain image features and semantic attributes such as gender or hair color.
However, its real-world application is only possible with GAN inversion by bridging the generated image space with real image domain. GAN inversion inverts a real image back into the latent space of a pre-trained GAN, extending the manipulation capability of the model to real images.  

\begin{figure}[t] %{r}{0.5\linewidth}
\centering
\includegraphics[width=\linewidth]{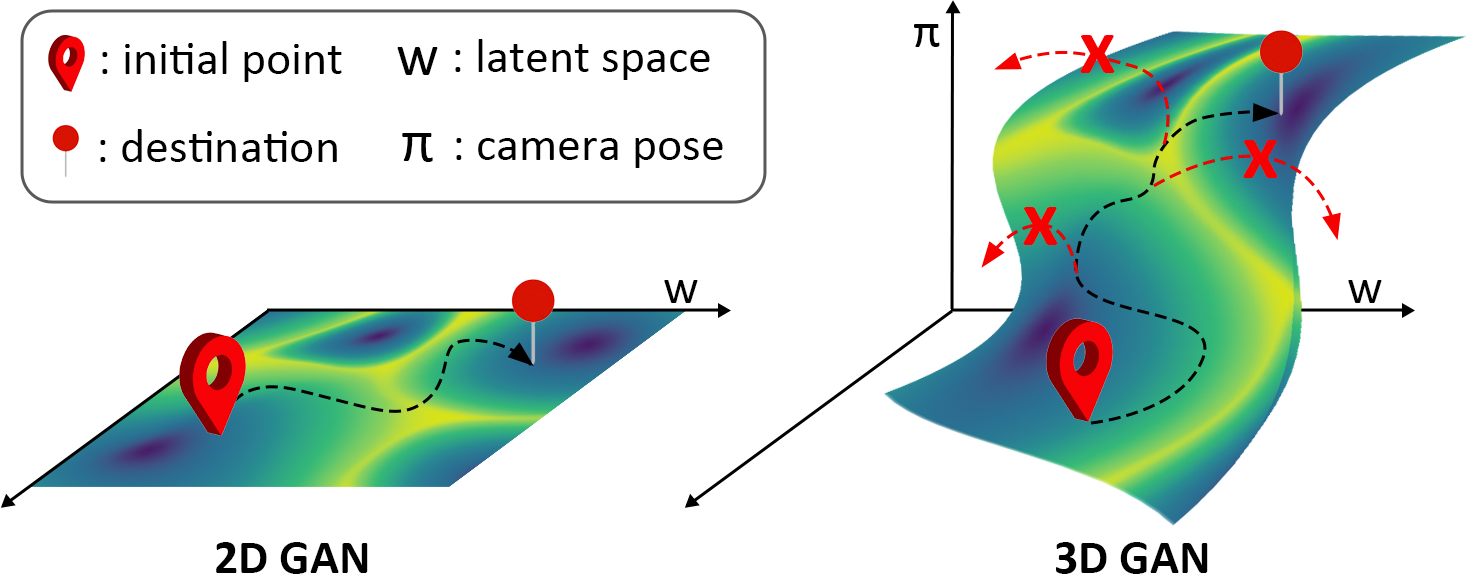}
\caption{\textbf{Comparison of the latent spaces of 2D GANs and 3D GANs.} The explicit control of the camera parameter is the main property of 3D GANs that takes latent space into higher dimensions, making the inversion process harder. Finding both latent features at the same time is much more prone to fall into the local minima.}
\label{fig:difficult3d}

\end{figure}

% However, image editing via GAN inversion is only effective as the expressiveness and editability the utilized GAN model provides.
However, editing an image by projecting it onto the latent space of a 2D GAN makes the task vulnerable to the same set of problems of 2D GANs. 
As training methods of 2D GANs do not take into account the underlying geometry of an object, they offer limited control over the geometrical aspects of the generated image. % Solving Inverse Problems with NeRFGANs
Thus, manipulating image viewpoint using the latent space of 2D GANs always runs into the issue of multi-view inconsistency. %Injecting 3D Perception of Controllable NeRF-GAN into StyleGAN for Editable Portrait Image Synthesis

On the other hand, 3D-aware image synthesis addresses this issue by integrating explicit 3D representations into the generator architecture and enabling explicit control over the camera pose. With the success of neural radiance fields (NeRF)~\cite{mildenhall2020nerf} in novel view synthesis, recent 3D-aware generation models employ a NeRF-based generator, which condition the neural representations on sample noise or disentangled appearance and shape codes in order to represent diverse 3D scenes or object. % Giraffe HD
More recent attempts address the quality gap between 3D GANs and 2D GANs by adopting 2D CNN-based upsampler or efficient point sampling strategy, which enables the generation of high-resolution and photorealistic images on par with 2D GANs.
%Very recently, \cite{Chan2022} presented a style-based 3D-aware GAN architecture, by incorporating the StyleGAN2 generator for 3D scene generalization, which enabled state-of-the-art visual diversity with realistic editing ability while also maintaining view consistency and geometry quality. 
%Very recently, several 3D-aware GANs adopted 2D CNN-based generators to upsample the low-resolution feature map produced by NeRF, and achieved high-resolution image generation with a high level of quality comparable to 2D GANs, while also being 3D controllable. 

% Despite the recent success of 3D-aware GANs and the obvious strengths over their 2D counterpart, leveraging pre-trained 3D-GANs into inverse problem: estimating latent feature and camera pose for given 2D image is especially difficult, since both attributes required simultaneously. Existing works that share the same problem settings optimize latent feature with paired ground-truth camera parameter~\cite{daras2021solving} or rest on the 3D morphable model to find the suitable camera direction~\cite{lin20223d}, both have limits on real-world application. The input 2D image often lack of ground-truth camera parameter, Thus easily fails on inferring its latent code.

Projecting a 2D image onto the learned manifold of these 3D GANs unlocks many opportunities in computer vision applications. Not only can it generate multi-view consistent images from the acquired latent code, but it can also gather the exact surface geometry from the image. Furthermore, recent 3D GANs adopted a style-based generator module to learn the disentangled representations of 3D geometry and appearance. Similar to the latent-based image editing tasks of 2D GANs, by manipulating the latent code of a style-based 3D-aware generator, we can manipulate the semantic attributes of the reconstructed 3D model. 
%Projecting a 2D image onto the learned manifold of these 3D GANs unlocks many opportunities in computer vision applications, especially for solving single-view 3D reconstruction problems. Most of the approaches mainly tried to resolve the inherent depth ambiguity problem by learning a geometric priors from image collections[], or simply rely on the off-the-shelf 

%comparatively tough
%very few papers conducted research on the attribute-controllablity of the latent space of the 3D GANs and user-based editing on real images by leveraging 3D GANs.
Despite its usefulness, few research have been conducted on 3D GAN inversion.
Since 3D-aware GANs initially require a random vector and camera pose for their image generation, an inversion process to reacquire the latent code of a given image necessitates the camera pose of the image, information which real-life images usually often lack. Most of the existing methods require ground-truth camera information or must rely on the off-the-shelf geometry and camera pose from the 3D morphable model, which limit their application to a single category.

In this work, we propose a 3D-GAN inversion method that iteratively optimizes both the latent code and 3D camera pose of a given image simultaneously.
We build upon the recently proposed 2D GAN inversion method that first inverts the given image into a pivot code, and then slightly tunes the generator based on the fixed pivot code (i.e. \textit{pivotal tuning}~\cite{roich2021pivotal}), which showed prominent results in both reconstruction and editability.
Similarly, we acquire both the latent code and camera pose simultaneously as a pivot and fine-tune the pre-trained 3D-GAN to alter the generator manifold into pivots. Note that this is non-trivial since shape and camera direction compromise each other during optimization. %We observe that imperfect camera direction for input image brings defects to latent code optimization.
%(e.g. imperfect camera direction causes undesirable shape deformation to fit in the input image, compromising the defects on pose estimation) 

Recognizing the interdependency between latent code and camera parameter, we use a hybrid of learning and optimization-based approach by first using an encoder to infer a rough estimate of the camera pose and latent code, and further refining it to an optimal destination. As can be seen in our experiments, giving a good initial point for optimizing pivots much less falls into the local minimum.
In order to further enforce the proximity of camera viewpoint, we introduce regularization loss that utilizes traditional depth-based image warping~\cite{Zhou_2017_CVPR}.
%Specifically, we first generate a set of images with a pre-trained 3D GAN model with a randomly given latent code and camera parameter. We feed these images to our encoder

%Leveraging the encoder to set a good initial point 
%In addition, during the fine-tuning of the generator, since the generative NeRF module becomes overfitted to a single view of a scene, we introduce an additional loss function to encourage the new geometry generated from the modified generator does not deviate too much from the likely 3D shape, and thus generate desirable images even from multiple views. 
%utilizing the geometry generated by 3D-GANs.  In order to enforce the proximity of camera viewpoint, we introduce regularization loss to

% To the best of our knowledge, our method is the first generalizable method that does not require a ground-truth camera pose during the inversion process on 3D-GANs. 
We demonstrate that our method enables high-quality reconstruction and editing while preserving multi-view consistency, and show that our results are applicable to a multitude of different categories. While we evaluate our proposed method on EG3D~\cite{Chan2022}, the current state-of-the-art 3D-aware GAN, our method is also relevant to other 3D-aware GANs that leverage NeRF for its 3D representation.

\section{Related Work}
\paragraph{Generative 3D-Aware Image Synthesis.}
3D-aware GANs aim to generate 3D-aware images from 2D image collections.
% Based on the main objective of GANs, they learn which is the most plausible shape regardless of the camera viewpoint. 
The first approaches utilize voxel-based representation~\cite{Nguyen-Phuoc_2019_ICCV}, which lacks fine details in image generation, due to memory inefficiency from its 3D representation.
Starting from~\cite{Schwarz_2020_NEURIPS}, several works achieved better quality by adopting NeRF-based representation, even though they struggle on generating high-resolution images due to the expensive computational cost of volumetric rendering. Some approaches proposed an efficient point sampling strategy~\cite{Schwarz_2020_NEURIPS, Deng_2022_CVPR, xiang2022_gram}, while others adopted 2D CNN-layers to efficiently upsample the volume rendered feature map~\cite{Niemeyer_2021_CVPR, Gu_stylenerf_2022_ICLR, zhou2021_cips3d}. Recently, other methods proposed hybrid representations to reduce computational burden from MLP layers, while achieving high-resolution image generation~\cite{Chan2022, Xu_2022_CVPR, Schwarz2022_voxgraf, Skorokhodov2022_epigraf}. Especially, our work is implemented on EG3D~\cite{Chan2022}, which achieved state-of-the-art image quality while preserving 3D-consistency. \vspace{-5pt}

\paragraph{2D GAN Inversion.}
The first step in applying latent-based image editing on real-world images is to project the image to the latent space of pre-trained GANs.
% Notably, it has been shown that StyleGAN creates a disentangled and semantically rich latent space for 2D image synthesis. Many recent works in 2D GAN Inversion have focused on this latent space. 
%In 2D GAN inversion, two latent spaces are typically used. 
%The W space is single style vector in StyleGAN. It has been shown to be limited expressiveness. To increase expressiveness, W+ space extends W space by using different W across layers~\cite{Abdal_2019_ICCV}. Although this extension is expressive enough to represent real images, it has less editability because of inverting images away from W space~\cite{tov2021designing}.
Existing 2D GAN inversion approaches can be categorized into optimization-based, learning-based, and hybrid methods. Optimization approaches~\cite{Abdal_2019_ICCV,Creswell2019InvertingTG} directly optimize the latent code for a single image. This method can achieve high reconstruction quality but is slow for inference. Unlike per-image optimization, learning-based approaches~\cite{richardson2021encoding,tov2021designing,alaluf2021restyle} use a learned encoder to project images. These methods have shorter inference time but fail to achieve high-fidelity reconstruction. Hybrid approaches are a proper mixture of the two aforementioned methods. \cite{guan2020collaborative,zhu2016generative} used the cooperative learning strategy for encoder and direct optimization. PTI~\cite{roich2021pivotal} fine-tunes StyleGAN parameters for each image after obtaining an initial latent code, solving the trade-off~\cite{tov2021designing} between reconstruction and editability. \vspace{-5pt}

\paragraph{3D GANs Inversion.}
3D GAN inversion approaches share the same goal as 2D GAN inversion with the additional need for the extrinsic camera parameters. Few existing methods solving inverse problems in 3D GANs propose effective training solutions of their own. \cite{daras2021solving} proposed regularization loss term to avoid generating unrealistic geometries by leveraging the popular CLIP~\cite{radford2021learning} model. \cite{lin20223d} can animate the single source image to resemble the target video frames, but is limited to human face as it requires off-the-shelf models~\cite{Feng_SIGGRAPH_2021} to extract expression, pose, and shape. \cite{Cai_2022_CVPR} proposes a joint distillation strategy for training encoder, which is inadequate for 3D GANs that contain mapping function. \vspace{-5pt}

\paragraph{Image Manipulation.}
Image manipulation can be conducted by changing the latent code derived from GAN inversion. Many works have examined semantic direction in the latent spaces of pre-trained GANs and then utilized it for editing. 
While some works~\cite{shen2020interpreting, 10.1145/3447648} use supervision in the form of semantic labels predicted by off-the-shelf attribute classifiers or annotated images, they are often limited to known attributes. 
Thus, other researchers resorted to using an unsupervised approach~\cite{harkonen2020ganspace} or contrastive learning based methods~\cite{yuksel2021latentclr, patashnik2021styleclip} to find meaningful directions. In this work, we leverage GANSpace~\cite{harkonen2020ganspace}, which performs principal component analysis in the latent space, to demonstrate latent-based manipulation of 3D shape.% using 3D GAN inversion. 

% Two approaches are being tried: supervised and unsupervised. The supervised methods need off-the-shelf attribute classifiers or annotated images for specific attributes. InterfaceGAN~\cite{shen2020interpreting} trained SVM to learn hyperplane classifying each binary attribute. StyleFlow~\cite{10.1145/3447648} learned reversible mapping through normalizing flow by using FaceAPI, a face attribute classifier. The Unsupervised methods are the same in finding meaningful direction, but they do it in a self-supervised manner. GANSpace~\cite{harkonen2020ganspace} performed PCA in latent space and used principal components as semantic direction. LatentCLR~\cite{yuksel2021latentclr} explored directions by contrastive learning. Further, StyleCLIP~\cite{patashnik2021styleclip} utilized CLIP, a contrastive language-image pre-training model to edit images by conditioning human words. In our paper, we demonstrate our 3D inversion approach by using these editing methods for latent-based manipulation of 3D shape. 

\section{Preliminaries}
\paragraph{2D GANs Inversion and Pivotal Tuning.}
Given a pretrained 2D generator $\mathcal{G}_\mathrm{2D}(\cdot\,;\theta)$ parameterized by weights $\theta$, 2D GANs inversion aims to find the latent representation $\mathbf{w}$ that can be passed to the generator to reconstruct a given image $x$:
\begin{equation}
    \mathbf{w}^* = \underset{\mathbf{w}}{\mathrm{argmin}}\;\mathcal{L}(x, \mathcal{G}_\mathrm{2D}(\mathbf{w};\theta)),
    \label{eq:inversion_via_optim}
\end{equation}
where the loss function $\mathcal{L}(\cdot,\cdot)$ is usually defined as pixel-wise reconstruction loss %~\cite{?} 
or perceptual loss~\cite{Zhang2018CVPR} between the given image $x$ and reconstructed image $\mathcal{G}_\mathrm{2D}(\mathbf{w};\theta)$. %while $\theta$ is fixed. 

To improve the performance, some other methods aim to optimize an encoder $\mathcal{E}(x;\theta_\mathcal{E})$ with parameters $\theta_\mathcal{E}$ that maps images to their latent representations such that:
\begin{equation}
    \theta^*_\mathcal{E} = \underset{\theta_\mathcal{E}}{\mathrm{argmin}}\;\mathcal{L}(x, \mathcal{G}_\mathrm{2D}(\mathcal{E}(x; \theta_\mathcal{E});\theta)).
    \label{eq:inversion_via_encoder}
\end{equation}
Some recent methods~\cite{zhu2016generative, Bau_2019_ICCV, zhu2020indomain} take a hybrid approach of leveraging the encoded latent representation $\theta_\mathcal{E}(x; \theta_\mathcal{E})$ with learned parameters $\theta_\mathcal{E}$ as an initialization for a subsequent optimization process for \equref{eq:inversion_via_optim}, resulting in a faster and more accurate reconstruction.

%We follow the procedure used in works such as \cite{zhu2016generative, Bau_2019_ICCV, zhu2020indomain}, in which the trained encoder $E$ is leveraged as a fast initialization for the optimization process for \equref{eq:inversion_via_optim}. 

Furthermore, it has been recently well studied that existing GANs inversion methods~\cite{tov2021designing,zhu2020improved,10.1145/3447648} struggle on the trade-offs between reconstruction and editability. To overcome this, \cite{roich2021pivotal} proposed a pivotal tuning stage in a manner that after finding the optimal latent representation $\mathbf{w}^*$, called the pivot code, the generator weights $\theta$ are fine-tuned so that the pivot code can more accurately reconstruct the given image while keeping its editability:
\begin{equation}
    \theta^* = \underset{\theta}{\mathrm{argmin}}\; \mathcal{L}(x, \mathcal{G}_\mathrm{2D}(\mathbf{w}^*; \theta)).
    \label{eq:2d_pivotal_tuning}
\end{equation}
By utilizing the pivot code $\mathbf{w}^*$ and the tuned weights $\theta^*$, the final reconstruction is obtained as $y^* =\mathcal{G}_\mathrm{2D}(\mathbf{w}^*; \theta^*)$. \vspace{-5pt}

\paragraph{NeRF and 3D-aware GANs.}
Neural Radiance Fields (NeRF) \cite{mildenhall2020nerf} achieves a novel view synthesis by employing a fully connected network to represent implicit radiance fields that maps location and direction $(\mathbf{x}, \mathbf{d})$ to color and density $(\mathbf{c}, \sigma)$. Specifically, along with each projected ray $r$ for a given pixel, $M$ points are sampled as $\{ t_i \}^M_{i=1}$, and with the estimated color and density $(\mathbf{c}_i, \sigma_i)$ of each sampled point, the RGB value $c(r)$ for each ray can be calculated by volumetric rendering as:
\begin{equation}
    c(r) = \sum_{i=1}^M T_i (1 - \mathrm{exp}(-\sigma_i \delta_i))\mathbf{c}_i,
\label{eq:volume_rendering}
\end{equation}
where $T_i = \mathrm{exp}\left(-\sum_{j=1}^{i-1} \sigma_j \delta_j\right)$, and $\delta_i$ is the distance between adjacent sampled points such that $\delta_i = t_{i+1} - t_{i}$.
Furthermore, per-ray depth $d(r)$ can also be approximated as
\begin{equation}
    d(r) = \sum_{i=1}^{M} T_i (1 - \mathrm{exp}(-\sigma_i \delta_i))t_i.
\label{eq:volume_rendering_depth}
\end{equation}
While NeRF trains a single MLP on multiple posed images of a single scene, NeRF-based generative models~\cite{Gu_stylenerf_2022_ICLR, Chan2022} often condition the MLP on latent style code $\mathbf{w}$ that represents individual latent features learned from unposed image collections. These style-based 3D GANs have been popularly used in 3D aware image generation~\cite{Gu_stylenerf_2022_ICLR,Chan2022,zhou2021_cips3d}, and we denote this generator $\mathcal{G}_\mathrm{3D}(\mathbf{w},\pi; \theta)$
with a given latent code $\mathbf{w}$, which can be formally represented as the conditional function: $\{c,d\}=\mathcal{G}_\mathrm{3D}(\mathbf{w},\pi; \theta)$, where $c$ is rendered $\mathit{RGB}$ image and $d$ is depth map.

\section{Method}
\subsection{Overview}
\begin{figure}[!t]
\centering
\includegraphics[width=.44\textwidth]{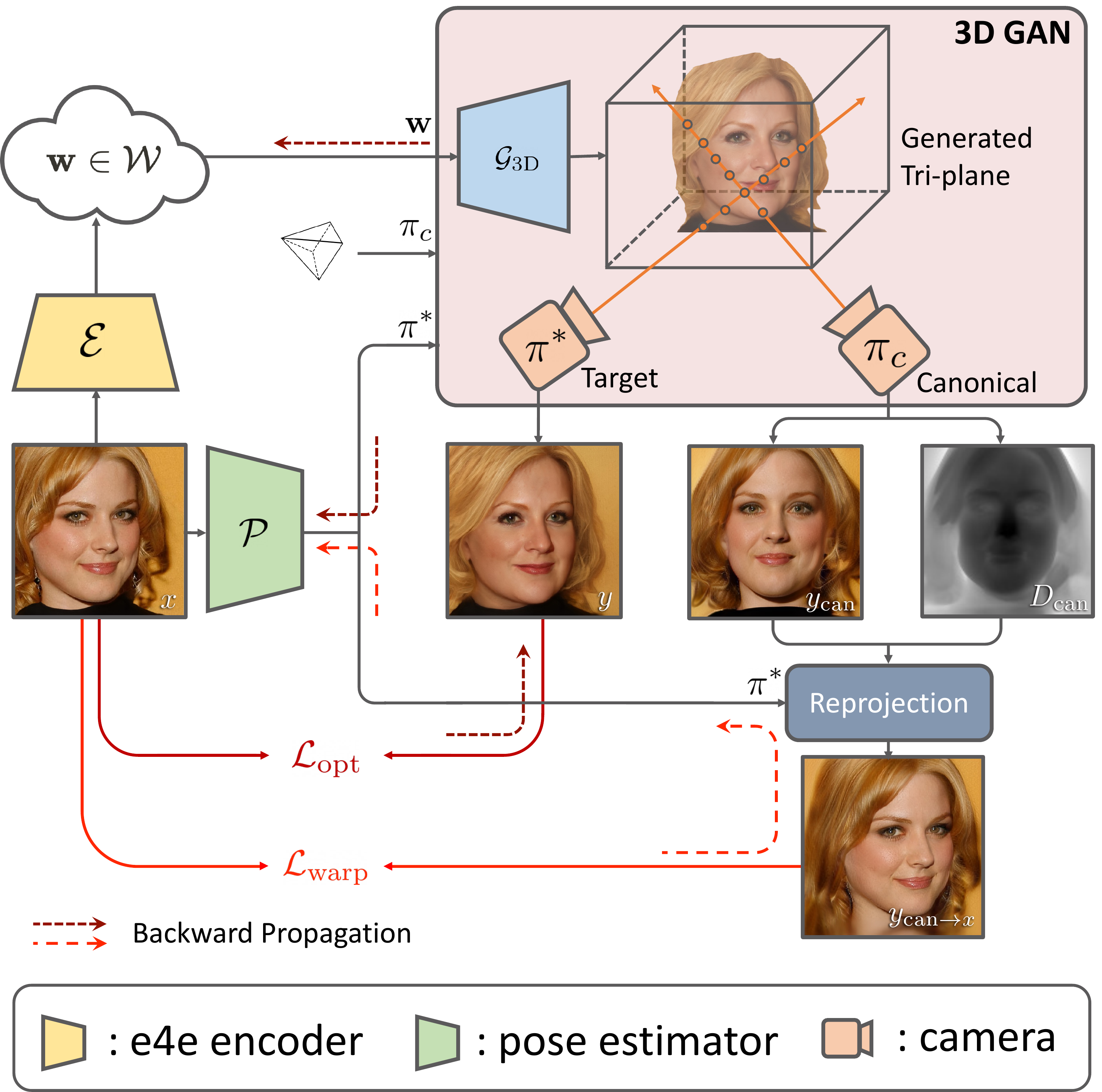}\hfill\\
\caption{\textbf{Overall architecture.} This figure shows our method using depth-based warping to optimize latent code and camera pose simultaneously~\ref{optimization_step}.}
\label{fig:overall}
\end{figure}

Our objective, which we call 3D GAN inversion, is to project a real photo into the learned manifold of the GAN model. However, finding the exact match for $\mathbf{w}^*$ and $\pi^*$ for a given image $x$ is a non-trivial task since one struggles to optimize if the other is drastically inaccurate. To overcome this, we follow \cite{zhu2016generative, Bau_2019_ICCV, zhu2020indomain} to first construct two encoders that approximately estimate the initial codes from $x$ by $\mathbf{w}=\mathcal{E}(x;\theta_\mathcal{E})$ and $\pi=\mathcal{P}(x;\theta_\mathcal{P})$~(Sec. \ref{pretrain_encoder}), and further solving optimization problem(Sec. \ref{optimization_step}).
%\equref{eq:3d_inversion}..
In particular, we introduce loss functions employed in (Sec. \ref{optimization_step}) and further discuss the effects and purposes of employing these loss functions(Sec. \ref{loss_functions}).
An overview of our method is shown in \figref{fig:overall}.
%(ADD overview...!)

% \input{Tables/algorithm}
% \input{Figures/pseudo}
%While we use the same loss functions for optimizing $w$ as in 2D GAN inversion, in order to iteratively refine $\pi$, we incorporate the depth-based reprojection method from unsupervised depth estimation literature~\cite{garg2016unsupervised,Zhou_2017_CVPR,monodepth17,Godard_2019_ICCV} into the optimization-based inversion approach used in 2D GAN inversion. 
%Additionally, as NeRF tends to overfit given only a single view and result in inconsistent geometry with distracting artifacts, we utilize geometry-aware regularizations for the pivotal tuning step. 

%The intuition of NeRF-based GANs is that the geometry prior instilled by the conditional NeRF is enough to disentangle the camera viewpoint given an unposed batch of images successfully. However, the CNN-based latent encoder $E$ has no such prior and is susceptible to inaccurate predictions when two different views of the same object are given. 
%For a better reconstruction, the estimated latent code has to be refined further from an optimal viewpoint. 
%Similarly, the camera pose should also be well initialized, if the given image is taken from a completely different viewpoint, and the essential attributes used to infer the latent code are occluded, the latent code cannot be optimized.

\begin{figure*}[!t]
\centering
\newcolumntype{M}[1]{>{\centering\arraybackslash}m{#1}}
\setlength{\tabcolsep}{1pt}
\renewcommand{\arraystretch}{0.5}
\begin{tabular}{M{0.113\linewidth}M{0.113\linewidth} @{\hskip 0.005\linewidth}|@{\hskip 0.005\linewidth} M{0.113\linewidth}M{0.113\linewidth}M{0.113\linewidth} @{\hskip 0.005\linewidth}|@{\hskip 0.005\linewidth} M{0.113\linewidth}M{0.113\linewidth}M{0.113\linewidth}}

\multicolumn{1}{c}{Input} 
& \multicolumn{1}{c}{Ours} 
&\multicolumn{1}{c}{SG2} 
&\multicolumn{1}{c}{SG2 $\mathcal{W}+$}
&\multicolumn{1}{c}{PTI} 
&\multicolumn{1}{c}{SG2$^\dagger$}
&\multicolumn{1}{c}{SG2$^\dagger$ $\mathcal{W}+$} & \multicolumn{1}{c}{PTI$^\dagger$ } \\

\includegraphics[width=\linewidth]{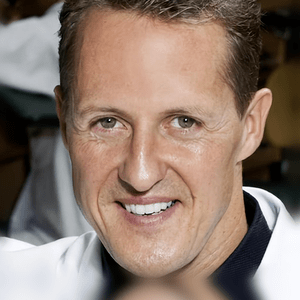}\hfill &

\includegraphics[width=\linewidth]{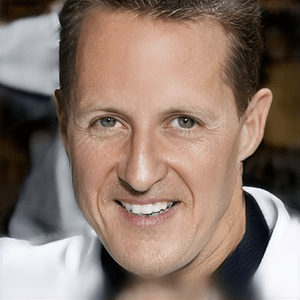}\hfill &
\includegraphics[width=\linewidth]{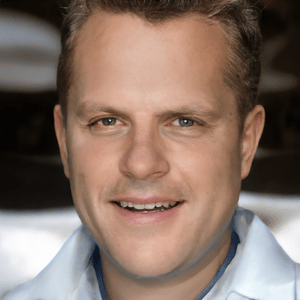}\hfill & 
\includegraphics[width=\linewidth]{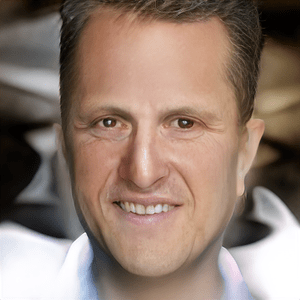}\hfill& 
\includegraphics[width=\linewidth]{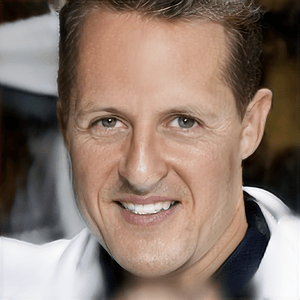}\hfill &
\includegraphics[width=\linewidth]{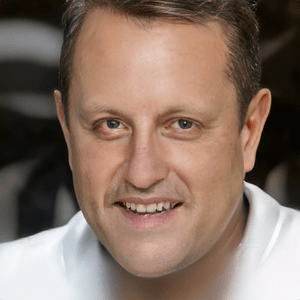}\hfill &
\includegraphics[width=\linewidth]{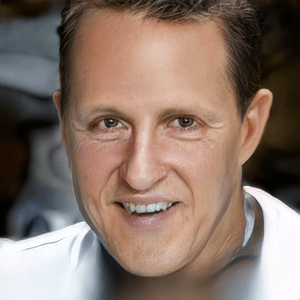}\hfill &
\includegraphics[width=\linewidth]{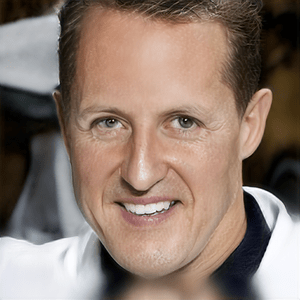}\hfill \\
&
\includegraphics[width=\linewidth]{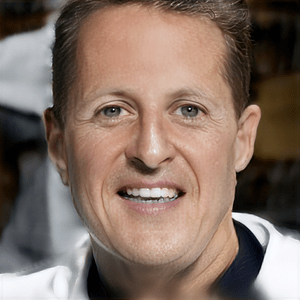}\hfill &
\includegraphics[width=\linewidth]{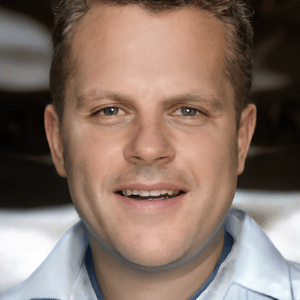}\hfill & 
\includegraphics[width=\linewidth]{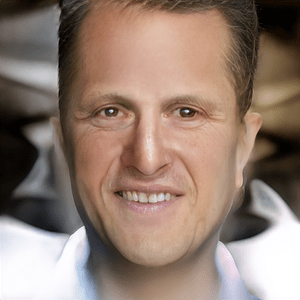}\hfill& 
\includegraphics[width=\linewidth]{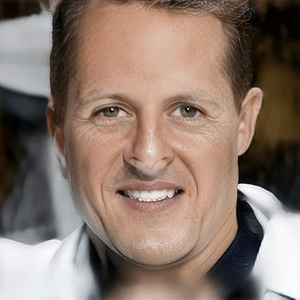}\hfill &
\includegraphics[width=\linewidth]{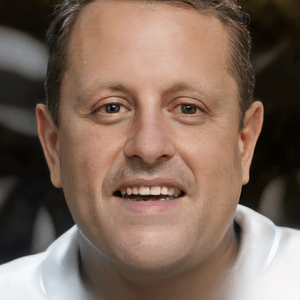}\hfill &
\includegraphics[width=\linewidth]{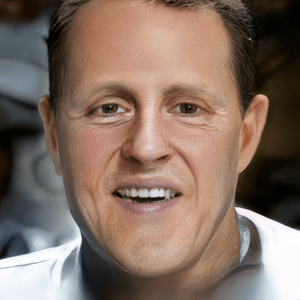}\hfill &
\includegraphics[width=\linewidth]{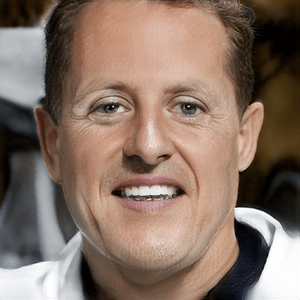}\hfill \\
&
\includegraphics[width=\linewidth]{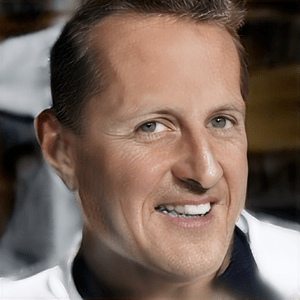}\hfill &
\includegraphics[width=\linewidth]{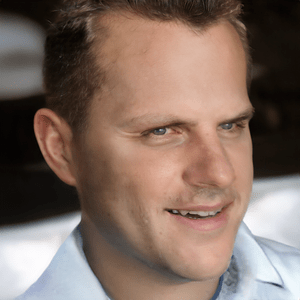}\hfill & 
\includegraphics[width=\linewidth]{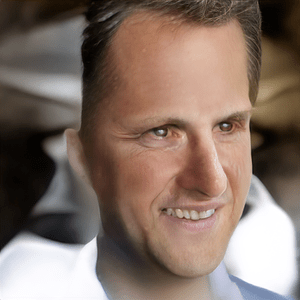}\hfill& 
\includegraphics[width=\linewidth]{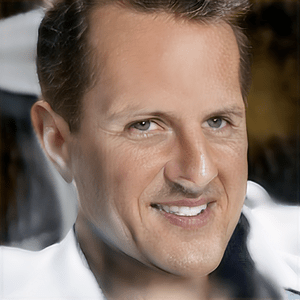}\hfill &
\includegraphics[width=\linewidth]{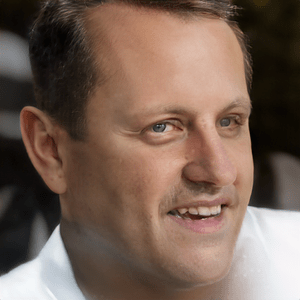}\hfill &
\includegraphics[width=\linewidth]{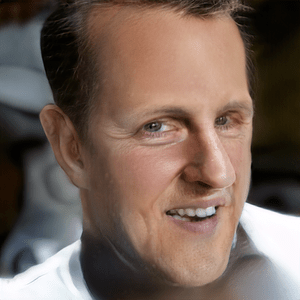}\hfill &
\includegraphics[width=\linewidth]{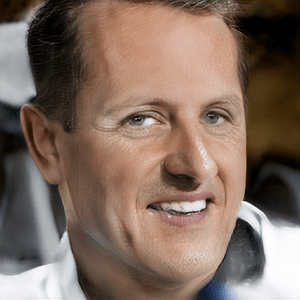}\hfill \\

\end{tabular}
\caption{\textbf{Comparison of novel view synthesis of out-of-domain samples.} Given the optimized camera pose $\hat{\pi}$ and latent code $\hat{\mathbf{w}}$ obtained by each method, we explicitly control the viewpoint of the generated facial scene, by differing $\pi$ for different camera viewpoint. We compare our 3D GAN inversion method to standard 2D GAN inversion methods by applying the gradient-based optimization to both the latent code and the camera pose. We also leverage the same methods only on the latent code with the given ground-truth camera pose, and show the results labeled with $^\dagger$. }
\label{fig:reconstruction_comparison_mesh}

\end{figure*}

\subsection{Latent Encoder $\mathcal{E}$ and Pose Estimator $\mathcal{P}$} \label{pretrain_encoder}
For better 3D GANs inversion, utilizing a well-trained estimator for initialization should be considered~\cite{Hsu2019QuatNetQH,valle2020multi}, but it is a solution limited to a single category. To obtain category-agnostic estimator, 
%we present to use a pre-trained neural renderer~\cite{?}. Specifically, %??
we first generate a pseudo dataset and its annotation pair $\{(\mathbf{w}_\mathrm{ps}, \pi_\mathrm{ps}), x_\mathrm{ps}\}$ to pre-train our encoders, where $x_\mathrm{ps} = \mathcal{G}_\mathrm{3D}(\mathbf{w}_\mathrm{ps},\pi_\mathrm{ps}; \theta)$. Thanks to the generation power of 3D-aware GANs, we can generate nearly unlimited numbers of pairs within the generator's manifold.
%as exemplified in \figref{fig:pseudo}. 
More specifically, for given latent encoder $\mathcal{E}$, let $\bigtriangleup\mathbf{w} = \mathcal{E}(x_\mathrm{ps};\theta_\mathcal{E})$ denote the output of the encoder, where $\mathbf{w}\in\mathbb{R}^{1\times512}$. Following the training strategy of \cite{tov2021designing}, we employ the generator $\mathcal{G}_\mathrm{3D}$ and its paired discriminator $\mathcal{D}$ to guide encoder to find the best replication of $x_\mathrm{ps}$ with $\bar{\mathbf{w}}+\bigtriangleup\mathbf{w}$, where $\bar{\mathbf{w}}$ is an average embeddings of $\mathcal{G}_\mathrm{3D}$. 
We provide more detailed implementation procedure of pre-training each network in the Appendix.

\subsection{Optimization} \label{optimization_step}
% Similarly, the estimated camera pose also have to be refined, as a fine-grained camera pose is required to improve the pixelwise comparison reconstruction. 
% While viewpoint estimation is a widely studied task with recent works achieving SOTA performance on fine-grained pose estimation, and some not limited to human face, our intuition is that an geometry-aware optimization to obtain relative pose given two similar object but different viewpoints is a far more easier task with a high potential to be more accurate compared to a general viewpoint estimation trained on a range of different objects.
%In doing so, not only do we need to find the optimal latent representation, we also need to acquire the camera pose of the real photo.
After the pre-traininig step, given an image $x$, the learnable latent vector and camera pose are first initialized from trained estimators as $\mathbf{w}_\mathrm{init}=\bar{\mathbf{w}} + \mathcal{E}(x;\theta_\mathcal{E})$ and $\pi_\mathrm{init}=\mathcal{P}(x;\theta_\mathcal{P})$. Subsequently, they are further refined for a more accurate reconstruction. 
% Subsequently, they are further refined to reach the optimal code $\mathbf{w}^*$ and $\pi^*$. 
In this stage, we reformulate optimization step in \eqref{eq:inversion_via_optim} into the 3D GAN inversion task, in order to optimize the latent code and camera viewpoint starting from each initialization $\{\mathbf{w}_\mathrm{init}, \pi_\mathrm{init} \}$, such that:
\begin{equation}
    \mathbf{w}^*, \pi^*, n^* = \underset{\mathbf{w}, \pi, n}{\mathrm{argmin}}\;\mathcal{L}^\mathrm{opt}(x, \mathcal{G}_\mathrm{3D}(\mathbf{w}, \pi, n;\theta)),
    \label{eq:3d_inversion}
\end{equation}
where $n$ denotes the per-layer noise inputs of the generator and $\mathcal{L}^\mathrm{opt}$ contains employed loss functions on the optimization step.
Note that, following the latent code optimization method in \cite{roich2021pivotal}, we use the native latent space $\mathcal{W}$ which provides the best editability.

In addition, in the pivotal-tuning step, using the optimized latent code $\mathbf{w}^*$ and optimized camera pose $\pi^*$, we augment the generator's manifold to include the image by slightly tuning $\mathcal{G}_\mathrm{3D}$ with following reformulation of \eqref{eq:2d_pivotal_tuning}:
\begin{equation}
    \theta^* = \underset{\theta}{\mathrm{argmin}}\;\mathcal{L}^\mathrm{pt}(x, \mathcal{G}_\mathrm{3D}(\mathbf{w}^*, \pi^*, n^*;\theta)).
    \label{eq:3d_inversion_tuning}
\end{equation}
In this optimization, following \cite{roich2021pivotal}, we unfreeze the generator and tune it to reconstruct the input image $x$ with given $\mathbf{w}^*$ and $\pi^*$, which are both constant. We also implement the same locality regularization in \cite{roich2021pivotal}.
Again, $\mathcal{L}^\mathrm{pt}$ denotes a combination of loss functions on the pivotal-tuning step.
%We measure reconstruction using LPIPS loss function~\cite{Zhang2018CVPR}, and define the objective function for latent code optimization as:
%Along with \eqref{eq:objective_function}, we utilized additional loss term to regularize pose optimization.  
%최종 pivot들 언급하고 다음이랑 이어지게?
%As illustrated in Fig(), the background of the generated images usually considered as a flat surface at the end of tri-plane. Warping every single pixel into other viewpoints can leads to negative effects for optimization due to the unrealistic shape representation. Therefore, we also generate a discrete mask to filter the loss from the background pixels, whose threshold is the mean value of depth distance.

%\subsection{Pivotal Tuning of the 3D Generator}
%2D GAN inversion methods often use the $\mathcal{W}+$ space for more accurate reconstruction but at the cost of compromised editability. 
%\cite{roich2021pivotal} avoids this problem by tuning the pretrained generator and augmenting the latent space $\mathcal{W}$ to include the image. 
%We apply the same technique to our 3D GAN inversion, by fine-tuning the conditional NeRF so that a more accurate reconstruction can be obtained given the same optimized latent code and camera pose.

\subsection{Loss functions} \label{loss_functions}
\paragraph{LPIPS and MSE Loss.}
To reconstruct the given image $x$, we adopt commonly used LPIPS loss for both optimization and pivotal tuning step. As stated in \eqref{eq:3d_inversion}, the loss is used to train the both latent code $\mathbf{w}$ and camera pose $\pi$. Additional mean square error is given only at the pivotal tuning step, which is commonly used to regularize the sensitivity of LPIPS to adversarial examples. Formally, our losses can be defined by:
\begin{equation} %lpips
    \mathcal{L}_\mathrm{lpips} = \mathcal{L}_\mathrm{lpips}(x, \mathcal{G}^\mathrm{c}_\mathrm{3D}(\mathbf{w}, \pi, n;\theta)),
    \label{eq:lpips_loss}
\end{equation} \vspace{-10pt}
\begin{equation} %L2
    \mathcal{L}_{L2} = \mathcal{L}_{L2}(x, \mathcal{G}^\mathrm{c}_\mathrm{3D}(\mathbf{w}, \pi, n;\theta)).
    \label{eq:L2_loss}
\end{equation}
%\kyusun{
%In practice, we downsample both the target image and the synthesized image to $256 \times 256$ for increased performance and stability. 
%}

%\kyusun{
%Furthermore, for style-based generators such as EG3D~\cite{Chan2022}, the per-layer noise inputs of the generator should be of marginal importance in the final visual appearance. Thus, we apply the same noise regularization term of \cite{Karras2019stylegan2} to prevent the noise vector from carrying coherent signal. The regularization enforces the auto-correlation coefficients of the noise inputs to match unit Gaussian noise, and is denoted as $\mathcal{L}_n(n)$.
%}
\vspace{-5pt}

\paragraph{Depth-based Warping Loss.}
Similar to~\cite{Zhou_2017_CVPR}, every point on the target image can be warped into other viewpoints. We consider the shape representation of $\mathbf{w}$ latent code plausible enough to fit in the target image. 
%Note that the shape priors of pre-trained generator, the optimizable w latent code within the generator manifold represents plausible geometry. 
Given a canonical viewpoint $\pi_\mathrm{can}$, we generate a pair of image and depth map $\{y_\mathrm{can}, D_\mathrm{can} \} = \mathcal{G}_\mathrm{3D}(\mathbf{w}, \pi_\mathrm{can}; \theta)$. 
Let $y_\mathrm{can}(r)$ denote the homogeneous coordinates of a pixel $\gamma$ in the generated image of a canonical view $\pi_\mathrm{can}$ using \eqref{eq:volume_rendering}. Also for each $y_\mathrm{can}(r)$ we obtain the depth value $D_\mathrm{can}(r)$ by using \eqref{eq:volume_rendering_depth}.

Then we can obtain $y_\mathrm{can}(r)$'s projected coordinates onto the source view $\pi_x$ denoted as $\hat{y}_x(r)$ by 
\begin{equation}
    \hat{y}_x(r) \sim K\hat{\pi}_{\mathrm{can} \rightarrow x} D_\mathrm{can}(r) K^{-1} y_\mathrm{can}(r),
\end{equation}
where K is the camera intrinsic matrix, and $\hat{\pi}_{\mathrm{can} \rightarrow x}$ is the predicted relative camera pose from canonical to source. 
As $\hat{x}(r)$ for every pixel $r$ are continuous values, following \cite{Zhou_2017_CVPR}, we exploit the differentiable bilinear sampling mechanism proposed in \cite{jaderberg2015spatial} to obtain the projected 2D coordinates. 

For simplicity of notation, from a generated image $y_\mathrm{can}=\mathcal{G}^\mathrm{c}_\mathrm{3D}(\mathbf{w}, \pi_\mathrm{can}; \theta)$, 
we denote the projected image as $y_{\mathrm{can} \rightarrow x} = y_\mathrm{can} \langle proj(D_\mathrm{can}, \pi_{\mathrm{can} \rightarrow x} ,K) \rangle $, where $proj(\cdot)$ is the resulting 2D image using the depth map $D_\mathrm{can}$ and $	\langle\cdot \rangle$ denotes the bilinear sampling operator, and define the objective function to calculate $\pi_{\mathrm{can} \rightarrow x}$:
\begin{equation}
% \pi_{c \rightarrow x} = \underset{\pi }{\mathrm{argmin}} \; 
\mathcal{L}_\mathrm{warp} = \mathcal{L}_\mathrm{lpips}\left(x, y_\mathrm{can} \langle proj(D_\mathrm{can}, \pi,K) \rangle \right),
\end{equation}
again using a LPIPS loss to compare the two images. \vspace{-5pt}
% or
% \begin{equation}
% \begin{split}
% \hat{\theta}_E, \hat{\theta}_P = \underset{\theta_E, \theta_P}{\mathrm{argmin}} \;\mathcal{L} (x,
% \mathcal{R}(\pi_c, G(E(x; \theta_E); \theta))  \\    \langle proj(\mathcal{D}(\pi_c, G(E(x; \theta_E); \theta)), P(x; \theta_P), K) \rangle)
% \end{split}
% \end{equation}

\paragraph{Depth Regularization Loss.}
Neural radiance field is infamous for its poor performance when only one input view is available. Although tuning the parameters of 2D GANs seem to retain its latent editing capabilities, we found the NeRF parameters to be much more delicate, and tuning them to a single view degrades the 3D structure before reaching the desired expressiveness, resulting in low-quality renderings at novel views. 

% of NeRF-GANs and show undesirable artifacts or bizarre distortions on the estimated depth. Pivotal tuning overfits the NeRF parameters to the given view, and result in 

To mitigate this problem, we take advantage of the geometry regularization used in \cite{Niemeyer2021Regnerf} and encourage the generated depth to be smooth, even from unobserved viewpoints. The regularization is based on the real-world observation that real geometry or depth tends to be smooth, and is more likely to be flat, and formulated such that depth for each pixel should not be too different from those of neighboring pixels.
We enforce the smoothness of the generated depth $D(r)$ for each pixel $r$ with the depth regularization loss:
\begin{equation}
\begin{split}
    \mathcal{L}_{DR}(D) =  \sum_{i, j = 1}^{H-1, W-1} \left( \left(D(r_{i,j}) - D(r_{i+1, j}) \right)^2  \right. \\ + \left. ( \left(D(r_{i,j}) - D(r_{i, j+1}) \right)^2         \right),
\end{split}
\end{equation}
where $H$ and $W$ are the height and width of the generated depth map, and $r_{i,j}$ indicates the ray through pixel (i, j). 
Note that while \cite{Niemeyer2021Regnerf} implements the geometry regularization by comparing overlapping patches, we utilize the full generated depth map $D$ for our implementation.

\paragraph{Overall Loss Function.}
Ultimately, we define the entire optimization step with a generated image and depth $\{y, D\} = \mathcal{G}_\mathrm{3D}(\mathbf{w}, \pi, n;\theta)$:
\begin{align}
\begin{split}
 \mathcal{L}^\mathrm{opt} = & \;\,\mathcal{L}_{\mathrm{lpips}}(x, y)\\ & + \lambda_\mathrm{warp}\mathcal{L}_{\mathrm{warp}}(x, y_\mathrm{can}, D) + \lambda_{n}\mathcal{L}_{n}(n),
\end{split}
\end{align}
and the pivotal tuning process is defined by:
\begin{align}
\begin{split}
 \mathcal{L}^{\mathrm{pt}} = & \;\,\mathcal{L}_{\mathrm{lpips}}(x, y)\\ & + \lambda_{L2}\mathcal{L}_{L2}(x, y) + \lambda_{DR}\mathcal{L}_{DR}(D),
\end{split}
\end{align}
where $\mathcal{L}_{n}$ denotes the noise regularization loss proposed in \cite{Karras2019stylegan2}, which prevents the noise from containing crucial signals of the target image.
\section{Experimental Results}
\subsection{Experimental Settings}
\begin{figure*}
\centering
\newcolumntype{M}[1]{>{\centering\arraybackslash}m{#1}}
\setlength{\tabcolsep}{1pt}
\renewcommand{\arraystretch}{0.5}
\begin{tabular}{M{0.113\linewidth}M{0.113\linewidth} @{\hskip 0.005\linewidth}|@{\hskip 0.005\linewidth} M{0.113\linewidth}M{0.113\linewidth}M{0.113\linewidth} @{\hskip 0.005\linewidth}|@{\hskip 0.005\linewidth} M{0.113\linewidth}M{0.113\linewidth}M{0.113\linewidth}}

\multicolumn{1}{c}{Input} 
& \multicolumn{1}{c}{Ours} 
&\multicolumn{1}{c}{SG2} 
&\multicolumn{1}{c}{SG2 $\mathcal{W}+$}
&\multicolumn{1}{c}{PTI} 
&\multicolumn{1}{c}{SG2$^\dagger$}
&\multicolumn{1}{c}{SG2$^\dagger$ $\mathcal{W}+$} & \multicolumn{1}{c}{PTI$^\dagger$ } \\

\includegraphics[width=\linewidth]{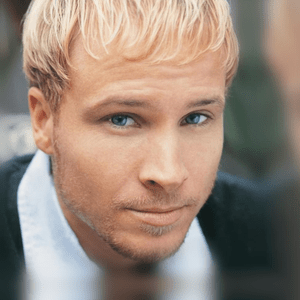}\hfill &
\includegraphics[width=\linewidth]{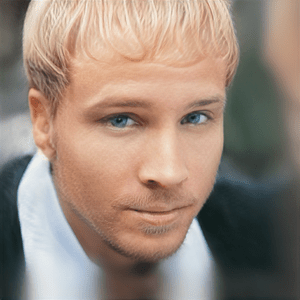}\hfill &
\includegraphics[width=\linewidth]{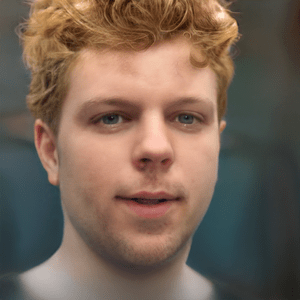}\hfill & 
\includegraphics[width=\linewidth]{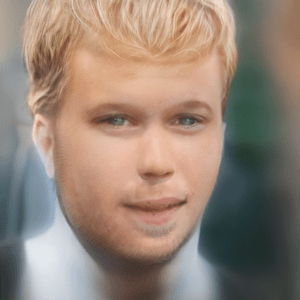}\hfill& 
\includegraphics[width=\linewidth]{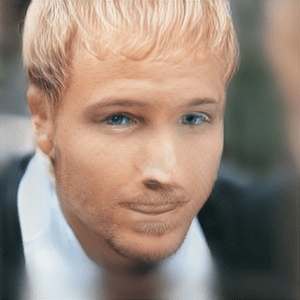}\hfill &
\includegraphics[width=\linewidth]{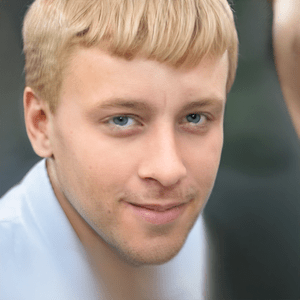}\hfill &
\includegraphics[width=\linewidth]{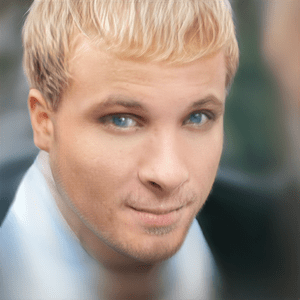}\hfill &
\includegraphics[width=\linewidth]{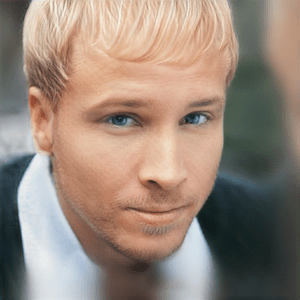}\hfill \\

&
\includegraphics[trim=540 40 260 0,clip,width=\linewidth, height=\linewidth]{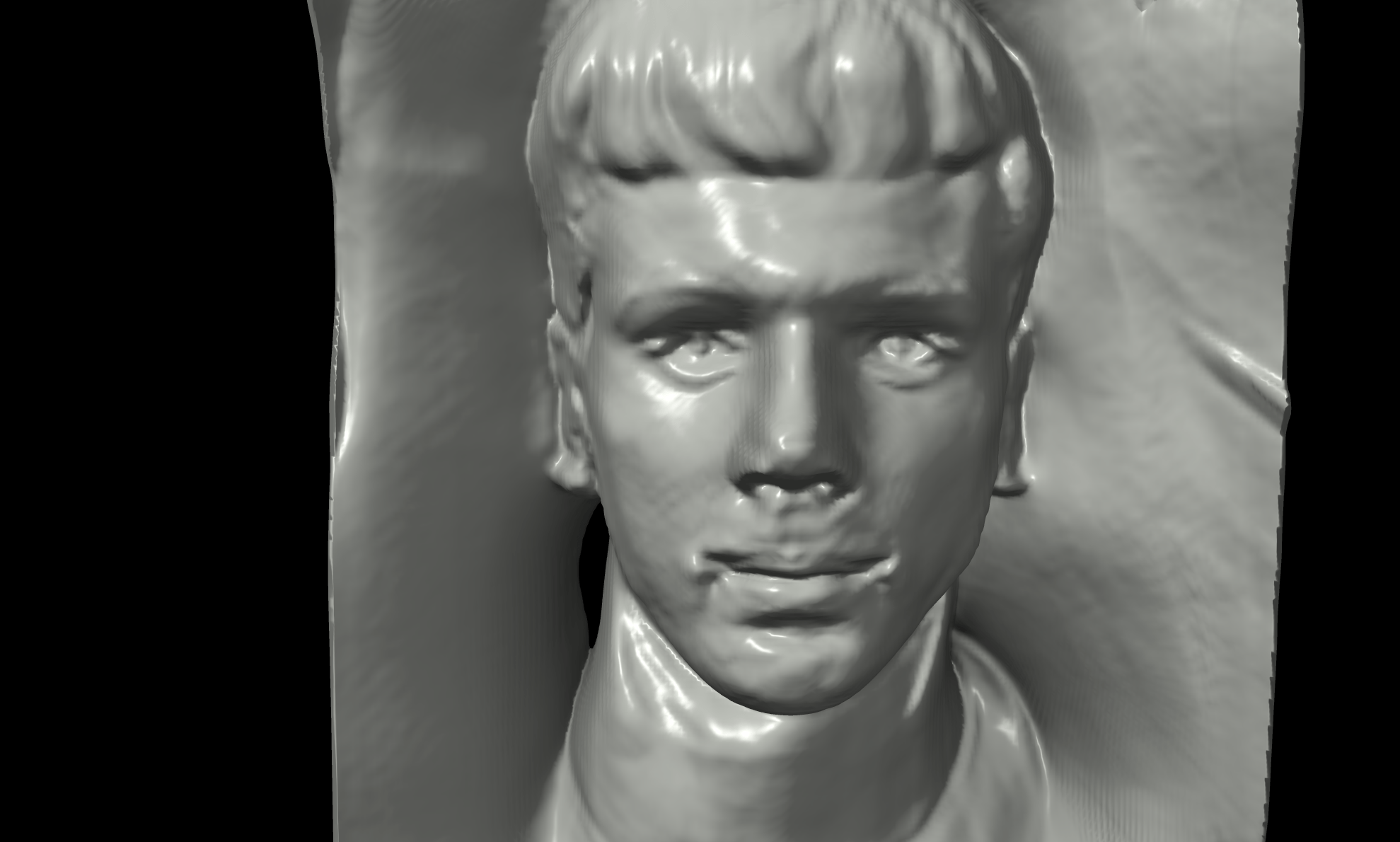}\hfill &
\includegraphics[trim=540 40 260 0,clip,width=\linewidth, height=\linewidth]{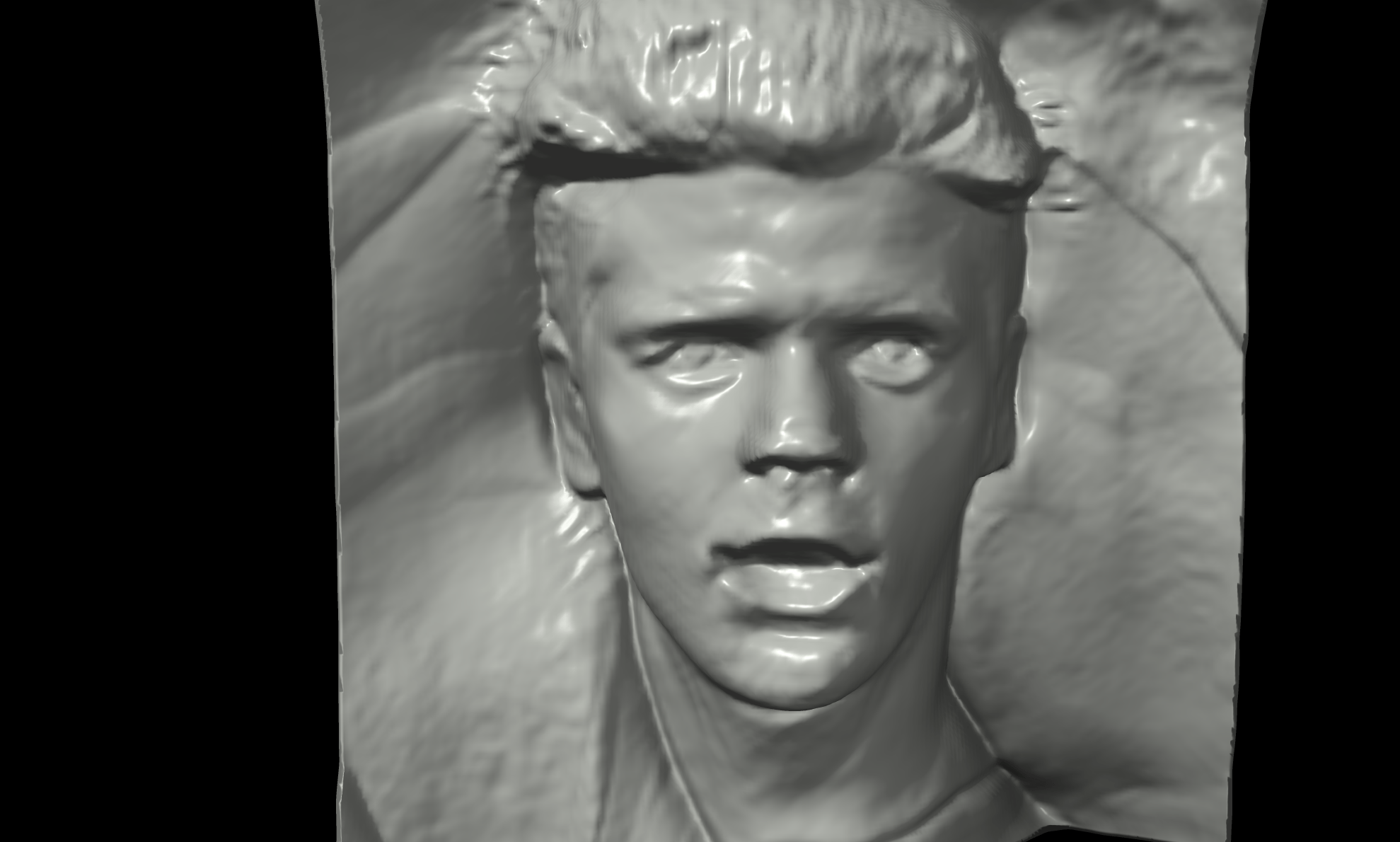}\hfill &
\includegraphics[trim=540 40 260 0,clip,width=\linewidth, height=\linewidth]{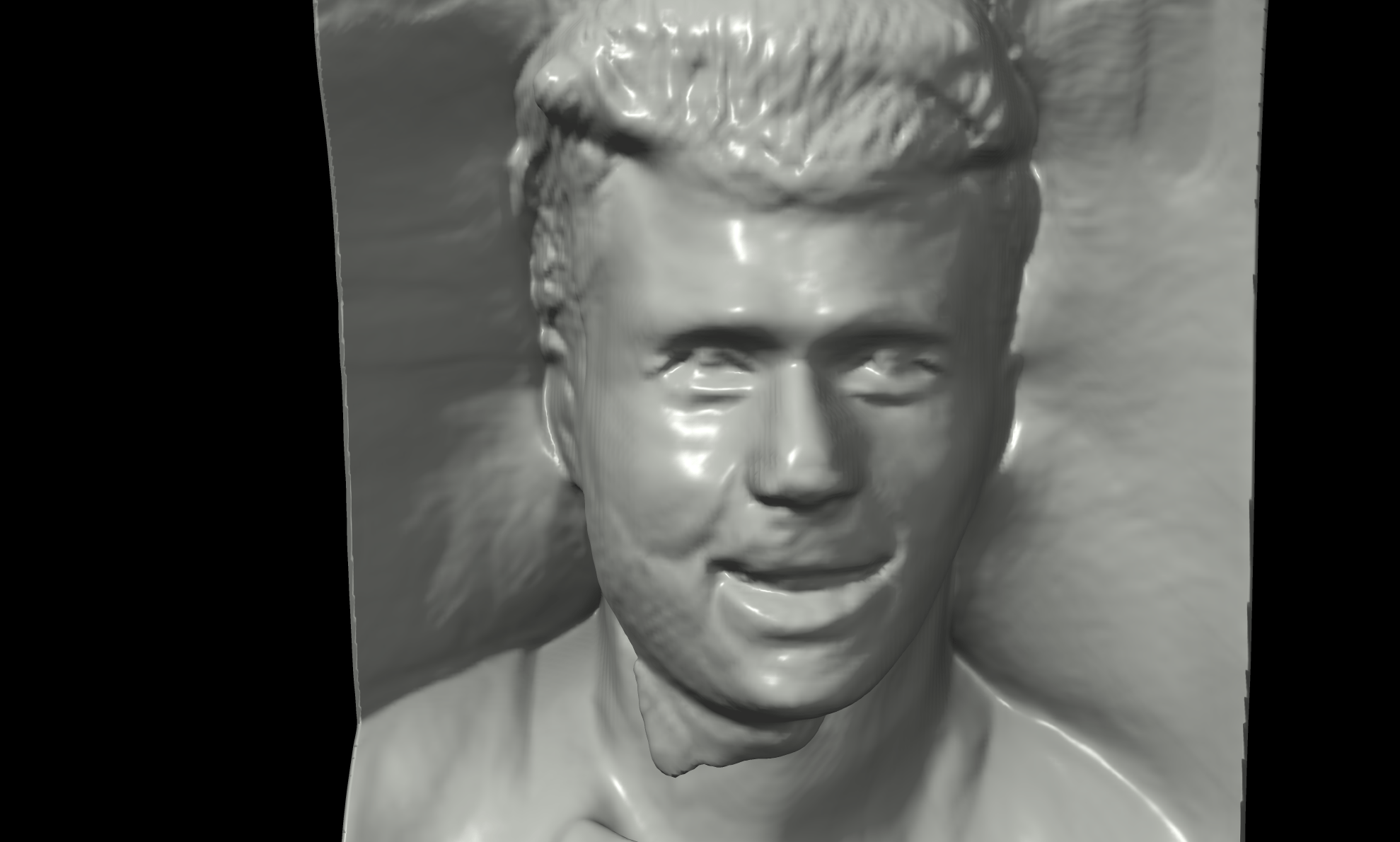}\hfill &
\includegraphics[trim=540 40 260 0,clip,width=\linewidth, height=\linewidth]{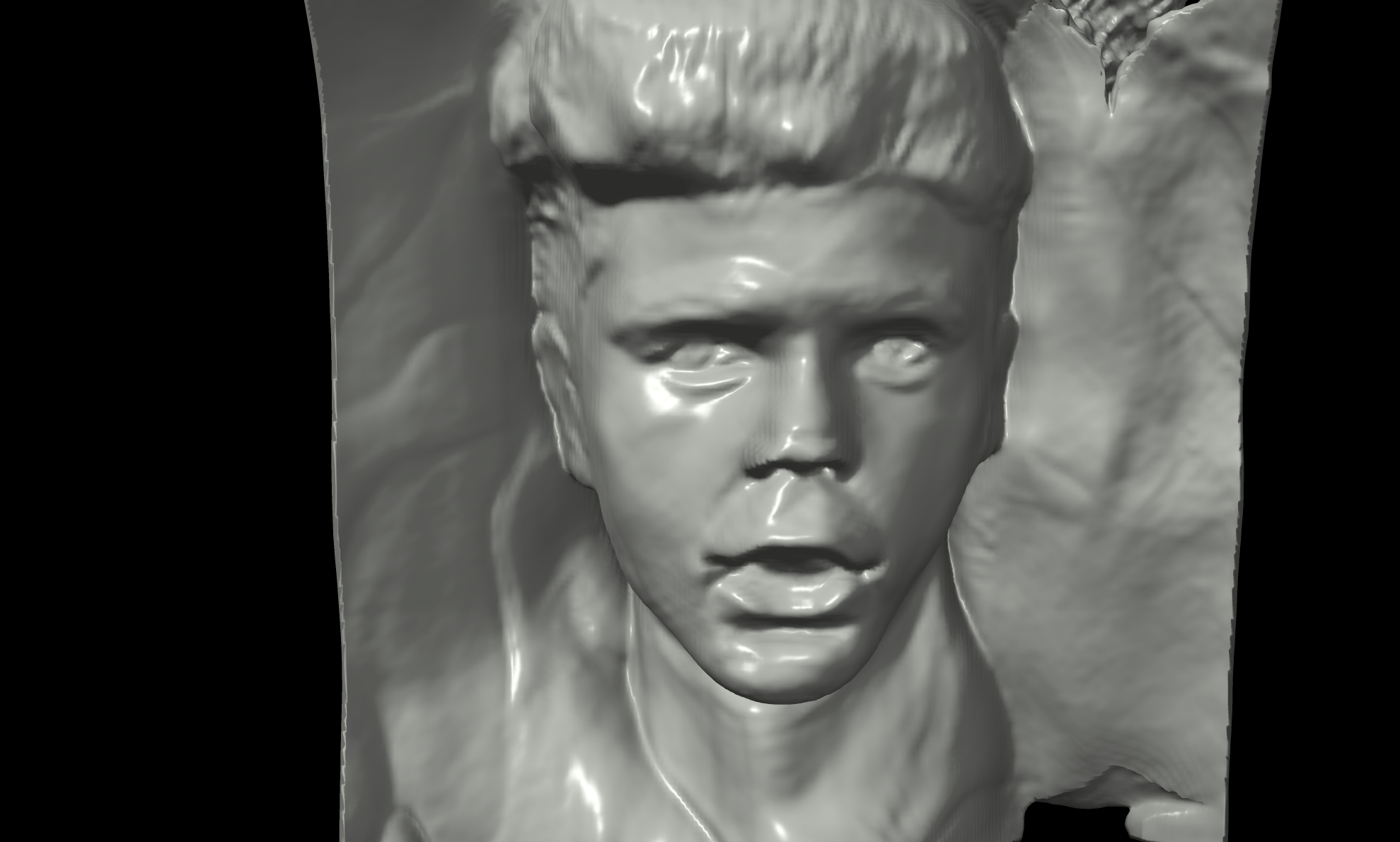}\hfill &
\includegraphics[trim=540 40 260 0,clip,width=\linewidth, height=\linewidth]{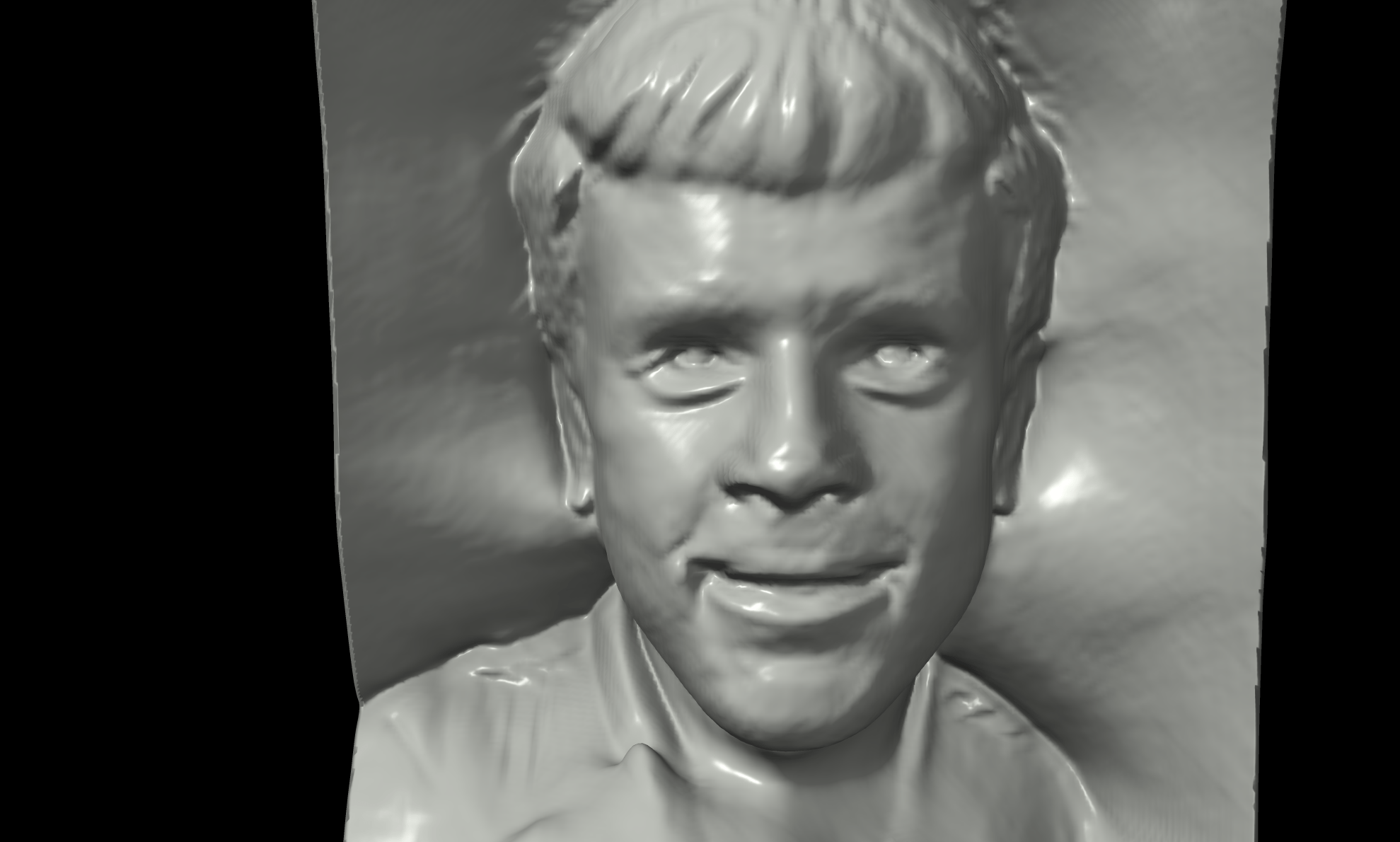}\hfill &
\includegraphics[trim=540 40 260 0,clip,width=\linewidth, height=\linewidth]{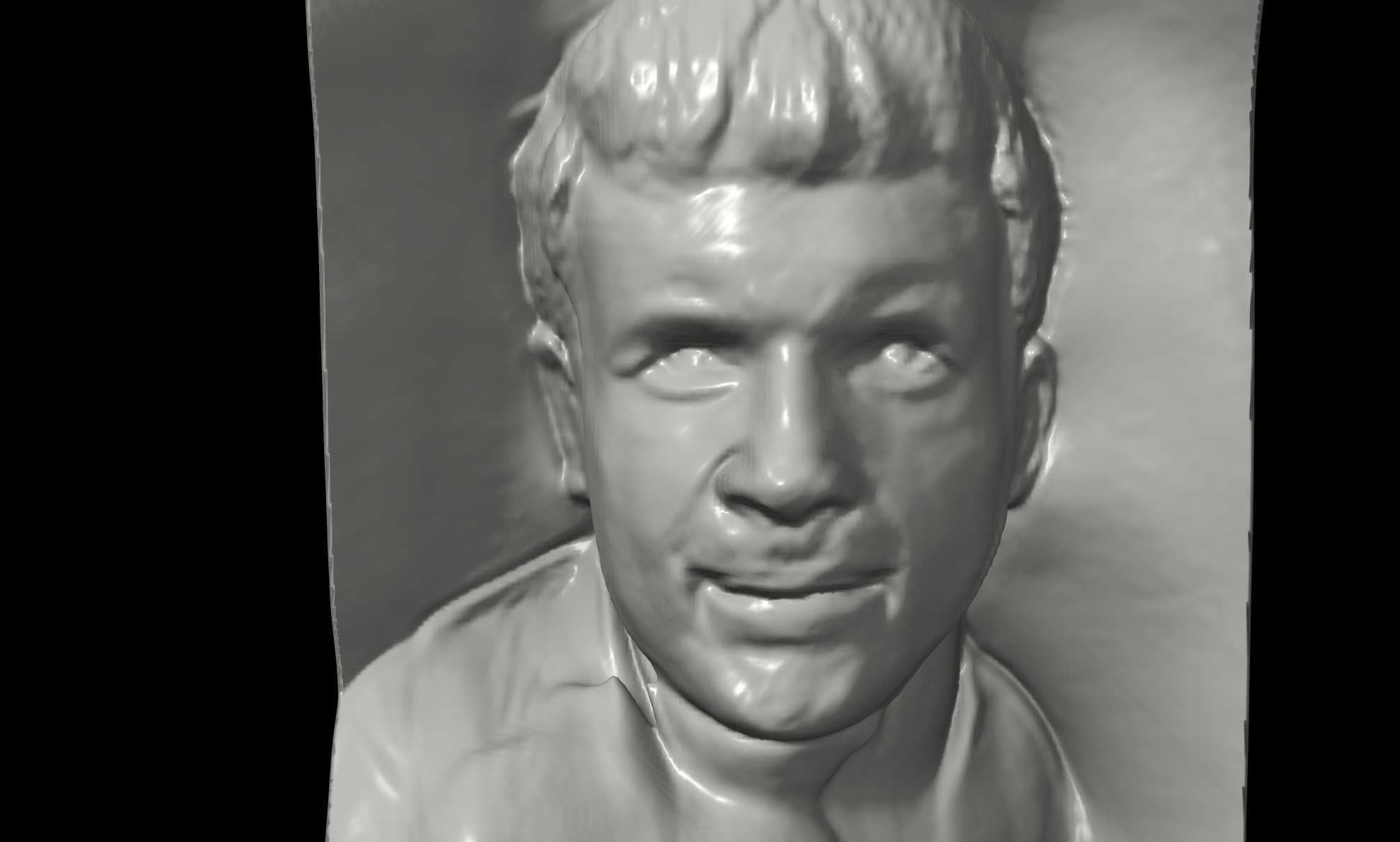}\hfill &
\includegraphics[trim=540 40 260 0,clip,width=\linewidth, height=\linewidth]{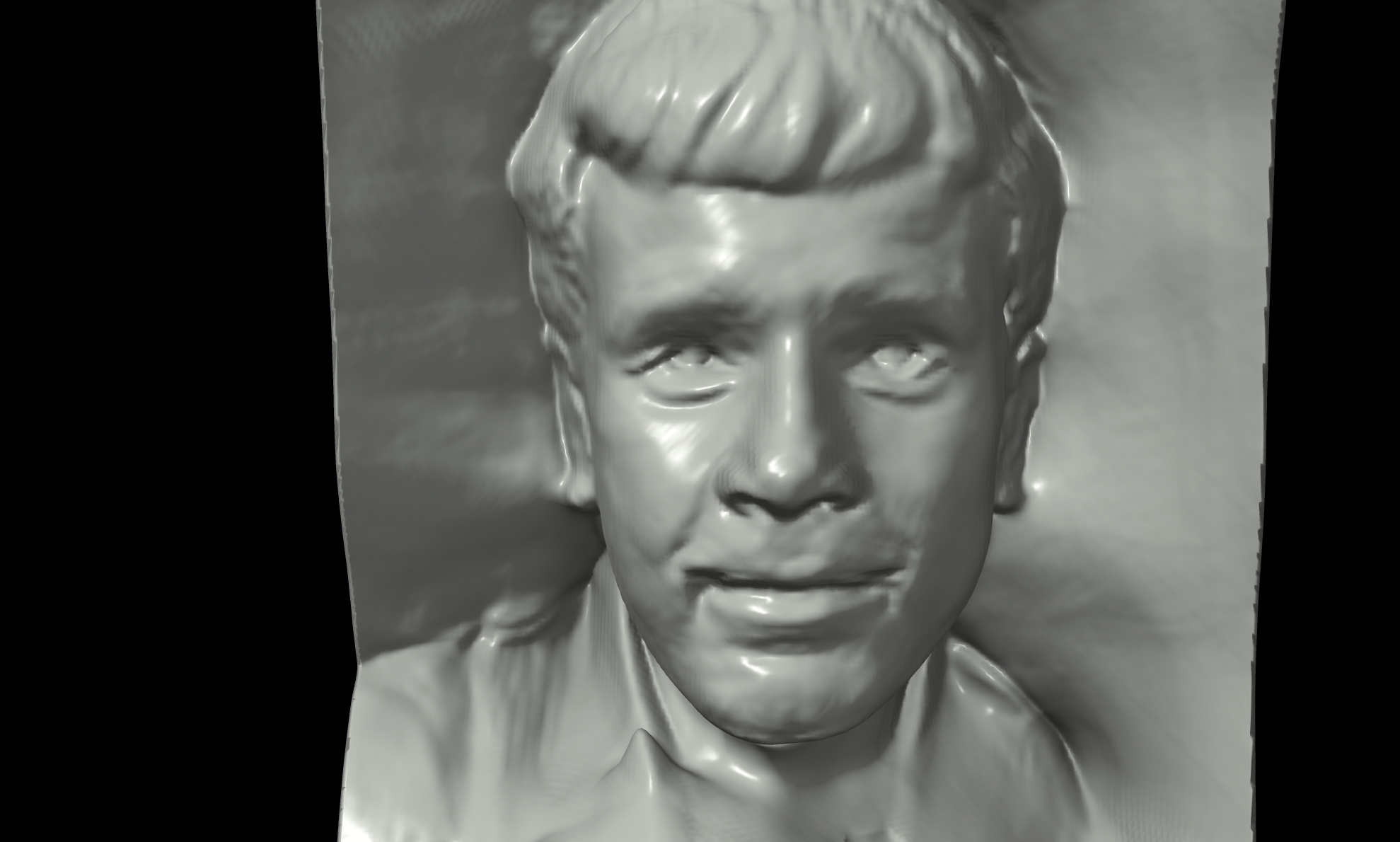}\hfill \\ %trim=285 130 285 80

% Input & \makecell{SG2 \\ GT pose} & \makecell{SG2 \\ Optim. pose} &
% \makecell{SG2 $\mathcal{W}+$ \\ GT pose} & \makecell{SG2 $\mathcal{W}+$ \\ Optim. pose }&
% \makecell{PTI \\ GT pose} & \makecell{PTI \\ Optim. pose} & Ours \\
\end{tabular}
% \vspace{-10pt}
\caption{\textbf{2D and 3D Reconstruction of out-of-domain samples.} We compare both the image reconstruction and 3D reconstruction capabilities of each method, where the 3D shapes are iso-surfaces extracted from the density field using marching cubes. Methods labeled with $\dagger$ use ground-truth camera pose.}
\label{fig:reconstruction_comparison}
\end{figure*}
\paragraph{Datasets.} % Paragraph Split following hyeperinverter
We conduct the experiments on two 3D object types, \textit{human faces} and \textit{cat faces}, as they are the two most popular tasks in GAN inversion.  % Hyperinverter
For all experiments, we employ the pre-trained EG3D~\cite{Chan2022} generator. For \textit{human faces}, we use the weights pre-trained on the cropped FFHQ dataset~\cite{karras2019style}, and we evaluate our method with the CelebA-HQ validation dataset~\cite{karras2017progressive, liu2015faceattributes}. We also use the pre-trained weights on the AFHQ dataset~\cite{choi2020starganv2} for \textit{cat faces} and evaluate on the AnimalFace10 dataset~\cite{liu2019few}.
%to show our method is not limited to facial edits.  % Pivotal Tuning

\paragraph{Baselines.}
%At the time of submission, both of the two existing methods for finding the latent vector of a 3D-aware generator, \cite{daras2021solving} which requires ground-truth pose during inference, and \cite{lin20223d} which leverages 3D morphable models for facial manipulation, 
Since the current works~\cite{daras2021solving, lin20223d} do not provide public source code for reproduction and comparison of their work,
we mainly compare our methods with the popular 2D GAN inversion methods: The direct optimization scheme proposed by \cite{Karras2019stylegan2} to invert real images to $\mathcal{W}$ denoted as SG2, a similar method but extended to $\mathcal{W}+$ space~\cite{Abdal_2019_ICCV} denoted as SG2 $\mathcal{W}+$, and the PTI method from \cite{roich2021pivotal}. We adopt these methods to work with the pose-requiring 3D-aware GANs, either by providing the ground-truth camera pose during optimization or using the same gradient descent optimization method for the camera pose.
%While we do provide comparisons using our method with the results provided in the respective papers in the supplementary material,

% \paragraph{Implementation Details.}
% To implement the latent code encoder $\mathcal{E}$, we follow both the implementation and training strategy of $e4e$~\cite{tov2021designing}, which is a well-proven encoder architecture to map the input image into the distribution of $\mathcal{W}+$. We manipulate the output dimension of the encoder into $R^{1\times512}$ to fit in our method. For the camera pose estimator $\mathcal{P}$, we manipulate the simple $resnet34$~\cite{he2016deep} encoder to find theta and phi angles which are further calculated into the extrinsic matrix. In the case of dataset which has an additional roll angle component in the rotation such as \textit{cat faces} we choose the 6D rotation representation proposed by~\cite{Zhou_2019_CVPR}.
% Although the target images are cropped and refined by~\cite{deng2019accurate}, there exists an additional camera translation that euler angles cannot thoroughly define. Thus, we set an additional coordinate variance on the camera position as a learnable parameter. See supplementary for details and qualitative evaluation of additional translation. 
\begin{table}[!t]
\centering
\small
\resizebox{\linewidth}{!}{
\begin{tabular}{c|cccc|c}
\toprule
Method & MSE$\downarrow$ & LPIPS$\downarrow$ & MS-SSIM$\uparrow$ & ID Sim.$\uparrow$ & FID $\downarrow$ \\
\midrule
SG2 & 0.0277 & 0.3109 & 0.5889 & 0.0957 & 36.0291\\
SG2 $\mathcal{W}+$  & 0.0163 & 0.2398 & 0.6833 & 0.2906 & 32.3971\\
PTI & 0.0036 & 0.0789 & 0.8221 & 0.6671 & 32.7366\\
% \cite{lin20223d} \\
SG2$^\dagger$  & 0.0232 & 0.2898 & 0.6151 & 0.1125 & 34.7612\\
SG2$^\dagger$ $\mathcal{W}+$ & 0.0117 & 0.2029 & 0.7349 & 0.3972& 31.1732 \\
PTI$^\dagger$  & \textbf{0.0033} & \textbf{0.0722} & \textbf{0.8309} & \underline{0.6737}& \textbf{28.5911}\\
Ours &  \underline{0.0035} & \underline{0.0777} & \underline{0.8280} & \textbf{0.7013} & \underline{30.1192}\\
% PTI & StyleGAN2 \\
\bottomrule
\end{tabular}
}
\vspace{-5pt}
\caption{\textbf{Qualitative reconstruction results} measured over the CelebA-HQ test set using various similarity metrics. We also measure the quality of random views of the reconstructed face. The \textbf{best} and \underline{runner-up} values are marked bold and underlined, respectively. Methods labeled with $\dagger$ use ground-truth camera pose.}
\label{tab:reconstruction}
\end{table} 
% \begin{table}[!t]
% \centering
% \small
% \begin{tabular}{cc cccc}
% Method & Architecture & MSE & LPIPS & MS-SSIM & ID Similarity
% PTI (GT Pose) & EG3D \\
% PTI (Random Pose) & EG3D \\
% \cite{lin20223d} & EG3D \\
% Ours & EG3D \\
% PTI & StyleGAN2 \\

% \end{tabular}
% \caption{Reconstruction comparison of the same pose.}
% \label{tab:reconstriuction}
% \end{table}
\begin{figure}[!t]
\centering
\newcolumntype{M}[1]{>{\centering\arraybackslash}m{#1}}
\setlength{\tabcolsep}{1pt}
\renewcommand{\arraystretch}{0.5}
\begin{tabular}{M{0.22\linewidth} M{0.05\linewidth} M{0.22\linewidth} M{0.22\linewidth} M{0.22\linewidth} }
\multirow{6}{*}{
\makecell{
% \vspace{-1pt} \\ 
\\[16pt]
\includegraphics[width=\linewidth]{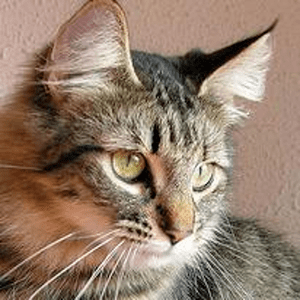} \\
Input }
}
&&
\includegraphics[width=\linewidth]{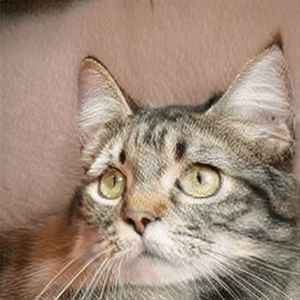}\hfill &
\includegraphics[width=\linewidth]{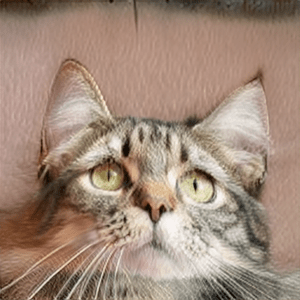}\hfill &
\includegraphics[width=\linewidth]{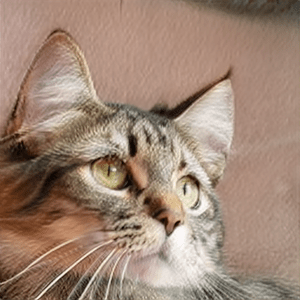}\hfill \\
&&
\includegraphics[width=\linewidth]{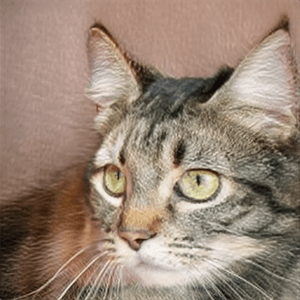}\hfill &
\includegraphics[width=\linewidth]{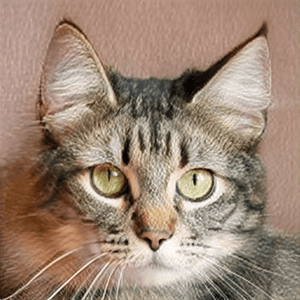}\hfill &
\includegraphics[width=\linewidth]{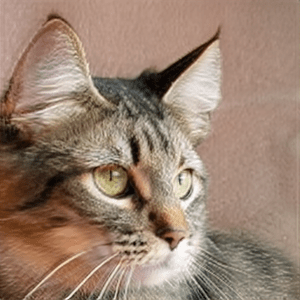}\hfill \\
&&
\includegraphics[width=\linewidth]{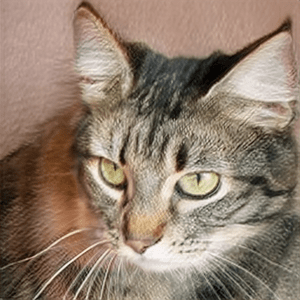}\hfill &
\includegraphics[width=\linewidth]{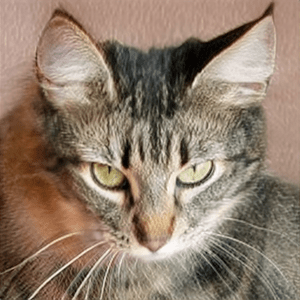}\hfill &
\includegraphics[width=\linewidth]{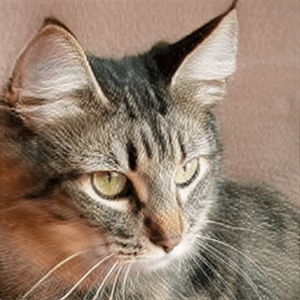}\hfill \\

%trim=300 50 300 50,clip,
\end{tabular}
\vspace{-8pt}

\caption{\textbf{Reconstruction and novel view synthesis on AnimalFace10 dataset.} Our inversion method is not limited to the facial domain and can be applied to other datasets. }
\label{fig:reconstruction_cat}
\end{figure}
%It extensively validates that our method is applicable to generators trained by other datasets.
\begin{figure*}[!t]
\centering
\newcolumntype{M}[1]{>{\centering\arraybackslash}m{#1}}
\setlength{\tabcolsep}{1pt}
\renewcommand{\arraystretch}{0.5}
\begin{tabular}{M{0.015\linewidth}M{0.11\linewidth}M{0.11\linewidth} @{\hskip 0.005\linewidth}|@{\hskip 0.005\linewidth} M{0.11\linewidth}M{0.11\linewidth}M{0.11\linewidth} @{\hskip 0.005\linewidth}|@{\hskip 0.005\linewidth} M{0.11\linewidth}M{0.11\linewidth}M{0.11\linewidth}}

& \multicolumn{1}{c}{Input} 
& \multicolumn{1}{c}{Ours} 
&\multicolumn{1}{c}{SG2} 
&\multicolumn{1}{c}{SG2 $\mathcal{W}+$}
&\multicolumn{1}{c}{PTI} 
&\multicolumn{1}{c}{SG2$^\dagger$}
&\multicolumn{1}{c}{SG2$^\dagger$ $\mathcal{W}+$} & \multicolumn{1}{c}{PTI$^\dagger$ } \\

\rotatebox[origin=c]{90}{\hspace{5pt} $-$Age} &
\includegraphics[width=\linewidth]{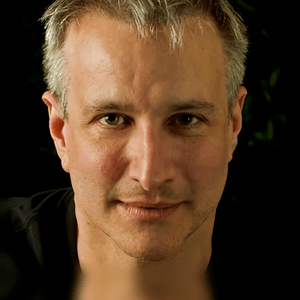}\hfill &
\includegraphics[width=\linewidth]{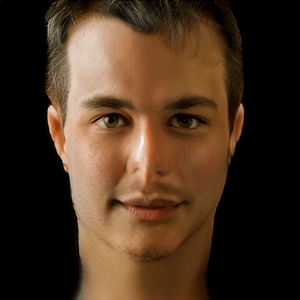}\hfill &
\includegraphics[width=\linewidth]{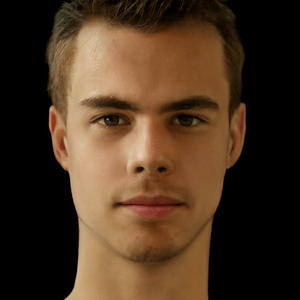}\hfill &
\includegraphics[width=\linewidth]{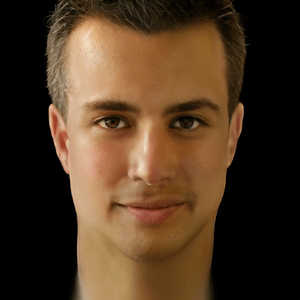}\hfill &
\includegraphics[width=\linewidth]{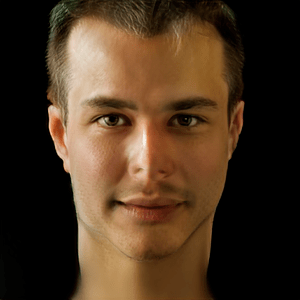}\hfill &
\includegraphics[width=\linewidth]{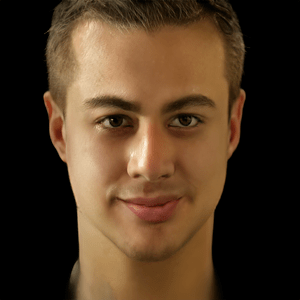}\hfill &
\includegraphics[width=\linewidth]{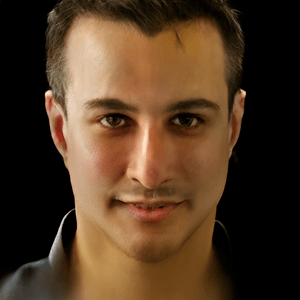}\hfill &
\includegraphics[width=\linewidth]{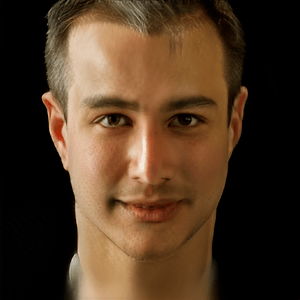}\hfill \\
& &
\includegraphics[trim=365 120 365 45,clip,width=\linewidth, height=\linewidth]{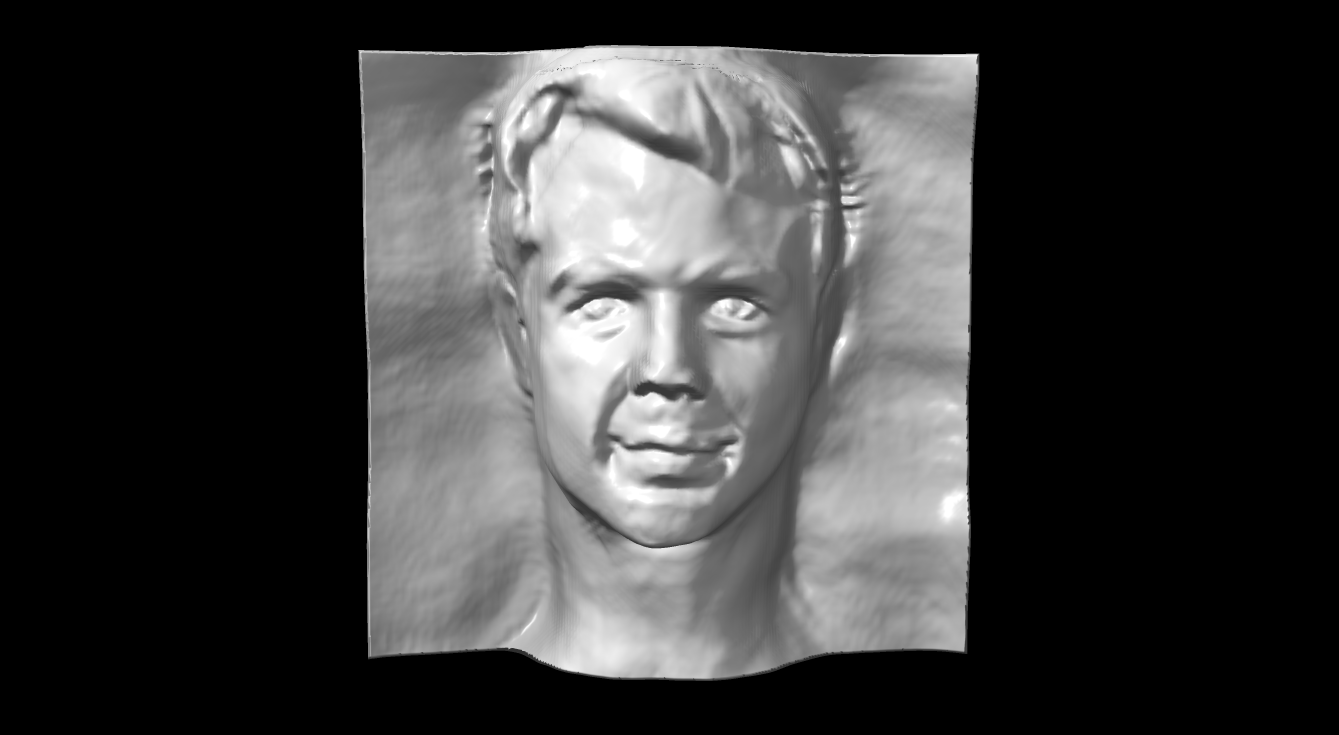}\hfill &
\includegraphics[trim=365 110 365 45,clip,width=\linewidth, height=\linewidth]{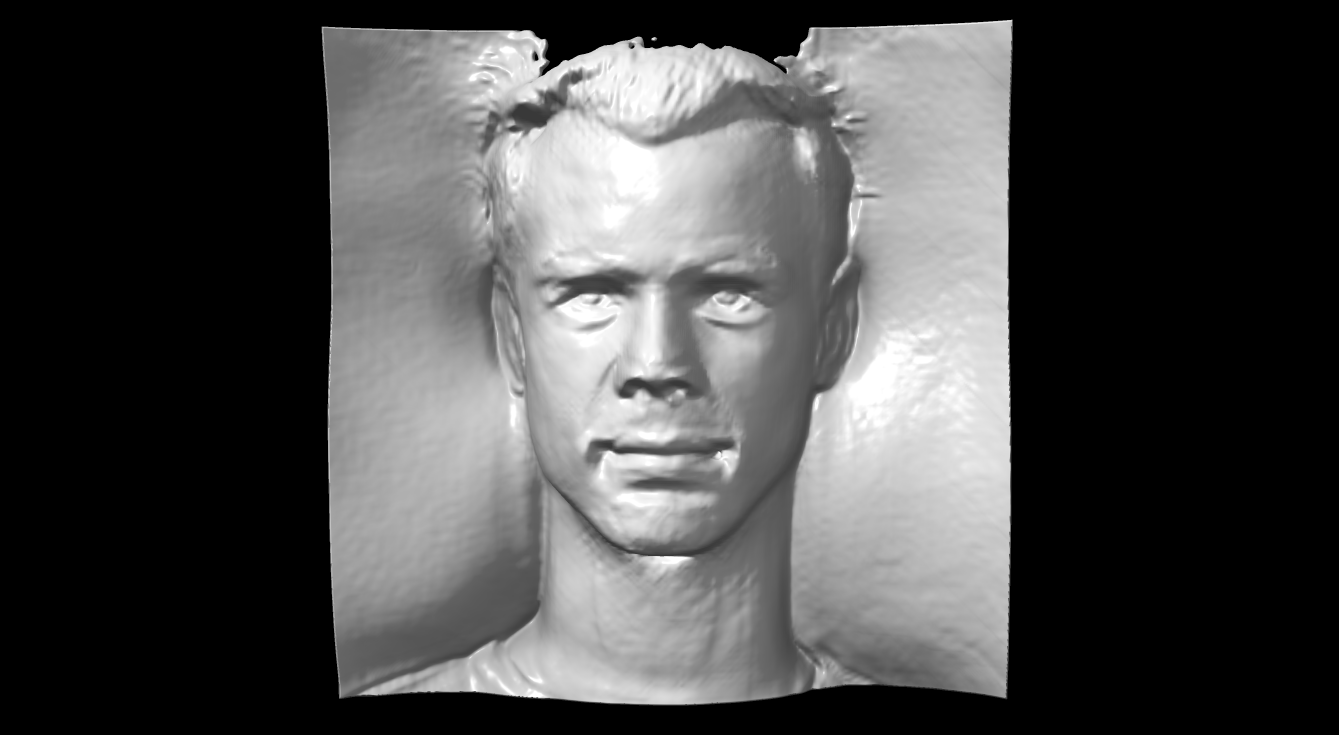}\hfill &
\includegraphics[trim=365 110 365 45,clip,width=\linewidth, height=\linewidth]{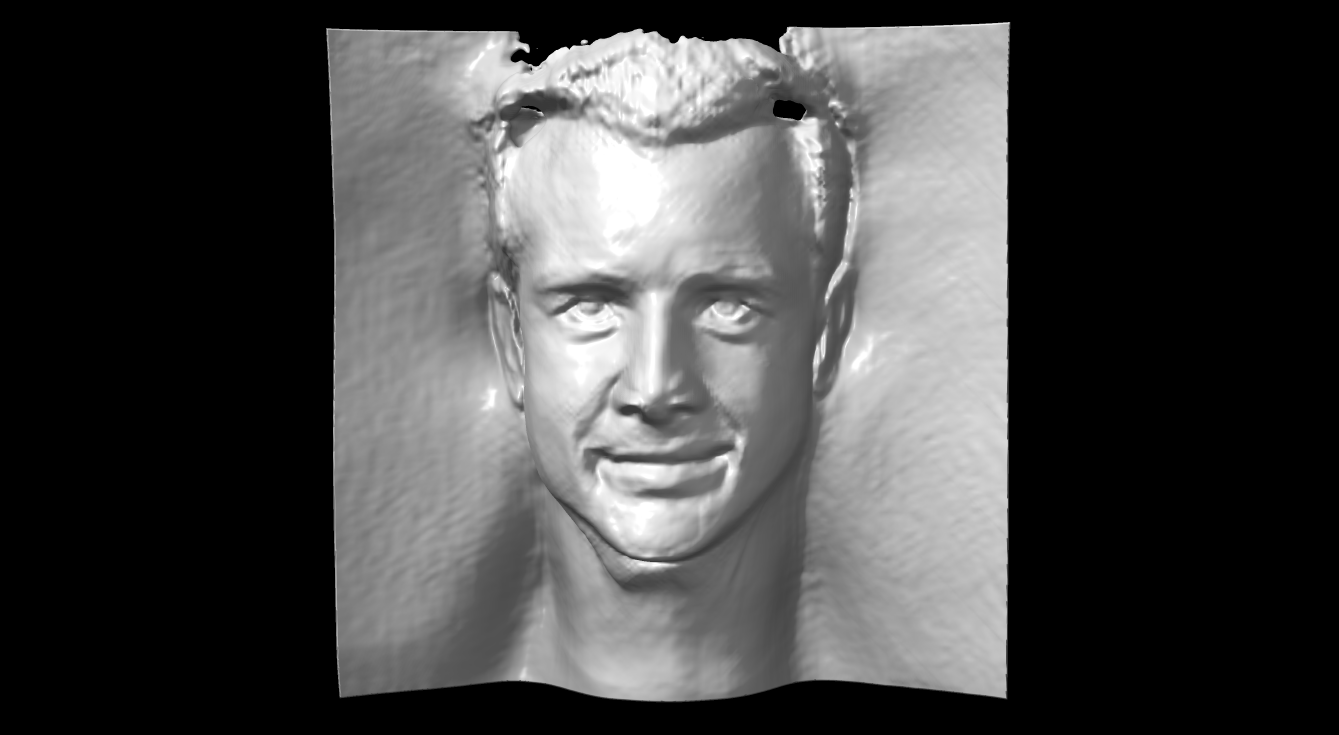}\hfill &
\includegraphics[trim=365 110 365 45,clip,width=\linewidth, height=\linewidth]{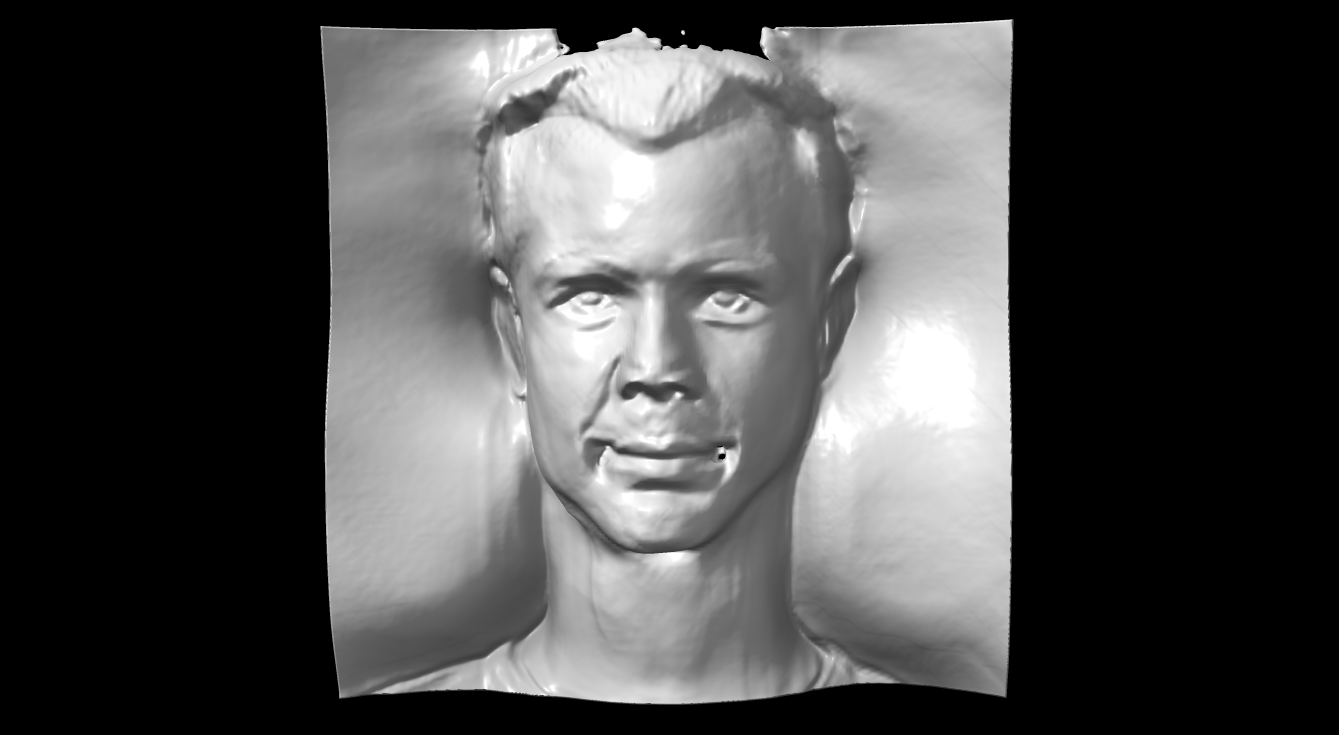}\hfill &
\includegraphics[trim=365 110 365 45,clip,width=\linewidth, height=\linewidth]{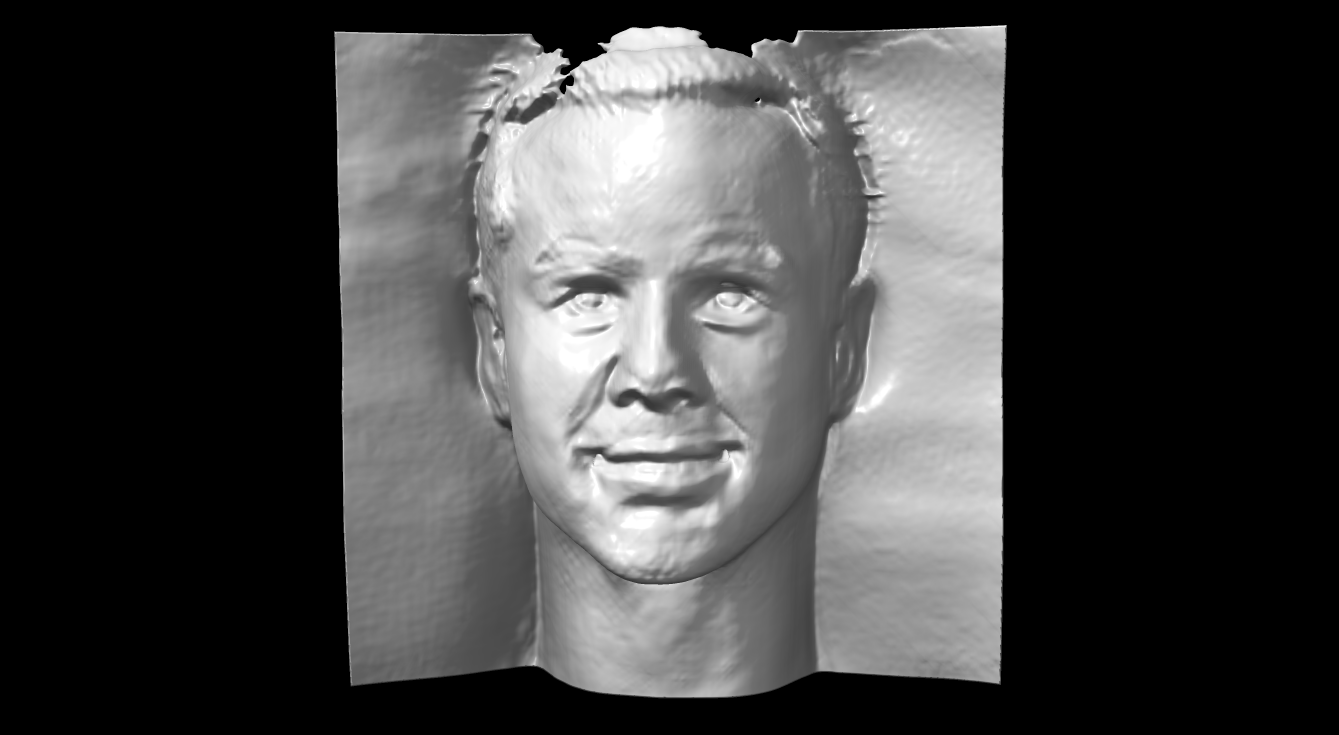}\hfill &
\includegraphics[trim=365 110 365 45,clip,width=\linewidth, height=\linewidth]{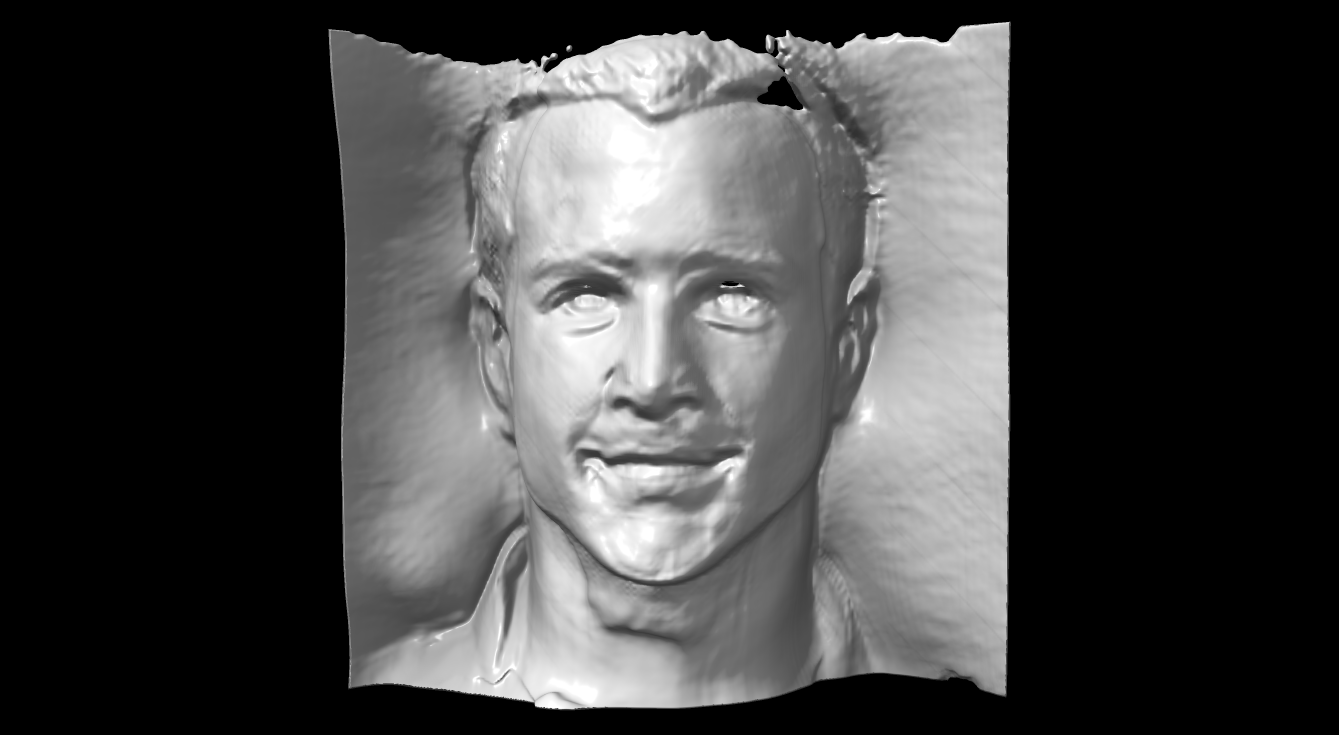}\hfill &
\includegraphics[trim=365 110 365 45,clip,width=\linewidth, height=\linewidth]{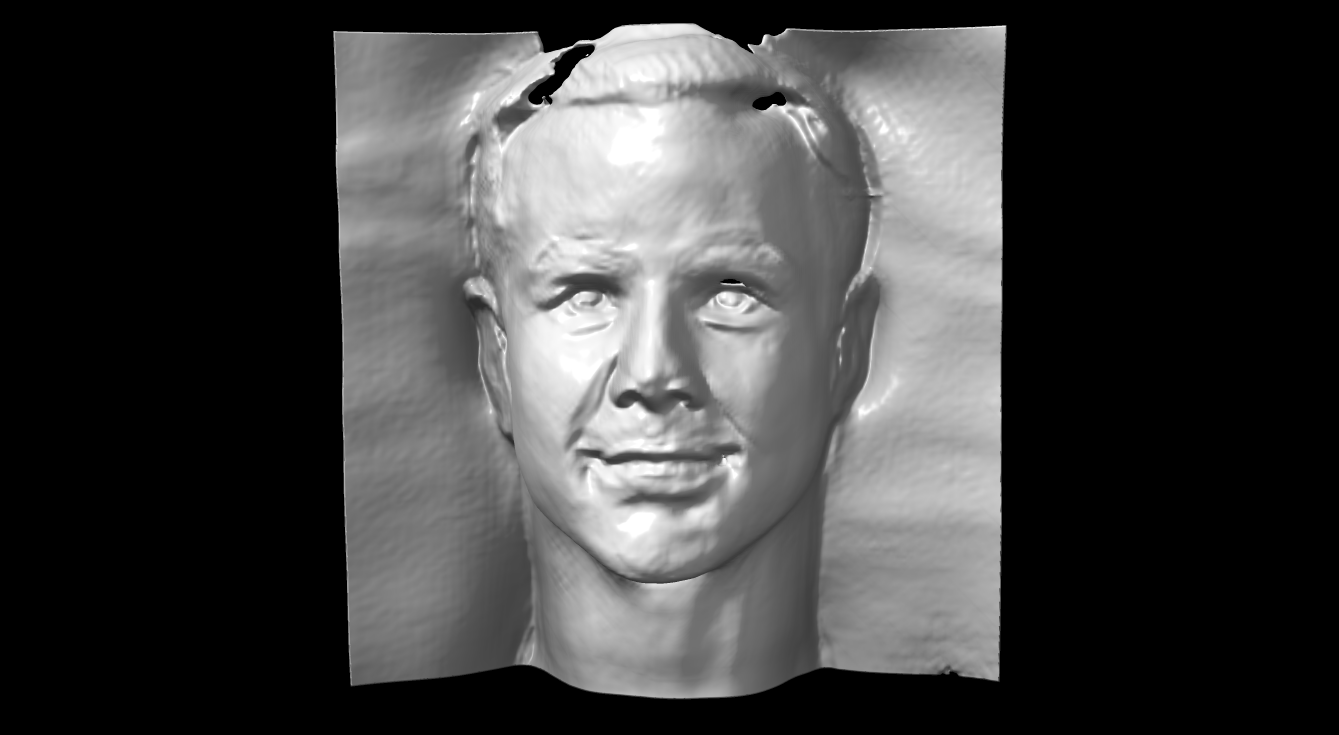}\hfill \\

\rotatebox[origin=c]{90}{\hspace{5pt} $+$Smile} &
\includegraphics[width=\linewidth]{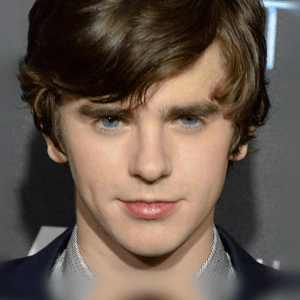}\hfill &
\includegraphics[width=\linewidth]{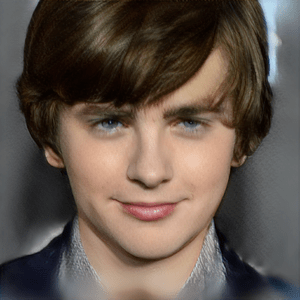}\hfill &
\includegraphics[width=\linewidth]{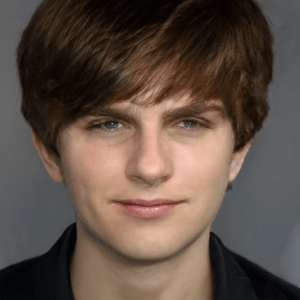}\hfill &
\includegraphics[width=\linewidth]{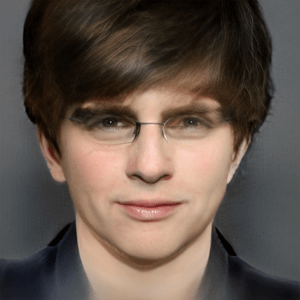}\hfill &
\includegraphics[width=\linewidth]{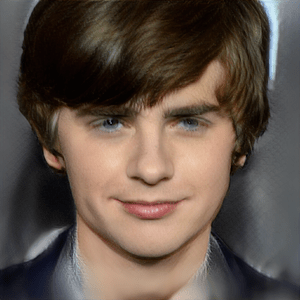}\hfill &
\includegraphics[width=\linewidth]{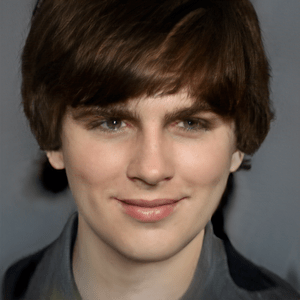}\hfill &
\includegraphics[width=\linewidth]{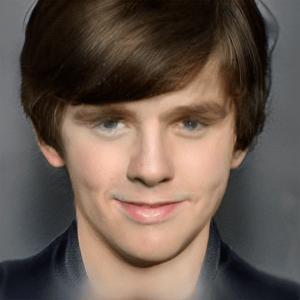}\hfill &
\includegraphics[width=\linewidth]{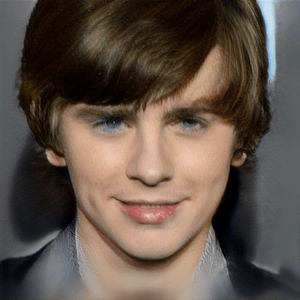}\hfill \\
&&
\includegraphics[trim=365 110 365 40,clip,width=\linewidth, height=\linewidth]{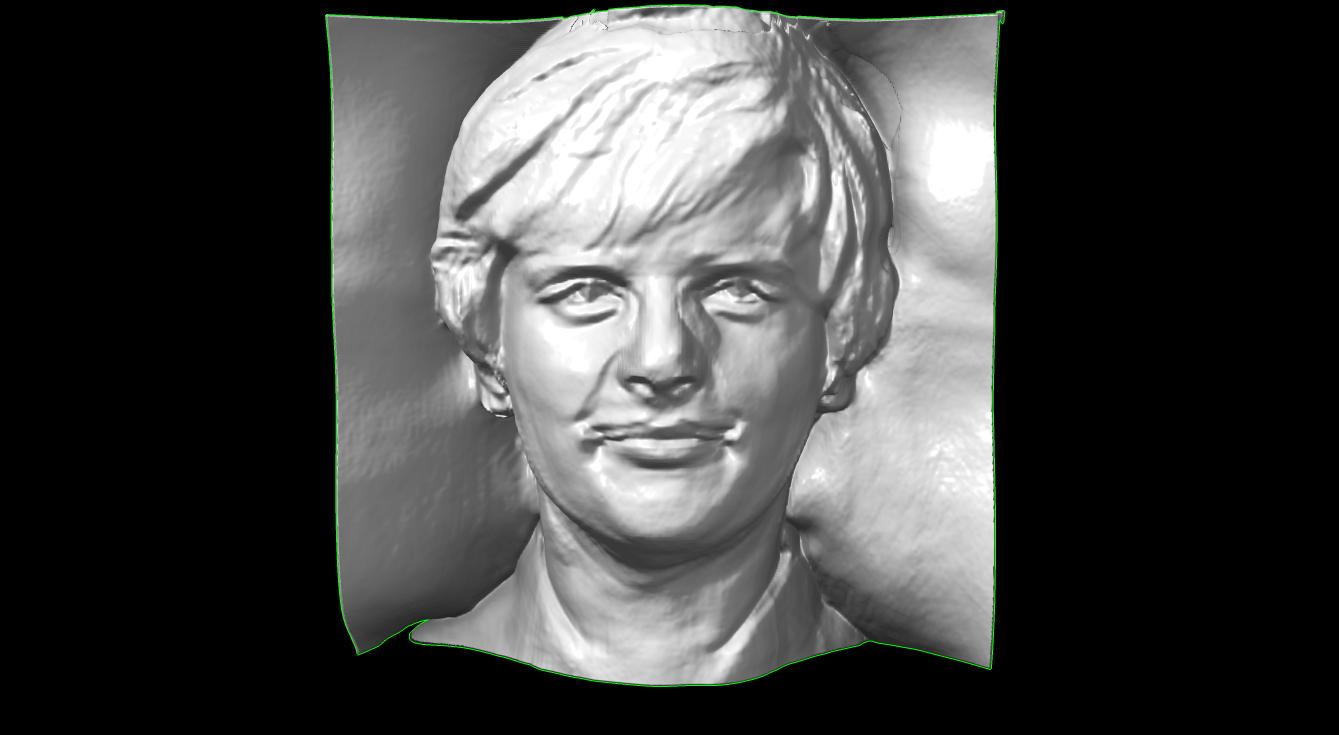}\hfill &
\includegraphics[trim=365 110 365 40,clip,width=\linewidth, height=\linewidth]{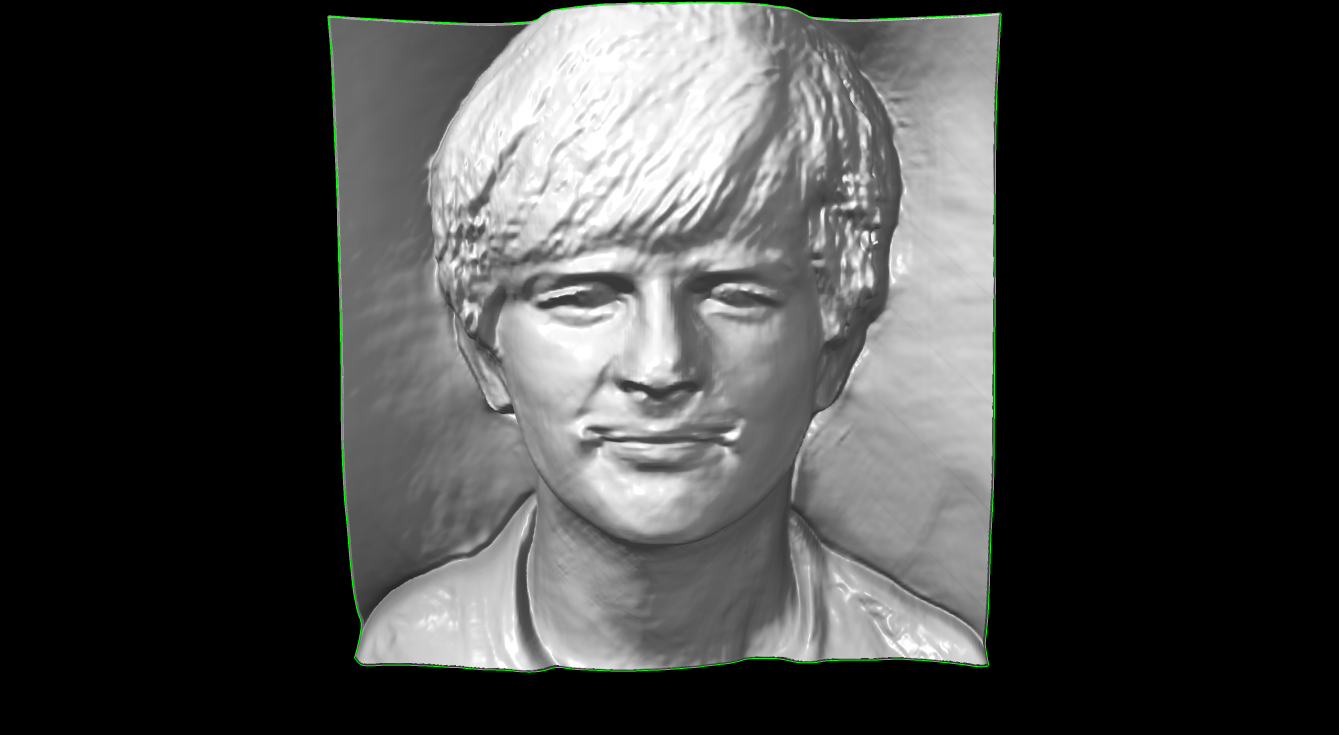}\hfill &
\includegraphics[trim=365 110 365 40,clip,width=\linewidth, height=\linewidth]{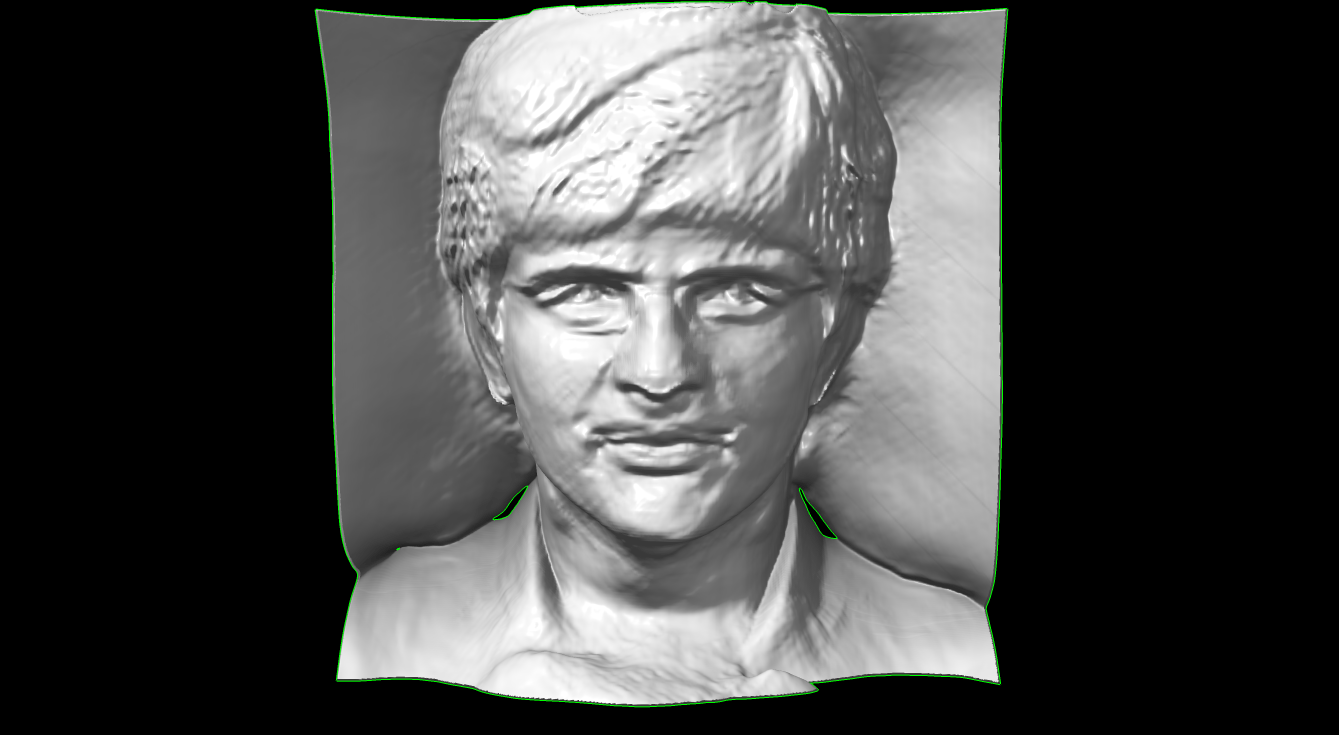}\hfill &
\includegraphics[trim=365 110 365 40,clip,width=\linewidth, height=\linewidth]{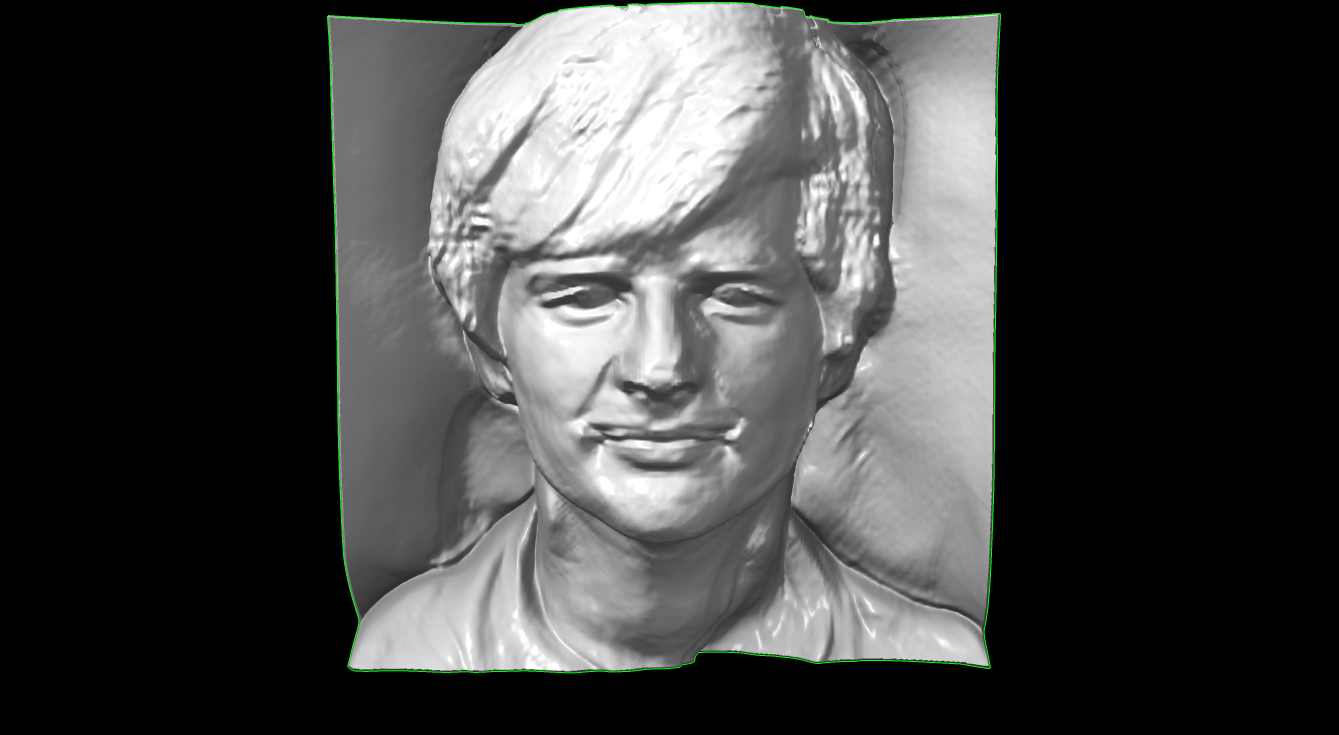}\hfill &
\includegraphics[trim=365 110 365 40,clip,width=\linewidth, height=\linewidth]{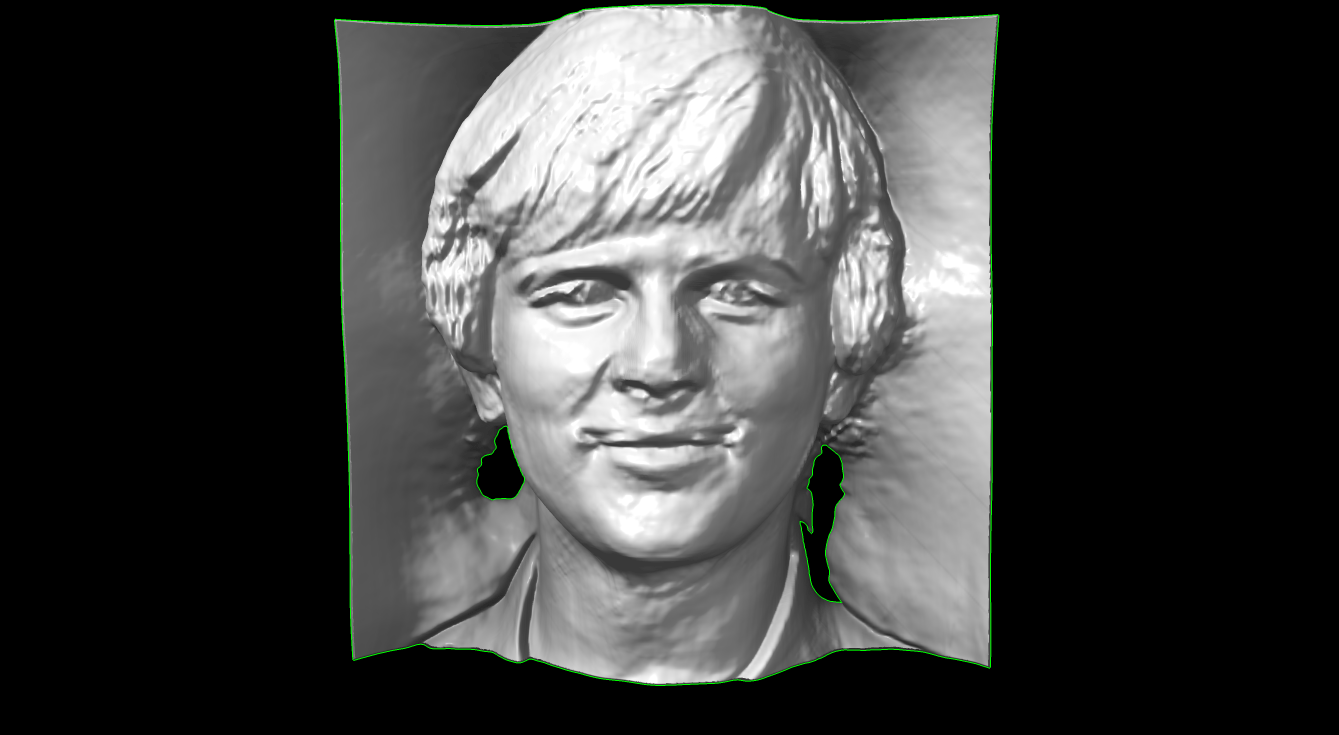}\hfill &
\includegraphics[trim=365 110 365 40,clip,width=\linewidth, height=\linewidth]{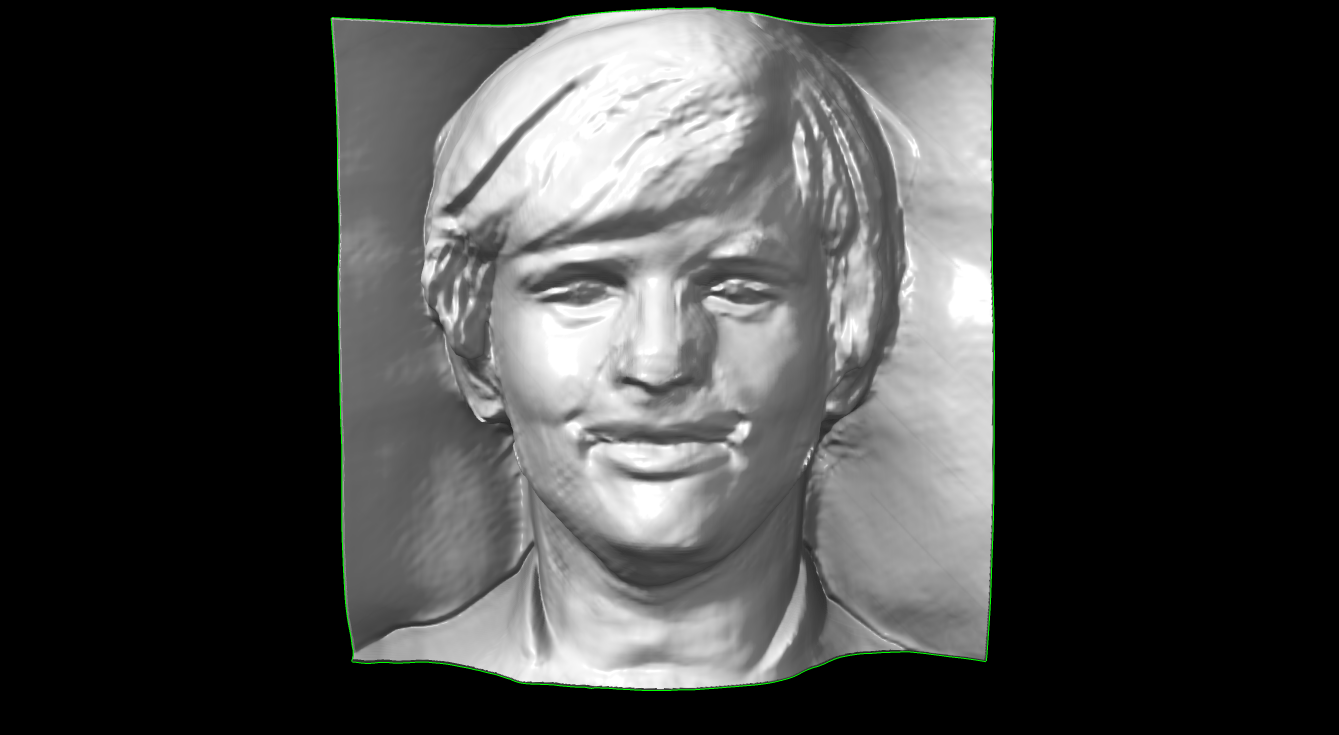}\hfill &
\includegraphics[trim=365 110 365 40,clip,width=\linewidth, height=\linewidth]{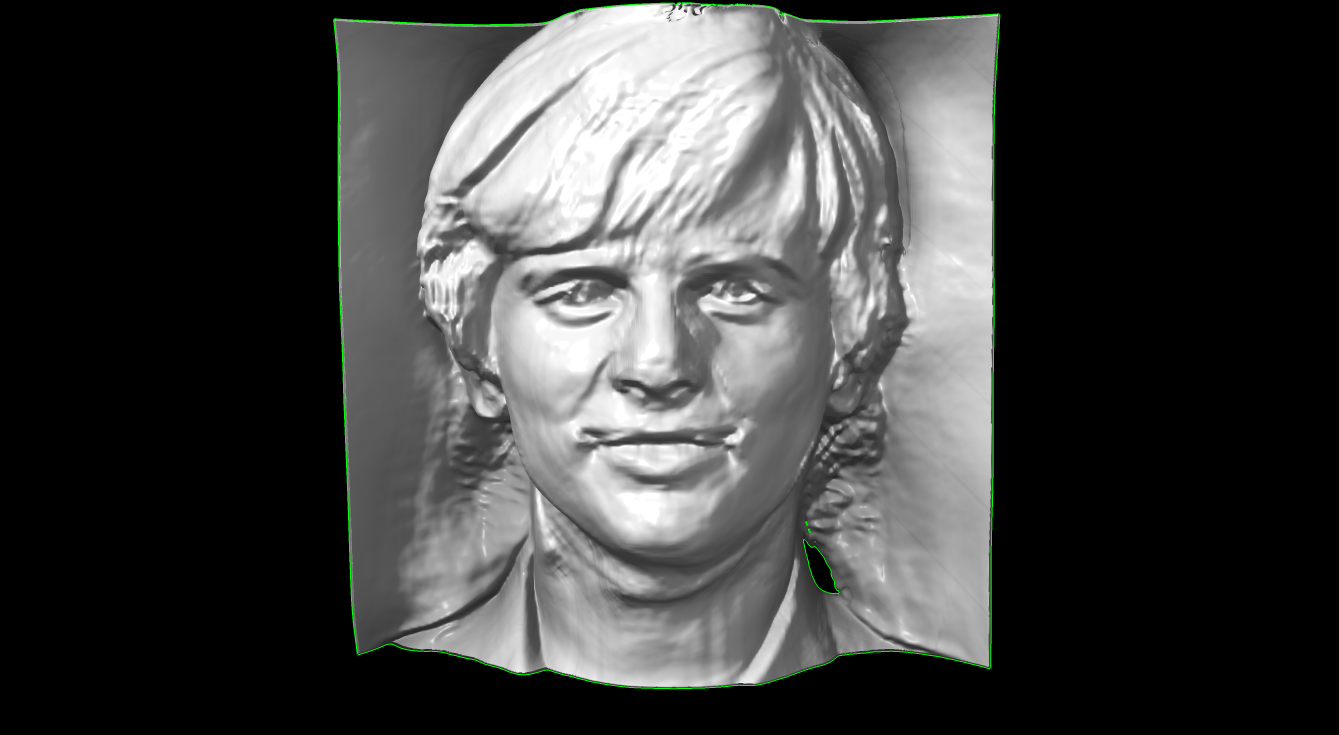}\hfill \\
\end{tabular}
\caption{\textbf{Editing Quality Comparison} We perform various edits~\cite{harkonen2020ganspace} over latent codes and camera pose acquired by each method. Benefiting from the capabilities of 3D-aware GANs, we also compare edited 3D shape generated from the edited latent codes. Our method achieves both realistic and accurate manipulation and is also more capable of preserving the identity and geometry of the original input. Methods labeled with $\dagger$ use ground-truth camera pose.}
\label{fig:editing_comparsion}
\end{figure*}

\subsection{Reconstruction}
\paragraph{Quantitative Evaluation.}
For quantitative evaluation, we reconstruct 2,000 validation images of CelebA-HQ and utilize the same standard metrics used in 2D GAN inversion literature: pixelwise L2 distance using MSE, perceptual similarity metric using LPIPS~\cite{Zhang2018CVPR} and structural similarity metric using MS-SSIM~\cite{wang2003multiscale}. % PTI
In addition, for facial reconstruction, we follow recent 2D GAN inversion works~\cite{dinh2022hyperinverter, alaluf2022hyperstyle} and measure identity similarity using a pre-trained facial recognition network of CurricularFace~\cite{huang2020curricularface}.
Furthermore, we also need to measure the 3D reconstruction ability of our method, as the clear advantage of 3D GAN inversion over 2D GAN inversion is that the former allows for novel view synthesis given a single input image. 
In other words, the latent code acquired by a successful inversion process should be able to produce realistic and plausible images at random views.
In order to measure image quality, we calculate the Frechet Inception Distance (FID)~\cite{heusel2017FID} between the original images and 2,000 additionally generated images from randomly sampled viewpoints.

The results are shown in \tabref{tab:reconstruction}. 
% While comparing our method with 2D GAN inversion methods directly on 3D GANs may not be a fair comparison, comparison with stochastic camera pose optimization %gradient-based optimization on image viewpoint 
% shows our geometry-aware pose optimization method is more reliable than the hit-or-miss optimization method, 
% and comparison with methods that uses GT camera pose proves our method successfully disentangles the geometry information of the given image. 
As can be seen, compared to the 2D GAN inversion methods that use the same gradient-based optimization for camera pose, using the depth-based warping method better guides the camera viewpoint to the desired angle, showing higher reconstruction metrics. 
Furthermore, while methods designed for high expressiveness such as SG2 $\mathcal{W}+$ and PTI achieve comparable pixel-wise reconstruction abilities, our method has the upper hand when it comes to 3D reconstruction, producing better quality images for novel views of the same face. 
Even compared to inversion methods using ground-truth camera pose, our method achieves competitive results without external data and reliably predicts the camera pose, showing similar reconstruction scores for each metric.

\paragraph{Qualitative Evaluation.}
We first visualize the reconstruction results in \figref{fig:reconstruction_comparison_mesh}, which illustrates reconstruction and novel views rendered with our method and the baseline. Notably, our method achieves comparable results to methods using ground-truth pose, while showing much better reconstruction compared to gradient-based camera estimation.
Not only do we provide qualitative comparison of the visual quality of inverted images, but we also show reconstructed 3D shape of the given image as a mesh using the Marching Cubes algorithm~\cite{lorensen1987marching}, as demonstrated in \figref{fig:reconstruction_comparison}. Different from 2D GAN inversion, we also compare the 3D geometry quality of each face by comparing the rendered views of different camera poses. 
Furthermore, we evaluate reconstruction and novel view synthesis on \textit{cat faces}  in \figref{fig:reconstruction_cat}
%It extensively validates that our method is applicable to generators trained by other datasets.

While SG2 $\mathcal{W}+$ and PTI show reasonable reconstruction result at the same viewpoint, when we steer the 3D model to different viewpoints, the renderings are incomplete and show visual artifacts with degraded 3D consistency. In contrast, we can see that by using our method, we can synthesize novel views with comparable quality to methods requiring ground-truth viewpoints.

\subsection{Editing Quality}
We employ GANSpace~\cite{harkonen2020ganspace} method to quantitatively evaluate the manipulation capability of the acquired latent code. 
In \figref{fig:editing_comparsion}, we compare latent-based editing results to the 2D GAN-inversion methods used directly to 3D-aware GANs.
Consistent with 2D GANs, the latent code found in the $\mathcal{W}+$ space produces more accurate reconstruction but fails to perform significant edits, while latent codes in $\mathcal{W}$ space show subpar reconstruction capability. Using pivotal tuning~\cite{roich2021pivotal} preserves the identity while maintaining the manipulation ability of $\mathcal{W}$ space. Reinforcing \cite{roich2021pivotal} with our more reliable pose estimation and regularized geometry, our method best preserves the 3D-aware identity while successfully performing meaningful edits. We also provide quantitative evaluation in the supplementary material.

\subsection{Comparison with 2D GANs}
%\begin{figure}[!t]
%\centering
%\newcolumntype{M}[1]{>{\centering\arraybackslash}m{#1}}
%\setlength{\tabcolsep}{1pt}
%\renewcommand{\arraystretch}{0.5}
%\begin{tabular}{M{0.22\linewidth} M{0.05\linewidth} M{0.22\linewidth} %M{0.22\linewidth} M{0.22\linewidth} }
%\multirow{2}{*}{
%\makecell{
% \vspace{-1pt} \\
%\includegraphics[width=\linewidth]{Fig_vs2D/camera_manipulation/002292.png}}
%}
%& \rotatebox[origin=rl]{90}{\small{2D}} &
%\includegraphics[width=\linewidth]%{Fig_vs2D/camera_manipulation/002292_2D_sample0.png}\hfill &
%\includegraphics[width=\linewidth]%{Fig_vs2D/camera_manipulation/002292_2D_sample2.png}\hfill&
%\includegraphics[width=\linewidth]{Fig_vs2D/camera_manipulation/002292_2D_sample4.png}\hfill \\
%&\rotatebox[origin=rl]{90}{\small{3D}}&
%\includegraphics[width=\linewidth]{Fig_vs2D/camera_manipulation/left.png}\hfill&
%\includegraphics[width=\linewidth]{Fig_vs2D/camera_manipulation/inverse.png}\hfill&
%\includegraphics[width=\linewidth]{Fig_vs2D/camera_manipulation/right.png}\hfill \\

%trim=300 50 300 50,clip,
%\end{tabular}

%%\caption{\textbf{Comparison of pose manipulation of real images inverted with 2D GANs and 3D GANs.} As can be seen, pose manipulation using PTI on StyleGAN2 (first row) only allows for implicit control by the editing magnitude and larger step sizes result in undesired transformations. On the other hand, using our method on EG3D (second row), allows for explicit control and because the acquired latent representation are viewpoint independent, the edits geometrically consistent.}
%\label{fig:pose_manipulation_2d}
%\end{figure}

\begin{figure}[!t]
\centering
\newcolumntype{M}[1]{>{\centering\arraybackslash}m{#1}}
\setlength{\tabcolsep}{1pt}
\renewcommand{\arraystretch}{0.5}
\begin{tabular}{M{0.22\linewidth} M{0.06\linewidth} M{0.22\linewidth} M{0.22\linewidth} M{0.22\linewidth} }
\multirow{2}{*}{
\makecell{
% \vspace{-1pt} \\
\includegraphics[width=\linewidth]{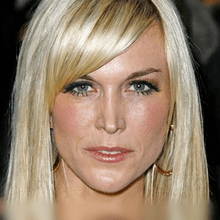}}
}
& \rotatebox[origin=rl]{90}{\small{2D}} &
\includegraphics[width=\linewidth]{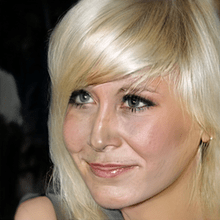}\hfill &
\includegraphics[width=\linewidth]{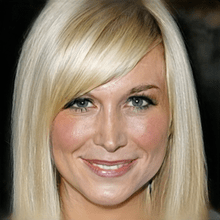}\hfill&
\includegraphics[width=\linewidth]{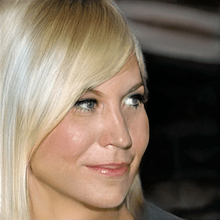}\hfill \\
&\rotatebox[origin=rl]{90}{\small{3D}}&
\includegraphics[width=\linewidth]{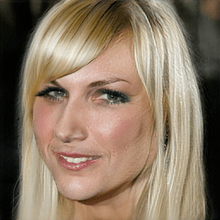}\hfill&
\includegraphics[width=\linewidth]{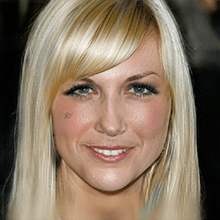}\hfill&
\includegraphics[width=\linewidth]{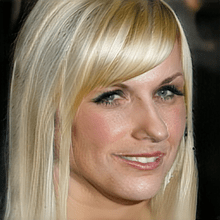}\hfill \\
\end{tabular}
\vspace{-5pt}
\caption{\textbf{Simultaneous attribute editing and viewpoint shift comparison of 2D and 3D GANs.} We compare editing results of applying attribute editing (smile) and viewpoint interpolation at the same time on the latent code acquired by PTI~\cite{roich2021pivotal} on StyleGAN2~\cite{Karras2019stylegan2} and the latent code acquired by our method on EG3D~\cite{Chan2022}}
\label{fig:simultaneous editing}
\end{figure}
\begin{table}[!t]
\centering
\newcolumntype{M}[1]{>{\centering\arraybackslash}m{#1}}
\resizebox{\linewidth}{!}{
\begin{tabular}{M{0.1\linewidth}M{0.1\linewidth}M{0.1\linewidth}|M{0.2\linewidth}M{0.2\linewidth}M{0.2\linewidth}M{0.2\linewidth}}
\toprule
$\mathcal{E} \& \mathcal{P}$ & $\mathcal{L}_\mathrm{warp}$ & $\mathcal{L}_{DR}$ & LPIPS$\downarrow$ & MS-SSIM$\uparrow$ &  ID Sim.$\uparrow$  &  FID$\downarrow$  \\
\midrule                                              % MSE LPIPS MSSSIM IDSIM FID
\xmark & \xmark & \xmark &0.0789 &0.8221 &0.6671 &32.7366   \\
\cmark & \xmark & \xmark &0.0780 &0.8259 &0.6823 &31.1518\\     
\xmark & \cmark & \xmark &0.0783 &0.8248 &0.6750 &31.3179 \\      
\cmark & \cmark & \xmark &0.0771 &0.8295 &0.7005 &30.6272 \\ 
\cmark & \cmark & \cmark &0.0777 &0.8280 &0.7013 &30.1198\\
\bottomrule
\end{tabular}}
\caption{\textbf{Reconstruction metric comparison of various combinations of proposed methods.} We evaluate our proposed optimization scheme by experimenting with certain combinations of proposed methods. We mark (\cmark) when the given method is employed and (\xmark) when it is not.}
\label{tab:full_ablation}
\end{table}

We demonstrate the effectiveness and significance of 3D-aware GAN inversion by comparing the viewpoint manipulation using the explicit camera pose control for EG3D and latent space. 
Even though \cite{kwak2022injecting} points out that 3D GANs lack the ability to manipulate semantic attributes, recent advancements of NeRF-based generative architectures have achieved a similar level of expressiveness and editability to 2D GANs.
As recent 3D-aware GANs such as EG3D can generate high-fidelity images with controllable semantic attributes while maintaining a reliable geometry, image editing using 3D GANs offers consistent multi-view results which are more useful. 

In \figref{fig:simultaneous editing}, we compare the simultaneous manipulation abilities of 2D and 3D-aware GANs. While pose manipulation of 2D GANs only allows for implicit control by the editing magnitude, 3D-aware GANs enable explicit control of the viewpoint. Also, the edited images in 2D GANs are not view consistent and large editing factors result in undesired transformations of the identity and editing quality. In contrast, pose manipulation of 3D GANs is always multi-view consistent, thus produces consistent pose interpolation even for an edited scene.

\subsection{Ablation Study}
In \tabref{tab:full_ablation} we show the importance of the various components of our approach by turning them on and off in turn. To further investigate the efficacy of our methodology, we further show some visual results for more intuitive comparisons. 

\begin{figure}[!t]
\centering
\small
\newcolumntype{M}[1]{>{\centering\arraybackslash}m{#1}}
\setlength{\tabcolsep}{0.2pt}
\renewcommand{\arraystretch}{0.1}
\begin{tabular}{M{0.15\linewidth}  M{0.15\linewidth} M{0.15\linewidth} M{0.15\linewidth} M{0.15\linewidth} M{0.05\linewidth} M{0.15\linewidth}}
&& 0 iter & 150 iter & 300 iter && tuned \\ \\ 

\multirow{3}{*}{
\makecell{
 \\[7pt]
\includegraphics[width=\linewidth]{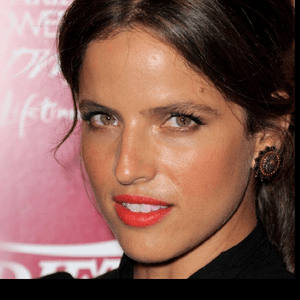} \\
Input}
} &
$\mathcal{E}$ & 
\includegraphics[width=\linewidth]{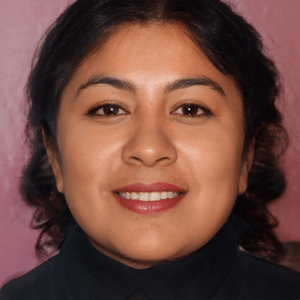}\hfill &
\includegraphics[width=\linewidth]{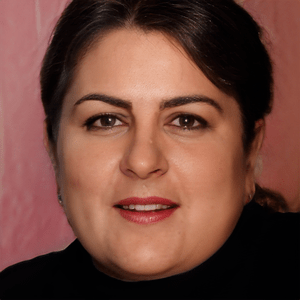}\hfill &
\includegraphics[width=\linewidth]{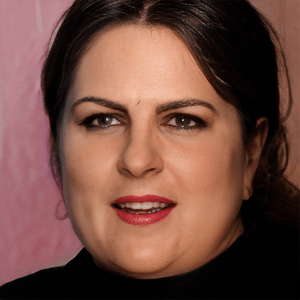}\hfill &
... &
\includegraphics[width=\linewidth]{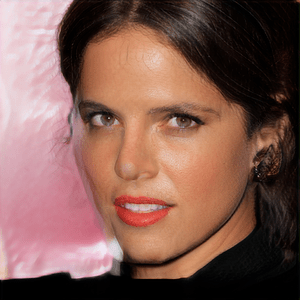}\hfill\\

&$\mathcal{P}$ & 
\includegraphics[width=\linewidth]{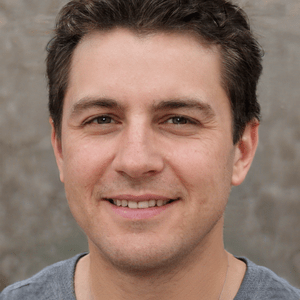}\hfill &
\includegraphics[width=\linewidth]{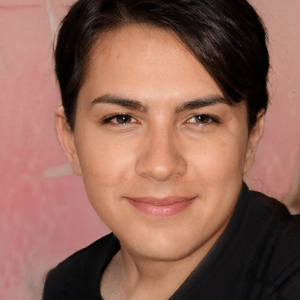}\hfill & 
\includegraphics[width=\linewidth]{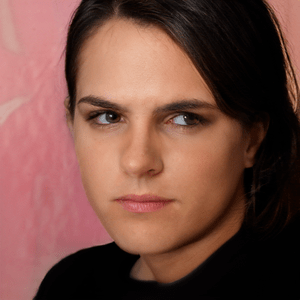}\hfill &
... &
\includegraphics[width=\linewidth]{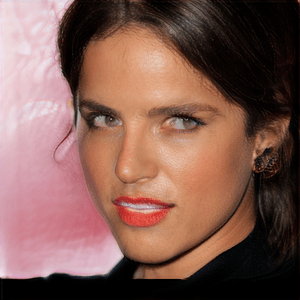}\hfill \\

&$\mathcal{E}$ \& $\mathcal{P}$ & 
\includegraphics[width=\linewidth]{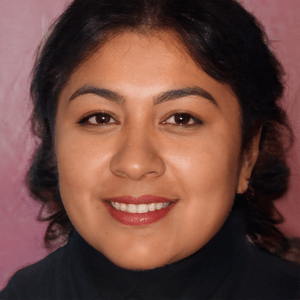}\hfill &
\includegraphics[width=\linewidth]{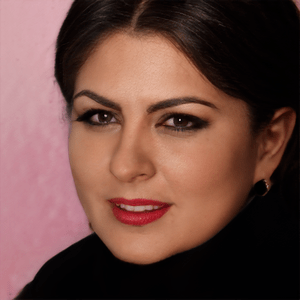}\hfill &
\includegraphics[width=\linewidth]{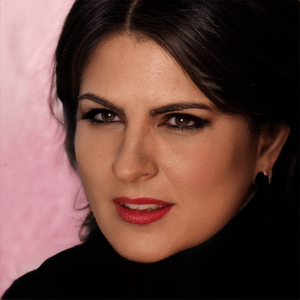}\hfill &
... &
\includegraphics[width=\linewidth]{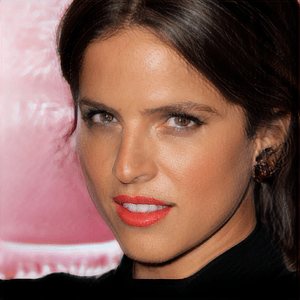}\hfill \\

\end{tabular}

\caption{\textbf{Importance of initialization} We selectively employ the latent encoder $\mathcal{E}$ and pose estimator $\mathcal{P}$ and compare their optimization process.}
\label{fig:inversion_method}
\end{figure}

\paragraph{Importance of initialization.}
We test the effectiveness of our design by comparing our full hybrid method to the single-encoder methods and show the results in \figref{fig:inversion_method} and \tabref{tab:inversion_method}. We show that using a hybrid approach consisting of learned encoder $\mathcal{E}$ and gradient-based optimization is the ideal approach when obtaining the latent code. Similarly, leveraging a pose estimator $\mathcal{P}$ for initialization for the pose refinement also shortens the optimization time.

\begin{table}[!t]
\centering
\resizebox{\linewidth}{!}{
\begin{tabular}{cc|cccccc}
\toprule
$\mathcal{E}$ & $\mathcal{P}$ & MSE$\downarrow$ & LPIPS$\downarrow$ & MS-SSIM$\uparrow$ & ID Sim.$\uparrow$ & $\theta$ & $\phi$\\
\midrule
\xmark & \cmark & 0.0036 & 0.0782 & 0.8263 & 0.6958 & 3.19 & 2.90 \\
\cmark & \xmark & 0.0038 & 0.0790 & 0.8219 & 0.6810 & 5.73 & 5.93 \\
\cmark & \cmark & 0.0035 & 0.0777 & 0.8280 & 0.7013 & 3.16 & 2.70 \\ 
\bottomrule
\end{tabular}
}
\caption{\textbf{Importance of initialization.} We state the importance of utilizing pre-trained latent encoder $\mathcal{E}$ and pose estimator $\mathcal{P}$ as an initialization for optimization, by comparing the optimization results that started from network outputs(\cmark) and those that started from random initialization(\xmark).}
\label{tab:inversion_method}
\end{table}

\begin{figure}
\centering
\newcolumntype{M}[1]{>{\centering\arraybackslash}m{#1}}
\setlength{\tabcolsep}{1pt}
\renewcommand{\arraystretch}{0.5}
\begin{tabular}{M{0.22\linewidth}M{0.22\linewidth}M{0.22\linewidth}M{0.22\linewidth}}
Input & $\lambda_{DR} = 0 $ & $\lambda_{DR} = 1 $ & $\lambda_{DR} = 50 $ \\
\includegraphics[width=\linewidth]{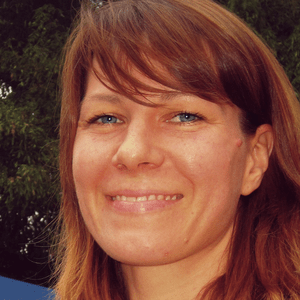}\hfill &
\includegraphics[width=\linewidth]{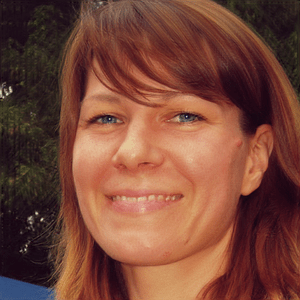}\hfill &
\includegraphics[width=\linewidth]{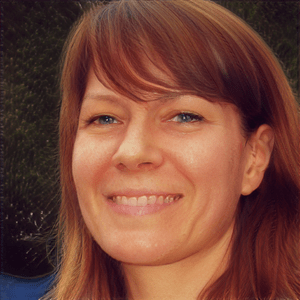}\hfill &
\includegraphics[width=\linewidth]{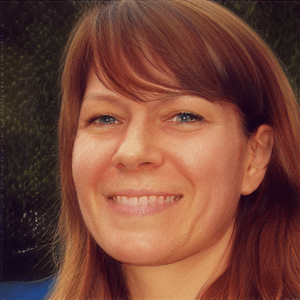}\hfill \\
 & \small{L2: 0.1233} & \small{L2: 0.1255} & \small{L2: 0.1257} \\ 
& \includegraphics[width=\linewidth]{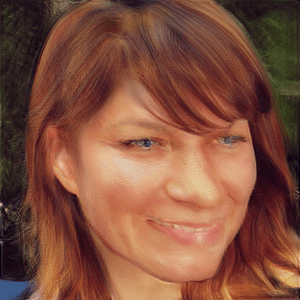}\hfill &
\includegraphics[width=\linewidth]{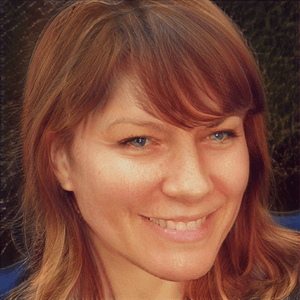}\hfill &
\includegraphics[width=\linewidth]{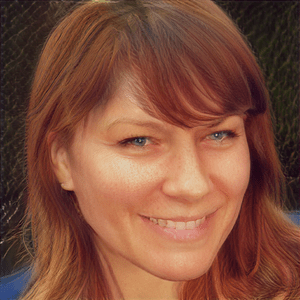}\hfill \\
& \includegraphics[trim=300 50 300 50,clip, width=\linewidth]{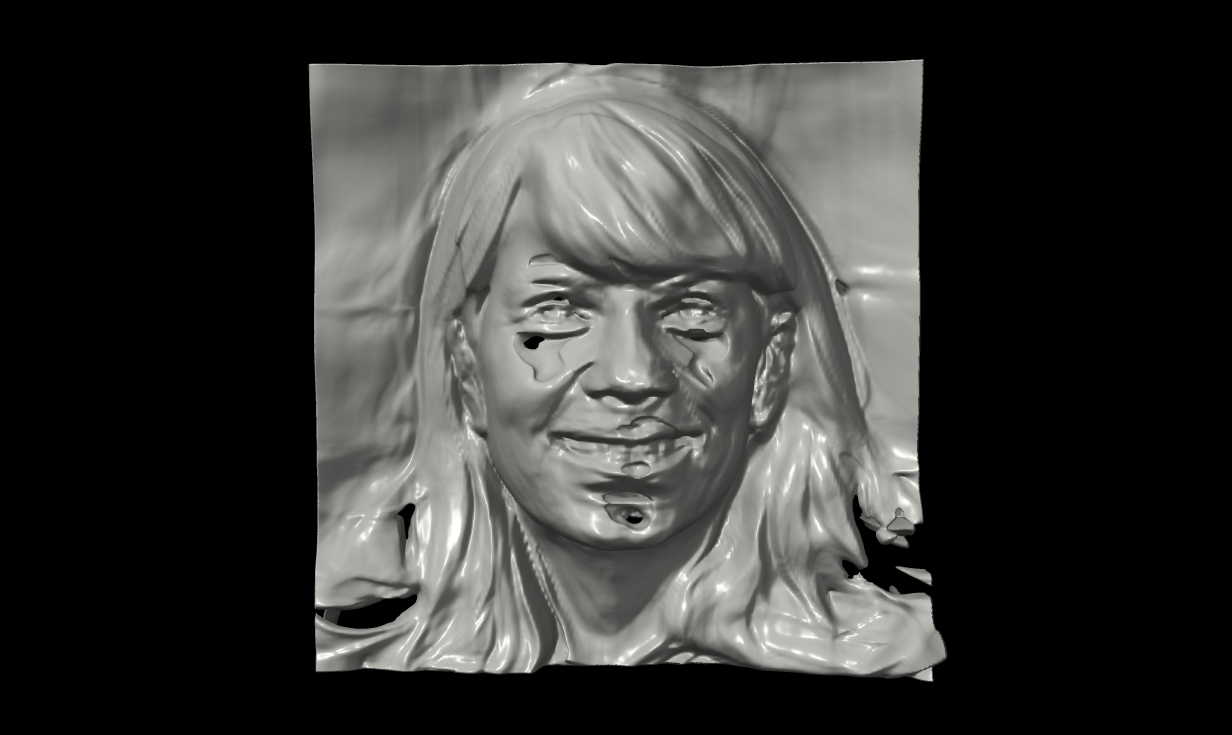}\hfill &
\includegraphics[trim=300 50 300 50,clip,width=\linewidth]{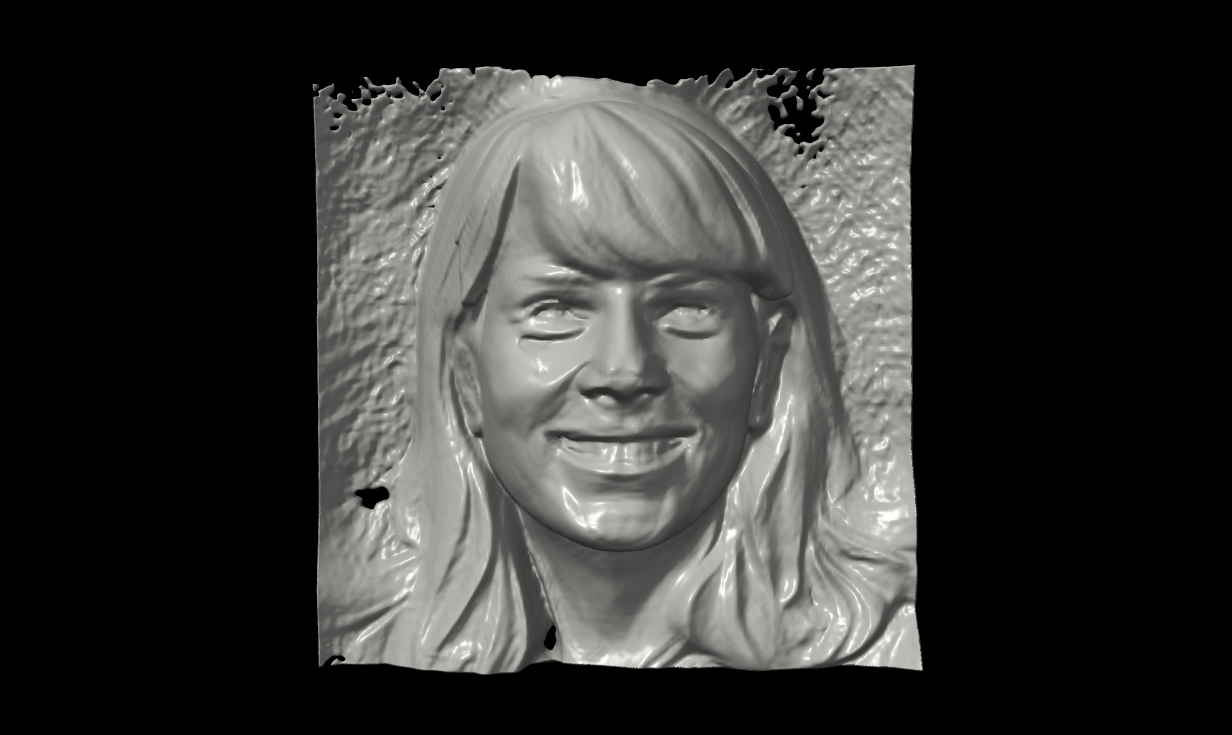}\hfill &
\includegraphics[trim=300 50 300 50,clip,width=\linewidth]{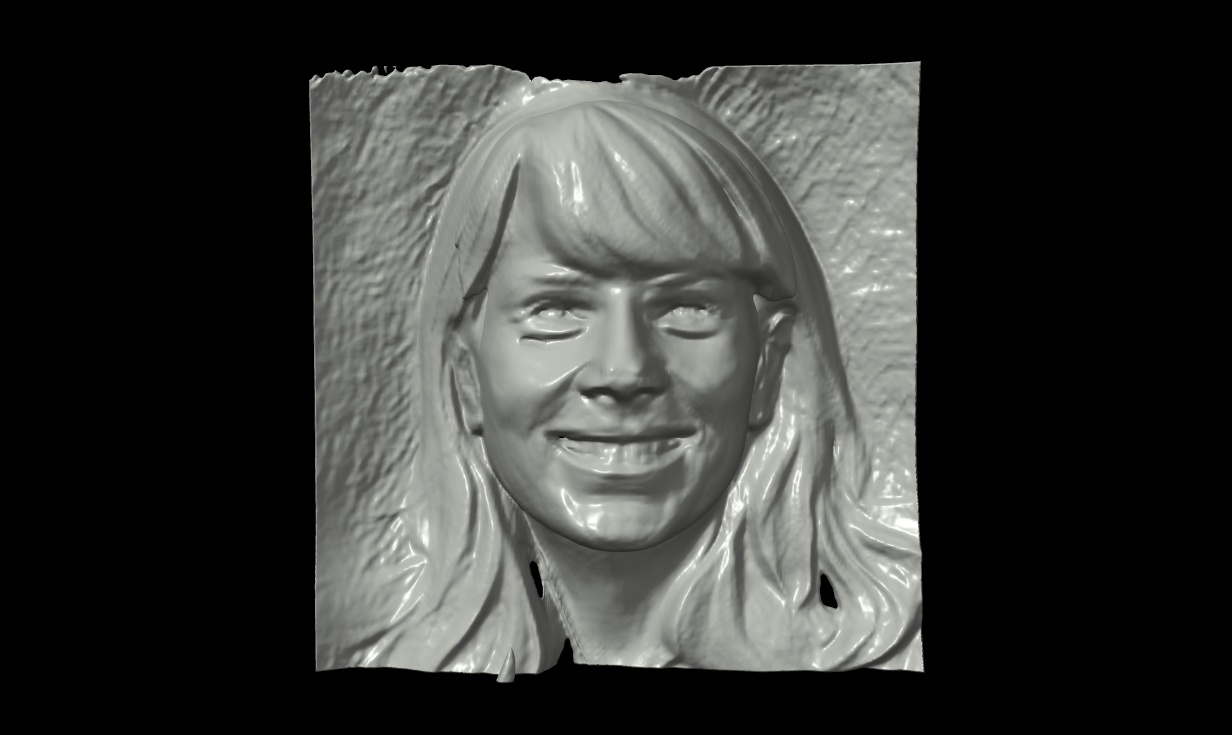}\hfill \\
\end{tabular}
% \vspace{-8pt}
\caption{\textbf{Comparison of different weights of depth smoothness loss during the pivotal tuning stage.} Although excluding depth smoothness shows the best reconstruction result (first row), the reconstructed geometry is distorted (third row) and provides malformed renderings for novel views (second row). Best viewed in zoom.}
\label{fig:depth_smoothness}
\end{figure}

\paragraph{Effectiveness of Geometry Regularization.}
We study the role of depth smoothness regularization in the pivotal tuning step by varying the weight $\lambda_{DR}$. We show the generated geometry and its pixelwise MSE value after fine-tuning the generator in \figref{fig:depth_smoothness}. While solely using the reconstruction loss leads to better quantitative results, novel views still contain floating artifacts and the generated geometry has holes and cracks.  % regnerf
In contrast, including the depth smoothness regularization with its weight as $\lambda_{DR}=1$ enforces solid and smooth surfaces while producing accurate scene geometry. It should be noted that a high weight for the depth smoothness blurs the fine segments of the generated geometry such as hair. 

% \subsection{Applications}

\section{Conclusion}
We present a geometric approach for inferring both the latent representation and camera pose of 3D GANs for a single given image. Our method can be widely applied to the 3D-aware generator, by utilizing hybrid optimization method with additional encoders trained by manually built pseudo datasets. In essence, this pre-training session helps the encoder acquire the representation power and geometry awareness of 3D GANs, thus finding a stable optimization pathway. 
Moreover, we utilize several loss functions and demonstrate significantly improved results in reconstruction and image fidelity both quantitatively and qualitatively. It should be noted that whereas previous methods used 2D GANs for editing, our work suggests the possibility of employing 3D GANs as an editing tool.
%Also, we leverage the depth-based warping loss function to effectively optimize the optimizable camera parameter. reconstruction, image fidelity
We hope that our approach will encourage further research on 3D GAN inversion, which will be further utilized with the single view 3D reconstruction and semantic attribute editings.

%------------------------------------------------------------------------

{\small
\bibliographystyle{ieee_fullname}
\bibliography{egbib}
 }

\newpage
\onecolumn
% \title{Pose-free 3D GAN Inversion \\ - Supplementary Material -} % Temporary 

% \and

% \maketitle
% \thispagestyle{empty}
\begin{center}
	\textbf{\large 3D GAN Inversion with Pose Optimization \\ - Supplementary Material-}
\end{center}
\vspace{5pt}

\appendix
\renewcommand{\thefigure}{\arabic{figure}}
\renewcommand{\theHfigure}{A\arabic{figure}}
\renewcommand{\thetable}{\arabic{table}}
\renewcommand{\theHtable}{A\arabic{table}}
\setcounter{figure}{0}
\setcounter{equation}{0}
\setcounter{table}{0}

\section{Implementation Details}
\subsection{Architectures and Hyperparameters}
To implement the latent code encoder $\mathcal{E}$, we follow both the implementation and training strategy of e4e~\cite{tov2021designing}, which is a well-proven encoder architecture to map the input image into the distribution of $\mathcal{W}+$. We manipulate the output dimension of the encoder into $\mathbb{R}^{1\times512}$ to fit in our method. For the camera pose estimator $\mathcal{P}$, we manipulate the simple resnet34~\cite{he2016deep} encoder to find theta and phi angles which are further calculated into the extrinsic matrix. In the case of dataset which has an additional roll angle component in the rotation such as \textit{cat faces}, we choose the 6D rotation representation proposed by~\cite{Zhou_2019_CVPR}.
Although the target images are cropped and refined by~\cite{deng2019accurate}, there exists an additional camera translation that euler angles cannot thoroughly define. Thus, we set an additional coordinate variance on the camera position as a learnable parameter. See supplementary for details and qualitative evaluation of additional translation. 

\subsection{Pre-training Latent Encoder $\mathcal{E}$ and Pose Estimator $\mathcal{P}$}
In order to train the encoder $\mathcal{E}$, we adopt LPIPS loss~\cite{Zhang2018CVPR} denoted as $\mathcal{L}^{\mathcal{E}}_\mathrm{lpips}$ in order to reconstruct the given image.
% which is defined by:
% \begin{equation}
%     \mathcal{L}^{\mathcal{E}}_\mathrm{lpips} =||x_\mathrm{ps} - \mathcal{G}^\mathrm{c}_\mathrm{3D}(\bar{\mathbf{w}} + \bigtriangleup\mathbf{w},\pi_\mathrm{ps};\theta)||^2_2.
%     \label{eq:encoder_train_lpips}
% \end{equation}
Additionally, we minimize the gap between $\bar{\mathbf{w}}+\bigtriangleup\mathbf{w}$ and the embedding space $\mathcal{W}$ of $\mathcal{G}_\mathrm{3D}$ by employing non-saturating GAN loss~\cite{goodfellow2014generative} and delta-regularization loss:
\begin{equation}
    \mathcal{L}^{\mathcal{E}}_\mathrm{adv}= -\mathbb{E}[\mathrm{log} \mathcal{D}(\mathcal{G}^\mathrm{c}_\mathrm{3D}((\bar{\mathbf{w}} + \bigtriangleup\mathbf{w}),\pi_\mathrm{ps};\theta))],
    \label{eq:encoder_train_adv}
\end{equation}
\begin{equation}
    \mathcal{L}^{\mathcal{E}}_\mathrm{reg} =||\bigtriangleup\mathbf{w}||^2_2.
    \label{eq:encoder_train_deltareg}
\end{equation}
Moreover, in the case of pose estimator $\mathcal{P}$, the predicted output $\hat\pi$ from a given image is directly compared with $\pi_\mathrm{ps}$. Let rotation $\hat{\mathrm{R}}\in\mathbb{R}^{3\times3}$, translation $\hat{t}\in\mathbb{R}^3$, and scale factor $\hat{s}\in\mathbb{R}^1$  denote as a decomposed set of $\pi$. Since the scale factor is given as a constant, we formulate our loss function as:
\begin{equation}
    \mathcal{L}^{\mathcal{P}}_\mathrm{rot} = || \mathrm{R}^{-1}_\mathrm{ps} \cdot \hat{\mathrm{R}} - \mathrm{I}_{3\times3}||_2,
    \label{eq:encoder_train_rot}
\end{equation}
\begin{equation}
    \mathcal{L}^{\mathcal{P}}_\mathrm{trans} =||t_\mathrm{ps} - \hat{t}||_2,
    \label{eq:encoder_train_trans}
\end{equation}
where $\mathrm{R}_\mathrm{ps}$ and $t_\mathrm{ps}$ are decomposition of $\pi_\mathrm{ps}$, and $\mathrm{I}_{3\times3}$ is identity matrix.

In summary, our total loss functions ($\mathcal{L}^{\mathcal{E}}$ for Latent Encoder $\mathcal{E}$ and $\mathcal{L}^{\mathcal{P}}$ for Pose Estimator $\mathcal{E}$) for pre-training the encoder are defined by:
\begin{equation}
 \mathcal{L}^{\mathcal{E}} = \mathcal{L}^{\mathcal{E}}_{\mathrm{lpips}} + \lambda_\mathrm{adv}\mathcal{L}^{\mathcal{E}}_\mathrm{adv} + \lambda_\mathrm{reg}\mathcal{L}^{\mathcal{E}}_\mathrm{reg},
\end{equation}
\begin{equation}
 \mathcal{L}^{\mathcal{P}} = \mathcal{L}^{\mathcal{P}}_{\mathrm{rot}} + \lambda_\mathrm{trans}\mathcal{L}^{\mathcal{P}}_\mathrm{trans}.
\end{equation}

\subsection{Pseudo Dataset}
\begin{figure}[!t]
\centering
\newcolumntype{M}[1]{>{\centering\arraybackslash}m{#1}}
\setlength{\tabcolsep}{0.2pt}
\renewcommand{\arraystretch}{0.1}
\begin{tabular}{M{0.16\linewidth}M{0.16\linewidth}M{0.16\linewidth}M{0.16\linewidth}M{0.16\linewidth}M{0.16\linewidth}}
\includegraphics[width=\linewidth]{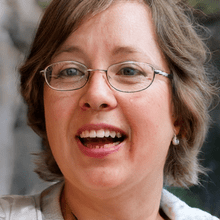}\hfill&
\includegraphics[width=\linewidth]{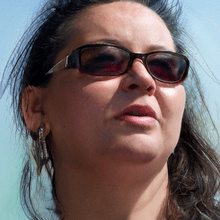}\hfill&
\includegraphics[width=\linewidth]{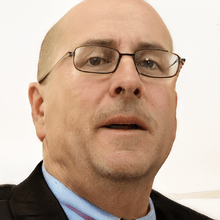}\hfill&
\includegraphics[width=\linewidth]{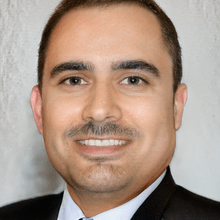}\hfill&
\includegraphics[width=\linewidth]{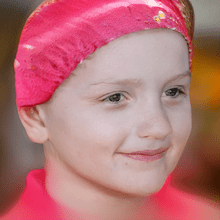}\hfill&
\includegraphics[width=\linewidth]{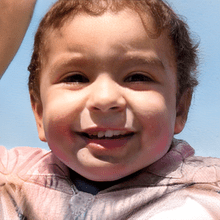}\hfill \\

\includegraphics[width=\linewidth]{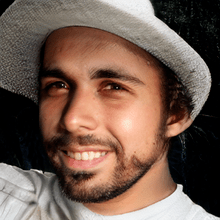}\hfill&
\includegraphics[width=\linewidth]{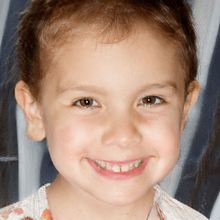}\hfill&
\includegraphics[width=\linewidth]{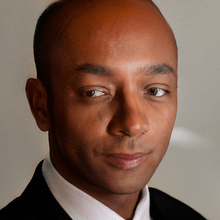}\hfill&
\includegraphics[width=\linewidth]{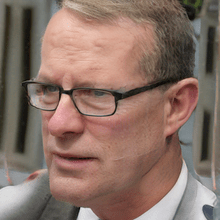}\hfill&
\includegraphics[width=\linewidth]{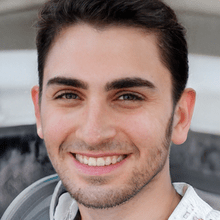}\hfill&
\includegraphics[width=\linewidth]{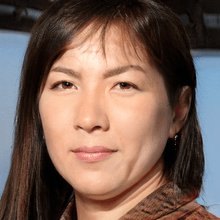}\hfill \\

\includegraphics[width=\linewidth]{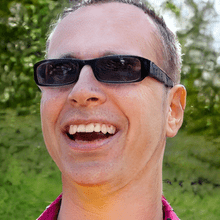}\hfill&
\includegraphics[width=\linewidth]{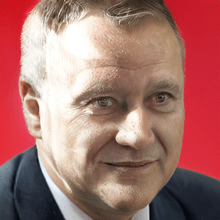}\hfill&
\includegraphics[width=\linewidth]{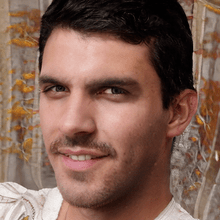}\hfill&
\includegraphics[width=\linewidth]{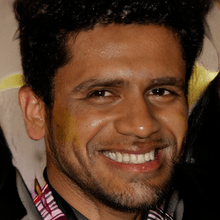}\hfill&
\includegraphics[width=\linewidth]{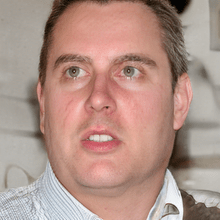}\hfill&
\includegraphics[width=\linewidth]{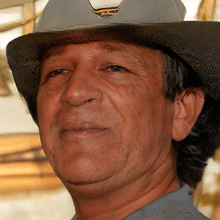}\hfill \\
\end{tabular}

\caption{\textbf{Generated pseudo ground-truth images for training latent encoder and pose estimator.} We utilize EG3D as a pseudo ground-truth generator to sample images paired with corresponding latent representations and camera viewpoint. These pseudo pairs $\{\mathbf{w}, I_\mathrm{pseudo} \}$ and $\{ \mathbf{\pi}, I_\mathrm{pseudo} \} $ are used as training data for latent encoder and pose estimator respectively. }
\label{fig:pseudo}
\end{figure}
We visualize a randomly selected pseudo dataset in \figref{fig:pseudo}. We construct 30,000 number of pseudo pairs for \textit{human faces}, and 10,000 for \textit{cat faces}. Each of the pseudo pairs rendered from a unique latent vector $\mathbf{w}_\mathrm{ps}$ with randomly sampled camera pose parameter $\pi_\mathrm{ps}$ within the camera pose boundary given by~\cite{Chan2022}. As can be seen from \figref{fig:pseudo}, the generative power of 3D GANs gives promising guidance both for the latent encoder and pose estimator. In the case of \textit{cat faces}, we additionally gave a \textit{roll} rotation to the pose parameter which gives the generated pseudo images to rotate on the image plane.

\subsection{Preparing Facial Images for GAN Inversion.}
EG3D~\cite{Chan2022} uses a slightly different cropping method for their training process. We follow the official code, found here: \url{https://github.com/NVlabs/eg3d/tree/FFHQ_preprocess} which uses face detection and pose-extraction pipeline~\cite{deng2019accurate} in order to identify and crop the face region and infer the camera viewpoint of the image. The inferred camera viewpoint is leveraged in our baseline 2D GAN inversion implementation and is also exploited as a ground-truth dataset to evaluate our pose estimation performance. 

\subsection{Implementation Details of Comparison Baselines.}
The following algorithms show our method, along with comparisons using 2D GAN inversion methods directly to 3D-aware GANs. 
\begin{center}
\begin{minipage}{0.8\linewidth}
\centering
\begin{algorithm}[H]
% \small
\SetAlgoLined
 \KwIn{real image $x$; canonical pose $\pi_c$; gradient-based optimizer $F'$; depth smoothness regularizer $DR$; depth based reprojection $proj(\cdot)$; camera intrinsics $K$ }
 \KwOut{the reconstructed image $\hat{y}$}
 Initialize() the code and pose $(w, \pi)$ = $(w', \pi ')$\;
 \While{not converged}{
  $Loss \leftarrow L\left(x, \mathcal{G}^\mathrm{c}_\mathrm{3D}(\mathbf{w}, \pi; \theta)\right)$\;
  $w \leftarrow w - \eta F'(\nabla_w Loss)$\;
  $\mathrm{projected} \leftarrow \mathcal{G}^\mathrm{c}_\mathrm{3D}(\mathbf{w}, \pi; \theta) \langle proj(\mathcal{G}^\mathrm{d}_\mathrm{3D}(\mathbf{w}, \pi; \theta), \pi,K)\rangle $ \;
  $Loss \leftarrow L\left(\mathcal{G}^\mathrm{c}_\mathrm{3D}(\mathbf{w}, \pi; \theta)),\mathrm{projected},\right)$\;
  $\pi \leftarrow \pi - \eta F'(\nabla_\pi Loss)$\;
  
 }
%   \If{pivotal tuning}{
\While{not converged}{
$Loss \leftarrow L\left(x,  \mathcal{G}^\mathrm{c}_\mathrm{3D}(\mathbf{w}, \pi; \theta))\right) + DR(\mathcal{G}^\mathrm{d}_\mathrm{3D}(\mathbf{w}, \pi; \theta))       $\;
$\theta \leftarrow \theta - \eta F'(\nabla_\theta Loss)$\;
}
%   }
%  \STATE
%  $Loss \leftarrow L\left(x, \mathcal{R}(\pi^*, G(\mathbf{w}; \theta))\right)$\;
%  $\theta \leftarrow \theta - \eta F'(\nabla_\theta L)$\;
%  }
$\hat{y}\leftarrow \mathcal{G}^\mathrm{c}_\mathrm{3D}(\mathbf{w}, \pi; \theta)$
  \caption{Our proposed method.}
 \label{alg:method}
\end{algorithm}
\begin{algorithm}[H]
% \small
\SetAlgoLined
 \KwIn{real image $x$ from viewpoint $\pi^*$; a pre-trained generator $\mathcal{R}(\cdot, G(\cdot;\theta))$; gradient-based optimizer $F'$.}
 \KwOut{the latent code $w$}
 Initialize() the code $w$ = $w'$\;
 \While{not converged}{
  $Loss \leftarrow L\left(x, \mathcal{G}^\mathrm{c}_\mathrm{3D}(\mathbf{w}, \pi^*; \theta)\right)$\;
  $w \leftarrow w - \eta F'(\nabla_w L)$\;
 }
  \If{pivotal tuning}{
        \While{not converged}{
         $Loss \leftarrow L\left(x, \mathcal{G}^\mathrm{c}_\mathrm{3D}(\mathbf{w}, \pi^*; \theta)\right)$\;
        $\theta \leftarrow \theta - \eta F'(\nabla_\theta L)$\;
        }
  }
%  \STATE
%  $Loss \leftarrow L\left(x, \mathcal{R}(\pi^*, G(w; \theta))\right)$\;
%  $\theta \leftarrow \theta - \eta F'(\nabla_\theta L)$\;
%  }
\caption{GT camera pose during optimization.}
 \label{alg:baseline1}
\end{algorithm}
\begin{algorithm}[H]
% \small
\SetAlgoLined
 \KwIn{real image $x$; a pre-trained generator $\mathcal{R}(\cdot, G(\cdot;\theta))$; gradient-based optimizer $F'$.}
 \KwOut{the latent code $w$}
 Initialize() the code and pose $(w, \pi)$ = $(w', \pi ')$\;
 \While{not converged}{
  $Loss \leftarrow L\left(x, \mathcal{G}^\mathrm{c}_\mathrm{3D}(\mathbf{w}, \pi; \theta)\right)$\;
  $(w, \pi) \leftarrow (w, \pi) - \eta F'(\nabla_{w, \pi} L)$\;
 }
  \If{pivotal tuning}{
        \While{not converged}{
        $Loss \leftarrow L\left(x,\mathcal{G}^\mathrm{c}_\mathrm{3D}(\mathbf{w}, \pi; \theta)\right)$\;
        $\theta \leftarrow \theta - \eta F'(\nabla_\theta L)$\;
        }
  }
%  \STATE
%  $Loss \leftarrow L\left(x, \mathcal{R}(\pi^*, G(w; \theta))\right)$\;
%  $\theta \leftarrow \theta - \eta F'(\nabla_\theta L)$\;
%  }
  \caption{Gradient descent to optimize camera.}
 \label{alg:baseline2}
\end{algorithm}

% \end{minipage}
% \begin{minipage}{0.35\linewidth}
% \begin{figure}[H]
% \includegraphics[width=\linewidth]{fig: edit/example-images/cat01.png}
% \caption{This is the second figure}
% \end{figure}
% \end{minipage}
\end{minipage}
\end{center}

\newpage
\section{Discussion}
\subsection{Difficulties of 3D GAN inversion}
\begin{figure}[h]
\centering
\newcolumntype{M}[1]{>{\centering\arraybackslash}m{#1}}
\setlength{\tabcolsep}{1pt}
\renewcommand{\arraystretch}{0.5}
\begin{tabular}{M{0.12\linewidth} @{\hskip 0.01\linewidth}| @{\hskip 0.01\linewidth}M{0.12\linewidth} @{\hskip 0.01\linewidth}| @{\hskip 0.01\linewidth}M{0.12\linewidth} @{\hskip 0.01\linewidth}| @{\hskip 0.01\linewidth}M{0.12\linewidth} M{0.12\linewidth} M{0.12\linewidth} @{\hskip 0.01\linewidth}| @{\hskip 0.01\linewidth}M{0.12\linewidth}}

Input &
Inv. &
Pivotal Tuning &
& Novel Views & &
Mesh \\

\includegraphics[width=\linewidth]{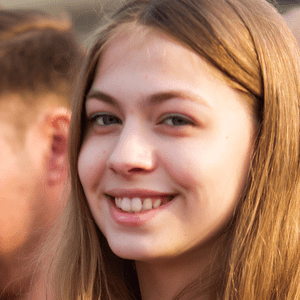} &
\includegraphics[width=\linewidth]{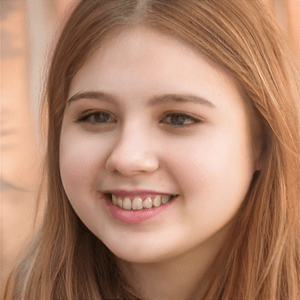} &
\includegraphics[width=\linewidth]{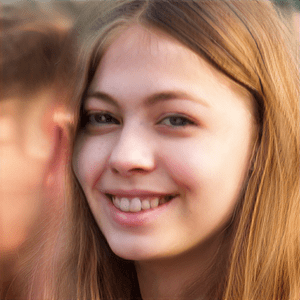} &
\includegraphics[width=\linewidth]{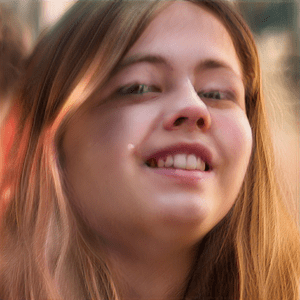} &
\includegraphics[width=\linewidth]{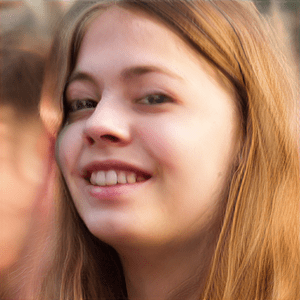} &
\includegraphics[width=\linewidth]{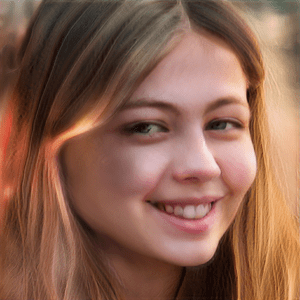} &
\includegraphics[trim=340 120 360 50,clip, width=\linewidth, height=\linewidth]{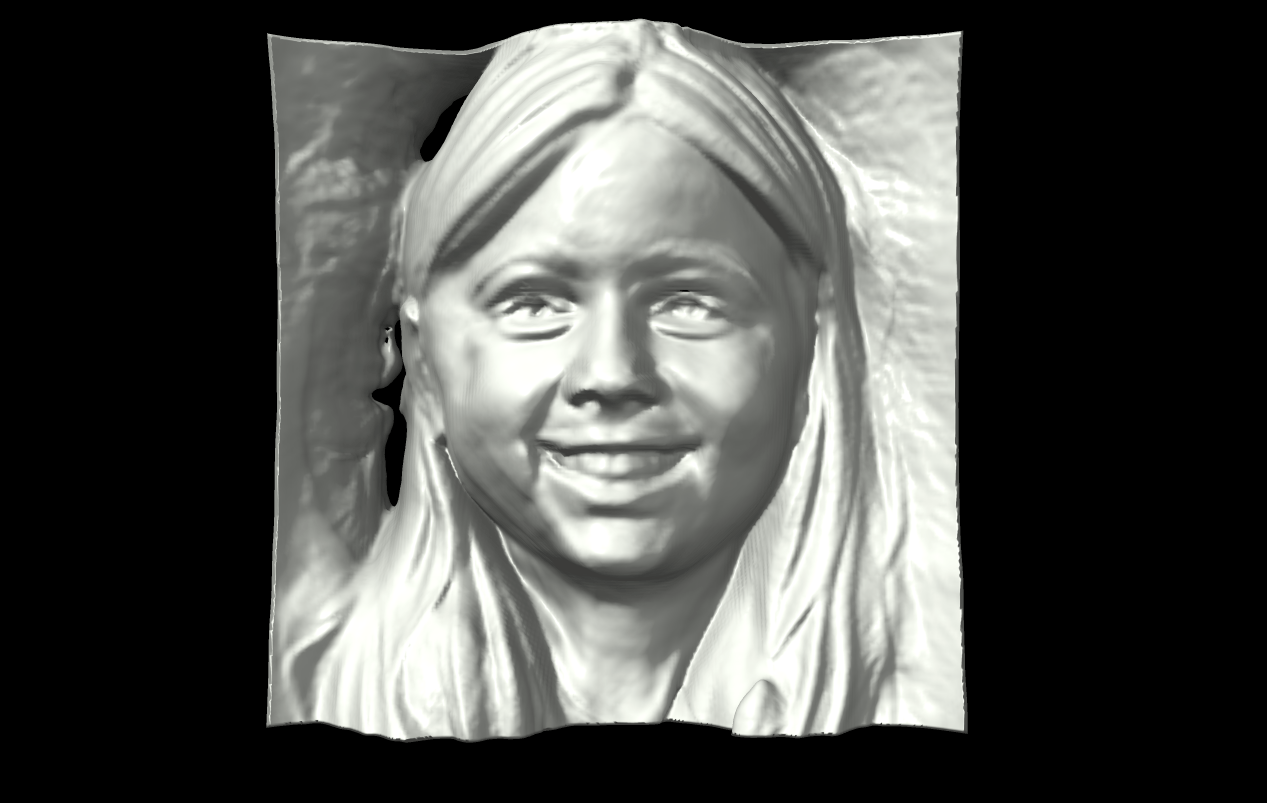} \\

\includegraphics[width=\linewidth]{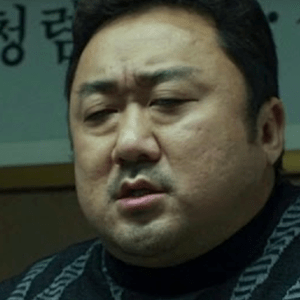} &
\includegraphics[width=\linewidth]{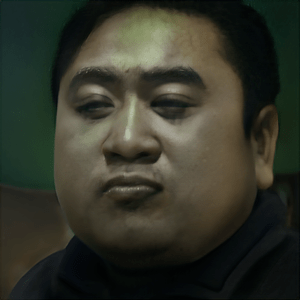} &
\includegraphics[width=\linewidth]{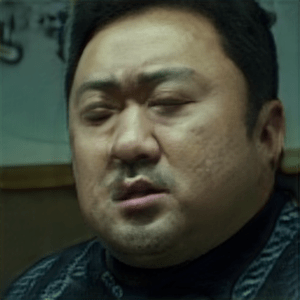} &
\includegraphics[width=\linewidth]{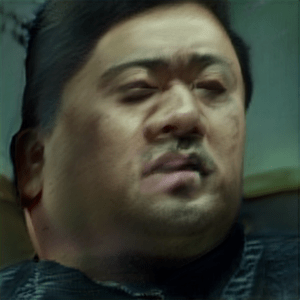} &
\includegraphics[width=\linewidth]{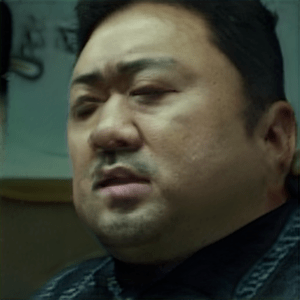} &
\includegraphics[width=\linewidth]{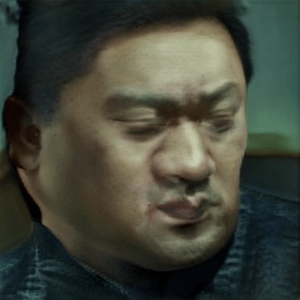} &
\includegraphics[trim=320 120 370 70,clip, width=\linewidth, height=\linewidth]{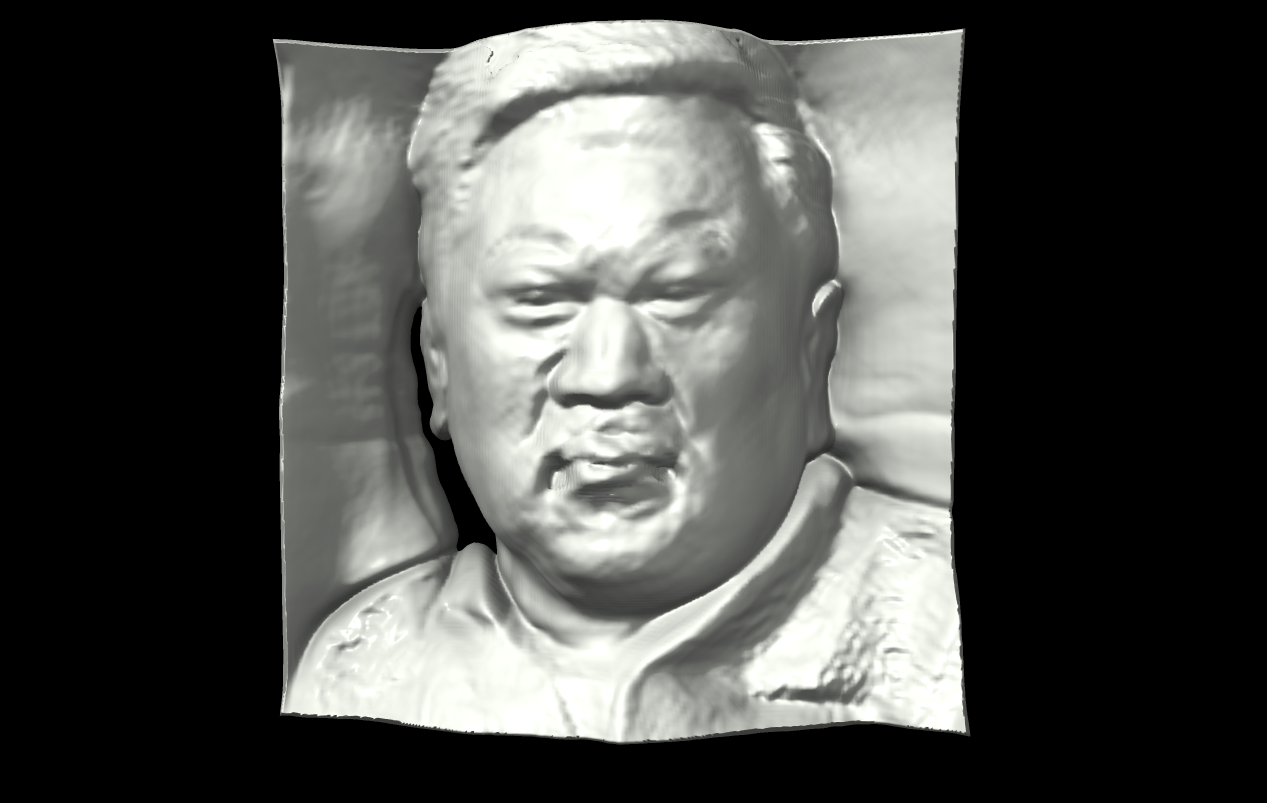} \\

\end{tabular}
\caption{Importance of accurate camera pose estimation. }
\label{fig:verydifficult3d}
\end{figure}

As demonstrated in the main paper, 3D GAN inversion is non-trivial, since one struggles to optimize if the other is inaccurate. Optimizing latent features in 3D GANs gives an additional consideration to pose optimization, which needs to be optimized simultaneously. Specifically, this becomes difficult when either latent feature or camera pose is imperfect. In \figref{fig:verydifficult3d}, we show failure cases to illustrate how the imperfect camera pose estimation leads to shape distortion. The second column of \figref{fig:verydifficult3d} is the final inversion result without pivotal tuning. As the pose estimation fails, the second column shows misalignment in the head pose. Even though the pivotal tuning step resolves the visual perception error, they often fail on novel view synthesis (4-6 columns). This is because the tuning procedure forces the implicit volume to fit into an input image with a misaligned camera parameter, which becomes distortions to the other viewpoints. 

\begin{figure}[!h]
\centering
\newcolumntype{M}[1]{>{\centering\arraybackslash}m{#1}}
\setlength{\tabcolsep}{1pt}
\renewcommand{\arraystretch}{0.5}
\begin{tabular}{M{0.11\linewidth} @{\hskip 0.01\linewidth} M{0.11\linewidth}  M{0.11\linewidth}  M{0.11\linewidth} M{0.11\linewidth} @{\hskip 0.01\linewidth} M{0.11\linewidth} @{\hskip 0.01\linewidth} M{0.11\linewidth}@{\hskip 0.01\linewidth} M{0.11\linewidth}}

\multirow{2}{*}{Input} & \multicolumn{4}{c}{iters} & \multirow{2}{*}{Tuned} & \multirow{2}{*}{Novel View} & \multirow{2}{*}{Mesh}  \\ 
\cmidrule(lr){2-5}  & 0 & 150 & 300 & 500 & & \\

\includegraphics[width=\linewidth]{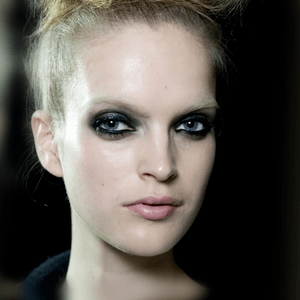} &
\includegraphics[width=\linewidth]{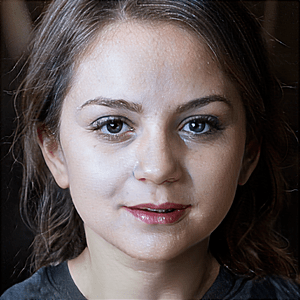} &
\includegraphics[width=\linewidth]{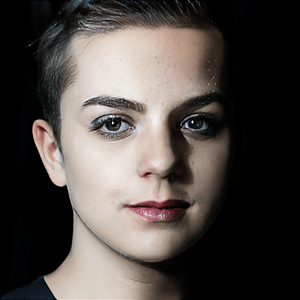} &
\includegraphics[width=\linewidth]{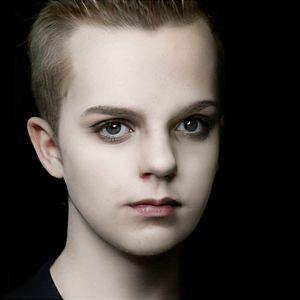} &
\includegraphics[width=\linewidth]{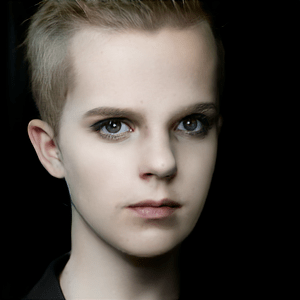} &
\includegraphics[width=\linewidth]{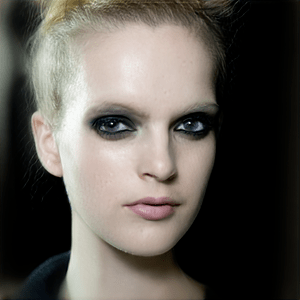} &
\includegraphics[width=\linewidth]{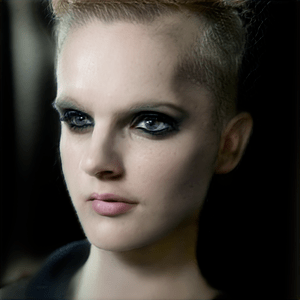} &
\includegraphics[trim=360 150 370 60,clip, width=\linewidth, height=\linewidth]{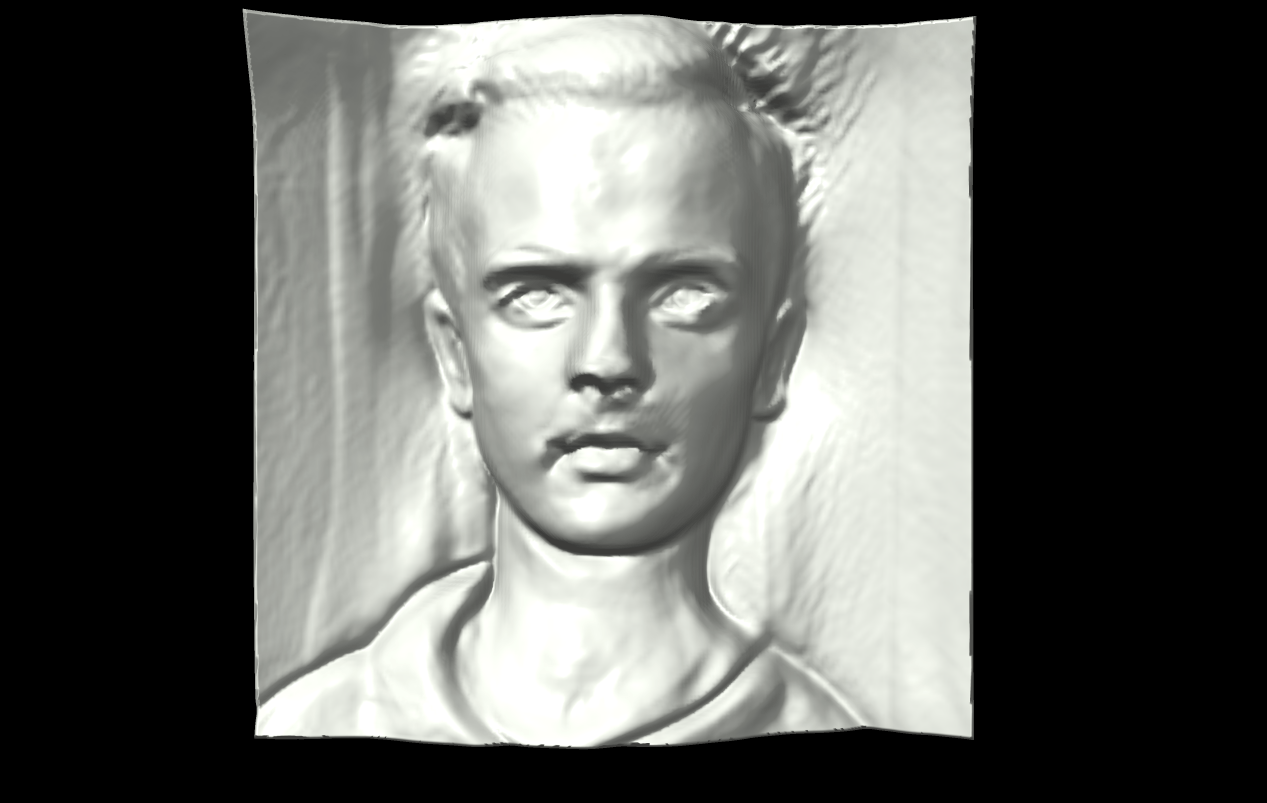} \\

\includegraphics[width=\linewidth]{AppendixFigure/why3d/042842_gt.png} &
\includegraphics[width=\linewidth]{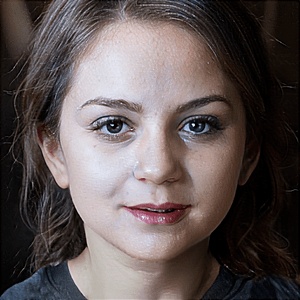} &
\includegraphics[width=\linewidth]{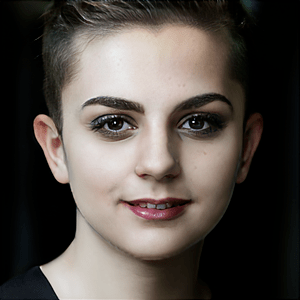} &
\includegraphics[width=\linewidth]{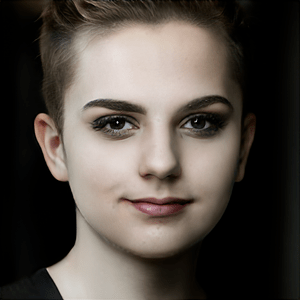} &
\includegraphics[width=\linewidth]{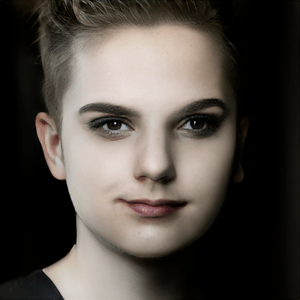} &
\includegraphics[width=\linewidth]{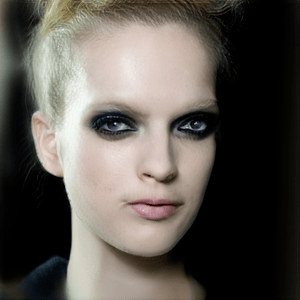} &
\includegraphics[width=\linewidth]{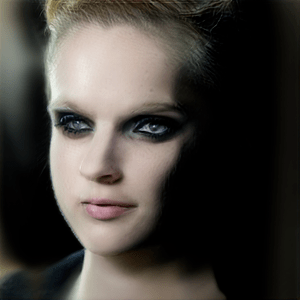} &
\includegraphics[trim=360 150 370 60,clip, width=\linewidth, height=\linewidth]{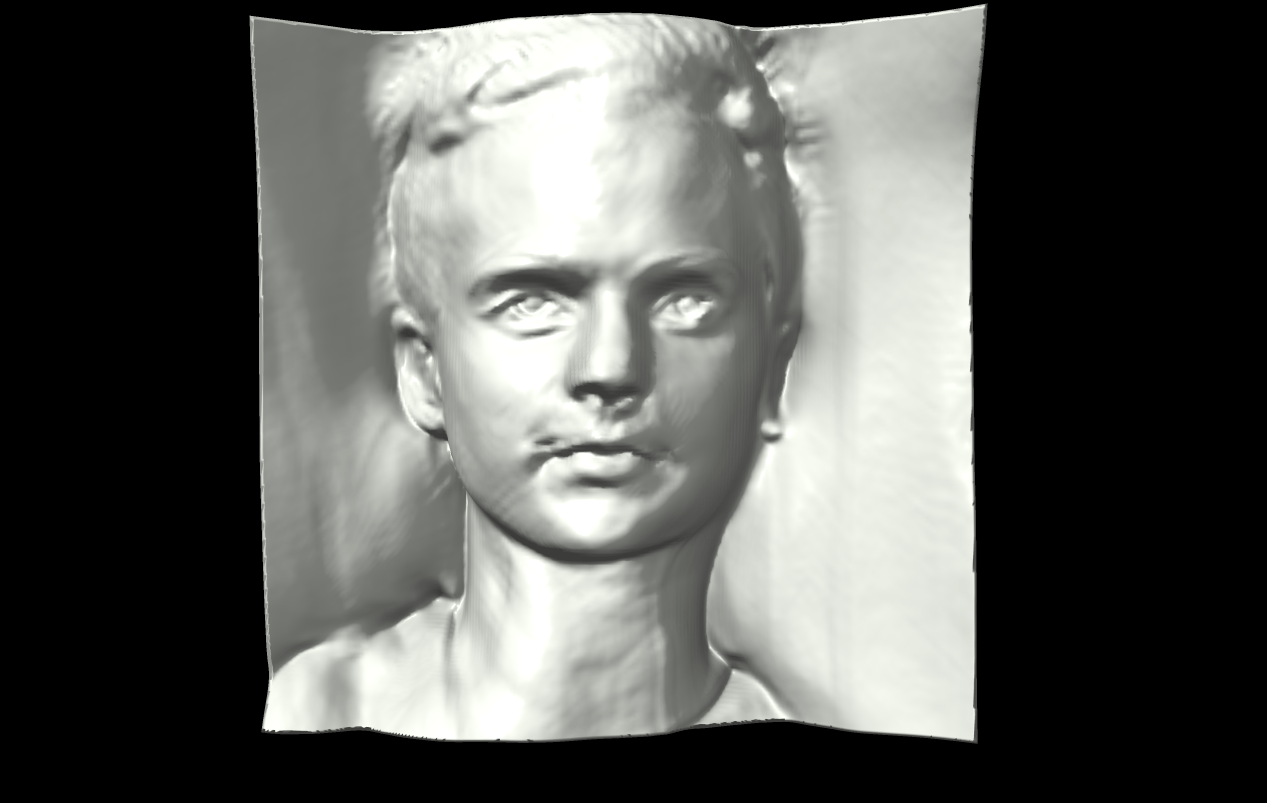} \\

\end{tabular}
\caption{\textbf{Importance of additional learnable translation parameter.} We compare our model(top row) with the model that doesn't have additional learnable translation parameters(bottom row). The top row shows both gradual changes in head location on the image plane and camera direction, while the other cannot converge to suitable camera rotation.}
\label{fig:translation}
\end{figure}

Furthermore, we conduct another ablation study on camera misalignment, illustrated in \figref{fig:translation}. The top row shows an inversion process using our training strategy, while the bottom row does not optimize the additional trainable translation vector. Following the preprocessing stage of~\cite{deng2019accurate}, they create a by-product of camera translation, which means every sample is not on the object-centric condition. Check for the optimization step (2-5 columns) of the first row to find the gradual change of its head position to the upper-left direction. The second row, however, cannot find its optimal direction because of the strict condition of the translation vector. Formally speaking, the lack of translation optimization makes the camera see always the center of the implicit space, thus limiting the optimizable path of the camera pose.

%\subsection{Prior Knowledge of Object Symmetry}
%We observed interesting findings on the prior knowledge of the 3D GAN model. 

% \subsection{Importance of 3D GAN Inversion}
% We provide additional comparison results 
% The rotation results may differ as the two generators employ different cropping procedures in order to generate their respective training 

\begin{figure}[!p]
\centering
\newcolumntype{M}[1]{>{\centering\arraybackslash}m{#1}}
\setlength{\tabcolsep}{1pt}
\renewcommand{\arraystretch}{0.5}
\begin{tabular}{M{0.12\linewidth} M{0.05\linewidth} M{0.12\linewidth} M{0.12\linewidth} M{0.12\linewidth}M{0.12\linewidth}M{0.12\linewidth}M{0.12\linewidth}}
\multirow{2}{*}{
\makecell{
% \vspace{-1pt} \\
\includegraphics[width=\linewidth]{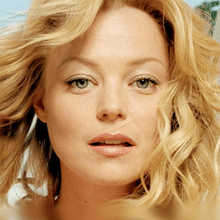}}
}
& \rotatebox[origin=rl]{90}{\small{2D}} &
\includegraphics[width=\linewidth]{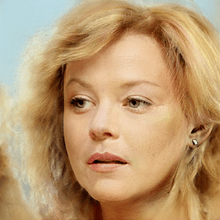}\hfill &
\includegraphics[width=\linewidth]{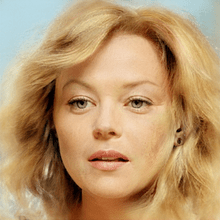}\hfill &
\includegraphics[width=\linewidth]{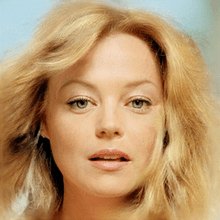}\hfill&
\includegraphics[width=\linewidth]{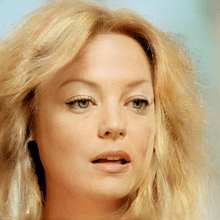}\hfill &
\includegraphics[width=\linewidth]{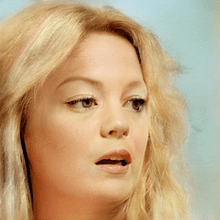}\hfill &
\includegraphics[width=\linewidth]{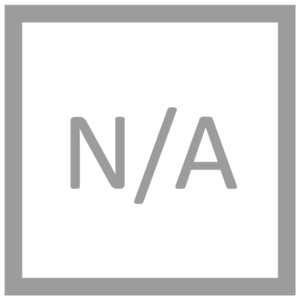}\hfill \\

&\rotatebox[origin=rl]{90}{\small{3D}}&
\includegraphics[width=\linewidth]{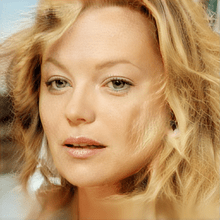}\hfill&
\includegraphics[width=\linewidth]{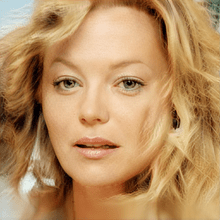}\hfill&
\includegraphics[width=\linewidth]{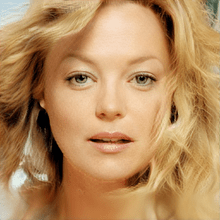}\hfill&
\includegraphics[width=\linewidth]{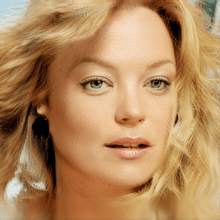}\hfill &
\includegraphics[width=\linewidth]{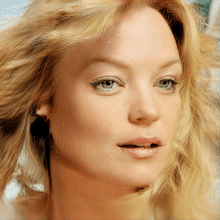}\hfill &
\includegraphics[trim=360 120 370 60,clip,width=\linewidth, height=\linewidth]{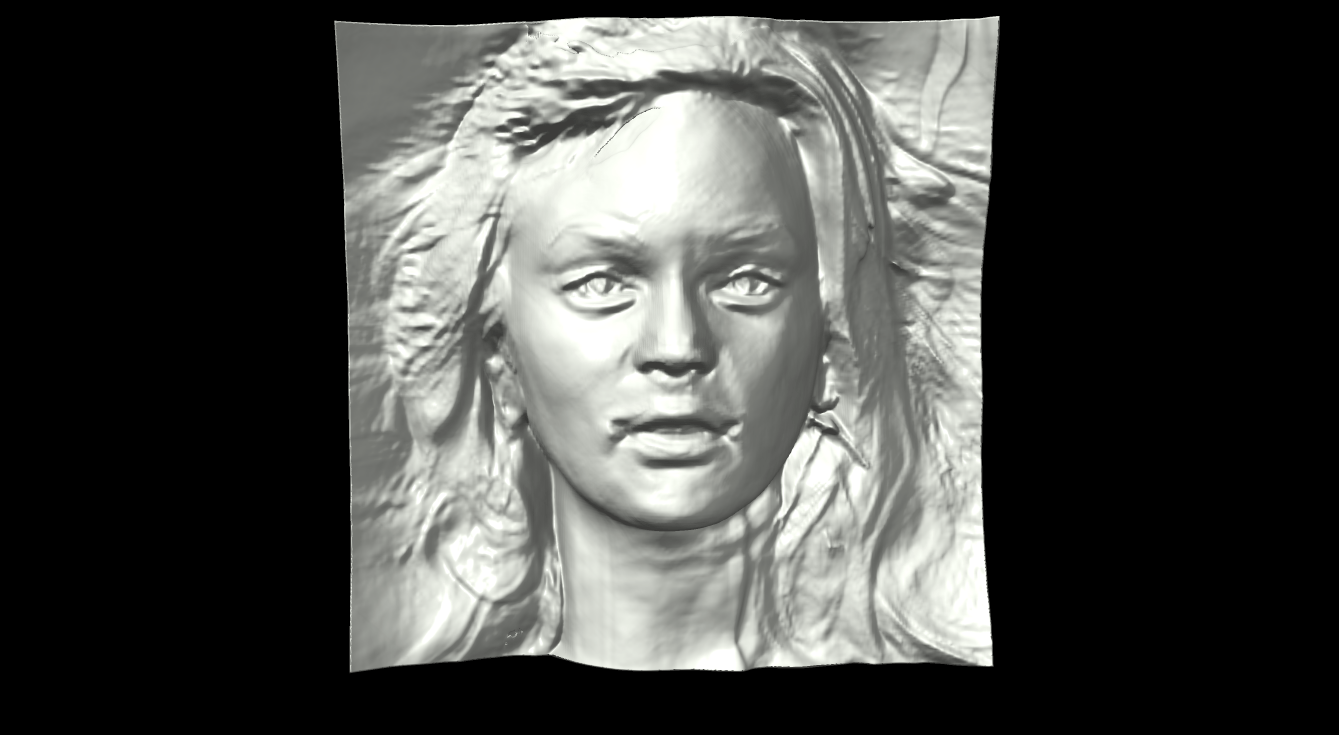} \\
\multirow{2}{*}{
\makecell{
% \vspace{-1pt} \\
\includegraphics[width=\linewidth]{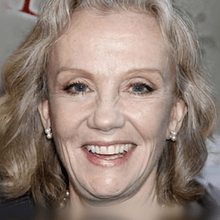}}
}
& \rotatebox[origin=rl]{90}{\small{2D}} &
\includegraphics[width=\linewidth]{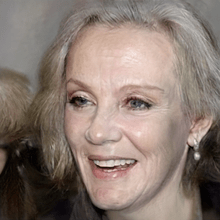}\hfill &
\includegraphics[width=\linewidth]{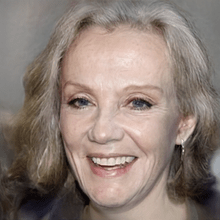}\hfill &
\includegraphics[width=\linewidth]{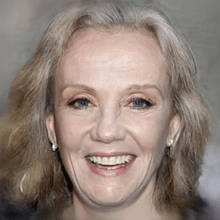}\hfill&
\includegraphics[width=\linewidth]{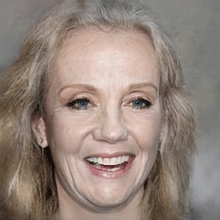}\hfill &
\includegraphics[width=\linewidth]{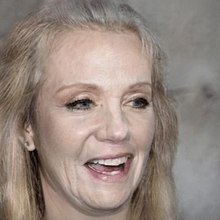}\hfill &
\includegraphics[width=\linewidth]{Fig_vs2D/NA.png}\hfill \\

&\rotatebox[origin=rl]{90}{\small{3D}}&
\includegraphics[width=\linewidth]{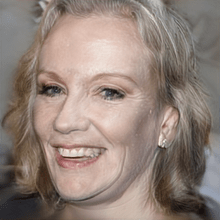}\hfill&
\includegraphics[width=\linewidth]{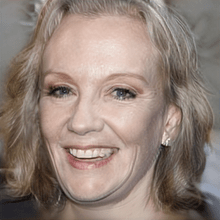}\hfill&
\includegraphics[width=\linewidth]{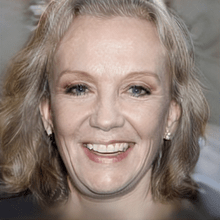}\hfill&
\includegraphics[width=\linewidth]{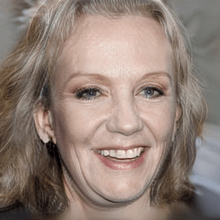}\hfill &
\includegraphics[width=\linewidth]{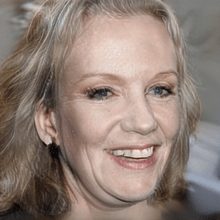}\hfill &
\includegraphics[trim=360 120 370 60,clip,width=\linewidth, height=\linewidth]{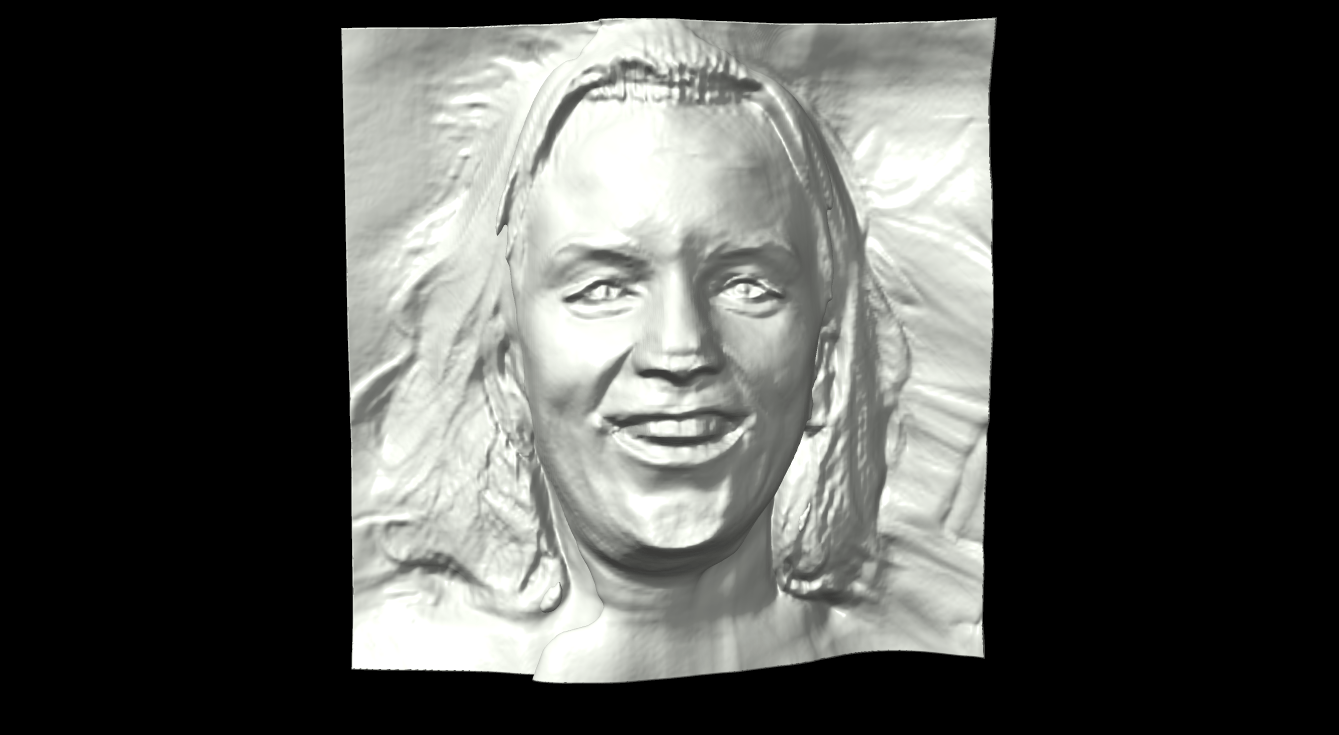} \\
\end{tabular}
\vspace{-5pt}
\caption{\textbf{Comparison of pose manipulation of real images inverted with 2D GANs and 3D GANs.} As can be seen, pose manipulation using PTI on StyleGAN2 (first row) only allows for implicit control by the editing magnitude and larger step sizes result in undesired transformations. On the other hand, using our method on EG3D (second row), allows for explicit control and because the acquired latent representation are viewpoint independent, the edits geometrically consistent.}
\label{appendix:pose_manipulation_2d}
% \end{figure}
% \begin{figure}[!h]
% \centering
% \newcolumntype{M}[1]{>{\centering\arraybackslash}m{#1}}
% \setlength{\tabcolsep}{1pt}
% \renewcommand{\arraystretch}{0.5}
\begin{tabular}{M{0.03\linewidth} M{0.12\linewidth} M{0.03\linewidth} M{0.12\linewidth} M{0.12\linewidth} M{0.12\linewidth}M{0.12\linewidth}M{0.12\linewidth}M{0.12\linewidth}}
\multirow{8}{*}{ \rotatebox[origin=rb]{90}{\small{+ Age}} }&
\multirow{2}{*}{\includegraphics[width=\linewidth]{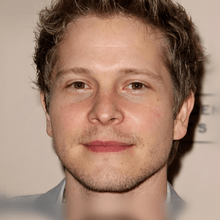}}
& \rotatebox[origin=rl]{90}{\small{2D}} &
\includegraphics[width=\linewidth]{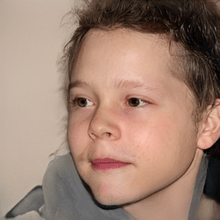}\hfill &
\includegraphics[width=\linewidth]{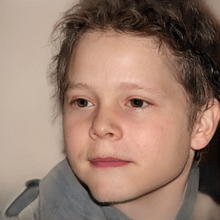}\hfill &
\includegraphics[width=\linewidth]{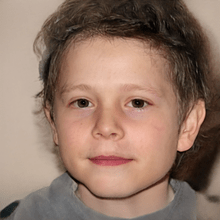}\hfill&
\includegraphics[width=\linewidth]{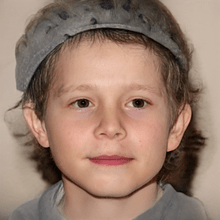}\hfill &
\includegraphics[width=\linewidth]{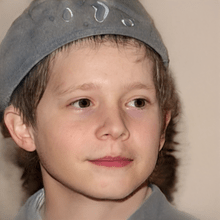}\hfill &
\includegraphics[width=\linewidth]{Fig_vs2D/NA.png}\hfill \\

&&\rotatebox[origin=rb]{90}{\small{3D}}&
\includegraphics[width=\linewidth]{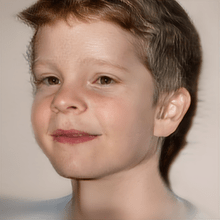}\hfill&
\includegraphics[width=\linewidth]{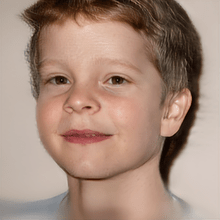}\hfill&
\includegraphics[width=\linewidth]{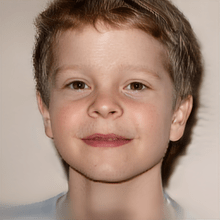}\hfill&
\includegraphics[width=\linewidth]{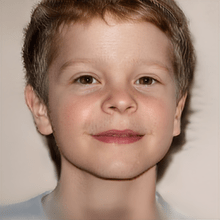}\hfill &
\includegraphics[width=\linewidth]{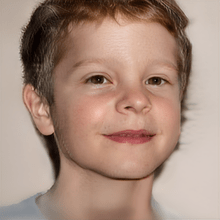}\hfill &
\includegraphics[trim=360 120 370 60,clip,width=\linewidth, height=\linewidth]{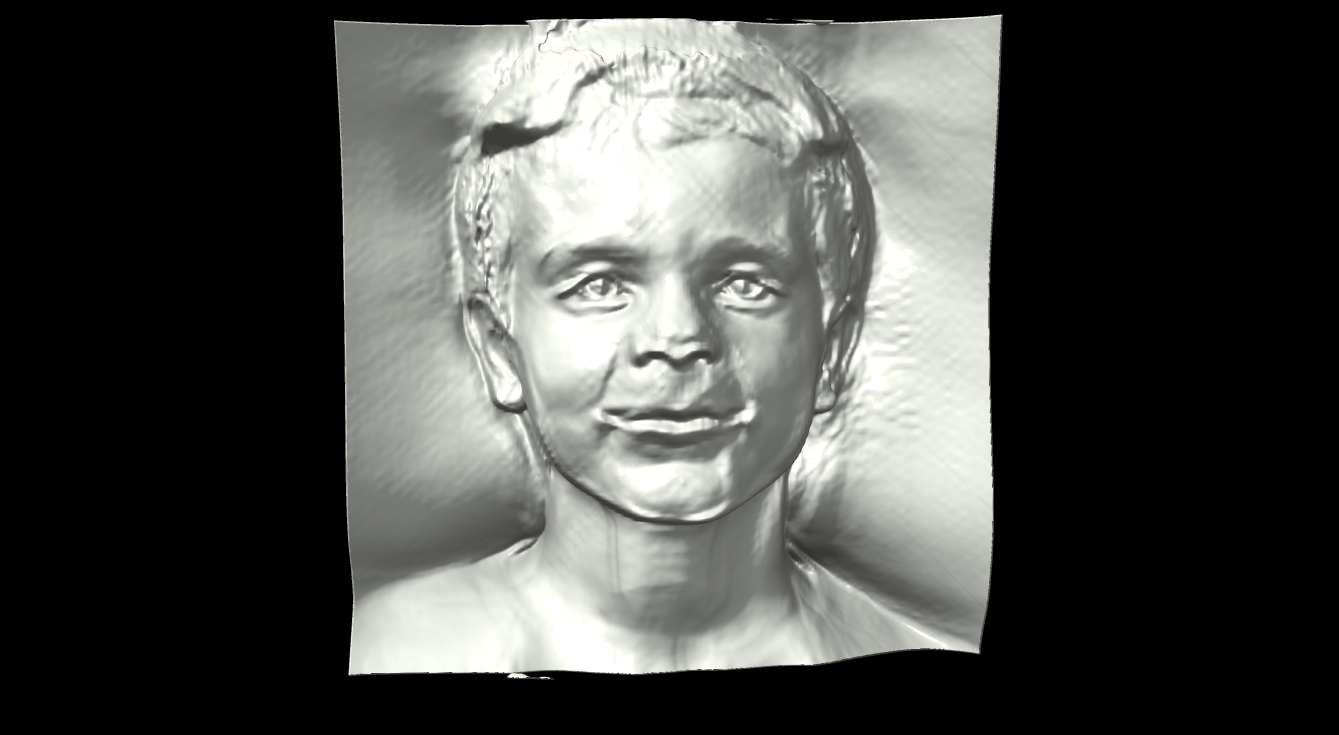} \\

\multirow{8}{*}{ \rotatebox[origin=rb]{90}{\small{+ gender}} }&
\multirow{2}{*}{\includegraphics[width=\linewidth]{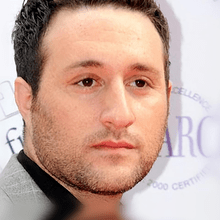}}
&  \rotatebox[origin=rl]{90}{\small{2D}} &
\includegraphics[width=\linewidth]{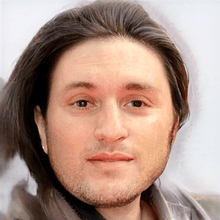}\hfill &
\includegraphics[width=\linewidth]{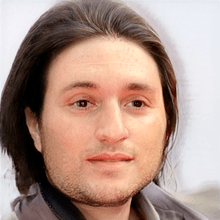}\hfill &
\includegraphics[width=\linewidth]{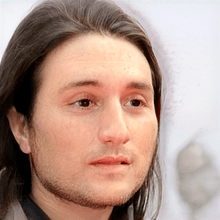}\hfill&
\includegraphics[width=\linewidth]{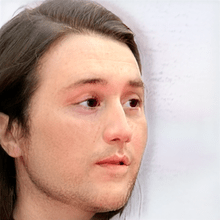}\hfill &
\includegraphics[width=\linewidth]{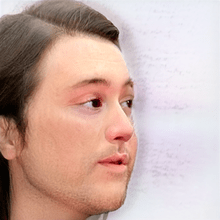}\hfill &
\includegraphics[width=\linewidth]{Fig_vs2D/NA.png}\hfill \\

&&\rotatebox[origin=rl]{90}{\small{3D}}&
\includegraphics[width=\linewidth]{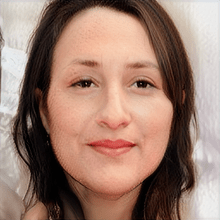}\hfill&
\includegraphics[width=\linewidth]{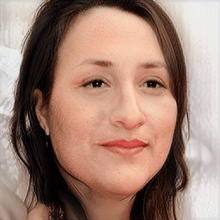}\hfill&
\includegraphics[width=\linewidth]{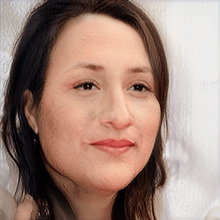}\hfill&
\includegraphics[width=\linewidth]{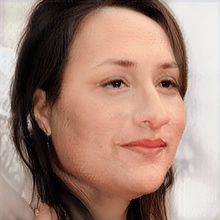}\hfill &
\includegraphics[width=\linewidth]{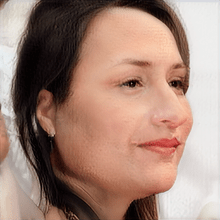}\hfill &
\includegraphics[trim=360 120 370 60,clip,width=\linewidth, height=\linewidth]{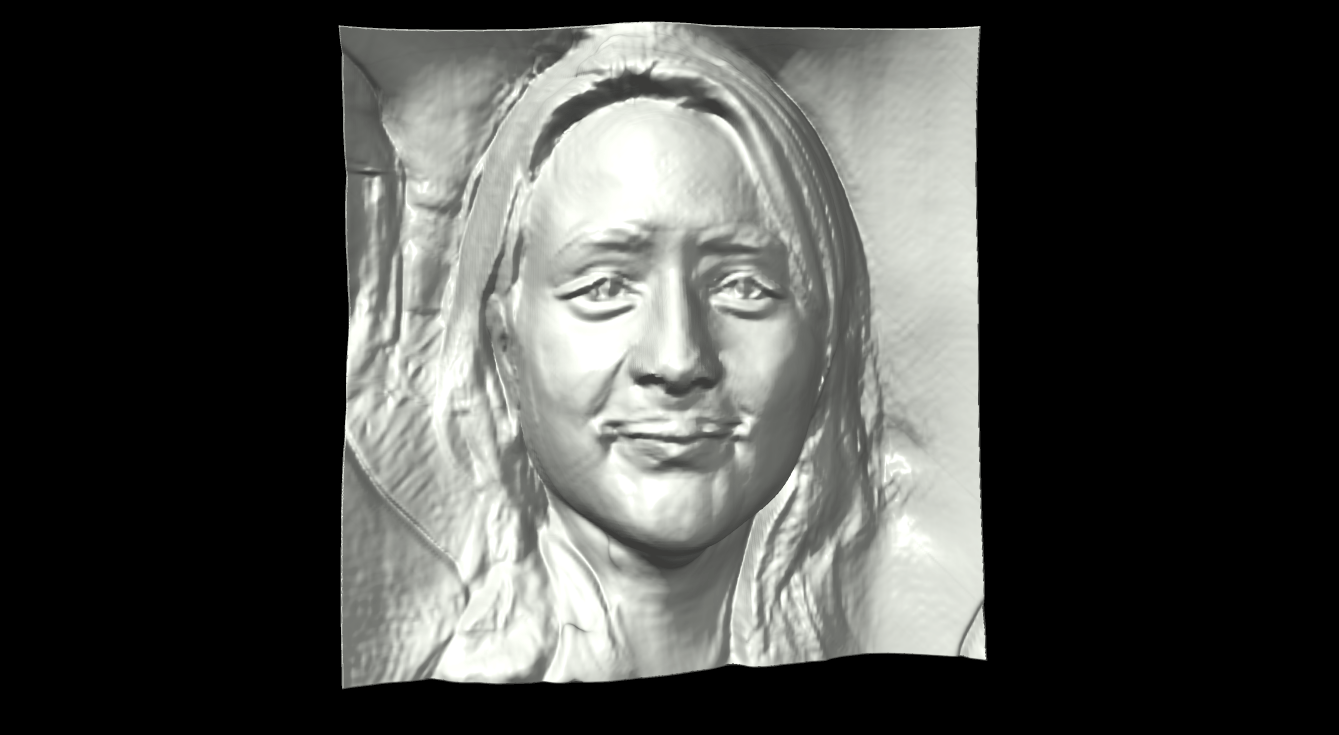} \\

\end{tabular}
\vspace{-5pt}
\caption{\textbf{Additional results of simultaneous attribute editing and viewpoint shift comparison of 2D and 3D GANs.} We compare editing results of applying attribute editing (smile) and viewpoint interpolation at the same time on the latent code acquired by PTI~\cite{roich2021pivotal} on StyleGAN2~\cite{Karras2019stylegan2} and the latent code acquired by our method on EG3D~\cite{Chan2022}}
\label{appendix:simultaneous editing}
\end{figure}

\newpage
\section{Qualitative Evaluation of Editing}

\begin{table}[!h]
\newcolumntype{M}[1]{>{\centering\arraybackslash}m{#1}}
\begin{subtable}[h]{\linewidth}
\centering
\resizebox{0.8\linewidth}{!}{
\begin{tabular}{M{0.05\linewidth} |M{0.08\linewidth}|  M{0.1\linewidth} M{0.1\linewidth}M{0.1\linewidth} M{0.1\linewidth} M{0.1\linewidth} M{0.1\linewidth} M{0.1\linewidth} }
\toprule
Attr. & $\gamma$ & SG2 & SG2 $\mathcal{W}+$ & PTI & SG2$^\dagger$ & SG2$^\dagger$ $\mathcal{W}+$ & PTI$^\dagger$ & Ours \\
\midrule
\multirow{7}{*}{
\rotatebox[origin=c]{90}{Age}
}
&-3.6&	-7.320&	-6.598&	-6.625&	-6.522&	-6.228&	-6.831&	-7.146 \\
&-2.4&	-5.455&	-4.804&	-4.526&	-4.834&	-4.518&	-4.726&	-4.922\\
&-1.2&	-3.02&	-2.557&	-2.3&	-2.737&	-2.476&	-2.518&	-2.584\\
&0&	0&	0&	0&	0&	0&	0&	0\\
&1.2&	3.142&	2.696&	2.656&	3.141&	3.075&	2.570&	2.725\\
&2.4&	6.464&	5.74&	5.506&	6.686&	6.414&	5.574&	5.929\\
&3.6&	9.868&	8.926&	8.558&	10.267&	10.077&	8.668&	9.296\\

\midrule
\multirow{7}{*}{
\rotatebox[origin=c]{90}{Smile}
}
&-2&	-3.150&	-2.392&	-2.715&	-2.627&	-2.111&	-2.107&	-2.711\\
&-1.33&	-2.210&	-1.597&	-1.838&	-1.829&	-1.501&	-1.447&	-1.882\\
&-0.67&	-1.129&	-0.798&	-0.920&	-0.940&	-0.764&	-0.714&	-0.937\\
&0&	0&	0&	0&	0&	0&	0&	0\\
&0.66&	0.964&	0.817&	0.850&	0.831&	0.711&	0.608&	0.919\\
&1.33&	1.925&	1.583&	1.734&	1.558&	1.271&	1.364&	1.854\\
&2&	2.762&	2.226&	2.524&	2.189&	1.861&	1.962&	2.694\\
\bottomrule
\end{tabular}}
\caption{Quantitative evaluation of manipulation capability}
\label{tab:magnipulation_capability}
\end{subtable}
\hfill
\begin{subtable}[h]{\linewidth}
\centering
\includegraphics[width=\linewidth]{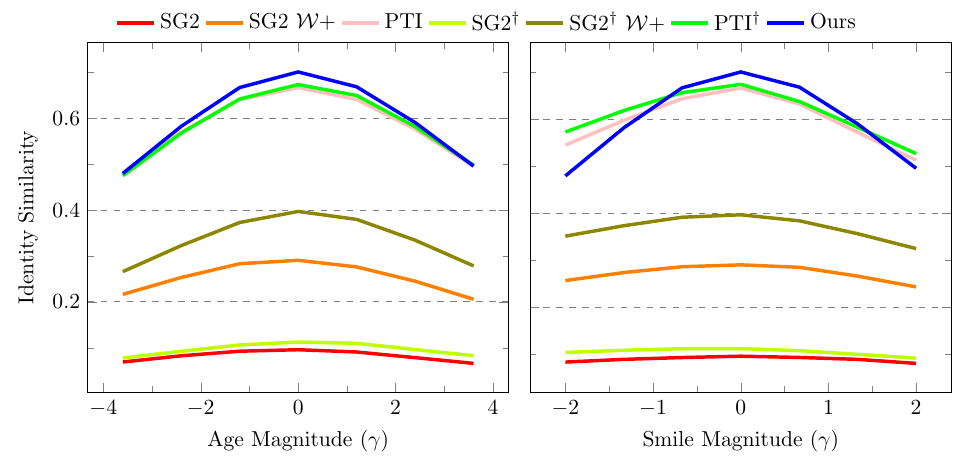}\hfill 
\caption{Quantitative evaluation of identity preservation}
\label{tab:edit_similarity}
\end{subtable}
\hfill
\caption{\textbf{Quantitative evaluation of editability.} In (a), we apply varying magnitudes $\gamma$ of age edit and smile edit to the latent codes acquired by each method and measure the amount of age change and difference of smile extent respectively. In (b), we also compare the identity similarity between the original image and edited images.}
\label{tab:temps}
\end{table}

A well-performed semantic edit should preserve the original identity of the object while performing meaningful modifications to the desired attributes and that attributes only. 
%While it has been proved that utilizing $\mathcal{W}+$ space in style-based generators results in better reconstruction quality but poorer manipulation performance. We follow the works of \cite{roich2021pivotal} and infer the latent code in the native $\mathcal{W}$ space not only to preserve its editing capacity, but also to generate plausible geometry by keeping the inferred $\mathbf{w}$ in the sampling distribution.
% We quantitatively compare the editability of inversion methods by calculating the identity preservation with respect to a range of different editing magnitudes.
From the latent code and camera pose derived by each inversion method, we first evaluate the editing capabilities by manipulating an attribute with the same magnitude and measuring the amount of variation between the original image and the edited image. 
While camera pose editing is a standard attribute to compare latent space manipulation, viewpoint editing cannot be used to compare the different methods as the generator in question controls geometric attributes explicitly. 
Instead, we chose \textit{age} and \textit{smile} for the attributes to compare latent space manipulation, using the trait-specific estimators, DEX VGG~\cite{Rothe-ICCVW-2015} for \textit{age} and the face attribute classifier used in \cite{lin2021anycost} for \textit{smile}. 
The results are shown in \tabref{tab:magnipulation_capability}. 
On the other hand, an ideal manipulation of the acquired latent code should achieve not only high editing ability but also high identity preservation for the unchanged attributes. For each inversion method, we compute the identity similarity between the original and edited images for different editing magnitudes. As before, we use the CurricularFace method~\cite{huang2020curricularface} to calculate identity similarity. The results are shown in \tabref{tab:edit_similarity}.

\newpage
\section{Additional Experimental Results}
\begin{figure}[!p]
\centering
\newcolumntype{M}[1]{>{\centering\arraybackslash}m{#1}}
\setlength{\tabcolsep}{1pt}
\renewcommand{\arraystretch}{0.5}
\begin{tabular}{M{0.1\linewidth}M{0.1\linewidth}M{0.1\linewidth}
@{\hskip 0.01\linewidth} M{0.1\linewidth}M{0.1\linewidth}M{0.1\linewidth} @{\hskip 0.01\linewidth} M{0.1\linewidth}M{0.1\linewidth}M{0.1\linewidth}}

& \includegraphics[width=\linewidth]{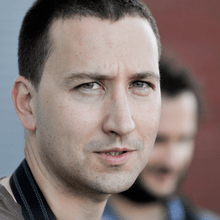}\hfill &
&& \includegraphics[width=\linewidth]{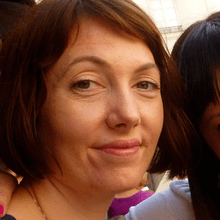}\hfill &
&& \includegraphics[width=\linewidth]{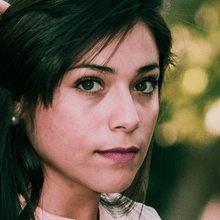}\hfill \\

\includegraphics[width=\linewidth]{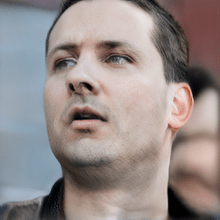}\hfill &
\includegraphics[width=\linewidth]{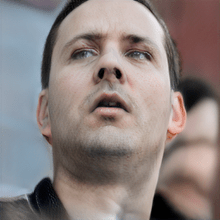}\hfill&
\includegraphics[width=\linewidth]{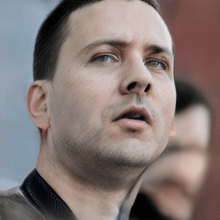}\hfill &

\includegraphics[width=\linewidth]{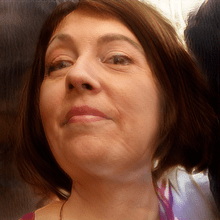}\hfill &
\includegraphics[width=\linewidth]{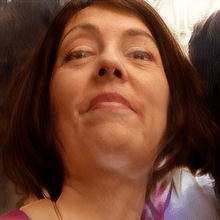}\hfill&
\includegraphics[width=\linewidth]{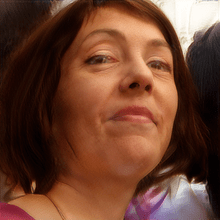}\hfill &

\includegraphics[width=\linewidth]{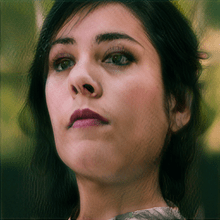}\hfill &
\includegraphics[width=\linewidth]{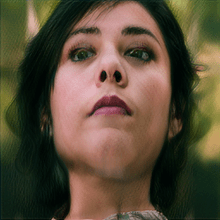}\hfill&
\includegraphics[width=\linewidth]{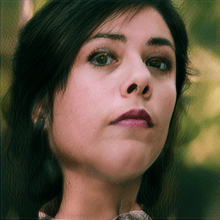}\hfill \\

\includegraphics[width=\linewidth]{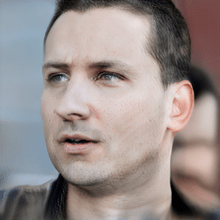}\hfill &
\includegraphics[width=\linewidth]{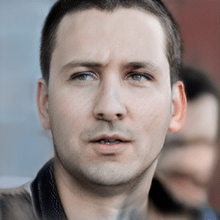}\hfill
\llap{\includegraphics[trim=650 90 380 120,clip,width=0.5\linewidth, height=\linewidth]{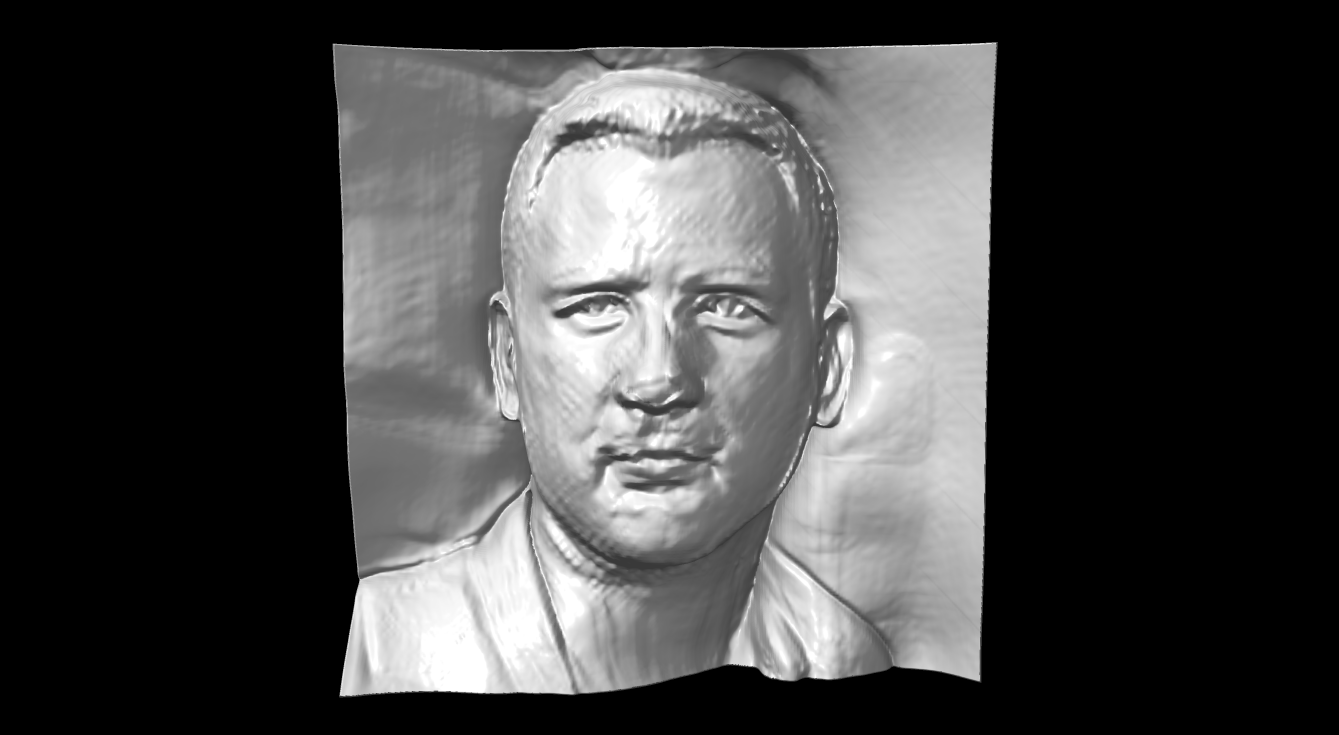}}&
\includegraphics[width=\linewidth]{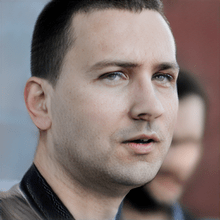}\hfill &

\includegraphics[width=\linewidth]{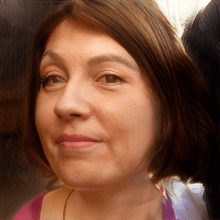}\hfill &
\includegraphics[width=\linewidth]{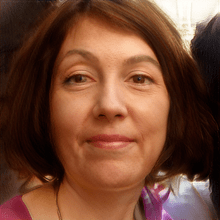}\hfill
\llap{\includegraphics[trim=650 90 380 100,clip,width=0.5\linewidth, height=\linewidth]{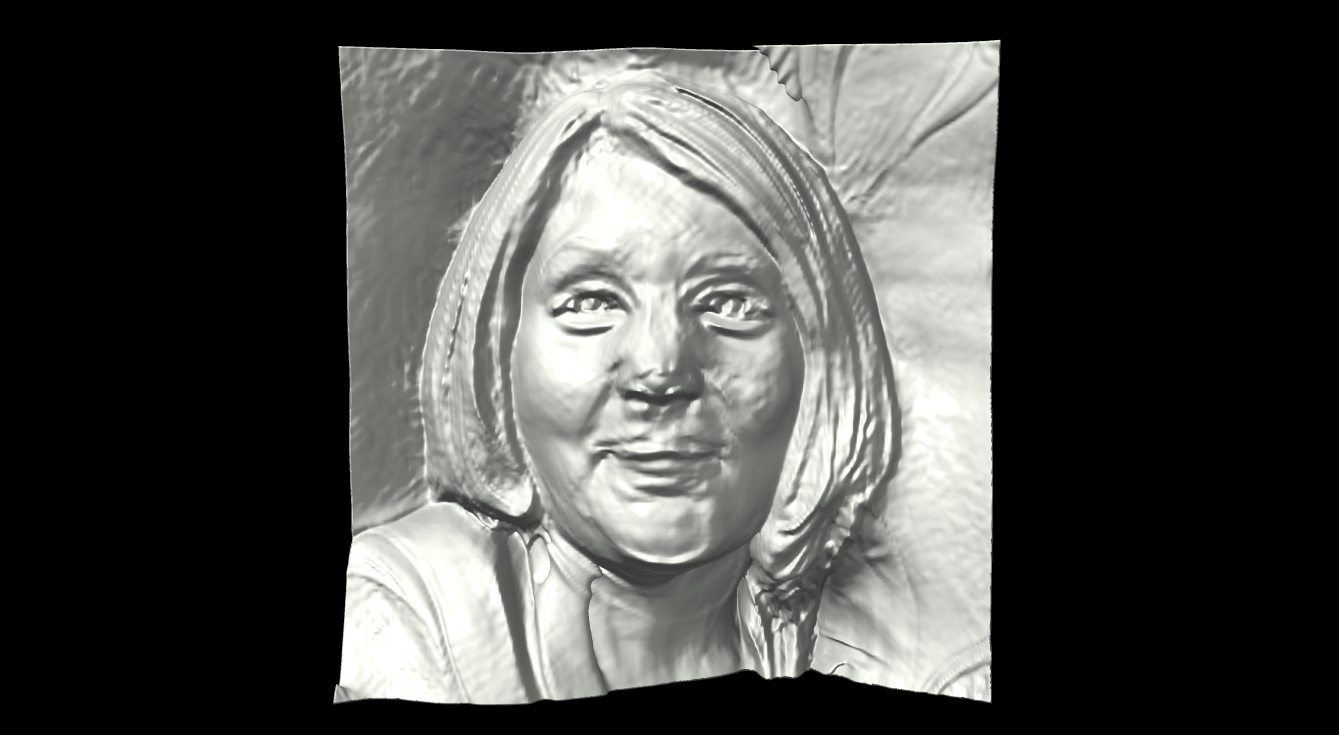}}&
\includegraphics[width=\linewidth]{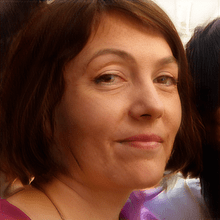}\hfill &

\includegraphics[width=\linewidth]{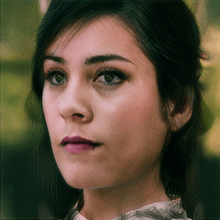}\hfill &
\includegraphics[width=\linewidth]{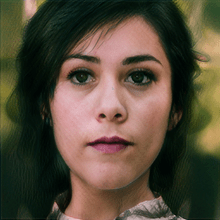}\hfill
\llap{\includegraphics[trim=650 80 380 100,clip,width=0.5\linewidth, height=\linewidth]{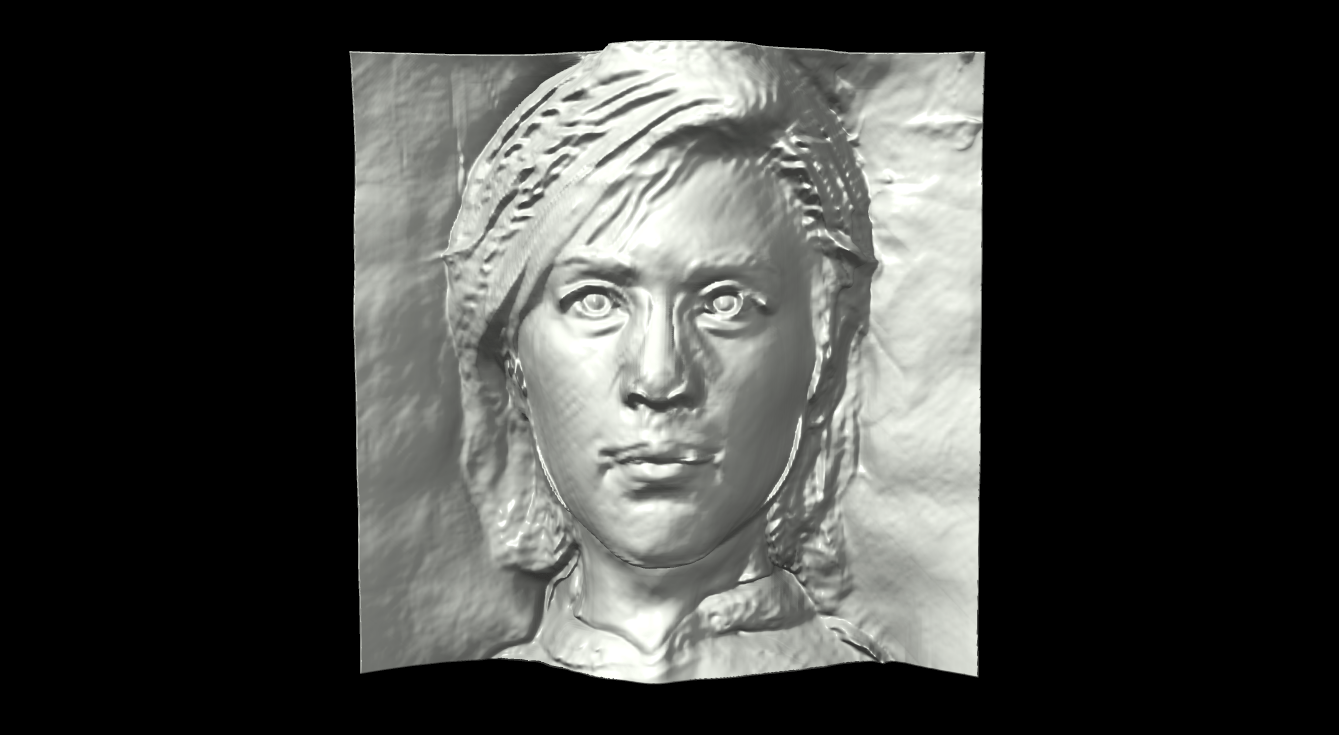}}&
\includegraphics[width=\linewidth]{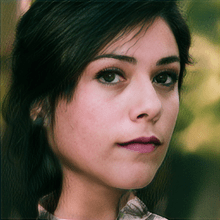}\hfill \\

\includegraphics[width=\linewidth]{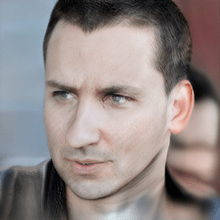}\hfill &
\includegraphics[width=\linewidth]{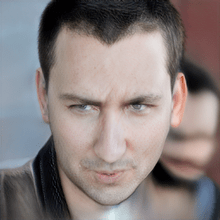}\hfill&
\includegraphics[width=\linewidth]{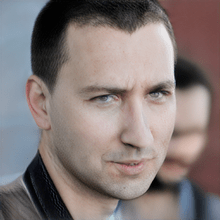}\hfill &

\includegraphics[width=\linewidth]{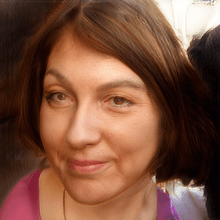}\hfill &
\includegraphics[width=\linewidth]{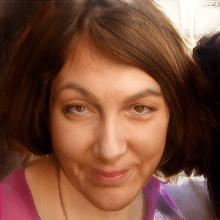}\hfill&
\includegraphics[width=\linewidth]{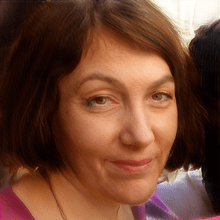}\hfill &

\includegraphics[width=\linewidth]{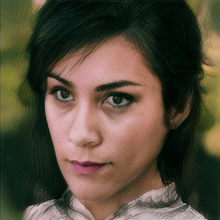}\hfill &
\includegraphics[width=\linewidth]{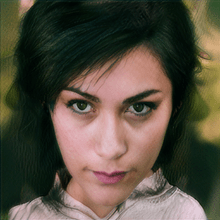}\hfill&
\includegraphics[width=\linewidth]{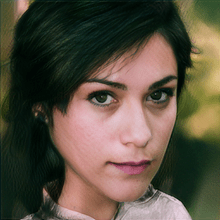}\hfill \\\\\\

& \includegraphics[width=\linewidth]{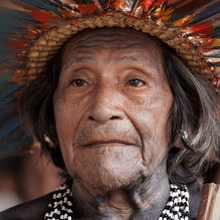}\hfill &
&& \includegraphics[width=\linewidth]{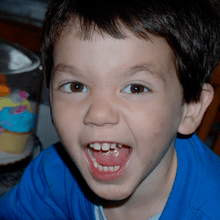}\hfill &
&& \includegraphics[width=\linewidth]{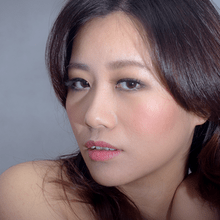}\hfill \\

\includegraphics[width=\linewidth]{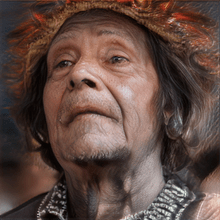}\hfill &
\includegraphics[width=\linewidth]{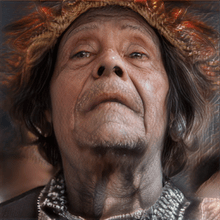}\hfill&
\includegraphics[width=\linewidth]{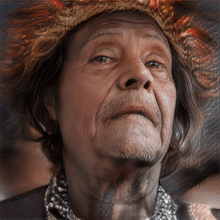}\hfill &

\includegraphics[width=\linewidth]{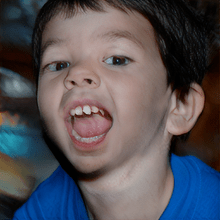}\hfill &
\includegraphics[width=\linewidth]{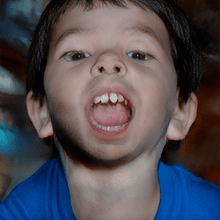}\hfill&
\includegraphics[width=\linewidth]{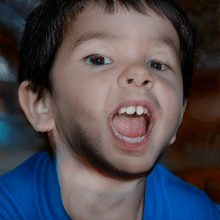}\hfill &

\includegraphics[width=\linewidth]{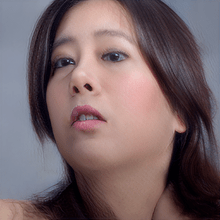}\hfill &
\includegraphics[width=\linewidth]{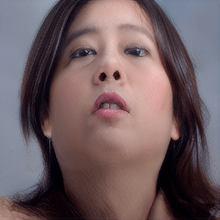}\hfill&
\includegraphics[width=\linewidth]{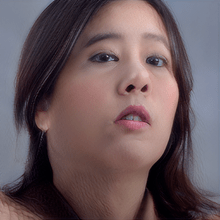}\hfill \\

\includegraphics[width=\linewidth]{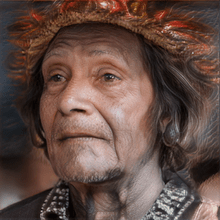}\hfill &
\includegraphics[width=\linewidth]{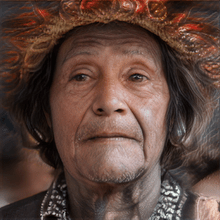}\hfill
\llap{\includegraphics[trim=650 100 380 110,clip,width=0.5\linewidth, height=\linewidth]{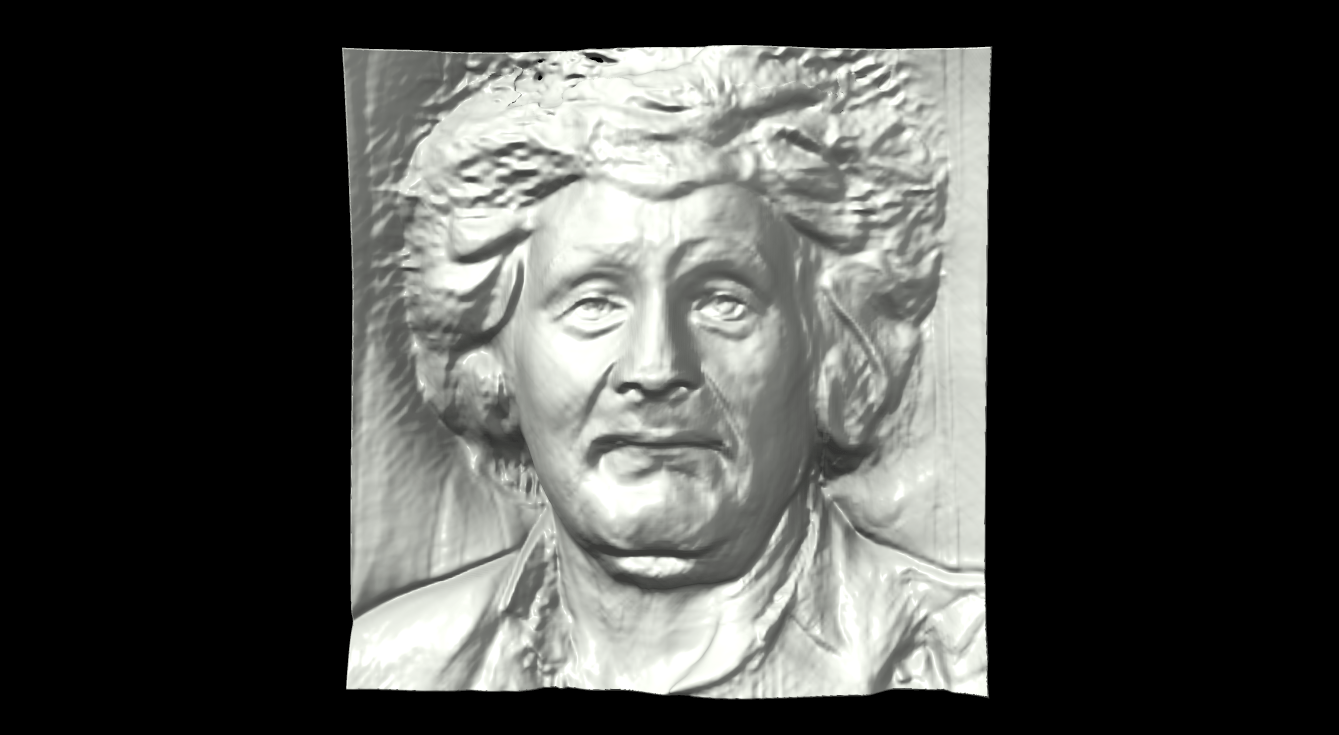}}&
\includegraphics[width=\linewidth]{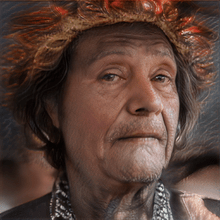}\hfill &

\includegraphics[width=\linewidth]{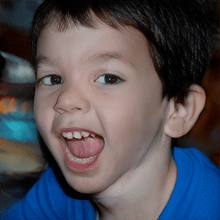}\hfill &
\includegraphics[width=\linewidth]{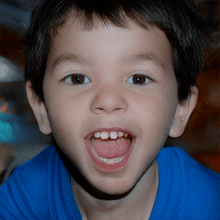}\hfill
\llap{\includegraphics[trim=650 100 380 130,clip,width=0.5\linewidth, height=\linewidth]{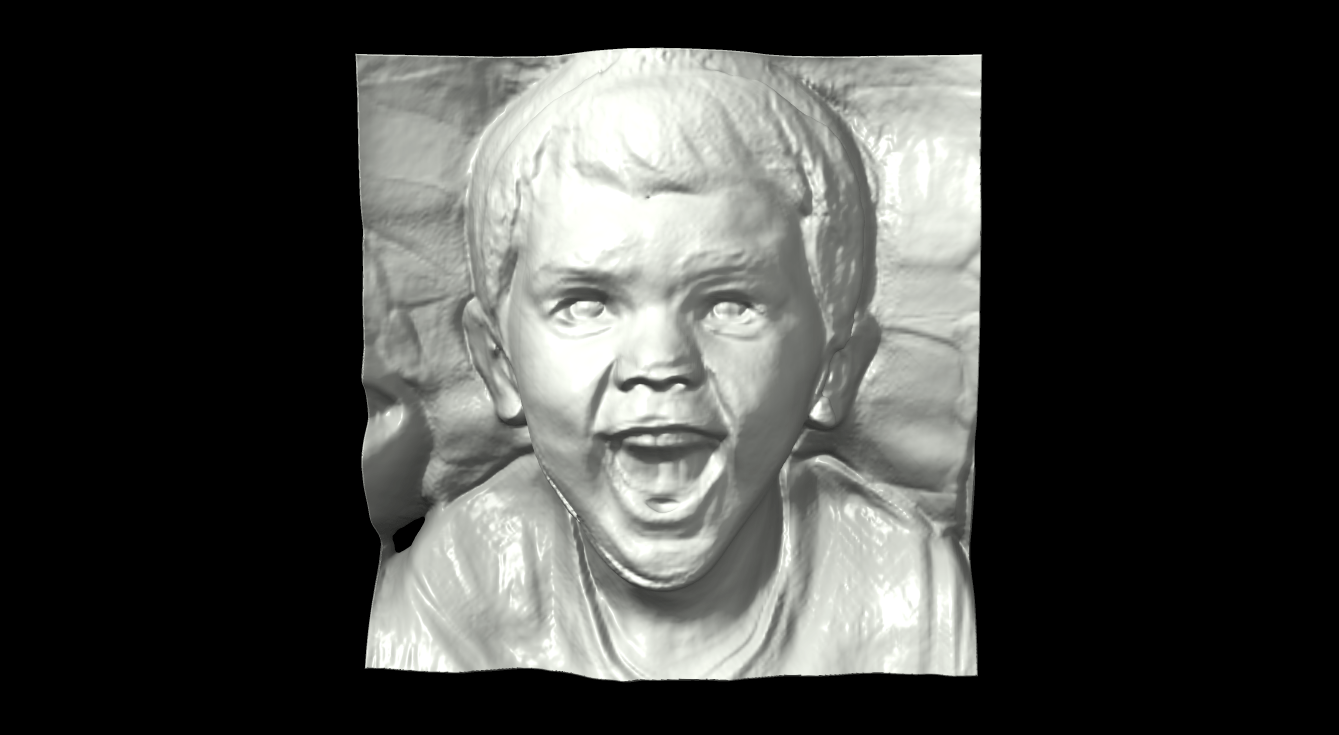}}&
\includegraphics[width=\linewidth]{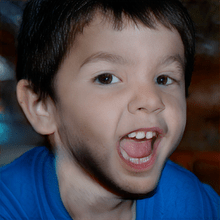}\hfill &

\includegraphics[width=\linewidth]{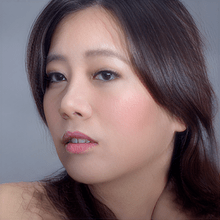}\hfill &
\includegraphics[width=\linewidth]{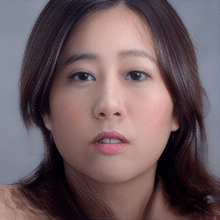}\hfill
\llap{\includegraphics[trim=650 90 380 100,clip,width=0.5\linewidth, height=\linewidth]{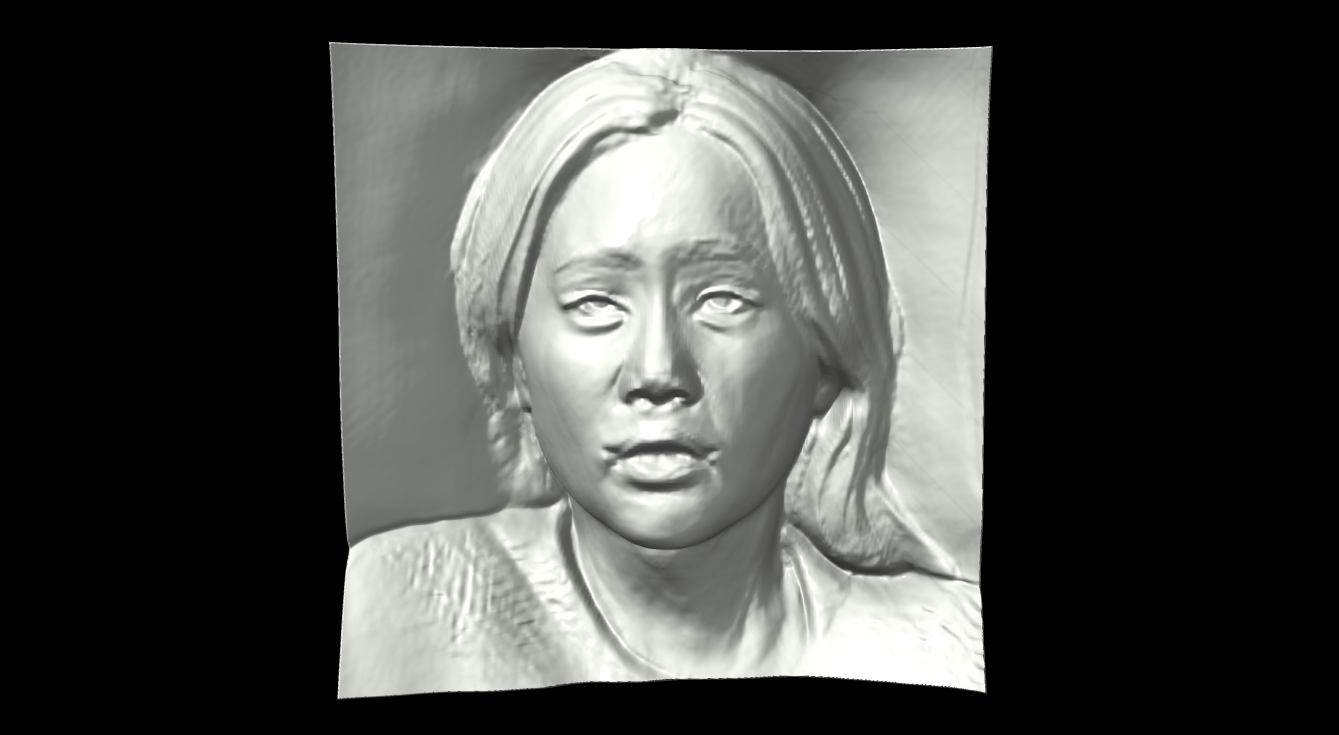}}&
\includegraphics[width=\linewidth]{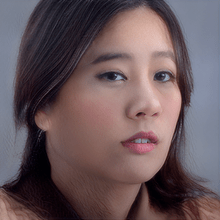}\hfill \\

\includegraphics[width=\linewidth]{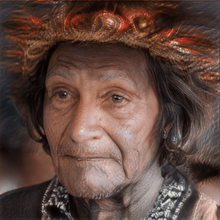}\hfill &
\includegraphics[width=\linewidth]{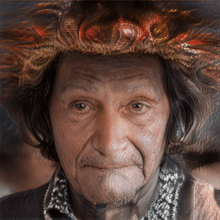}\hfill&
\includegraphics[width=\linewidth]{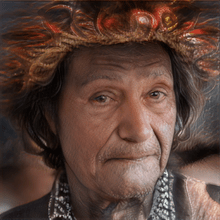}\hfill &

\includegraphics[width=\linewidth]{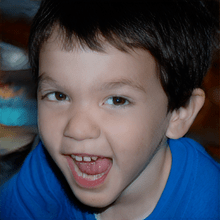}\hfill &
\includegraphics[width=\linewidth]{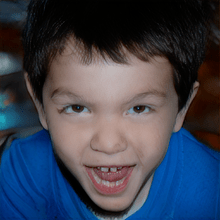}\hfill&
\includegraphics[width=\linewidth]{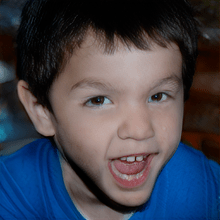}\hfill &

\includegraphics[width=\linewidth]{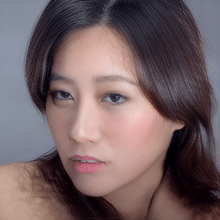}\hfill &
\includegraphics[width=\linewidth]{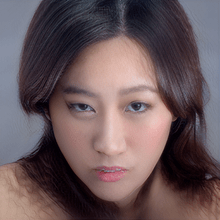}\hfill&
\includegraphics[width=\linewidth]{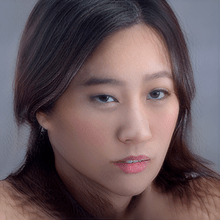}\hfill \\
\end{tabular}
\caption{\textbf{Additional qualitative results on in-domain dataset FFHQ~\cite{karras2019style}} }
%Simultaneous attribute editing and viewpoint shift comparison of 2D and 3D GANs. We compare editing results of applying attribute editing (smile) and viewpoint interpolation at the same time on the latent code acquired by PTI~\cite{roich2021pivotal} on StyleGAN2~\cite{Karras2019stylegan2} and the latent code acquired by our method on EG3D~\cite{Chan2022}}
\label{appendix:facial_inversion1}
\end{figure}

\begin{figure}[!p]
\centering
\newcolumntype{M}[1]{>{\centering\arraybackslash}m{#1}}
\setlength{\tabcolsep}{1pt}
\renewcommand{\arraystretch}{0.5}
\begin{tabular}{M{0.1\linewidth}M{0.1\linewidth}M{0.1\linewidth}
@{\hskip 0.01\linewidth} M{0.1\linewidth}M{0.1\linewidth}M{0.1\linewidth} @{\hskip 0.01\linewidth} M{0.1\linewidth}M{0.1\linewidth}M{0.1\linewidth}}

& \includegraphics[width=\linewidth]{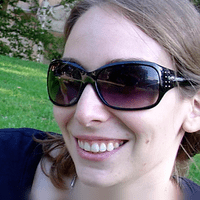}\hfill &
&& \includegraphics[width=\linewidth]{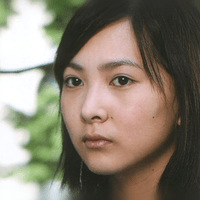}\hfill &
&& \includegraphics[width=\linewidth]{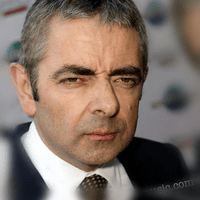}\hfill \\

\includegraphics[width=\linewidth]{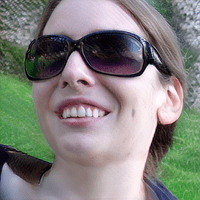}\hfill &
\includegraphics[width=\linewidth]{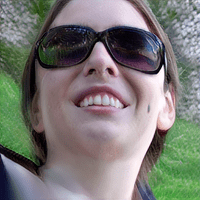}\hfill&
\includegraphics[width=\linewidth]{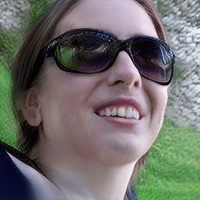}\hfill &

\includegraphics[width=\linewidth]{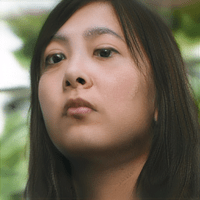}\hfill &
\includegraphics[width=\linewidth]{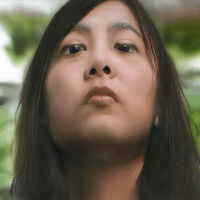}\hfill&
\includegraphics[width=\linewidth]{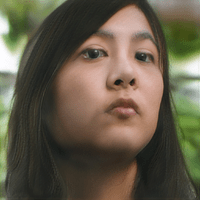}\hfill &

\includegraphics[width=\linewidth]{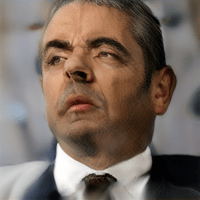}\hfill &
\includegraphics[width=\linewidth]{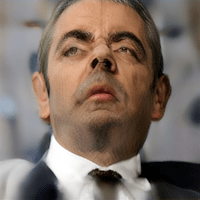}\hfill&
\includegraphics[width=\linewidth]{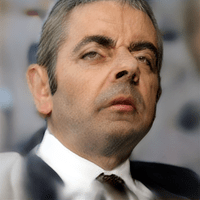}\hfill \\

\includegraphics[width=\linewidth]{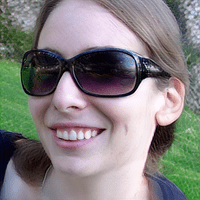}\hfill &
\includegraphics[width=\linewidth]{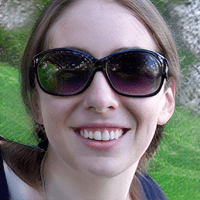}\hfill
\llap{\includegraphics[trim=650 90 380 120,clip,width=0.5\linewidth, height=\linewidth]{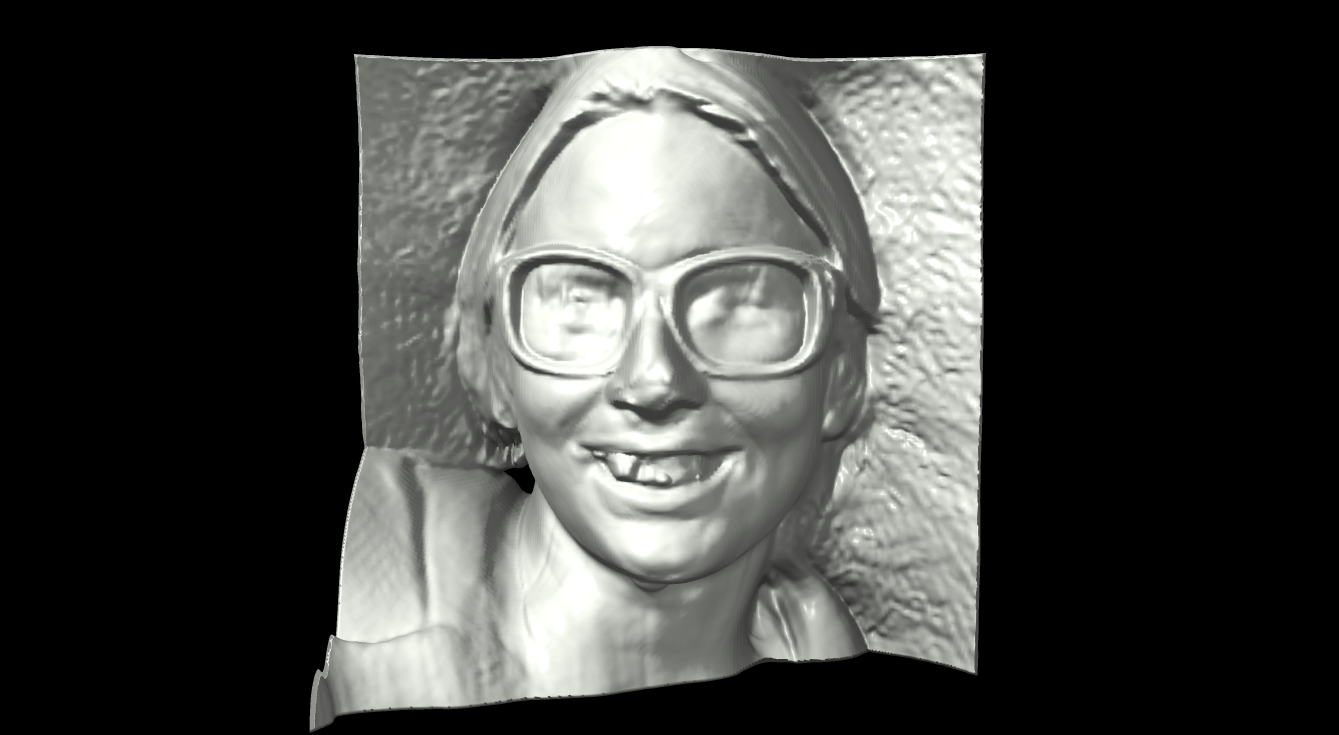}}&
\includegraphics[width=\linewidth]{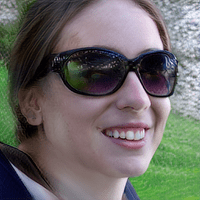}\hfill &

\includegraphics[width=\linewidth]{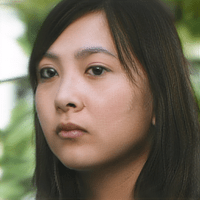}\hfill &
\includegraphics[width=\linewidth]{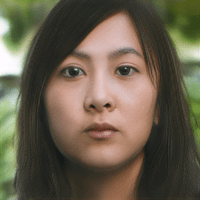}\hfill
\llap{\includegraphics[trim=650 90 380 120,clip,width=0.5\linewidth, height=\linewidth]{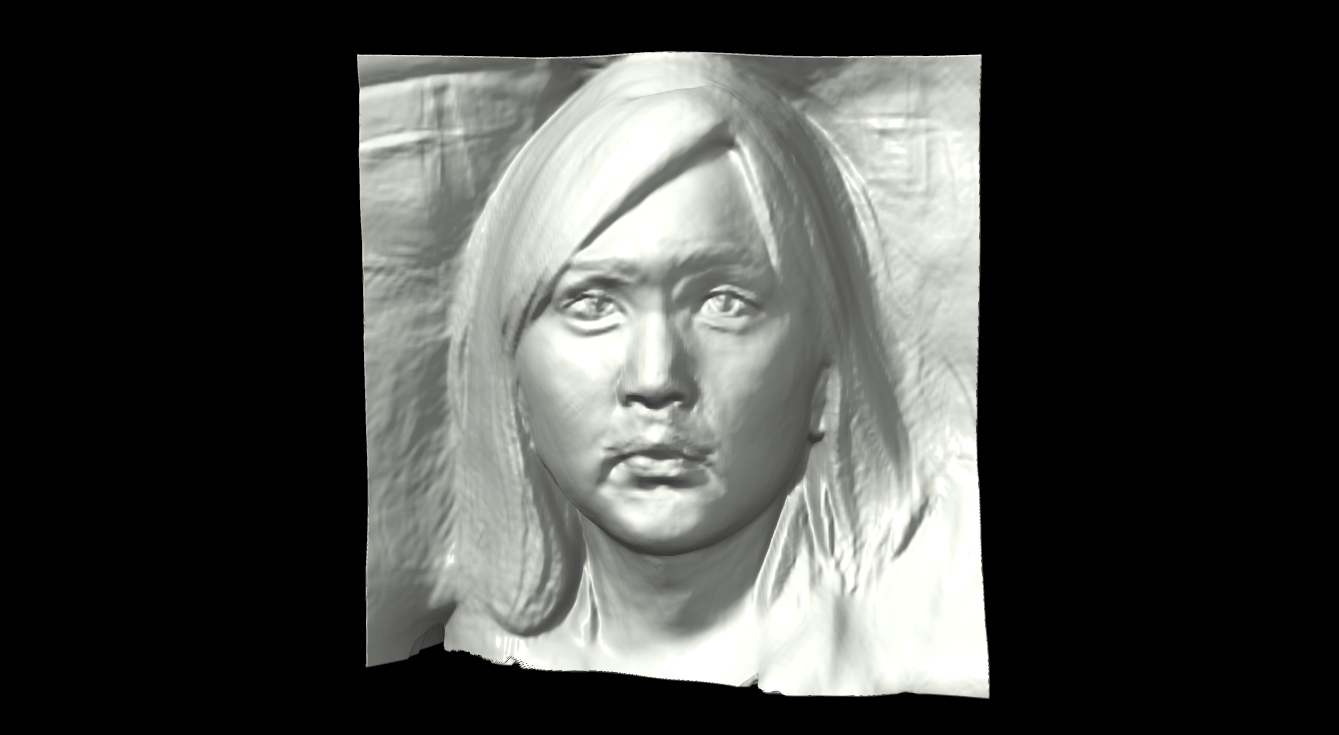}}&
\includegraphics[width=\linewidth]{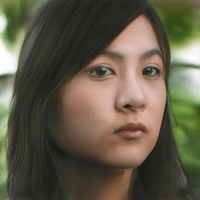}\hfill &

\includegraphics[width=\linewidth]{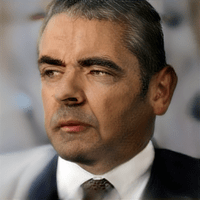}\hfill &
\includegraphics[width=\linewidth]{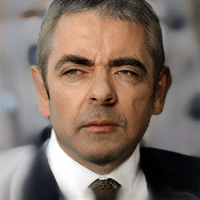}\hfill
\llap{\includegraphics[trim=650 90 380 120,clip,width=0.5\linewidth, height=\linewidth]{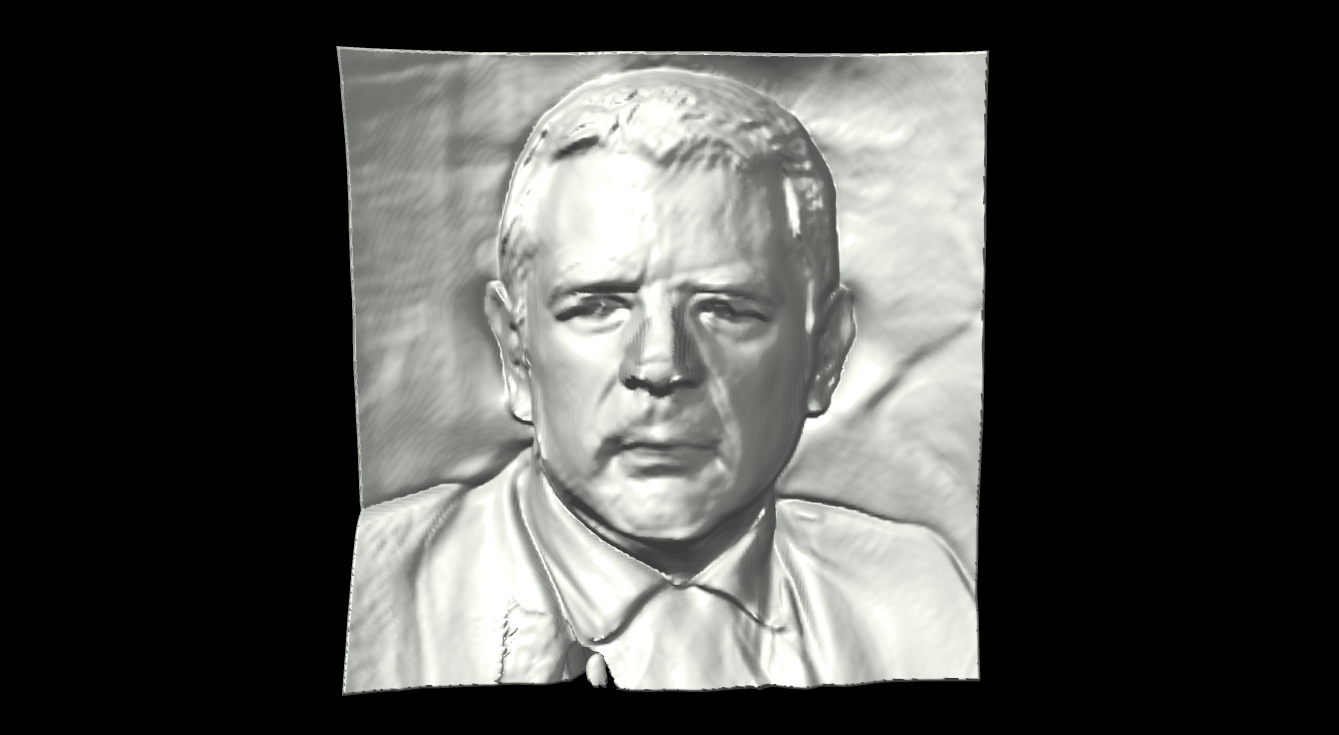}}&
\includegraphics[width=\linewidth]{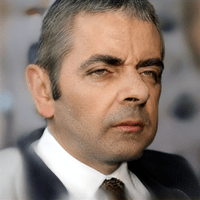}\hfill \\

\includegraphics[width=\linewidth]{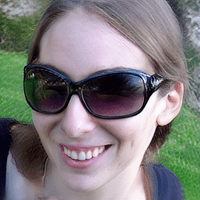}\hfill &
\includegraphics[width=\linewidth]{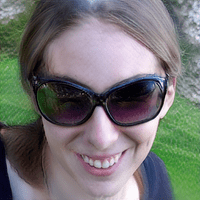}\hfill&
\includegraphics[width=\linewidth]{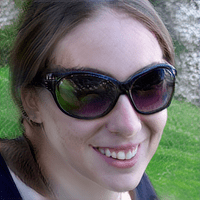}\hfill &

\includegraphics[width=\linewidth]{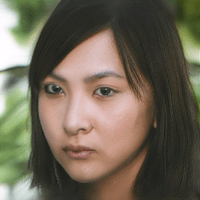}\hfill &
\includegraphics[width=\linewidth]{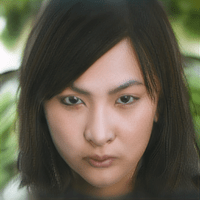}\hfill&
\includegraphics[width=\linewidth]{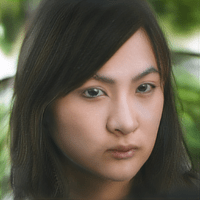}\hfill &

\includegraphics[width=\linewidth]{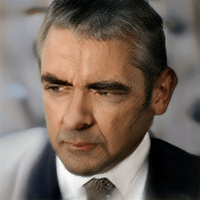}\hfill &
\includegraphics[width=\linewidth]{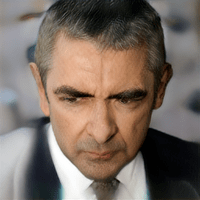}\hfill&
\includegraphics[width=\linewidth]{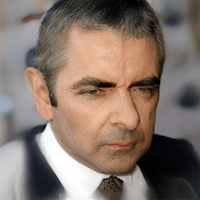}\hfill \\\\\\

& \includegraphics[width=\linewidth]{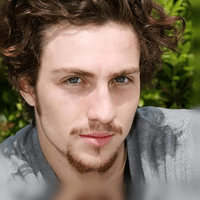}\hfill &
&& \includegraphics[width=\linewidth]{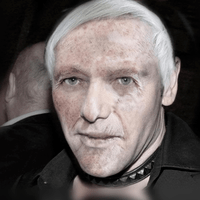}\hfill &
&& \includegraphics[width=\linewidth]{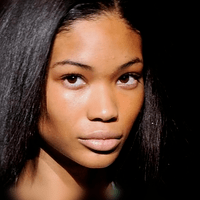}\hfill \\

\includegraphics[width=\linewidth]{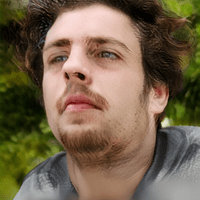}\hfill &
\includegraphics[width=\linewidth]{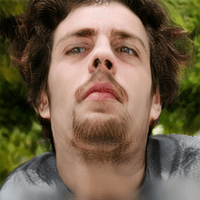}\hfill&
\includegraphics[width=\linewidth]{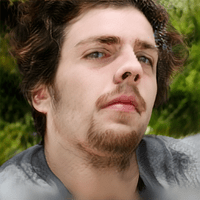}\hfill &

\includegraphics[width=\linewidth]{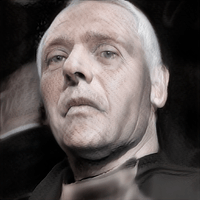}\hfill &
\includegraphics[width=\linewidth]{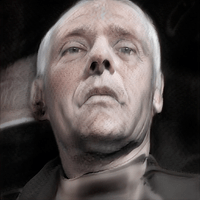}\hfill&
\includegraphics[width=\linewidth]{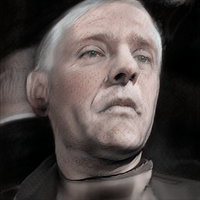}\hfill &

\includegraphics[width=\linewidth]{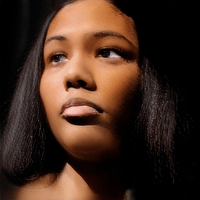}\hfill &
\includegraphics[width=\linewidth]{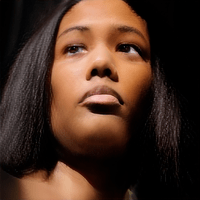}\hfill&
\includegraphics[width=\linewidth]{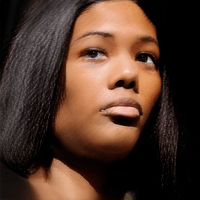}\hfill \\

\includegraphics[width=\linewidth]{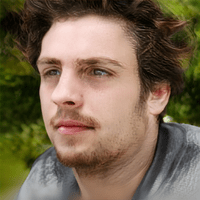}\hfill &
\includegraphics[width=\linewidth]{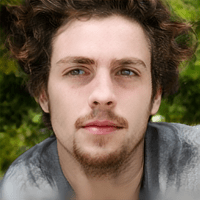}\hfill
\llap{\includegraphics[trim=650 100 380 110,clip,width=0.5\linewidth, height=\linewidth]{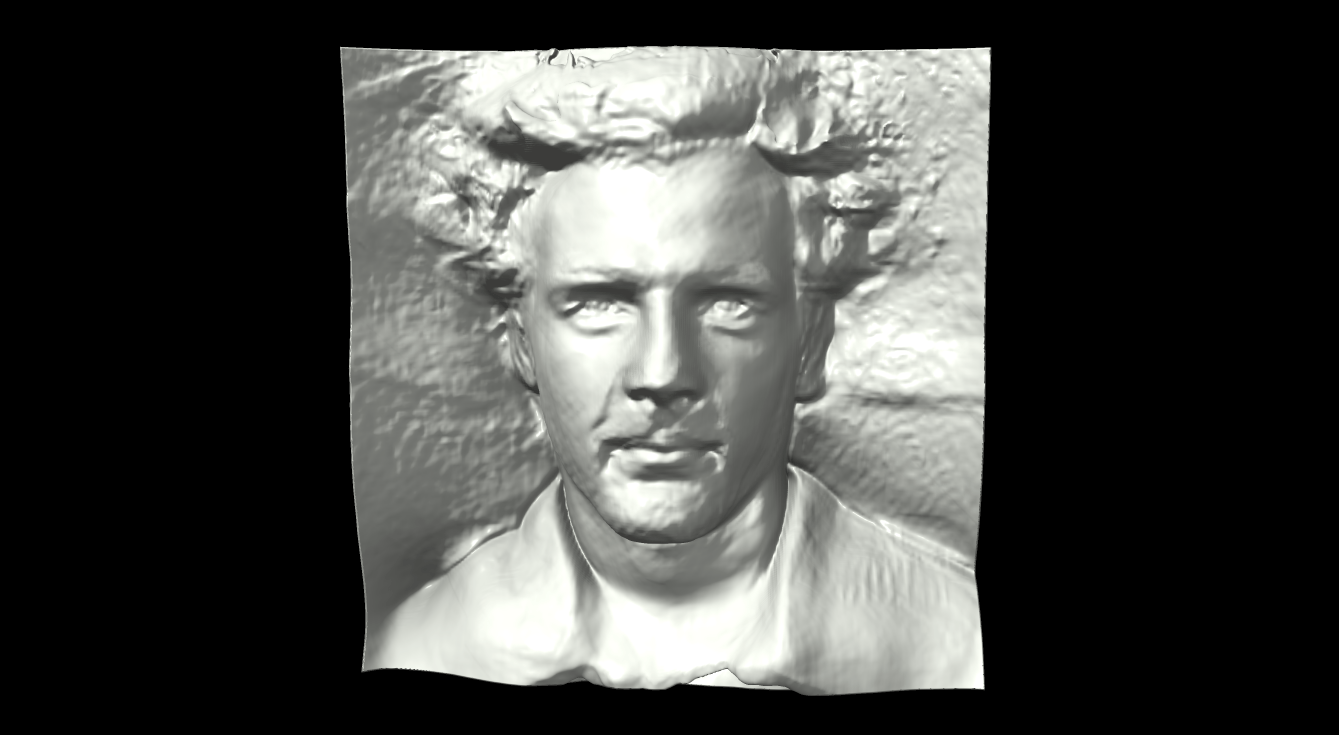}}&
\includegraphics[width=\linewidth]{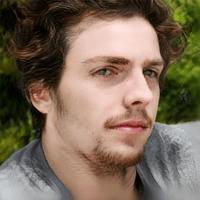}\hfill &

\includegraphics[width=\linewidth]{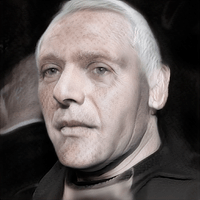}\hfill &
\includegraphics[width=\linewidth]{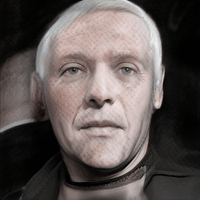}\hfill
\llap{\includegraphics[trim=650 90 380 120,clip,width=0.5\linewidth, height=\linewidth]{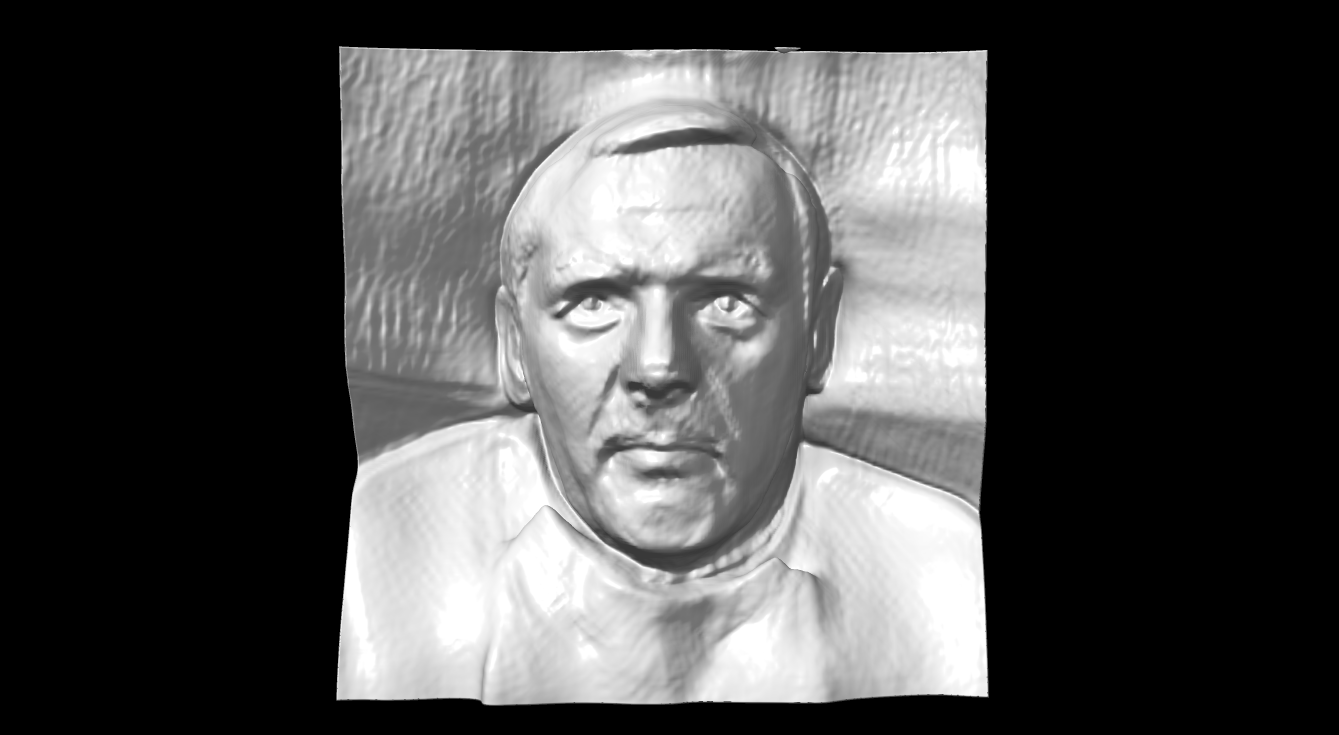}}&
\includegraphics[width=\linewidth]{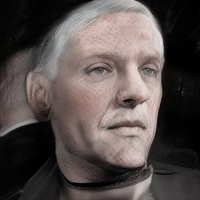}\hfill &

\includegraphics[width=\linewidth]{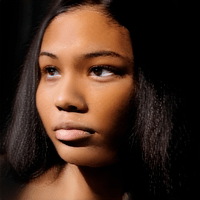}\hfill &
\includegraphics[width=\linewidth]{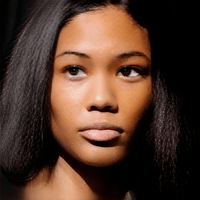}\hfill
\llap{\includegraphics[trim=650 100 380 110,clip,width=0.5\linewidth, height=\linewidth]{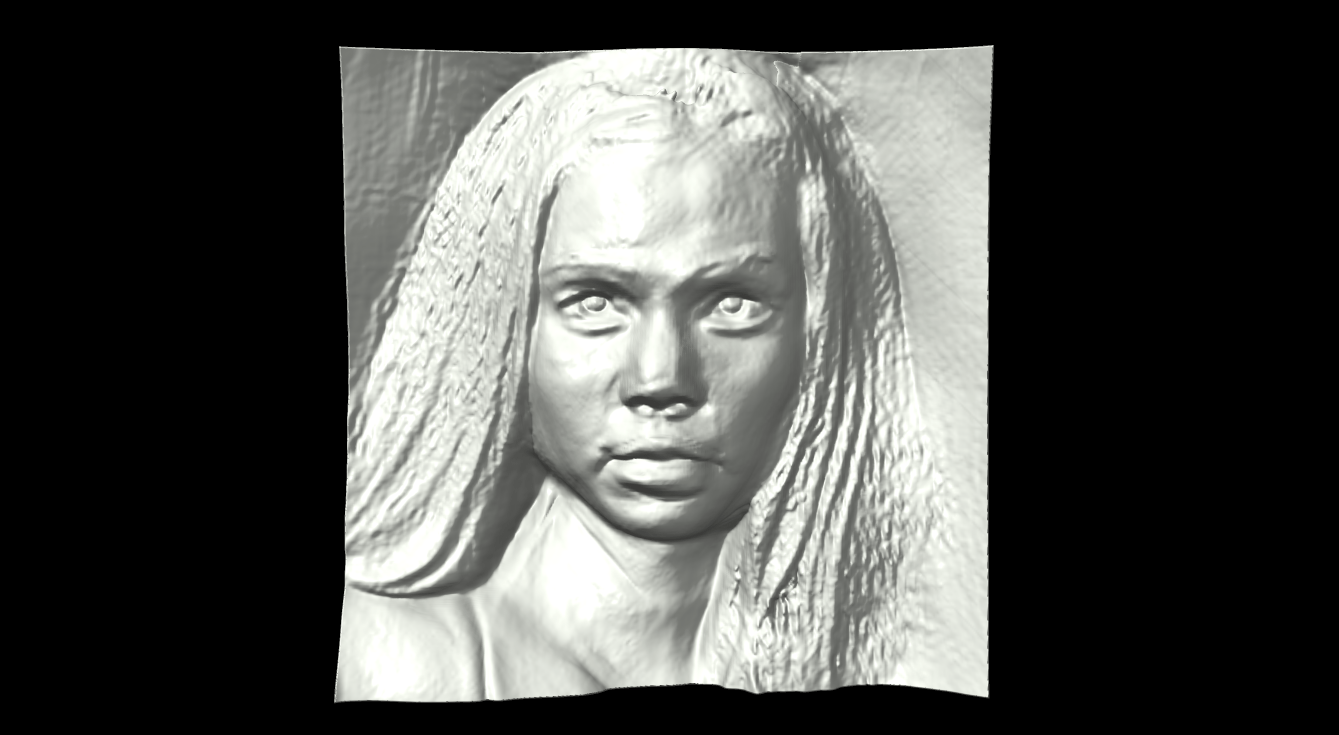}}&
\includegraphics[width=\linewidth]{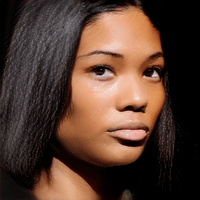}\hfill \\

\includegraphics[width=\linewidth]{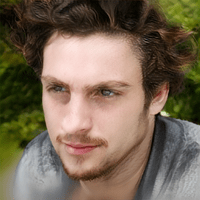}\hfill &
\includegraphics[width=\linewidth]{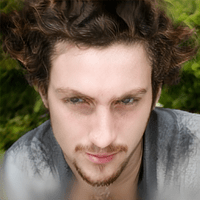}\hfill&
\includegraphics[width=\linewidth]{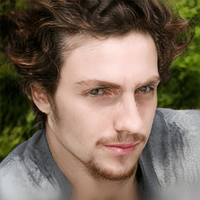}\hfill &

\includegraphics[width=\linewidth]{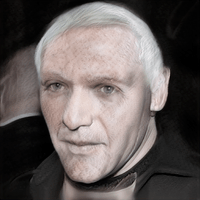}\hfill &
\includegraphics[width=\linewidth]{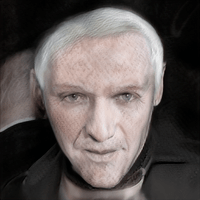}\hfill&
\includegraphics[width=\linewidth]{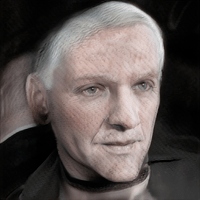}\hfill &

\includegraphics[width=\linewidth]{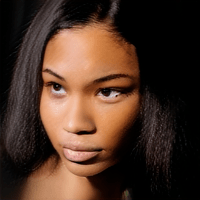}\hfill &
\includegraphics[width=\linewidth]{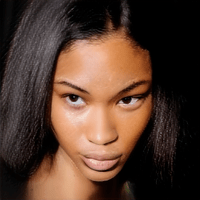}\hfill&
\includegraphics[width=\linewidth]{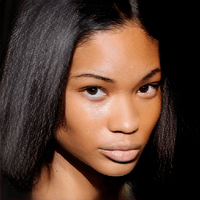}\hfill \\
\end{tabular}
\caption{\textbf{Additional qualitative results on out-of-domain dataset CelebA-HQ~\cite{karras2017progressive, liu2015faceattributes}} }
%Simultaneous attribute editing and viewpoint shift comparison of 2D and 3D GANs. We compare editing results of applying attribute editing (smile) and viewpoint interpolation at the same time on the latent code acquired by PTI~\cite{roich2021pivotal} on StyleGAN2~\cite{Karras2019stylegan2} and the latent code acquired by our method on EG3D~\cite{Chan2022}}
\label{appendix:facial_inversion2}
\end{figure}

\begin{figure}[!p]
\centering
\newcolumntype{M}[1]{>{\centering\arraybackslash}m{#1}}
\setlength{\tabcolsep}{1pt}
\renewcommand{\arraystretch}{0.5}
\begin{tabular}{M{0.1\linewidth}M{0.1\linewidth}M{0.1\linewidth}
@{\hskip 0.01\linewidth} M{0.1\linewidth}M{0.1\linewidth}M{0.1\linewidth} @{\hskip 0.01\linewidth} M{0.1\linewidth}M{0.1\linewidth}M{0.1\linewidth}}

& \includegraphics[width=\linewidth]{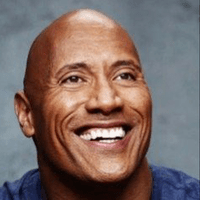}\hfill &
&& \includegraphics[width=\linewidth]{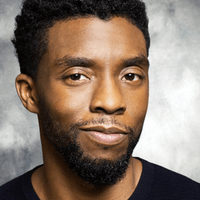}\hfill &
&& \includegraphics[width=\linewidth]{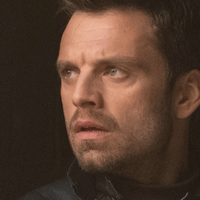}\hfill \\

\includegraphics[width=\linewidth]{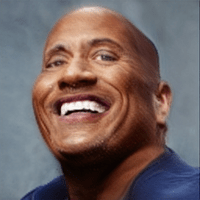}\hfill &
\includegraphics[width=\linewidth]{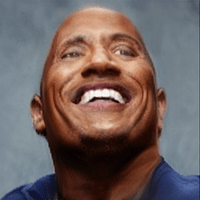}\hfill&
\includegraphics[width=\linewidth]{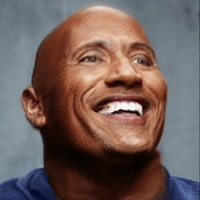}\hfill &

\includegraphics[width=\linewidth]{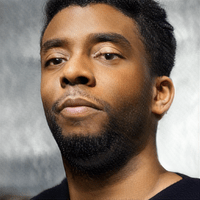}\hfill &
\includegraphics[width=\linewidth]{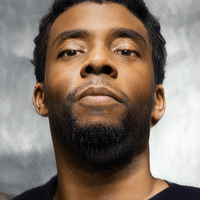}\hfill&
\includegraphics[width=\linewidth]{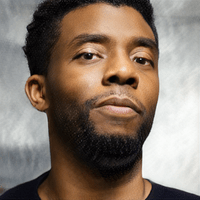}\hfill &

\includegraphics[width=\linewidth]{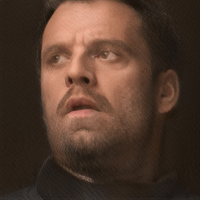}\hfill &
\includegraphics[width=\linewidth]{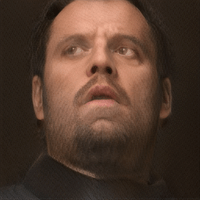}\hfill&
\includegraphics[width=\linewidth]{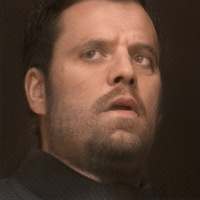}\hfill \\

\includegraphics[width=\linewidth]{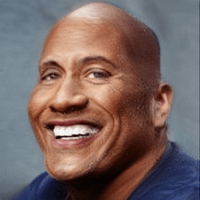}\hfill &
\includegraphics[width=\linewidth]{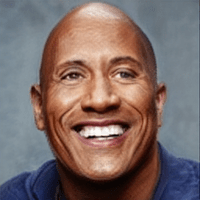}\hfill
\llap{\includegraphics[trim=650 90 380 120,clip,width=0.5\linewidth, height=\linewidth]{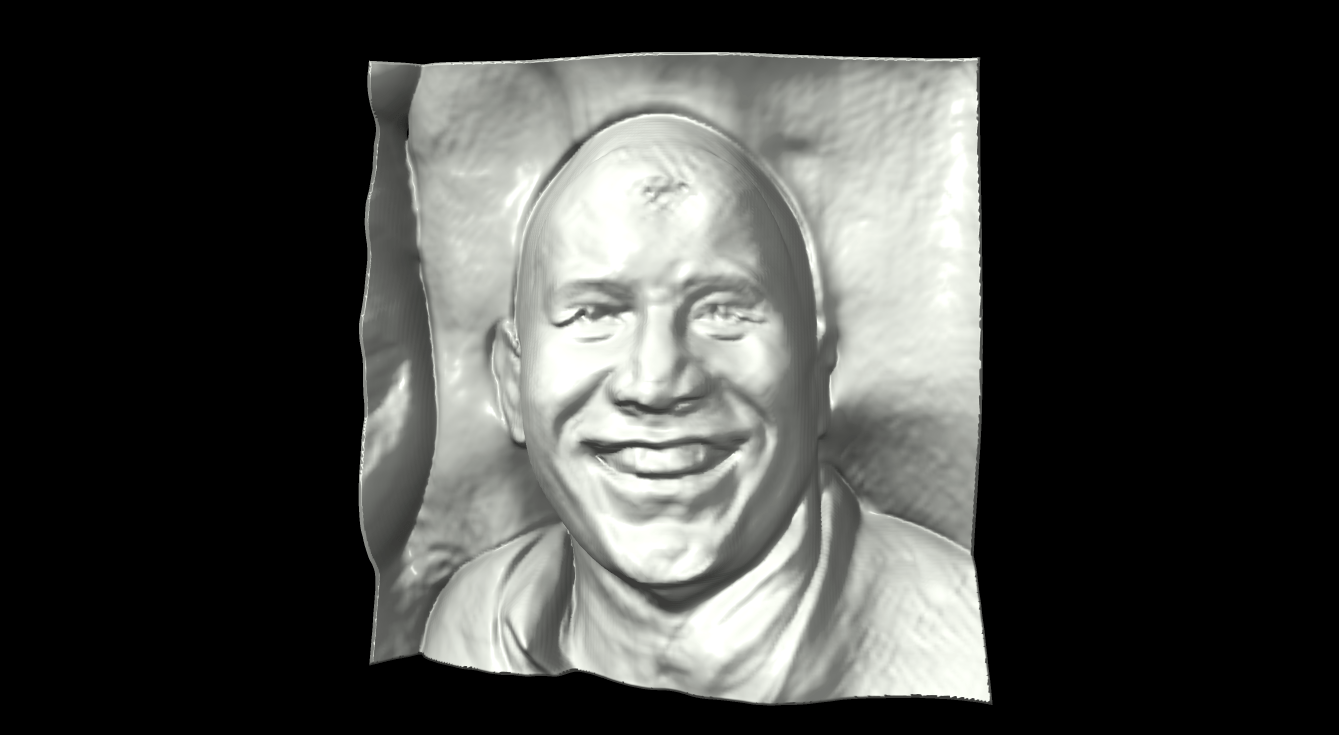}}&
\includegraphics[width=\linewidth]{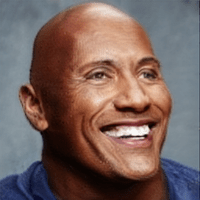}\hfill &

\includegraphics[width=\linewidth]{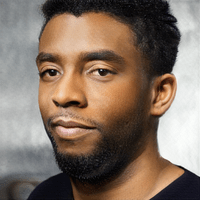}\hfill &
\includegraphics[width=\linewidth]{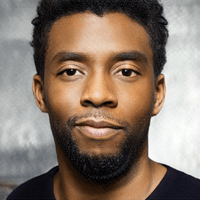}\hfill
\llap{\includegraphics[trim=650 90 380 120,clip,width=0.5\linewidth, height=\linewidth]{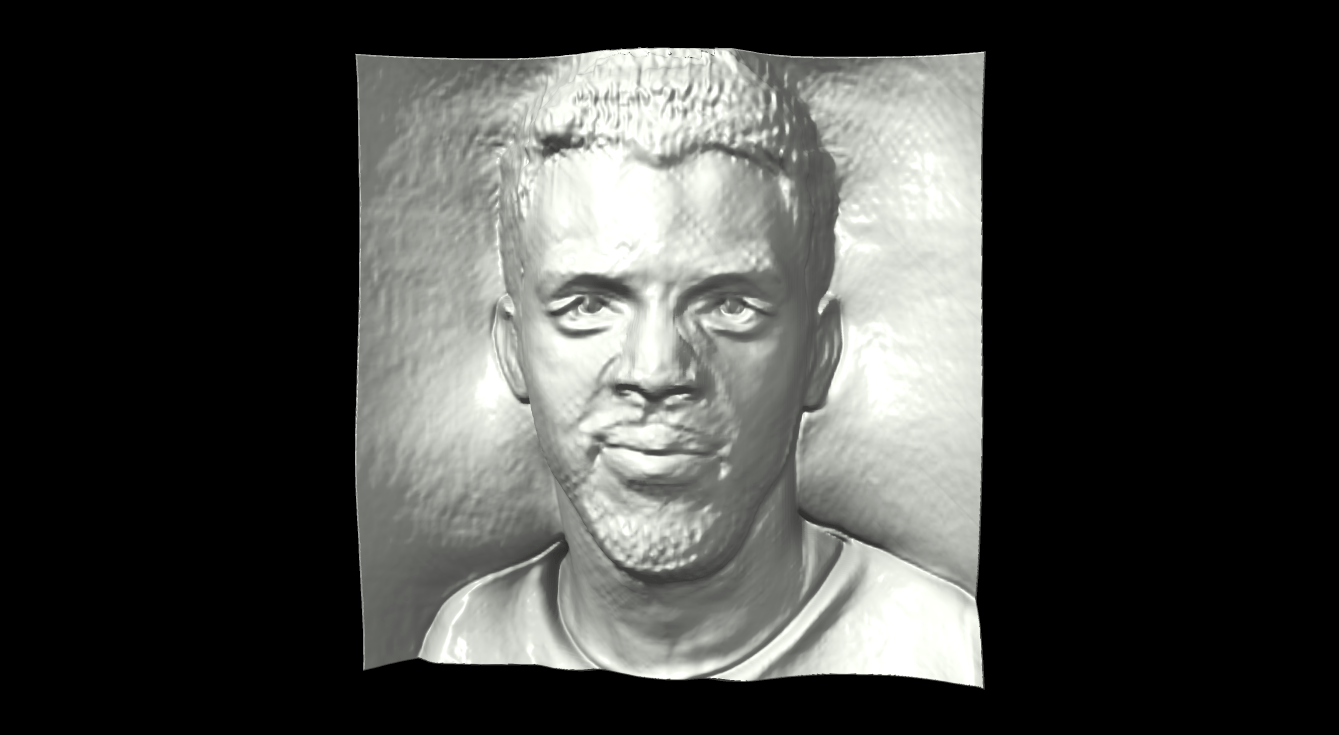}}&
\includegraphics[width=\linewidth]{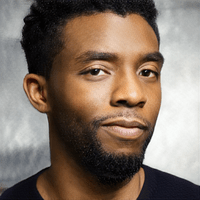}\hfill &

\includegraphics[width=\linewidth]{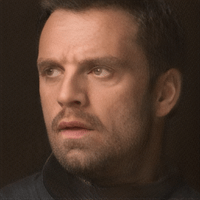}\hfill &
\includegraphics[width=\linewidth]{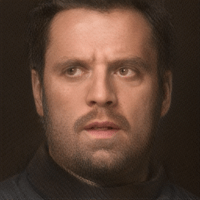}\hfill
\llap{\includegraphics[trim=650 90 380 120,clip,width=0.5\linewidth, height=\linewidth]{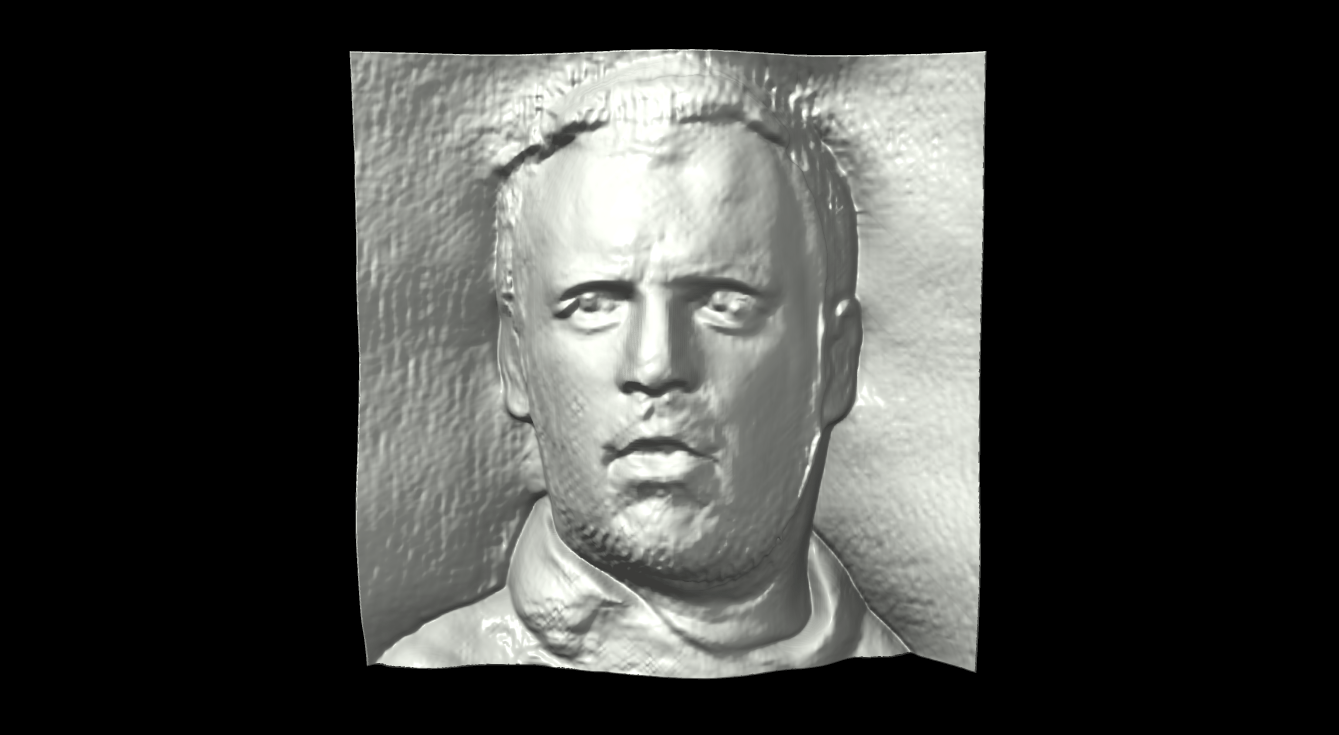}}&
\includegraphics[width=\linewidth]{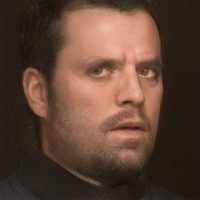}\hfill \\

\includegraphics[width=\linewidth]{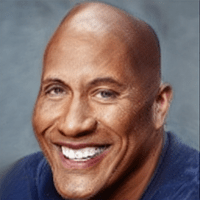}\hfill &
\includegraphics[width=\linewidth]{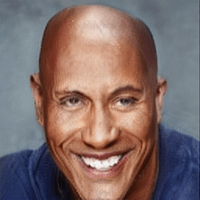}\hfill&
\includegraphics[width=\linewidth]{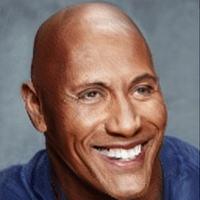}\hfill &

\includegraphics[width=\linewidth]{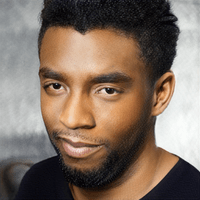}\hfill &
\includegraphics[width=\linewidth]{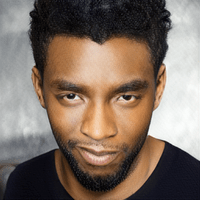}\hfill&
\includegraphics[width=\linewidth]{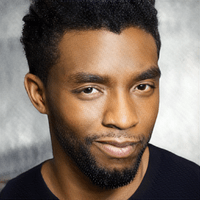}\hfill &

\includegraphics[width=\linewidth]{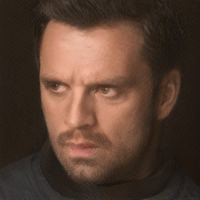}\hfill &
\includegraphics[width=\linewidth]{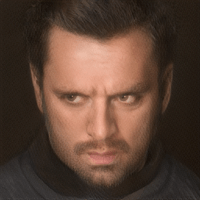}\hfill&
\includegraphics[width=\linewidth]{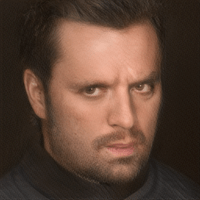}\hfill \\\\\\

& \includegraphics[width=\linewidth]{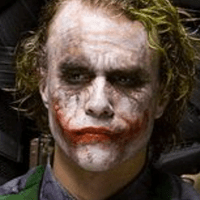}\hfill &
&& \includegraphics[width=\linewidth]{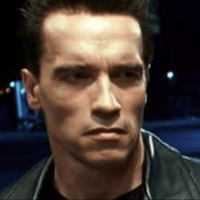}\hfill &
&& \includegraphics[width=\linewidth]{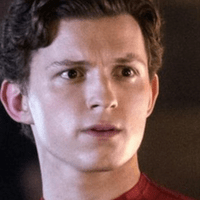}\hfill \\

\includegraphics[width=\linewidth]{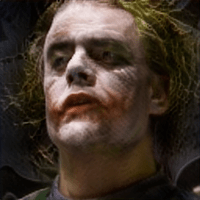}\hfill &
\includegraphics[width=\linewidth]{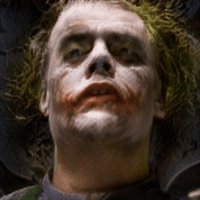}\hfill&
\includegraphics[width=\linewidth]{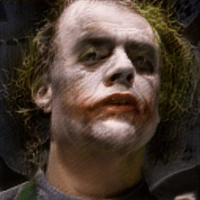}\hfill &

\includegraphics[width=\linewidth]{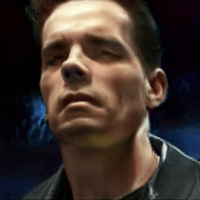}\hfill &
\includegraphics[width=\linewidth]{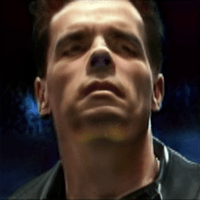}\hfill&
\includegraphics[width=\linewidth]{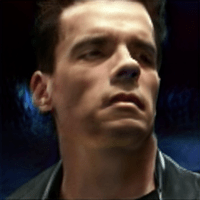}\hfill &

\includegraphics[width=\linewidth]{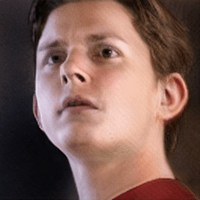}\hfill &
\includegraphics[width=\linewidth]{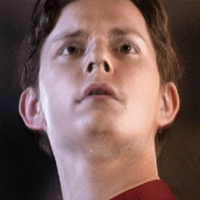}\hfill&
\includegraphics[width=\linewidth]{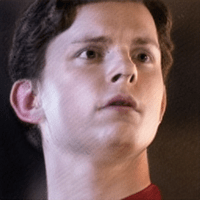}\hfill \\

\includegraphics[width=\linewidth]{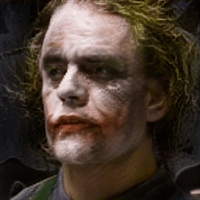}\hfill &
\includegraphics[width=\linewidth]{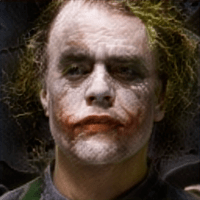}\hfill
\llap{\includegraphics[trim=650 90 380 120,clip,width=0.5\linewidth, height=\linewidth]{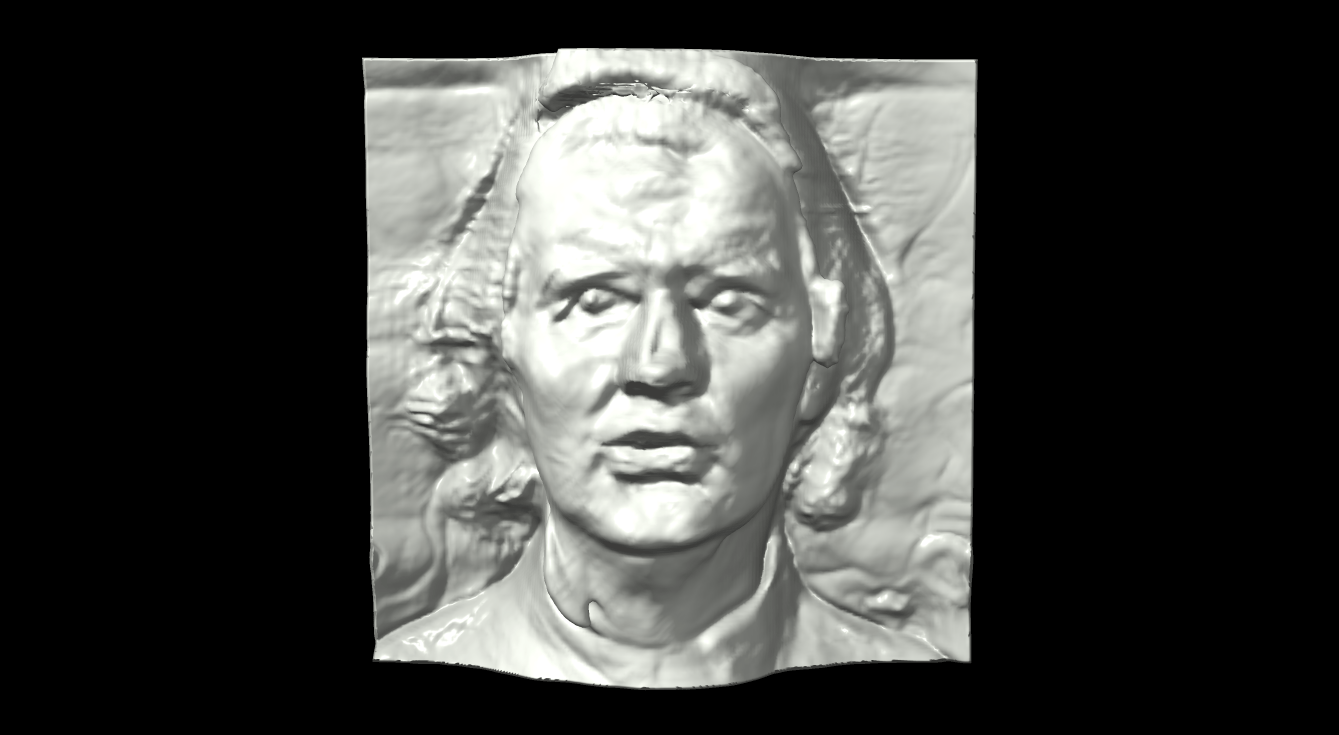}}&
\includegraphics[width=\linewidth]{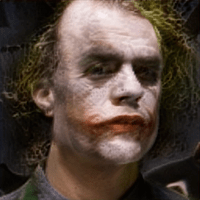}\hfill &

\includegraphics[width=\linewidth]{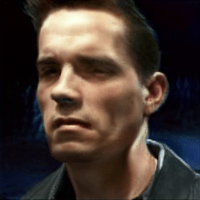}\hfill &
\includegraphics[width=\linewidth]{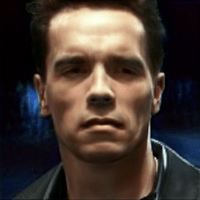}\hfill
\llap{\includegraphics[trim=650 90 380 120,clip,width=0.5\linewidth, height=\linewidth]{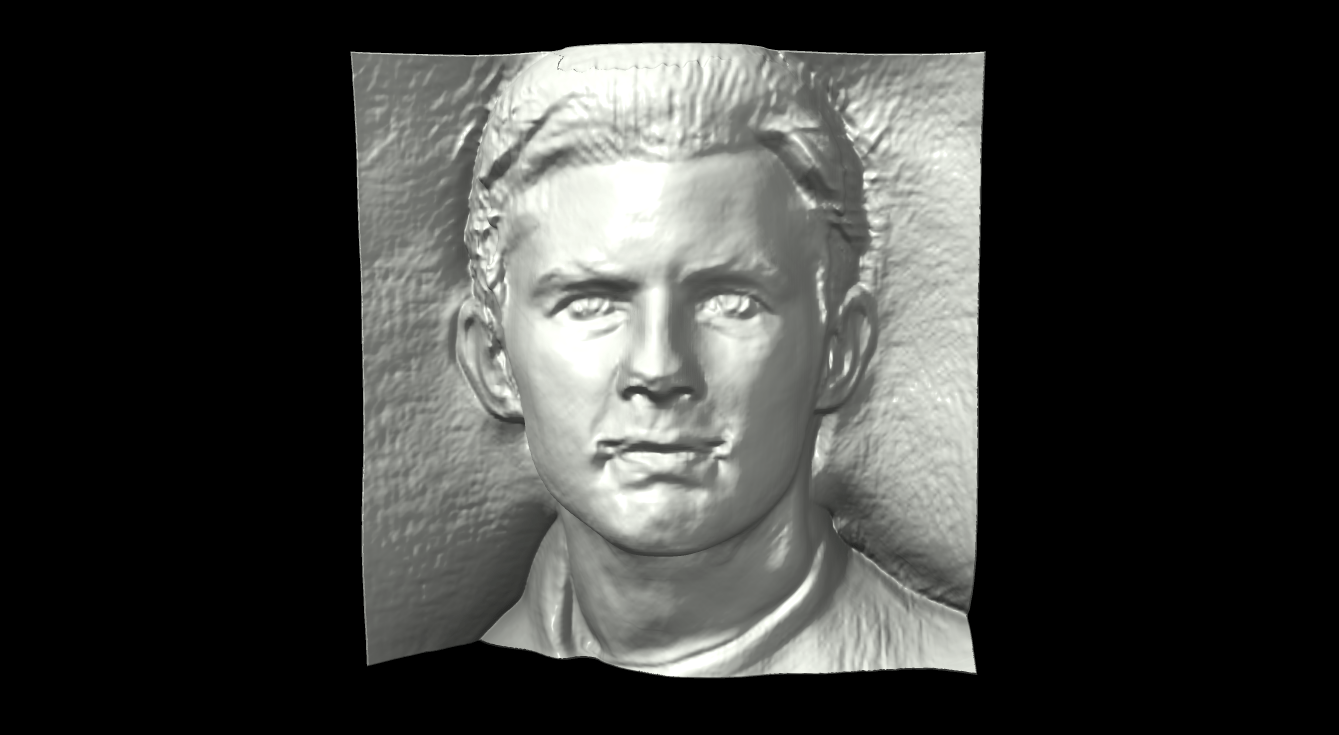}}&
\includegraphics[width=\linewidth]{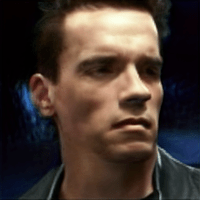}\hfill &

\includegraphics[width=\linewidth]{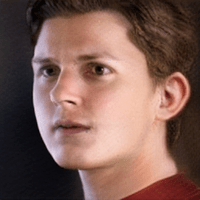}\hfill &
\includegraphics[width=\linewidth]{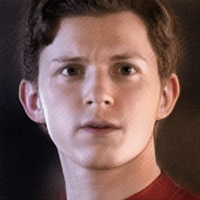}\hfill
\llap{\includegraphics[trim=660 100 380 110,clip,width=0.5\linewidth, height=\linewidth]{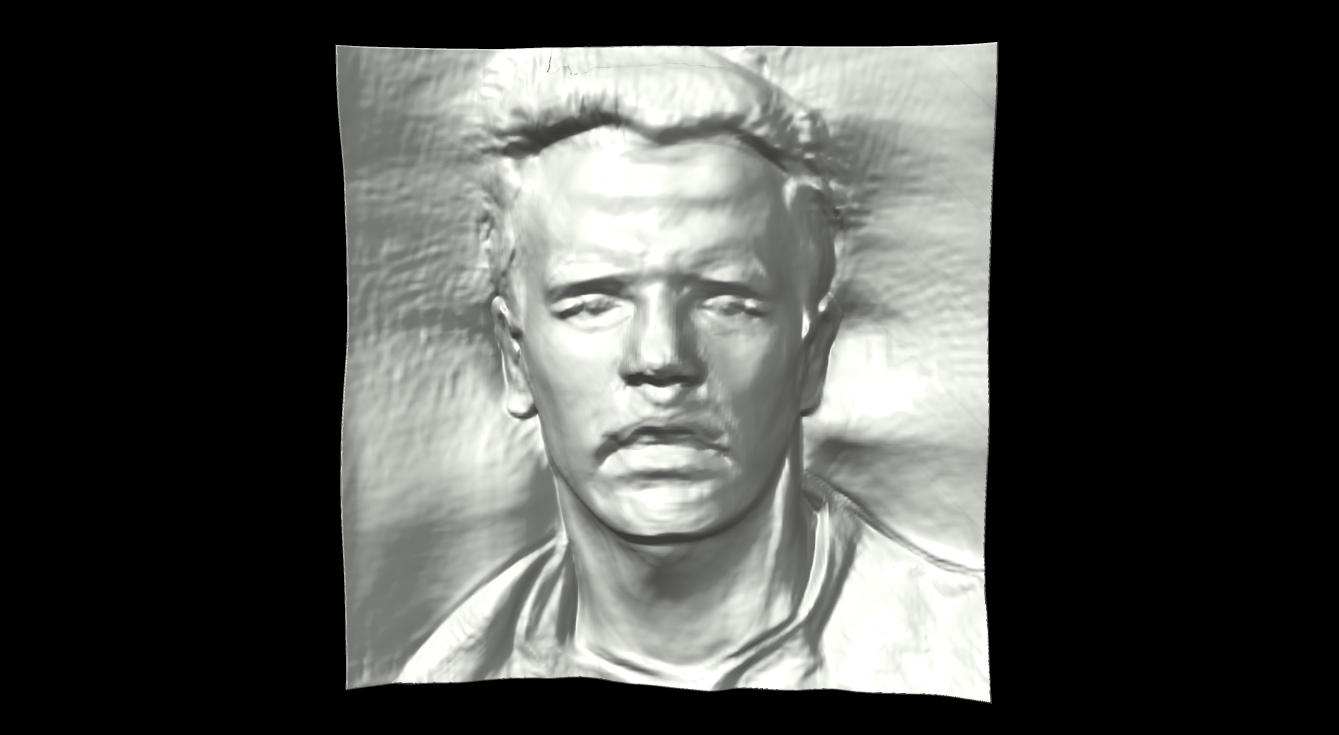}}&
\includegraphics[width=\linewidth]{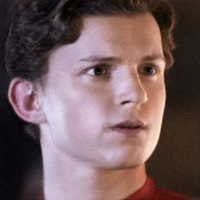}\hfill \\

\includegraphics[width=\linewidth]{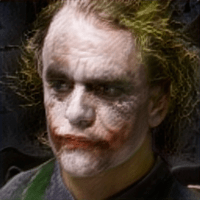}\hfill &
\includegraphics[width=\linewidth]{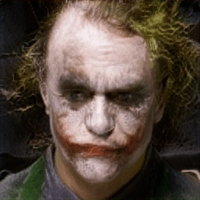}\hfill&
\includegraphics[width=\linewidth]{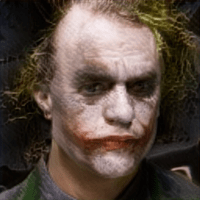}\hfill &

\includegraphics[width=\linewidth]{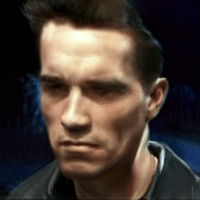}\hfill &
\includegraphics[width=\linewidth]{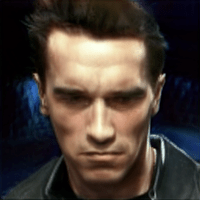}\hfill&
\includegraphics[width=\linewidth]{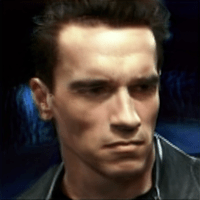}\hfill &

\includegraphics[width=\linewidth]{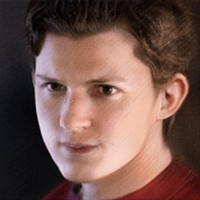}\hfill &
\includegraphics[width=\linewidth]{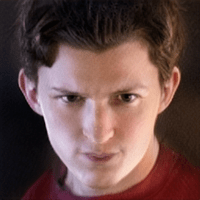}\hfill&
\includegraphics[width=\linewidth]{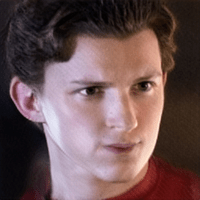}\hfill \\
\end{tabular}
\caption{\textbf{Movie Scene Novel view synthesis with 3D GAN Inversion.} We crop facial images from numerous famous movies and invert them into the latent space of EG3D~\cite{Chan2022}. We demonstrate novel view synthesis of these facial images along with visualization of 3D reconstructed mesh.}
% \caption{\textbf{Marvel Cinematic Universe EG3D Inversion.} We invert multiple images of the famous characters from movies of Marvel Studios into the EG3D latent space, and demonstrate novel view synthesis along with visualization of 3D reconstructed mesh.}
%Simultaneous attribute editing and viewpoint shift comparison of 2D and 3D GANs. We compare editing results of applying attribute editing (smile) and viewpoint interpolation at the same time on the latent code acquired by PTI~\cite{roich2021pivotal} on StyleGAN2~\cite{Karras2019stylegan2} and the latent code acquired by our method on EG3D~\cite{Chan2022}}
\label{appendix:facial_inversion3}
\end{figure}

\begin{figure}[!p]
\centering
\newcolumntype{M}[1]{>{\centering\arraybackslash}m{#1}}
\setlength{\tabcolsep}{1pt}
\renewcommand{\arraystretch}{0.5}
\begin{tabular}{M{0.1\linewidth}M{0.1\linewidth}M{0.1\linewidth}
@{\hskip 0.01\linewidth} M{0.1\linewidth}M{0.1\linewidth}M{0.1\linewidth} @{\hskip 0.01\linewidth} M{0.1\linewidth}M{0.1\linewidth}M{0.1\linewidth}}

& \includegraphics[width=\linewidth]{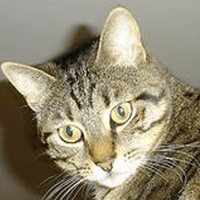}\hfill &
&& \includegraphics[width=\linewidth]{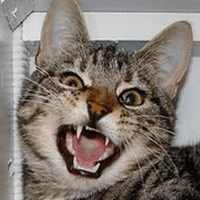}\hfill &
&& \includegraphics[width=\linewidth]{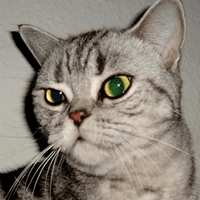}\hfill \\

\includegraphics[width=\linewidth]{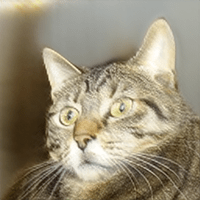}\hfill &
\includegraphics[width=\linewidth]{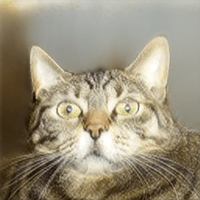}\hfill&
\includegraphics[width=\linewidth]{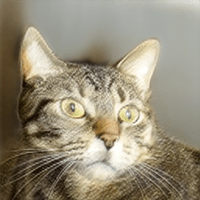}\hfill &

\includegraphics[width=\linewidth]{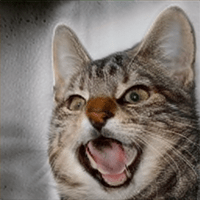}\hfill &
\includegraphics[width=\linewidth]{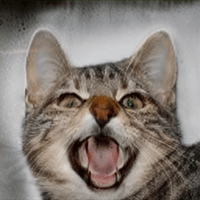}\hfill&
\includegraphics[width=\linewidth]{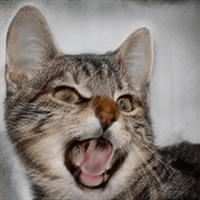}\hfill &

\includegraphics[width=\linewidth]{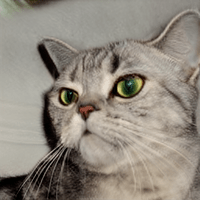}\hfill &
\includegraphics[width=\linewidth]{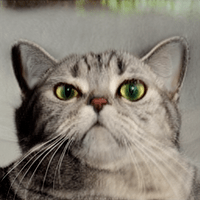}\hfill&
\includegraphics[width=\linewidth]{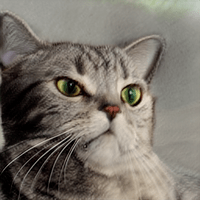}\hfill \\

\includegraphics[width=\linewidth]{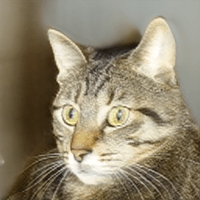}\hfill &
\includegraphics[width=\linewidth]{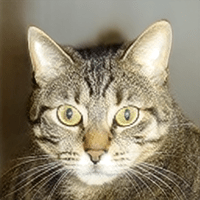}\hfill
\llap{\includegraphics[trim=650 100 400 60,clip,width=0.5\linewidth, height=\linewidth]{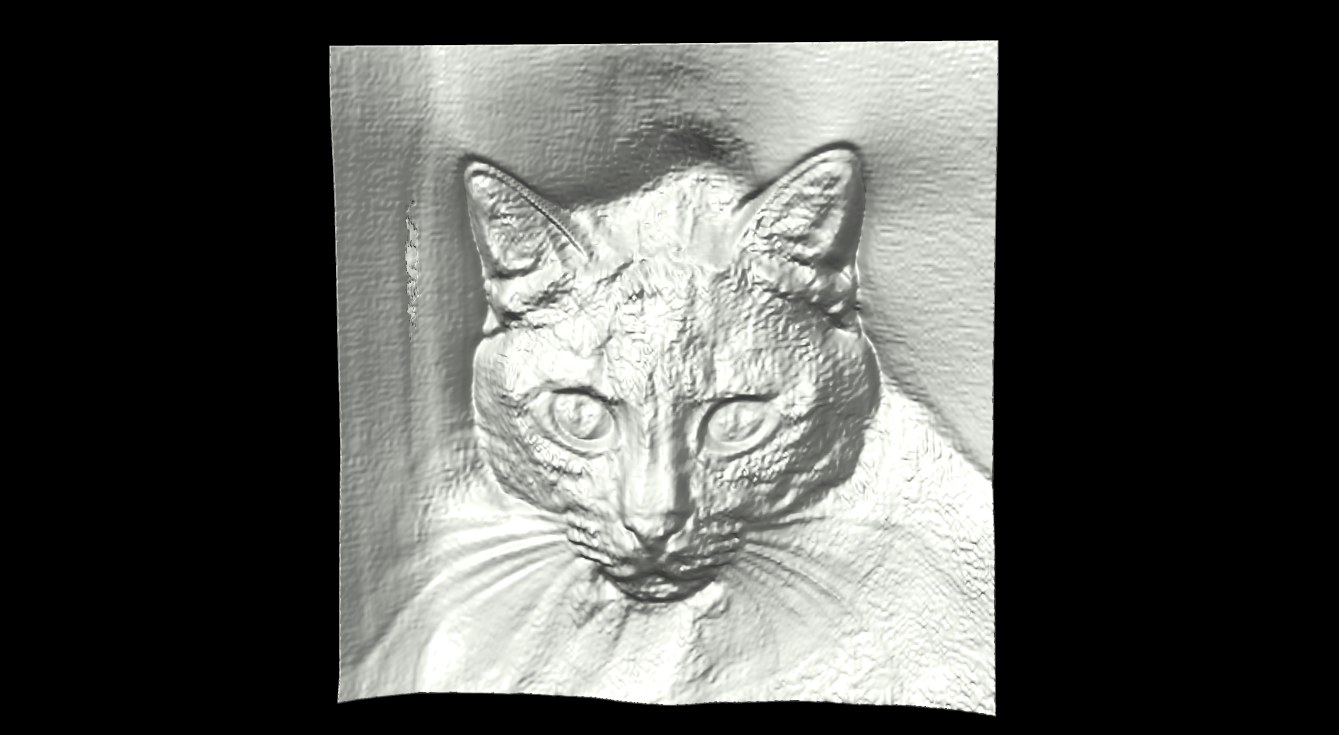}}&
\includegraphics[width=\linewidth]{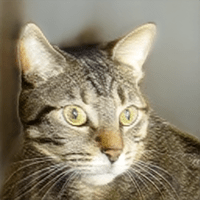}\hfill &

\includegraphics[width=\linewidth]{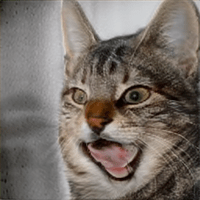}\hfill &
\includegraphics[width=\linewidth]{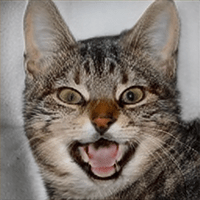}\hfill
\llap{\includegraphics[trim=650 100 380 60,clip,width=0.5\linewidth, height=\linewidth]{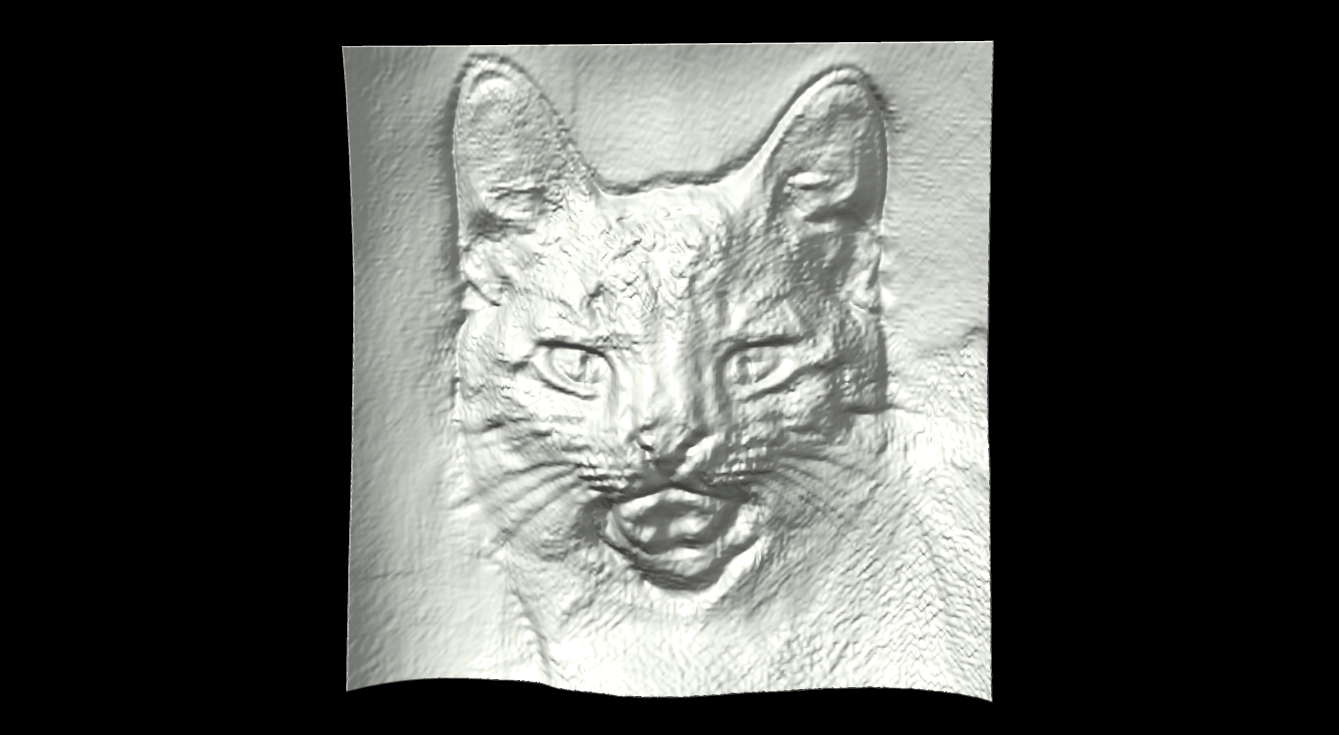}}&
\includegraphics[width=\linewidth]{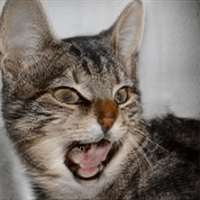}\hfill &

\includegraphics[width=\linewidth]{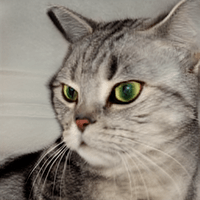}\hfill &
\includegraphics[width=\linewidth]{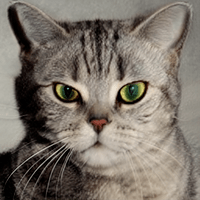}\hfill
\llap{\includegraphics[trim=650 90 380 100,clip,width=0.5\linewidth, height=\linewidth]{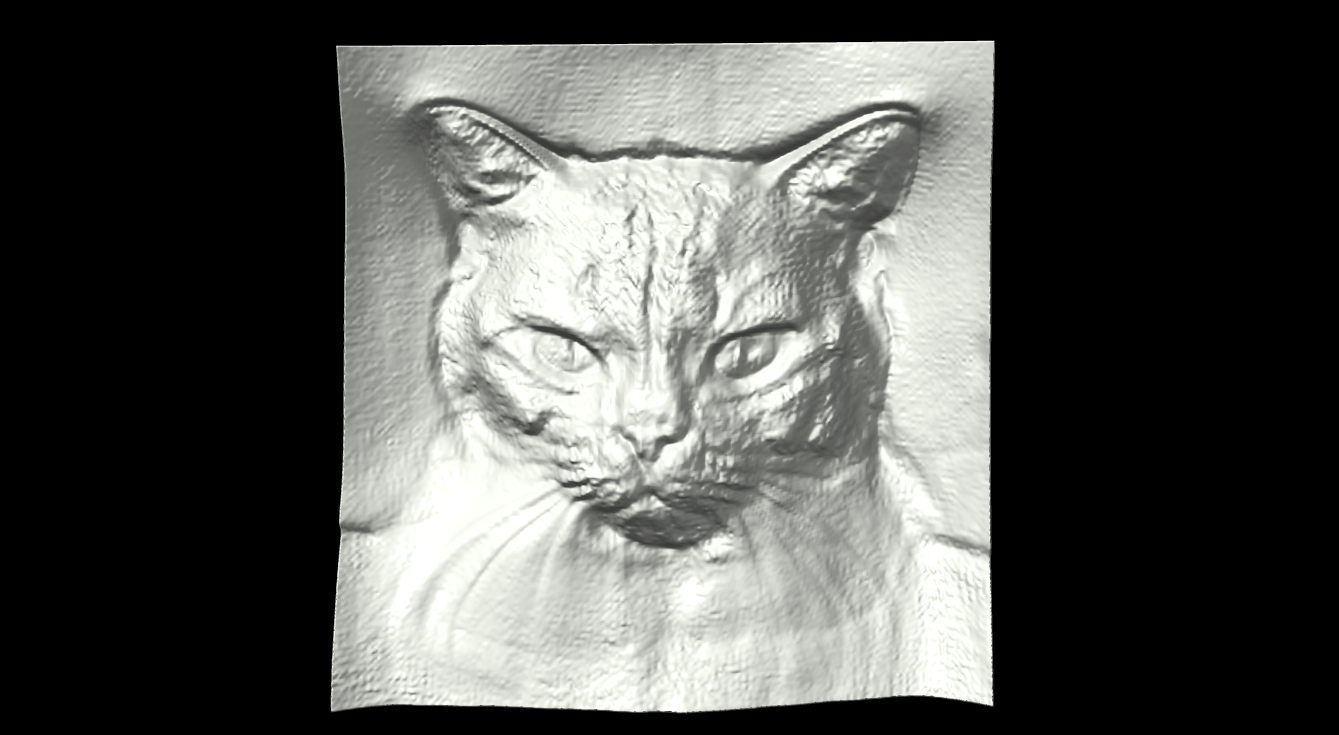}}&
\includegraphics[width=\linewidth]{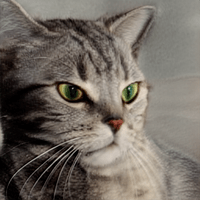}\hfill \\

\includegraphics[width=\linewidth]{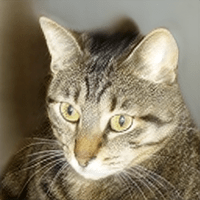}\hfill &
\includegraphics[width=\linewidth]{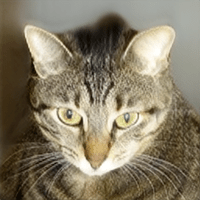}\hfill&
\includegraphics[width=\linewidth]{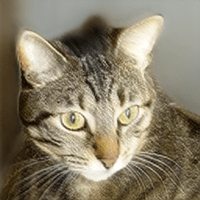}\hfill &

\includegraphics[width=\linewidth]{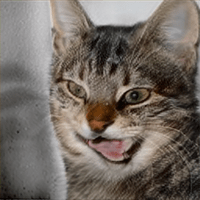}\hfill &
\includegraphics[width=\linewidth]{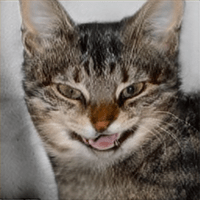}\hfill&
\includegraphics[width=\linewidth]{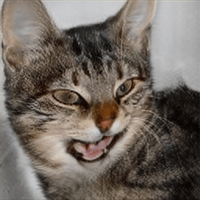}\hfill &

\includegraphics[width=\linewidth]{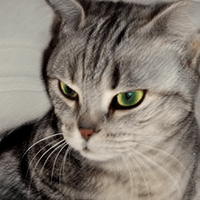}\hfill &
\includegraphics[width=\linewidth]{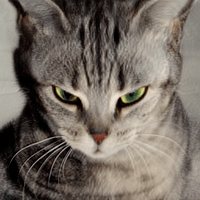}\hfill&
\includegraphics[width=\linewidth]{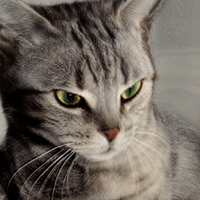}\hfill \\\\\\

& \includegraphics[width=\linewidth]{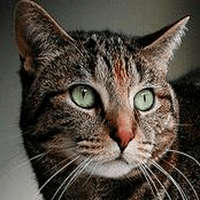}\hfill &
&& \includegraphics[width=\linewidth]{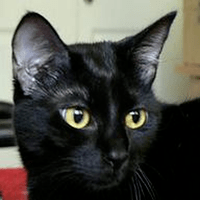}\hfill &
&& \includegraphics[width=\linewidth]{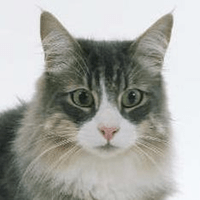}\hfill \\

\includegraphics[width=\linewidth]{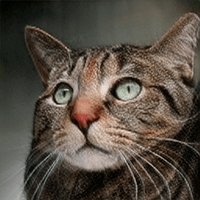}\hfill &
\includegraphics[width=\linewidth]{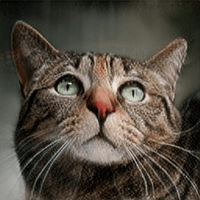}\hfill&
\includegraphics[width=\linewidth]{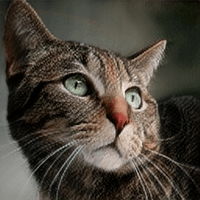}\hfill &

\includegraphics[width=\linewidth]{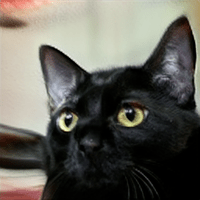}\hfill &
\includegraphics[width=\linewidth]{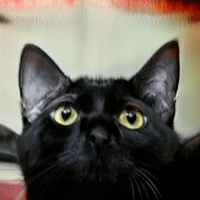}\hfill&
\includegraphics[width=\linewidth]{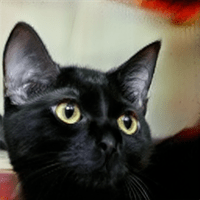}\hfill &

\includegraphics[width=\linewidth]{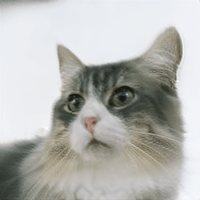}\hfill &
\includegraphics[width=\linewidth]{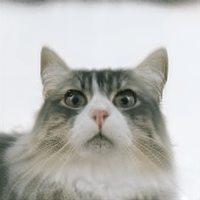}\hfill&
\includegraphics[width=\linewidth]{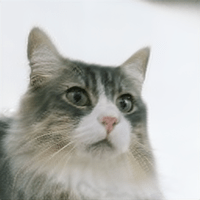}\hfill \\

\includegraphics[width=\linewidth]{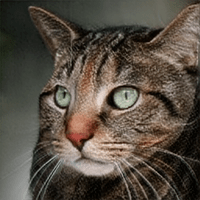}\hfill &
\includegraphics[width=\linewidth]{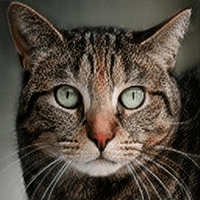}\hfill
\llap{\includegraphics[trim=650 90 380 60,clip,width=0.5\linewidth, height=\linewidth]{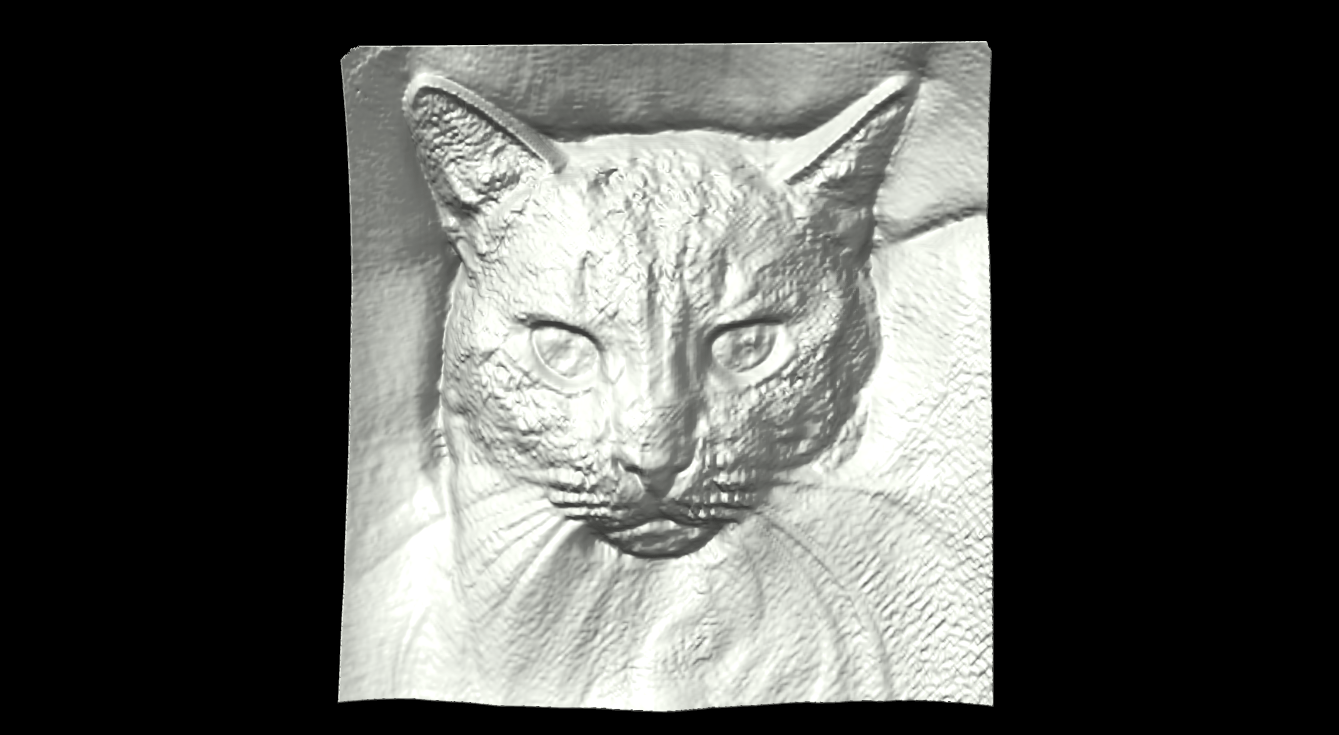}}&
\includegraphics[width=\linewidth]{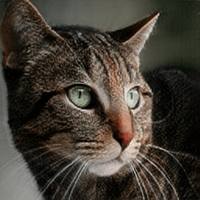}\hfill &

\includegraphics[width=\linewidth]{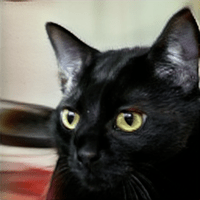}\hfill &
\includegraphics[width=\linewidth]{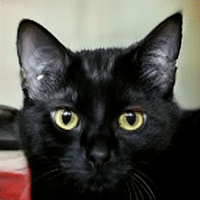}\hfill
\llap{\includegraphics[trim=650 90 380 100,clip,width=0.5\linewidth, height=\linewidth]{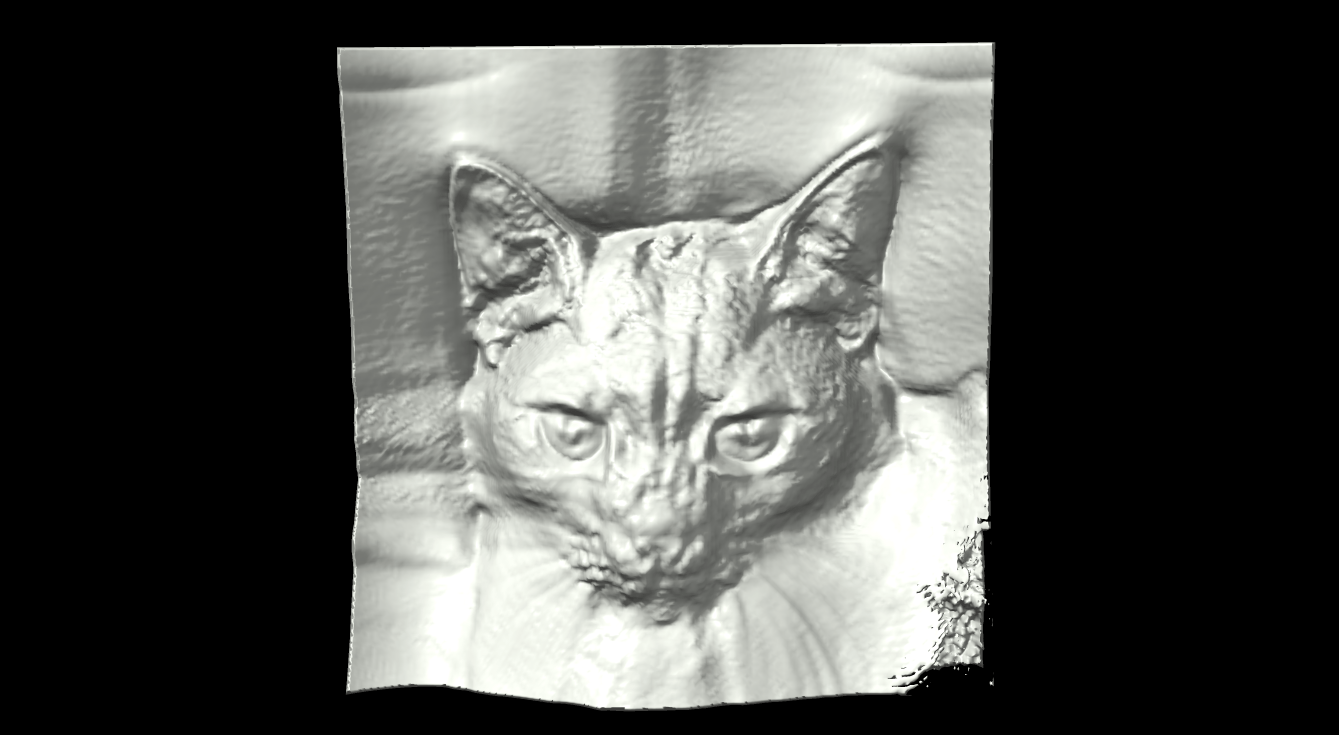}}&
\includegraphics[width=\linewidth]{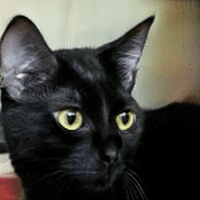}\hfill &

\includegraphics[width=\linewidth]{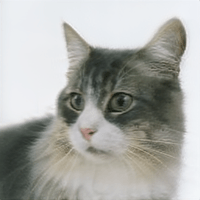}\hfill &
\includegraphics[width=\linewidth]{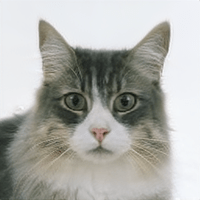}\hfill
\llap{\includegraphics[trim=650 90 380 100,clip,width=0.5\linewidth, height=\linewidth]{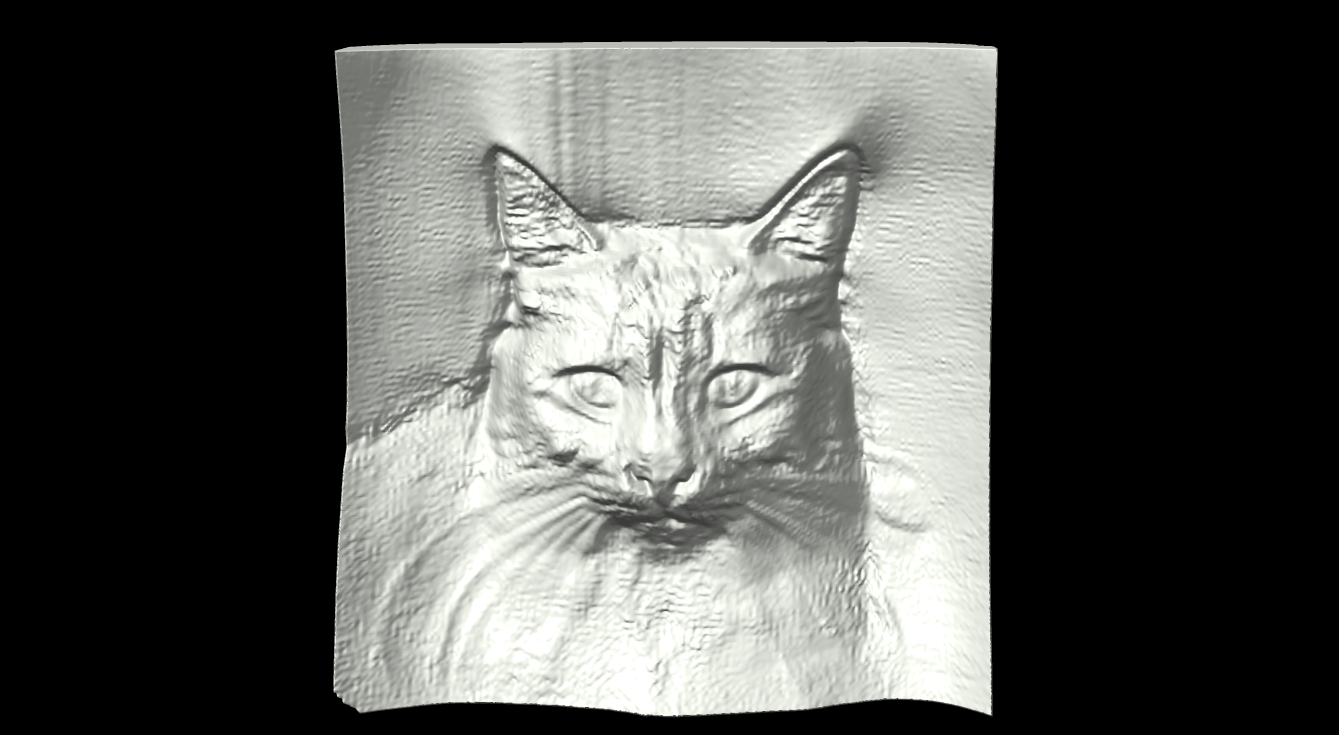}}&
\includegraphics[width=\linewidth]{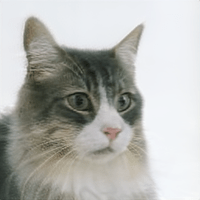}\hfill \\

\includegraphics[width=\linewidth]{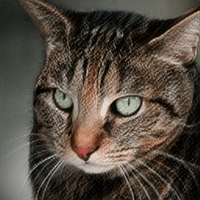}\hfill &
\includegraphics[width=\linewidth]{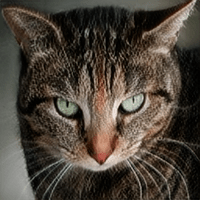}\hfill&
\includegraphics[width=\linewidth]{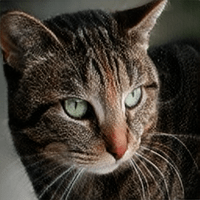}\hfill &

\includegraphics[width=\linewidth]{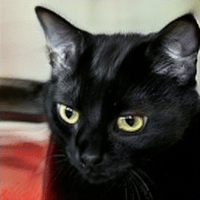}\hfill &
\includegraphics[width=\linewidth]{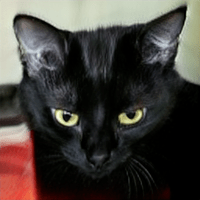}\hfill&
\includegraphics[width=\linewidth]{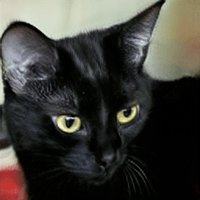}\hfill &

\includegraphics[width=\linewidth]{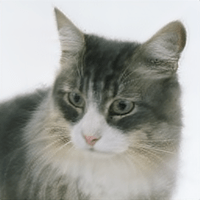}\hfill &
\includegraphics[width=\linewidth]{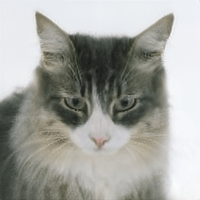}\hfill&
\includegraphics[width=\linewidth]{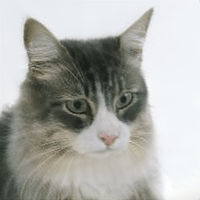}\hfill \\
\end{tabular}
\caption{\textbf{Additional qualitative results on out-of-domain dataset AnimalFace10 dataset~\cite{liu2019few}}}
\label{appendix:cat_inversion}
\end{figure}

Following the baseline comparison in our main paper, we provide additional inversion results in \figref{appendix:ablation} and also provide random viewpoints using the acquired latent code. 
\figref{appendix:facial_inversion1}, \figref{appendix:facial_inversion2} and  \figref{appendix:facial_inversion3} applies our inversion method on multiple facial datasets, and we demonstrate the effectiveness of 3D GAN inversion by providing the reconstructed mesh and novel views using the acquired latent code. 
\figref{appendix:cat_inversion} depicts our reconstruction ability on non-facial domain, namely cats in the AnimalFace10 dataset. 

Finally, additional GANspace~\cite{harkonen2020ganspace} editing results can be found in \figref{appendix:face_edit} and \figref{appendix:cat_edit}, where we use the edited latent code to generate 3D mesh and novel views and prove our method is capable of 3D shape editing.

\begin{figure*}[!p]
\centering
\newcolumntype{M}[1]{>{\centering\arraybackslash}m{#1}}
\setlength{\tabcolsep}{1pt}
\renewcommand{\arraystretch}{0.5}
\begin{tabular}{M{0.113\linewidth}M{0.113\linewidth} @{\hskip 0.005\linewidth}|@{\hskip 0.005\linewidth} M{0.113\linewidth}M{0.113\linewidth}M{0.113\linewidth} @{\hskip 0.005\linewidth}|@{\hskip 0.005\linewidth} M{0.113\linewidth}M{0.113\linewidth}M{0.113\linewidth}}

\multicolumn{1}{c}{Input} 
& \multicolumn{1}{c}{Ours} 
&\multicolumn{1}{c}{SG2} 
&\multicolumn{1}{c}{SG2 $\mathcal{W}+$}
&\multicolumn{1}{c}{PTI} 
&\multicolumn{1}{c}{SG2$^\dagger$}
&\multicolumn{1}{c}{SG2$^\dagger$ $\mathcal{W}+$} & \multicolumn{1}{c}{PTI$^\dagger$ } \\

\includegraphics[width=\linewidth]{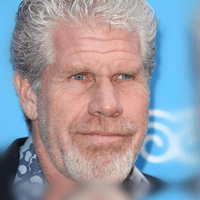}\hfill &

\includegraphics[width=\linewidth]{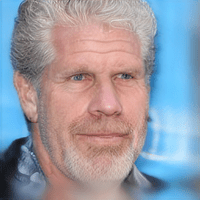}\hfill &
\includegraphics[width=\linewidth]{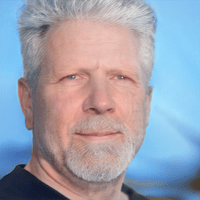}\hfill & 
\includegraphics[width=\linewidth]{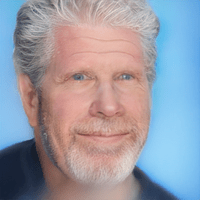}\hfill& 
\includegraphics[width=\linewidth]{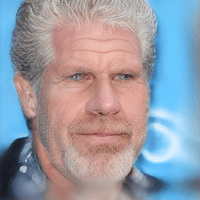}\hfill &
\includegraphics[width=\linewidth]{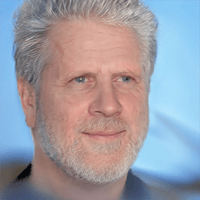}\hfill &
\includegraphics[width=\linewidth]{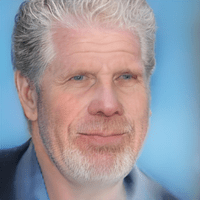}\hfill &
\includegraphics[width=\linewidth]{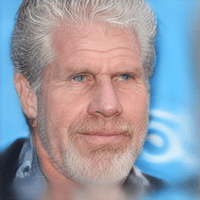}\hfill \\
&
\includegraphics[width=\linewidth]{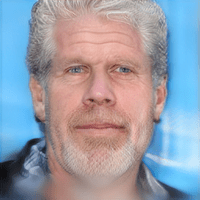}\hfill &
\includegraphics[width=\linewidth]{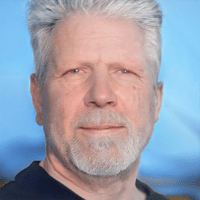}\hfill & 
\includegraphics[width=\linewidth]{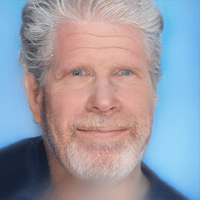}\hfill& 
\includegraphics[width=\linewidth]{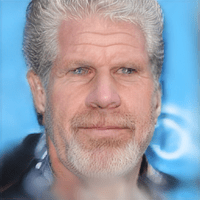}\hfill &
\includegraphics[width=\linewidth]{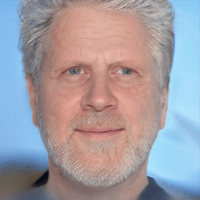}\hfill &
\includegraphics[width=\linewidth]{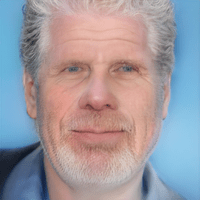}\hfill &
\includegraphics[width=\linewidth]{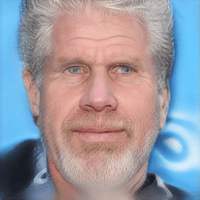}\hfill \\
&
\includegraphics[width=\linewidth]{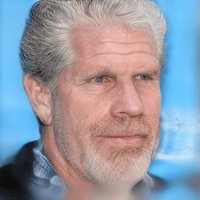}\hfill &
\includegraphics[width=\linewidth]{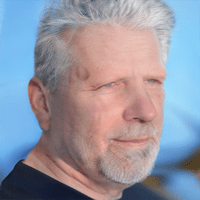}\hfill & 
\includegraphics[width=\linewidth]{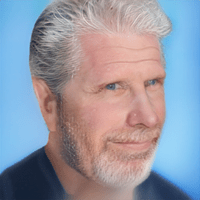}\hfill& 
\includegraphics[width=\linewidth]{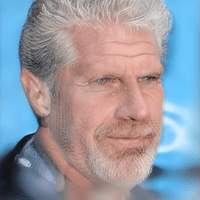}\hfill &
\includegraphics[width=\linewidth]{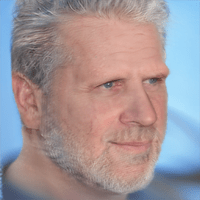}\hfill &
\includegraphics[width=\linewidth]{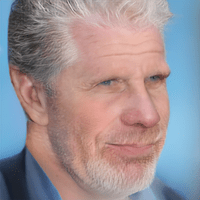}\hfill &
\includegraphics[width=\linewidth]{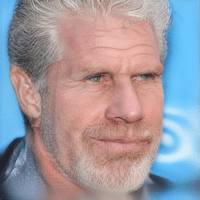}\hfill \\\\\\

\includegraphics[width=\linewidth]{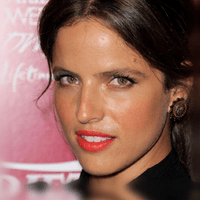}\hfill &

\includegraphics[width=\linewidth]{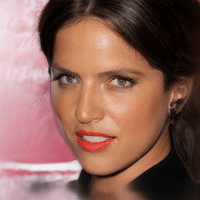}\hfill &
\includegraphics[width=\linewidth]{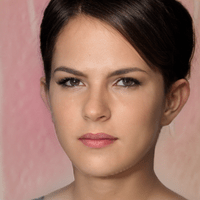}\hfill & 
\includegraphics[width=\linewidth]{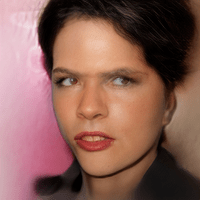}\hfill& 
\includegraphics[width=\linewidth]{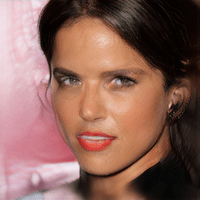}\hfill &
\includegraphics[width=\linewidth]{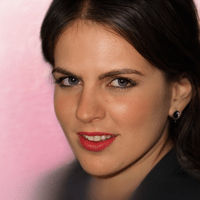}\hfill &
\includegraphics[width=\linewidth]{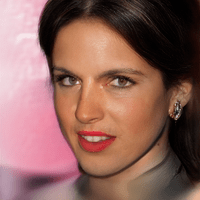}\hfill &
\includegraphics[width=\linewidth]{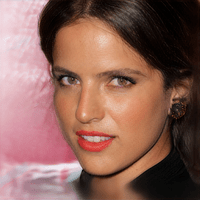}\hfill \\
&
\includegraphics[width=\linewidth]{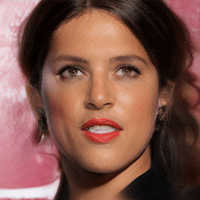}\hfill &
\includegraphics[width=\linewidth]{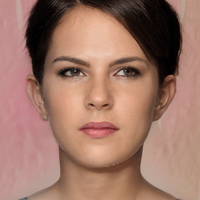}\hfill & 
\includegraphics[width=\linewidth]{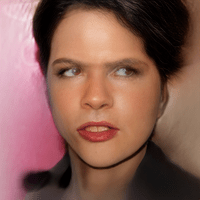}\hfill& 
\includegraphics[width=\linewidth]{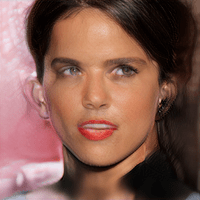}\hfill &
\includegraphics[width=\linewidth]{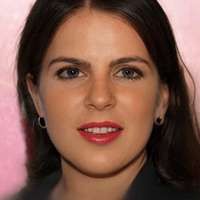}\hfill &
\includegraphics[width=\linewidth]{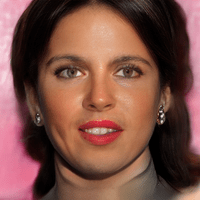}\hfill &
\includegraphics[width=\linewidth]{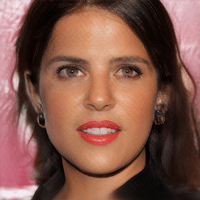}\hfill \\
&
\includegraphics[width=\linewidth]{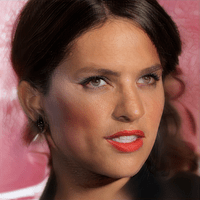}\hfill &
\includegraphics[width=\linewidth]{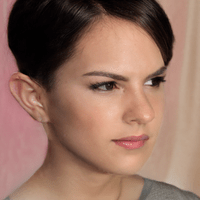}\hfill & 
\includegraphics[width=\linewidth]{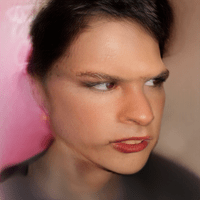}\hfill& 
\includegraphics[width=\linewidth]{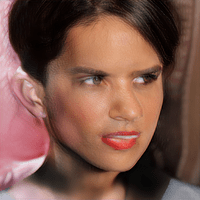}\hfill &
\includegraphics[width=\linewidth]{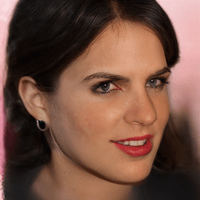}\hfill &
\includegraphics[width=\linewidth]{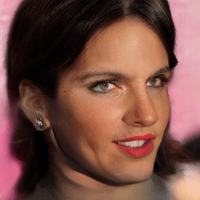}\hfill &
\includegraphics[width=\linewidth]{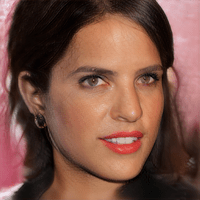}\hfill \\\\\\

\includegraphics[width=\linewidth]{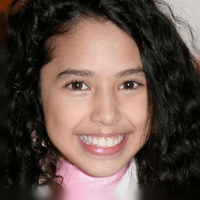}\hfill &

\includegraphics[width=\linewidth]{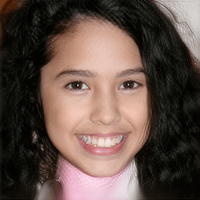}\hfill &
\includegraphics[width=\linewidth]{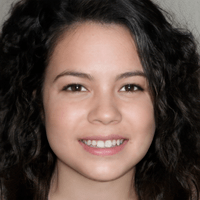}\hfill & 
\includegraphics[width=\linewidth]{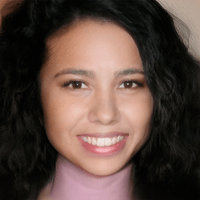}\hfill& 
\includegraphics[width=\linewidth]{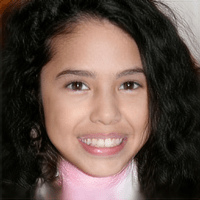}\hfill &
\includegraphics[width=\linewidth]{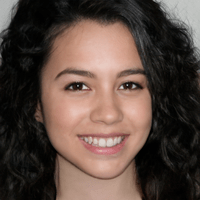}\hfill &
\includegraphics[width=\linewidth]{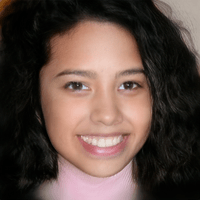}\hfill &
\includegraphics[width=\linewidth]{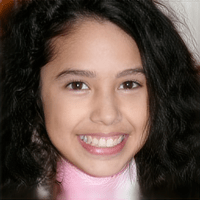}\hfill \\
&
\includegraphics[width=\linewidth]{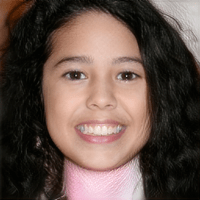}\hfill &
\includegraphics[width=\linewidth]{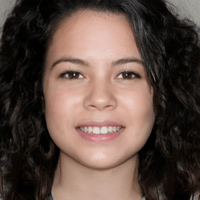}\hfill & 
\includegraphics[width=\linewidth]{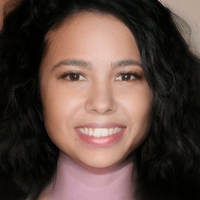}\hfill& 
\includegraphics[width=\linewidth]{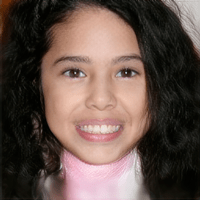}\hfill &
\includegraphics[width=\linewidth]{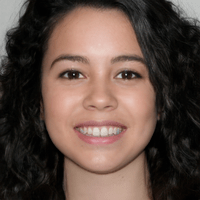}\hfill &
\includegraphics[width=\linewidth]{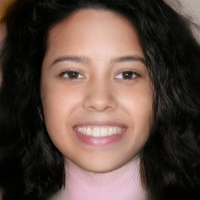}\hfill &
\includegraphics[width=\linewidth]{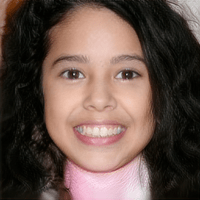}\hfill \\
&
\includegraphics[width=\linewidth]{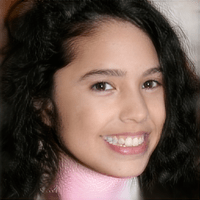}\hfill &
\includegraphics[width=\linewidth]{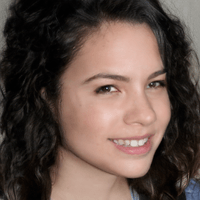}\hfill & 
\includegraphics[width=\linewidth]{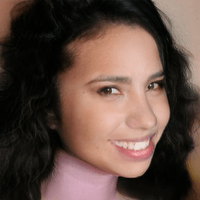}\hfill& 
\includegraphics[width=\linewidth]{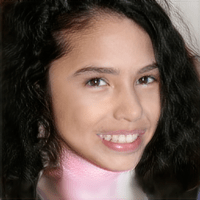}\hfill &
\includegraphics[width=\linewidth]{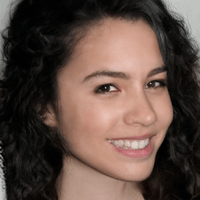}\hfill &
\includegraphics[width=\linewidth]{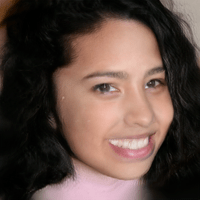}\hfill &
\includegraphics[width=\linewidth]{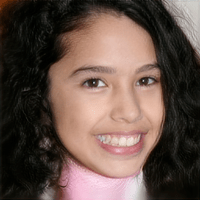}\hfill \\

\end{tabular}
\caption{\textbf{Additional ablation studies on CelebA-HQ~\cite{karras2017progressive, liu2015faceattributes} dataset}}
\label{appendix:ablation}
\end{figure*}
\begin{figure}[!p]
\centering
\newcolumntype{M}[1]{>{\centering\arraybackslash}m{#1}}
\setlength{\tabcolsep}{1pt}
\renewcommand{\arraystretch}{0.5}
\begin{tabular}{M{0.1\linewidth}M{0.1\linewidth}M{0.1\linewidth}
@{\hskip 0.01\linewidth} M{0.1\linewidth}M{0.1\linewidth}M{0.1\linewidth} @{\hskip 0.01\linewidth} M{0.1\linewidth}M{0.1\linewidth}M{0.1\linewidth}}

& \includegraphics[width=\linewidth]{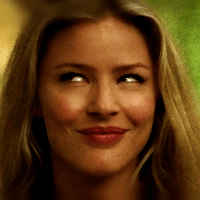}\hfill &
&& \includegraphics[width=\linewidth]{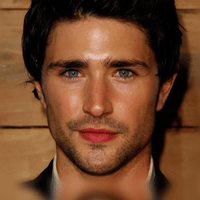}\hfill &
&& \includegraphics[width=\linewidth]{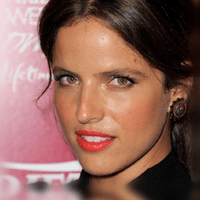}\hfill \\

\includegraphics[width=\linewidth]{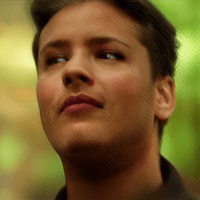}\hfill &
\includegraphics[width=\linewidth]{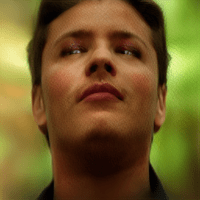}\hfill&
\includegraphics[width=\linewidth]{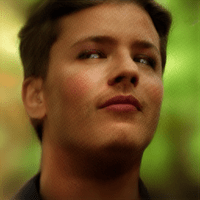}\hfill &

\includegraphics[width=\linewidth]{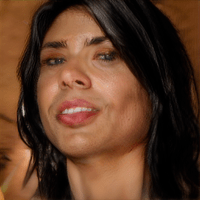}\hfill &
\includegraphics[width=\linewidth]{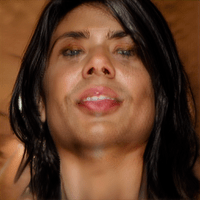}\hfill&
\includegraphics[width=\linewidth]{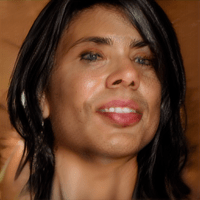}\hfill &

\includegraphics[width=\linewidth]{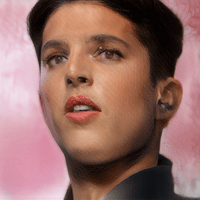}\hfill &
\includegraphics[width=\linewidth]{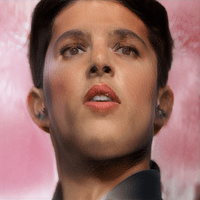}\hfill&
\includegraphics[width=\linewidth]{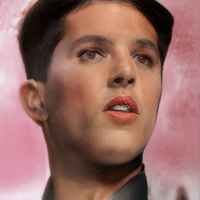}\hfill \\

\includegraphics[width=\linewidth]{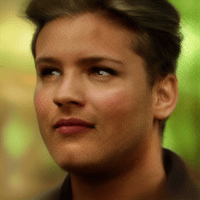}\hfill &
\includegraphics[width=\linewidth]{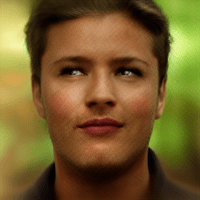}\hfill
\llap{\includegraphics[trim=670 100 360 90,clip,width=0.5\linewidth, height=\linewidth]{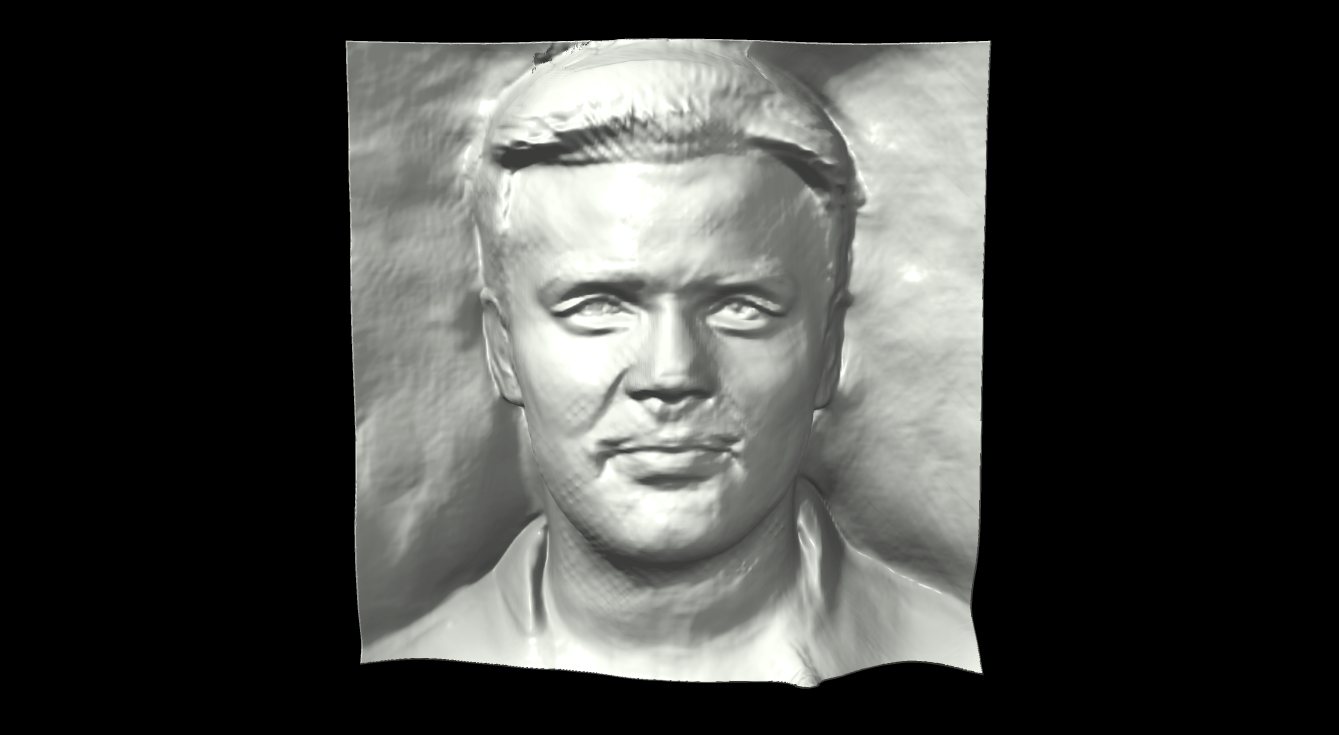}}&
\includegraphics[width=\linewidth]{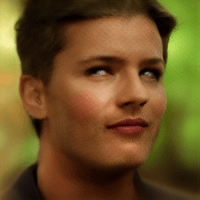}\hfill &

\includegraphics[width=\linewidth]{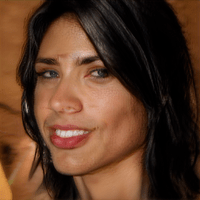}\hfill &
\includegraphics[width=\linewidth]{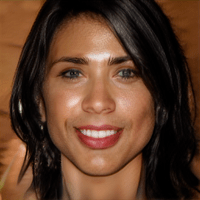}\hfill
\llap{\includegraphics[trim=670 100 360 90,clip,width=0.5\linewidth, height=\linewidth]{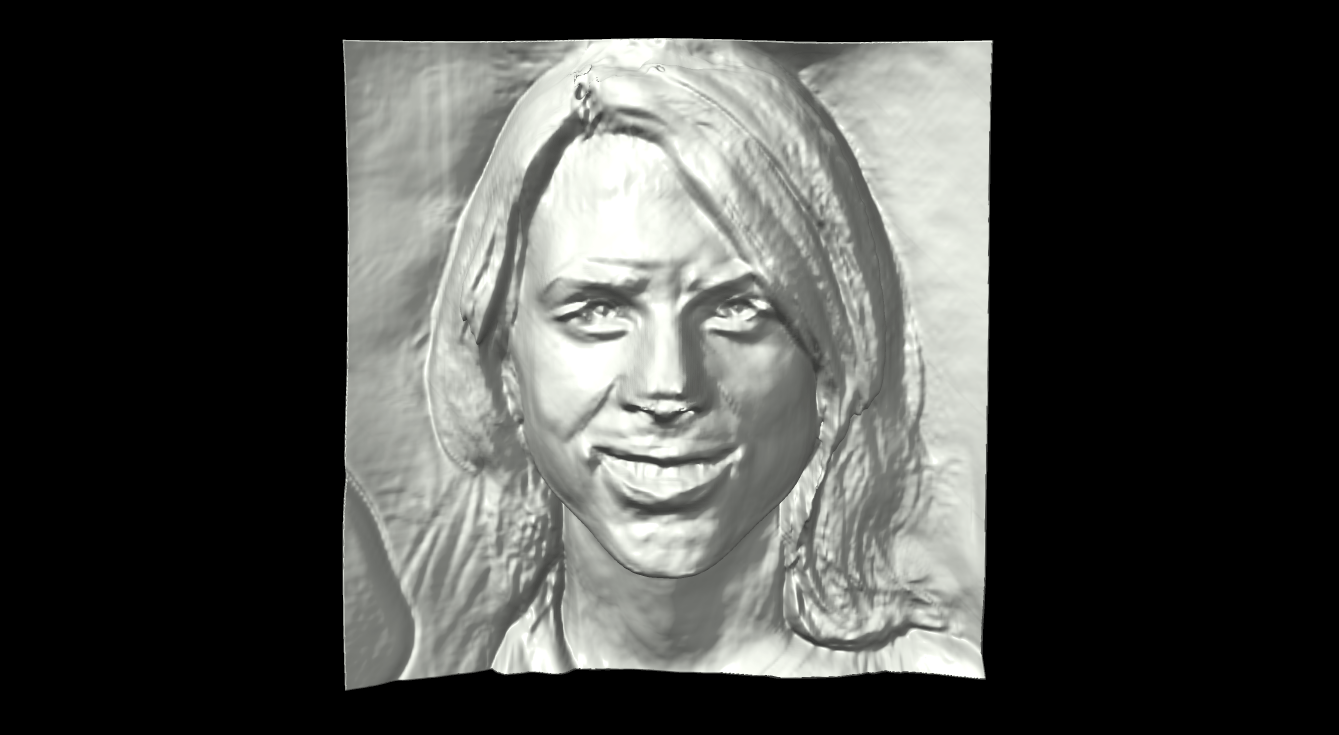}}&
\includegraphics[width=\linewidth]{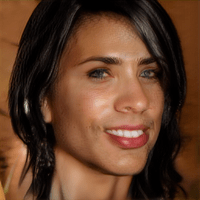}\hfill &

\includegraphics[width=\linewidth]{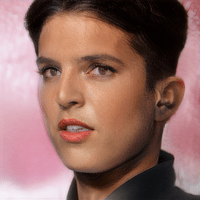}\hfill &
\includegraphics[width=\linewidth]{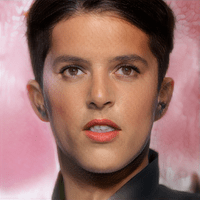}\hfill
\llap{\includegraphics[trim=670 100 360 90,clip,width=0.5\linewidth, height=\linewidth]{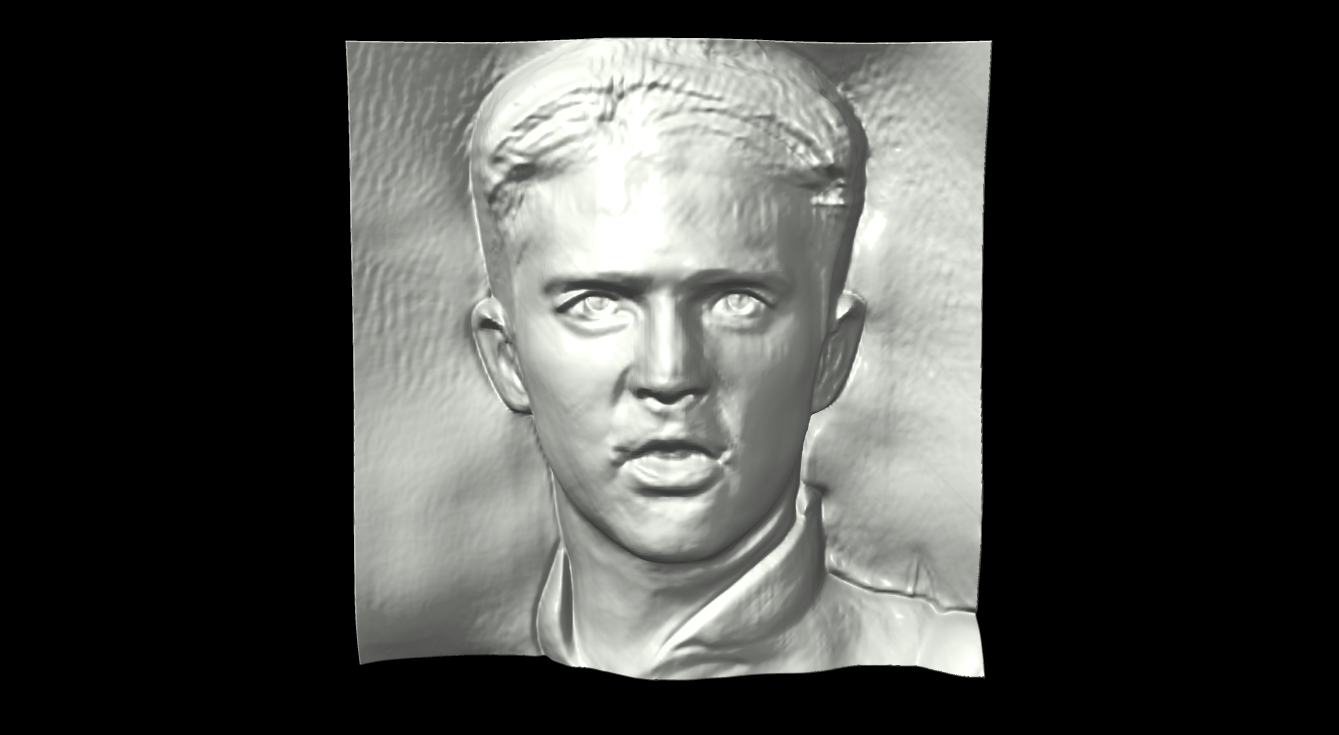}}&
\includegraphics[width=\linewidth]{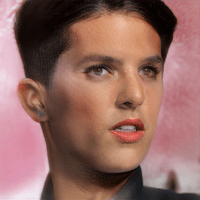}\hfill \\

\includegraphics[width=\linewidth]{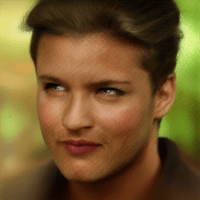}\hfill &
\includegraphics[width=\linewidth]{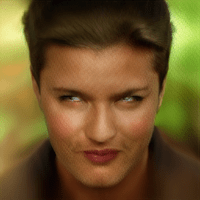}\hfill&
\includegraphics[width=\linewidth]{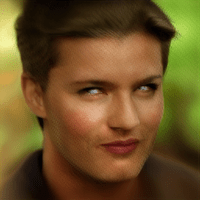}\hfill &

\includegraphics[width=\linewidth]{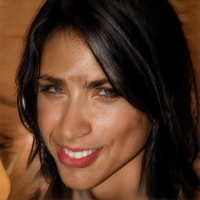}\hfill &
\includegraphics[width=\linewidth]{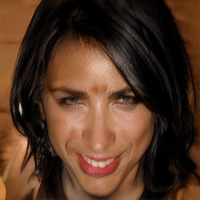}\hfill&
\includegraphics[width=\linewidth]{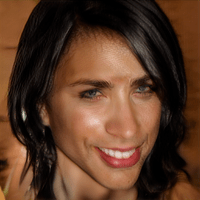}\hfill &

\includegraphics[width=\linewidth]{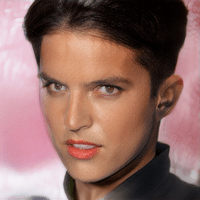}\hfill &
\includegraphics[width=\linewidth]{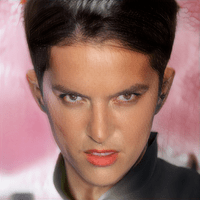}\hfill&
\includegraphics[width=\linewidth]{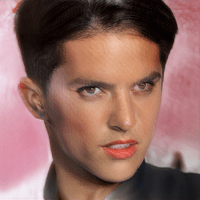}\hfill \\\\\\

\includegraphics[width=\linewidth]{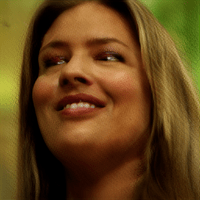}\hfill &
\includegraphics[width=\linewidth]{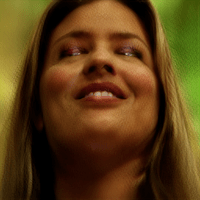}\hfill&
\includegraphics[width=\linewidth]{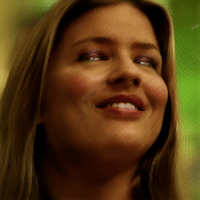}\hfill &

\includegraphics[width=\linewidth]{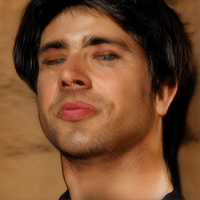}\hfill &
\includegraphics[width=\linewidth]{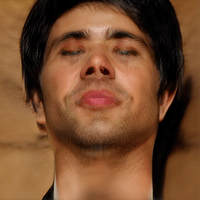}\hfill&
\includegraphics[width=\linewidth]{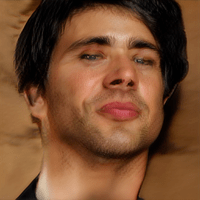}\hfill &

\includegraphics[width=\linewidth]{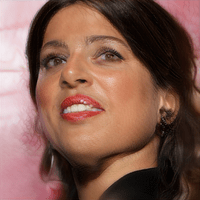}\hfill &
\includegraphics[width=\linewidth]{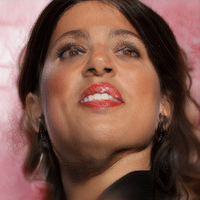}\hfill&
\includegraphics[width=\linewidth]{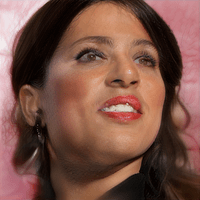}\hfill \\

\includegraphics[width=\linewidth]{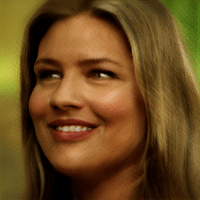}\hfill &
\includegraphics[width=\linewidth]{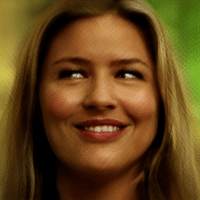}\hfill
\llap{\includegraphics[trim=670 110 360 90,clip,width=0.5\linewidth, height=\linewidth]{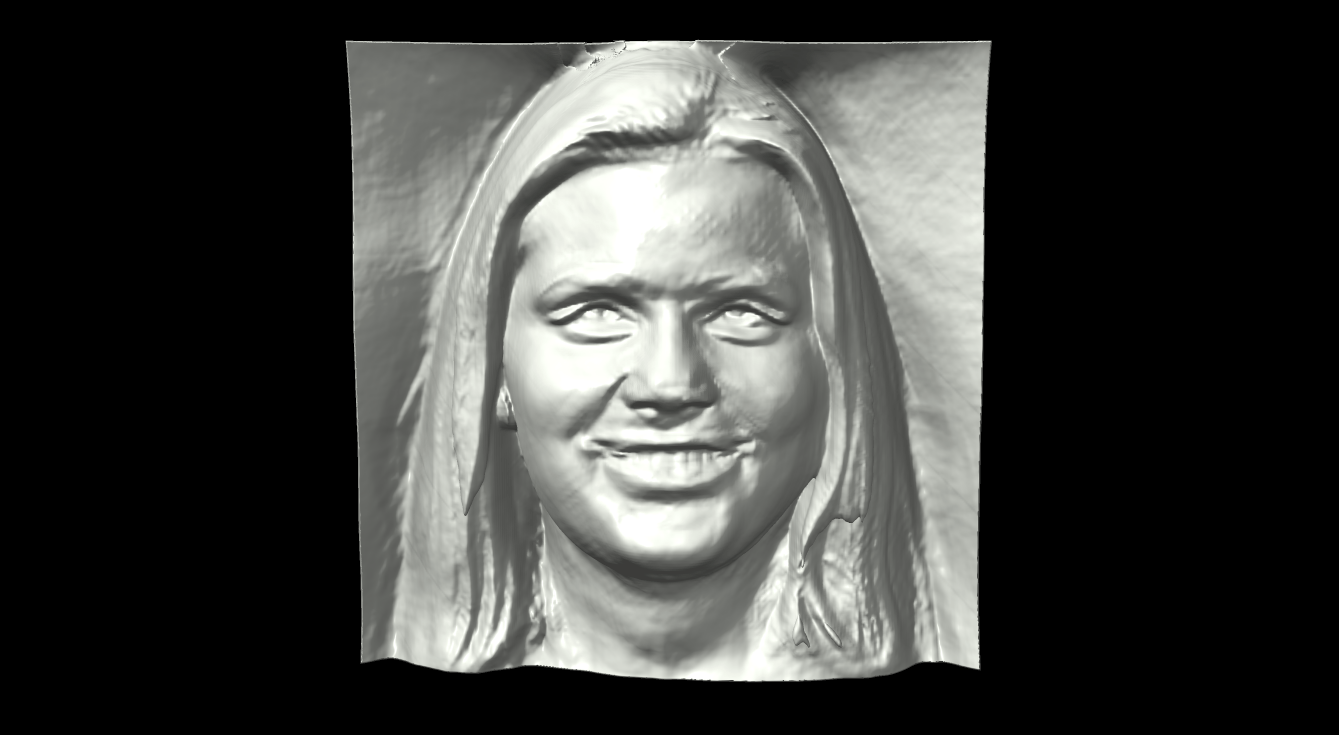}}&
\includegraphics[width=\linewidth]{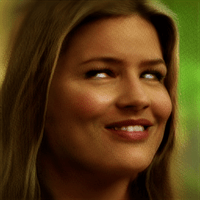}\hfill &

\includegraphics[width=\linewidth]{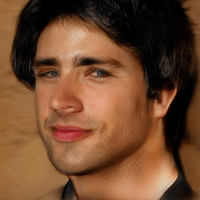}\hfill &
\includegraphics[width=\linewidth]{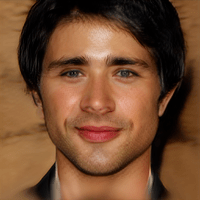}\hfill
\llap{\includegraphics[trim=670 110 360 90,clip,width=0.5\linewidth, height=\linewidth]{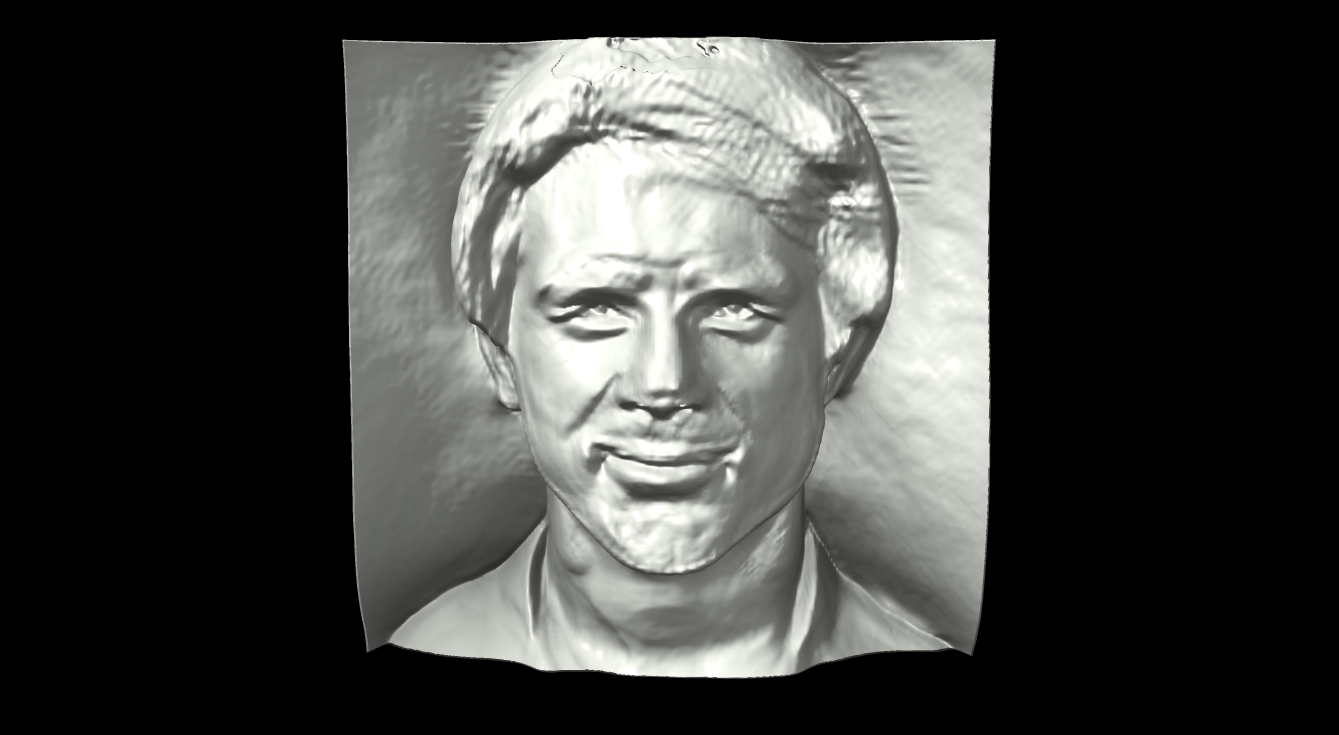}}&
\includegraphics[width=\linewidth]{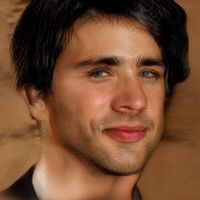}\hfill &

\includegraphics[width=\linewidth]{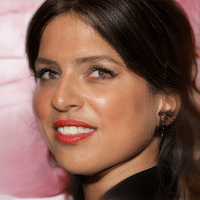}\hfill &
\includegraphics[width=\linewidth]{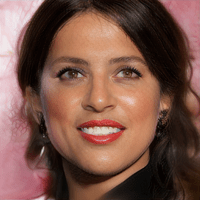}\hfill
\llap{\includegraphics[trim=670 110 360 90,clip,width=0.5\linewidth, height=\linewidth]{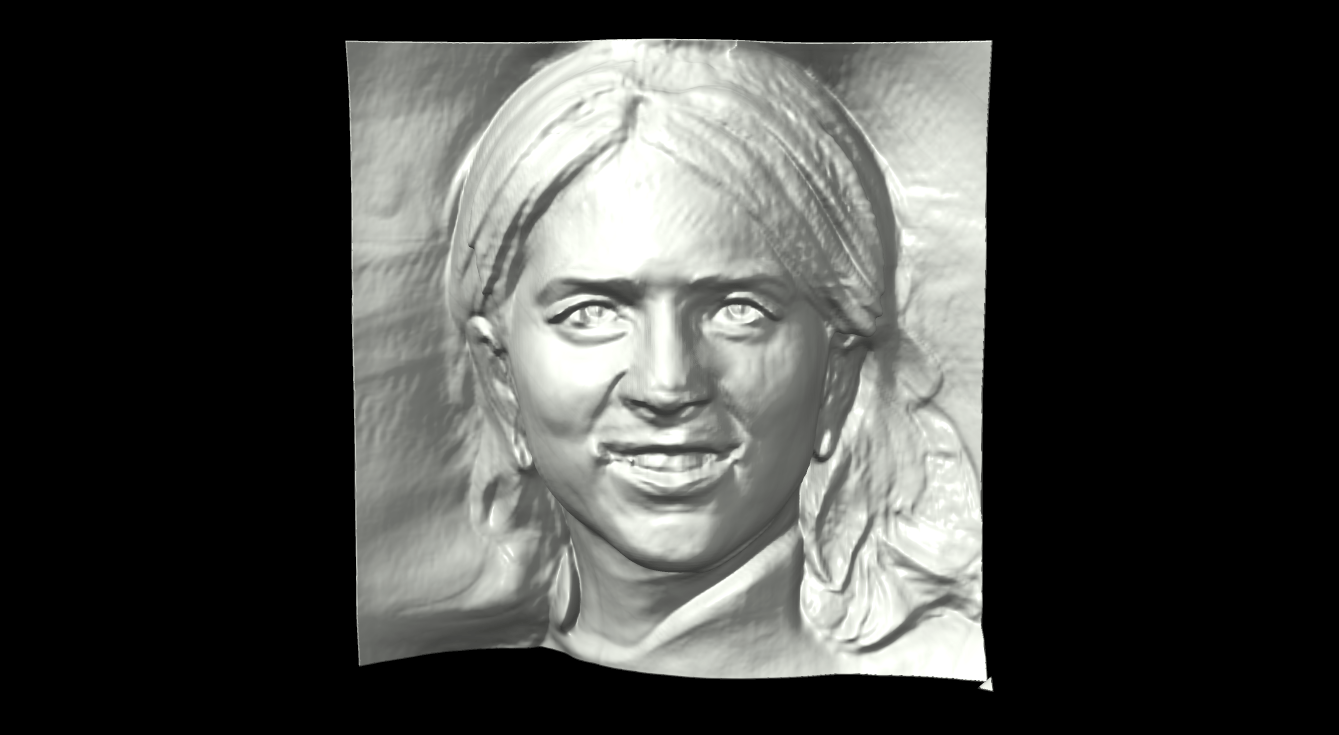}}&
\includegraphics[width=\linewidth]{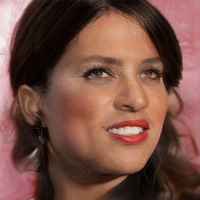}\hfill \\

\includegraphics[width=\linewidth]{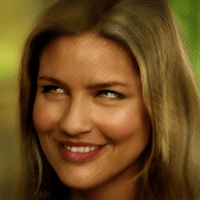}\hfill &
\includegraphics[width=\linewidth]{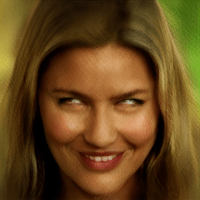}\hfill&
\includegraphics[width=\linewidth]{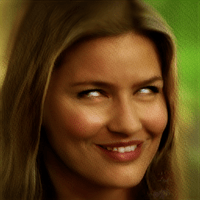}\hfill &

\includegraphics[width=\linewidth]{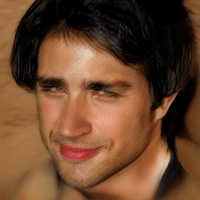}\hfill &
\includegraphics[width=\linewidth]{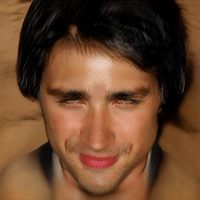}\hfill&
\includegraphics[width=\linewidth]{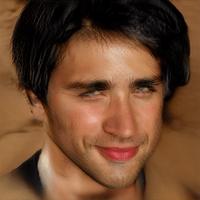}\hfill &

\includegraphics[width=\linewidth]{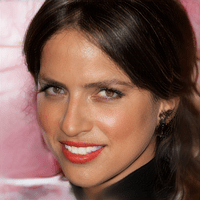}\hfill &
\includegraphics[width=\linewidth]{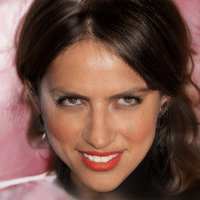}\hfill&
\includegraphics[width=\linewidth]{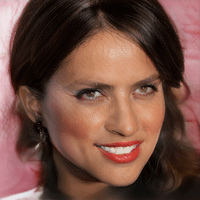}\hfill \\\\\\
\end{tabular}
\vspace{-10pt}
\caption{\textbf{3D facial Editing.} We demonstrate the style-aware geometry editing: gender(first row) and smile(second row). }
\label{appendix:face_edit}
% \end{figure}
% \begin{figure}[!p]
% \centering
% \newcolumntype{M}[1]{>{\centering\arraybackslash}m{#1}}
% \setlength{\tabcolsep}{1pt}
% \renewcommand{\arraystretch}{0.5}
\vspace{10pt}
\begin{tabular}{M{0.1\linewidth}M{0.1\linewidth}M{0.1\linewidth}
@{\hskip 0.01\linewidth} M{0.1\linewidth}M{0.1\linewidth}M{0.1\linewidth} @{\hskip 0.01\linewidth} M{0.1\linewidth}M{0.1\linewidth}M{0.1\linewidth}}
& \includegraphics[width=\linewidth]{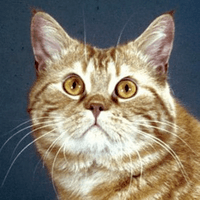}\hfill &
&& \includegraphics[width=\linewidth]{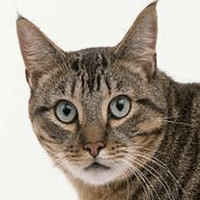}\hfill &
&& \includegraphics[width=\linewidth]{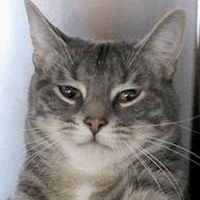}\hfill \\

\includegraphics[width=\linewidth]{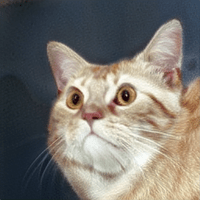}\hfill &
\includegraphics[width=\linewidth]{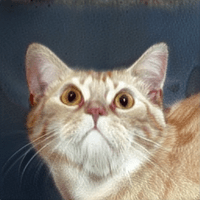}\hfill&
\includegraphics[width=\linewidth]{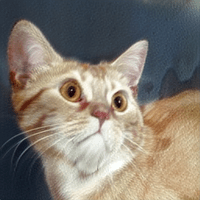}\hfill &

\includegraphics[width=\linewidth]{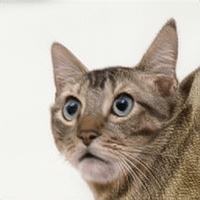}\hfill &
\includegraphics[width=\linewidth]{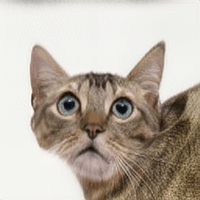}\hfill&
\includegraphics[width=\linewidth]{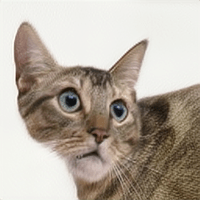}\hfill &

\includegraphics[width=\linewidth]{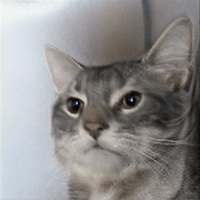}\hfill &
\includegraphics[width=\linewidth]{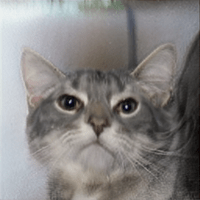}\hfill&
\includegraphics[width=\linewidth]{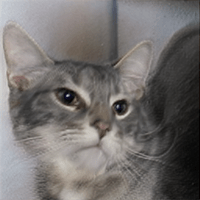}\hfill \\

\includegraphics[width=\linewidth]{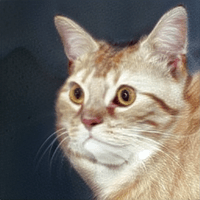}\hfill &
\includegraphics[width=\linewidth]{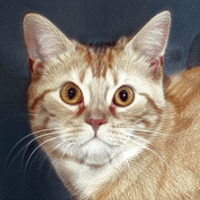}\hfill \llap{\includegraphics[trim=650 130 400 90,clip,width=0.5\linewidth, height=\linewidth]{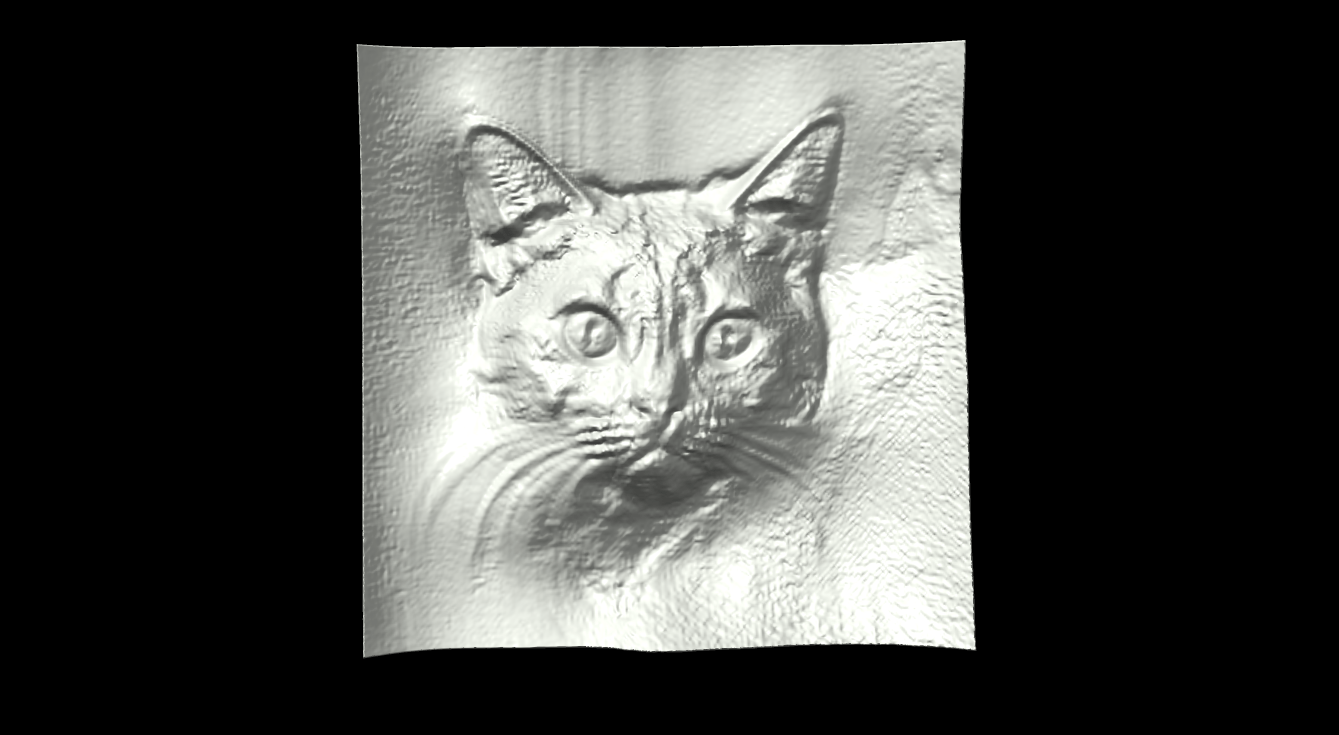}}&
\includegraphics[width=\linewidth]{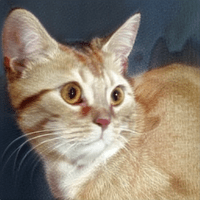}\hfill & 

\includegraphics[width=\linewidth]{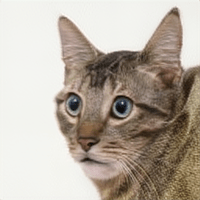}\hfill &
\includegraphics[width=\linewidth]{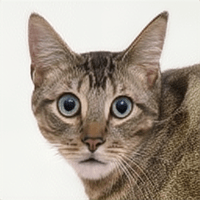}\hfill \llap{\includegraphics[trim=650 130 400 90,clip,width=0.5\linewidth, height=\linewidth]{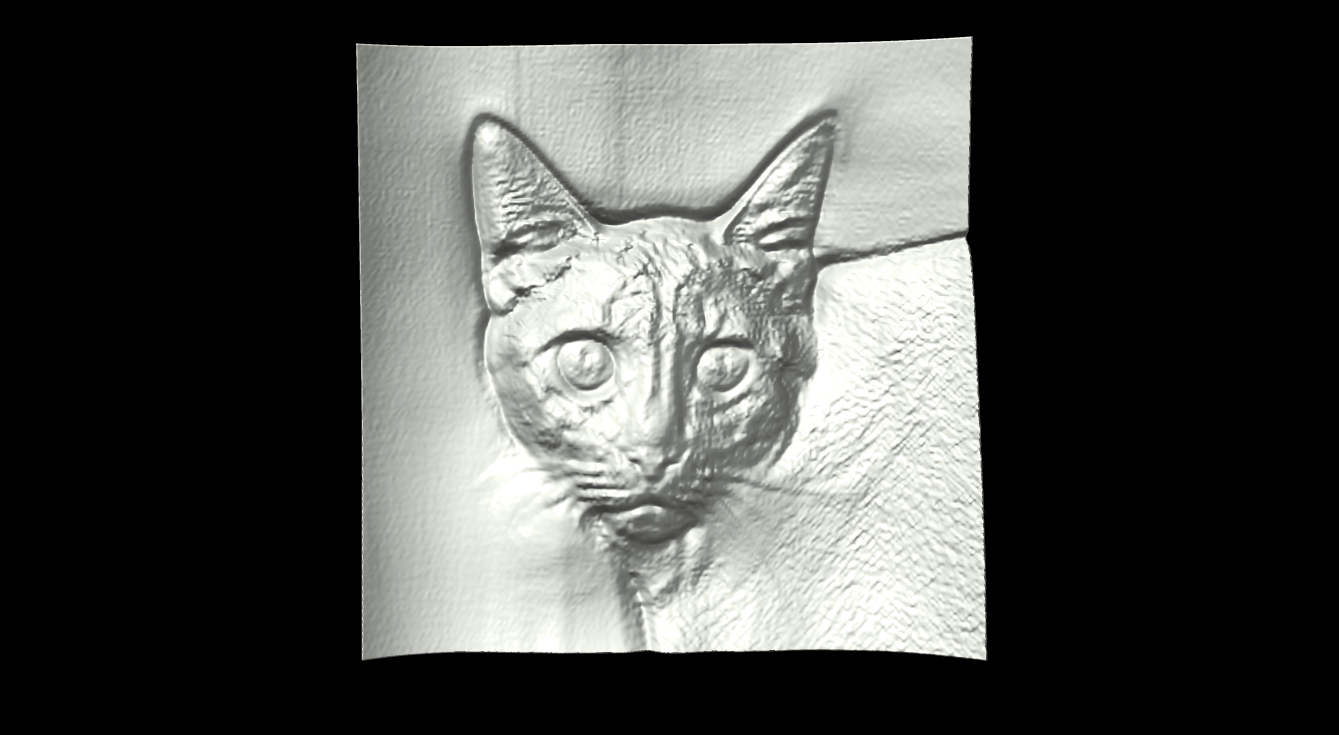}}&
\includegraphics[width=\linewidth]{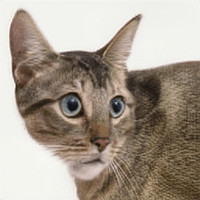}\hfill & 

\includegraphics[width=\linewidth]{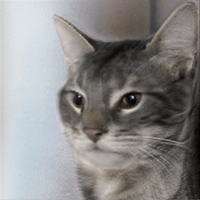}\hfill &
\includegraphics[width=\linewidth]{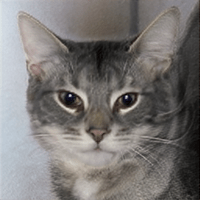}\hfill \llap{\includegraphics[trim=650 130 390 90,clip,width=0.5\linewidth, height=\linewidth]{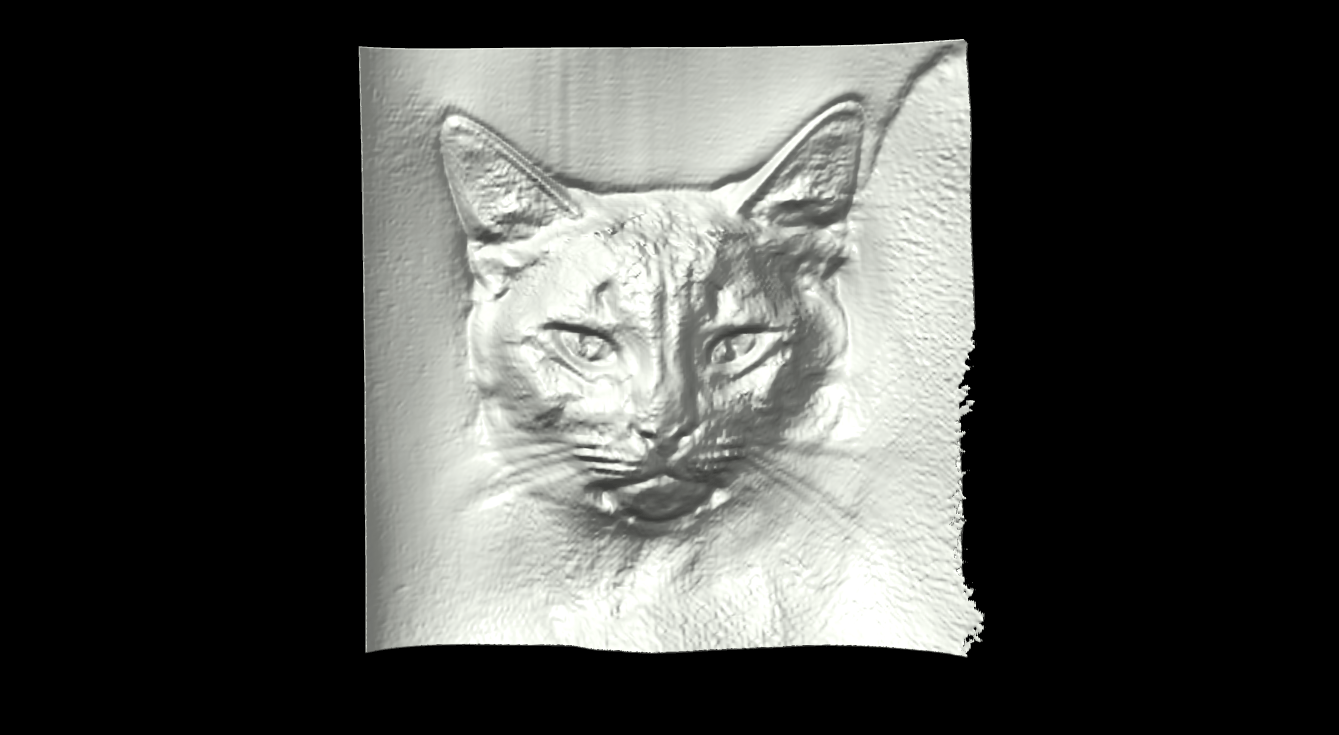}}&
\includegraphics[width=\linewidth]{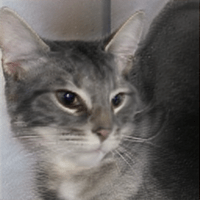}\hfill \\

\includegraphics[width=\linewidth]{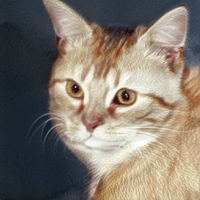}\hfill &
\includegraphics[width=\linewidth]{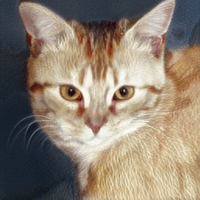}\hfill &
\includegraphics[width=\linewidth]{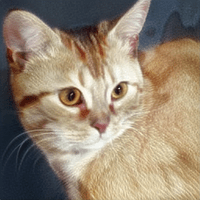}\hfill &

\includegraphics[width=\linewidth]{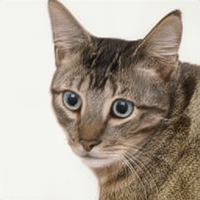}\hfill &
\includegraphics[width=\linewidth]{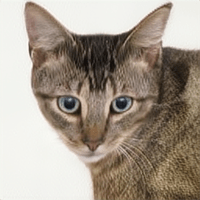}\hfill &
\includegraphics[width=\linewidth]{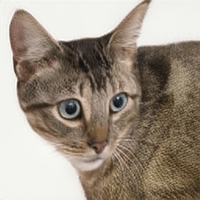}\hfill &

\includegraphics[width=\linewidth]{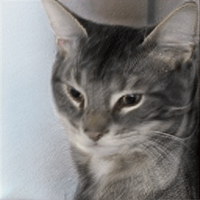}\hfill &
\includegraphics[width=\linewidth]{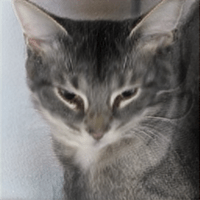}\hfill &
\includegraphics[width=\linewidth]{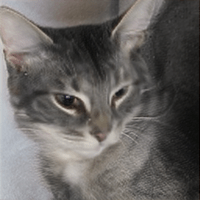}\hfill \\\\\\
\end{tabular}
\vspace{-10pt}
\caption{\textbf{Editing of cats in AnimalFace10 dataset~\cite{liu2019few} We evaluate 3D-aware edits: pupil size, and demonstrate our method enables latent-based editing for domains other than human face.}}
\label{appendix:cat_edit}
\end{figure}

% \include{AppendixFigure/catedit}
% \newpage
% {\small
% \bibliographystyle{ieee_fullname}
% \bibliography{egbib}
%  }

\end{document}